\pdfoutput=1  
\PassOptionsToPackage{table}{xcolor}
\documentclass[logo,address]{trfr}          

\usepackage[utf8]{inputenc}

\usepackage{amsmath,amssymb}
\usepackage{siunitx}
\usepackage[ruled,vlined,linesnumbered]{algorithm2e}

\usepackage{graphicx}
\usepackage{caption}
\usepackage{subcaption}
\usepackage{wrapfig}
\usepackage{float}
\usepackage{pdflscape}                  
\usepackage{rotating}                   
\graphicspath{{figures/}{./}}

\usepackage{booktabs}                   
\usepackage{multirow}                   
\usepackage{array}                      
\usepackage{makecell}
\usepackage{colortbl}                   
\usepackage{adjustbox}                  

\usepackage[table]{xcolor}

\usepackage[most]{tcolorbox}
\usepackage[shortlabels]{enumitem}
\usepackage{pifont}                     
\usepackage{soulutf8}                   
\usepackage{multicol}

\usepackage{tikz}
\usetikzlibrary{positioning,arrows.meta,fit,backgrounds,shadows,calc,tikzmark}
\usetikzlibrary{shapes.geometric}
\usetikzlibrary{matrix}
\usepackage{pgfplots}
\pgfplotsset{compat=1.18}
\usepgfplotslibrary{groupplots}

\pgfdeclarelayer{lband}
\pgfdeclarelayer{lbox}
\pgfsetlayers{lband,lbox,background,main}

\usepackage{hyperref}
\usepackage[numbers,sort&compress]{natbib}
\definecolor{trlightamber}{RGB}{248,234,221}
\definecolor{trdarkamber} {RGB}{212,121, 42}
\definecolor{trlightsky}  {RGB}{227,241,253}
\definecolor{trdarksky}   {RGB}{  8,116,227}
\definecolor{trlightteal} {RGB}{227,243,238}
\definecolor{trdarkteal}  {RGB}{ 77,178,153}
\definecolor{trgrayone}   {RGB}{249,247,245}
\definecolor{trgraytwo}   {RGB}{229,229,229}
\definecolor{trgraythree} {RGB}{159,159,159}
\definecolor{trgrayfour}  {RGB}{122,122,122}
\definecolor{trgraphite}  {RGB}{ 33, 34, 35}

\definecolor{cInk}    {HTML}{22262B}     
\definecolor{cMute}   {HTML}{6B7280}     
\definecolor{cPlan}   {HTML}{2F5D8C}     
\definecolor{cPlanBg} {HTML}{E9F0F7}
\definecolor{cRes}    {HTML}{1F7A6B}     
\definecolor{cResBg}  {HTML}{E7F3F0}
\definecolor{cComp}   {HTML}{7A4FA3}     
\definecolor{cCompBg} {HTML}{F2EAF9}
\definecolor{cTool}   {HTML}{B06A16}     
\definecolor{cToolBg} {HTML}{FCF2E3}
\definecolor{cRep}    {HTML}{455A64}     
\definecolor{cRepBg}  {HTML}{ECEFF1}

\definecolor{panelbg}     {HTML}{F5F5F0}
\definecolor{panelborder} {HTML}{9A9A8C}
\definecolor{excerpthl}   {HTML}{FFE896}
\definecolor{querytitle}  {HTML}{2F5C99}
\definecolor{reporttitle} {HTML}{5C5C4A}
\definecolor{scorebg}     {HTML}{FBE3E3}
\definecolor{scoreborder} {HTML}{C86464}
\definecolor{scoretitle}  {HTML}{A83232}
\definecolor{addressed}   {HTML}{2E7D46}
\definecolor{supported}   {HTML}{2E7D46}

\definecolor{colBase}{HTML}{2A78D6}      
\definecolor{colE2E} {HTML}{EB6834}      
\definecolor{colTie} {HTML}{BFBFBF}      

\definecolor{winbg}     {RGB}{223,244,232}
\definecolor{wintext}   {RGB}{ 27,122, 68}
\definecolor{lossbg}    {RGB}{252,228,228}
\definecolor{losstext}  {RGB}{176, 42, 42}
\definecolor{flattext}  {RGB}{110,110,110}
\definecolor{mutedtext} {RGB}{130,130,130}
\definecolor{headergray}{RGB}{110,110,110}

\definecolor{jJudge}{HTML}{3E6FA8}
\definecolor{jVerif}{HTML}{2E8B6B}

\newcolumntype{C}{>{\centering\arraybackslash}m{2.3cm}}
\newcolumntype{H}{>{\columncolor{cToolBg}}c}   

\newcounter{pillar}
\renewcommand{\thepillar}{\Roman{pillar}}
\newcommand{\pillaritem}[1]{%
  \refstepcounter{pillar}%
  \textsc{\thepillar. #1}%
}

\newcommand{\note}[1]{\,\textsuperscript{\scriptsize\itshape(#1)}}

\newcommand{\metric}[3]{%
  \begingroup
  \ifdim#3pt>0pt
    \colorbox{winbg}{\textcolor{wintext}{\bfseries +#3~pts}}%
  \else
    \colorbox{lossbg}{\textcolor{losstext}{\bfseries #3~pts}}%
  \fi
  \par\vspace{0pt}
  {\tiny\textcolor{mutedtext}{T1 #1\ /\ Comp #2}}
  \endgroup
}

\newlength{\hbarw}
\newlength{\hbarh}
\newlength{\hsegA}
\newlength{\hsegT}
\newlength{\hsegB}

\newif\ifdraft
\draftfalse                             

\newif\ifagdraft
\agdraftfalse                           

\newtcolorbox{agclaim}{
  colback=blue!4, colframe=blue!45!black, boxrule=0.5pt,
  left=8pt, right=8pt, top=5pt, bottom=5pt, arc=2pt}

\newtcolorbox[auto counter, number within=section]{agexample}[2][]{
  colback=gray!3, colframe=gray!55!black, boxrule=0.5pt,
  left=6pt, right=6pt, top=5pt, bottom=5pt, arc=2pt,
  fonttitle=\bfseries\small, title={Example~\thetcbcounter: #2}, #1}

\newcommand{\hlurl}[1]{{\sethlcolor{orange!30}\hl{#1}}}

\newcommand{\hlfab}[1]{{\sethlcolor{orange!30}\hl{#1}}}

\title{Thomson: Continual Learning of Frontier Models for SovereignAI}

\newcommand{\authorgroupbreak}{\\[4pt]}
\author[*]{Shengzhuang~Chen}
\author[*]{Jerrod~Parker}
\author[*]{Yejin~Bang}
\author[*]{Andrew~M.~Bean}
\author[*]{Nabeel~Seedat}
\author[\dagger]{\authorgroupbreak Stefan~Winzeck}
\author[\dagger]{Daniil~Glazko}
\author[\dagger]{Jannik~Zgraggen}
\author[\dagger]{Fangyi~Yu}
\author[\dagger]{Scott~Arnott}
\author[\dagger]{Dietrich~Trautmann}
\author[\dagger]{Luca~Ciuffreda}
\author[\dagger]{Guglielmo~Bonifazi}
\author[\dagger]{Davide~Romano}
\author[\dagger]{Bradley~Bell}
\author[\dagger]{Kirsty~Fielding}
\author[\ddagger]{\authorgroupbreak Daniele~Giofr\`e}
\author[\ddagger]{Tom~Zielund}
\author[\ddagger]{Ipshita~Chatterjee}
\author[\ddagger]{Sneha~Murthy~Ghantasala}
\author[\ddagger]{Manpreet~Nanreh}
\author[\ddagger]{John~Scoville}
\author[\ddagger]{Maciej~Sakowicz}
\author[\ddagger]{Wassim~Seifeddine}
\author[\ddagger]{Lukas~Thede}
\author[\P]{\authorgroupbreak Jonathan~Richard~Schwarz}

\contribution[*]{Primary Authors}
\contribution[\dagger]{Core Contributors}
\contribution[\ddagger]{Contributors}
\contribution[\P]{Project Lead}

\correspondence{{First.Last}@thomsonreuters.com, jschwarz@ic.ac.uk}

\abstract{
  The development of frontier models is commonly perceived to be in the exclusive remit of a small number of heavily funded players, creating an information, economic and power asymmetry between developers and the diverse user base of modern AI. Recent public discourse acknowledges this concern, calling for SovereignAI (an organisation's capability to independently build, deploy and govern AI use), but often providing little concrete advice on how this can be achieved in the short term under a diversity of funding settings.

In this report, we argue that frontier performance can be achieved by a wide range of institutions through Continual Learning on readily available open-weight models. As opposed to existing limited approaches such as small-scale fine-tuning, prompt engineering, or tool-augmentation with a frozen model, our Continual Learning approach takes advantage of the effectiveness of a modern mid- \& post-training stack while introducing safeguards preserving both plasticity and stability at each training stage and seeking to make the minimal number of high-impact interventions on the parameters. This strategy results in model improvements comparable to the gains typically seen across multiple successive model generations. Crucially, such results are achievable with compute and personnel budgets substantially lower than commonly thought, making ownership of large parts of the SovereignAI stack (model, tool infrastructure, values \& data privacy) viable for a wider range of actors.

To demonstrate this, we introduce \texttt{Thomson}, a new general-purpose frontier model trained with an enhanced focus on high-stakes professional work: domains commonly predicted to undergo large productivity improvements through AI. Through a unique focus on Continual Learning, data-centricity, and efficiency, we demonstrate that \texttt{Thomson} performs competitively with recent frontier models on a wide range of domains and capabilities, ranging from agentic tasks to safety, legal, tax \& multilingualism, to comprehensive large-scale Deep Research. Thorough evaluations show
a distinctive $\pi$-shaped pattern: distinct improvements across a wide range of capabilities (including those not explicitly targeted), while almost completely eliminating the forgetting problem common to narrow domain adaptation.
\newline

\noindent
In partnership with:
\includegraphics[height=1em]{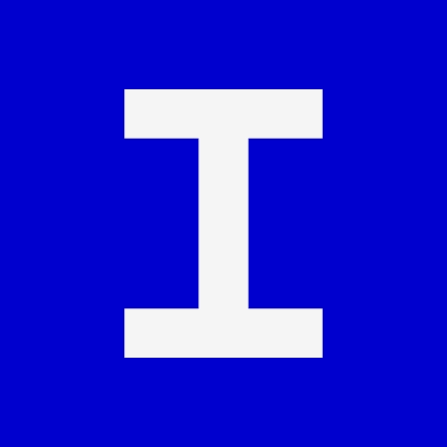}~Imperial College London\quad
\includegraphics[height=1em]{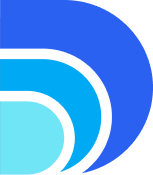}~DatologyAI\quad
\includegraphics[height=1em]{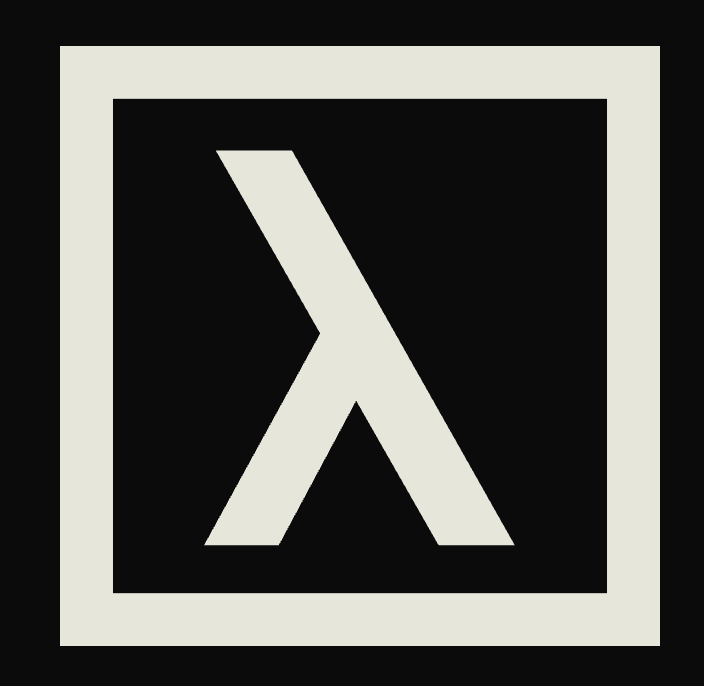}~Lambda\\[4pt]
\includegraphics[height=1em]{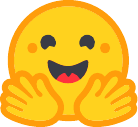}\ \texttt{Open-weight model:} \href{https://huggingface.co/thomsonreuters/Thomson-1.0-Small}{\texttt{Thomson-1.0-Small} (35b)}

}

\begin{document}

\maketitle

\section{Introduction}
\subsection{SovereignAI through Continual Learning}
\label{ssec:sovereign_ai_through_continual_learning}

\begin{figure}[t]
    \centering
    \begin{subfigure}[b]{0.55\textwidth}
        \centering
        \includegraphics[width=\linewidth]{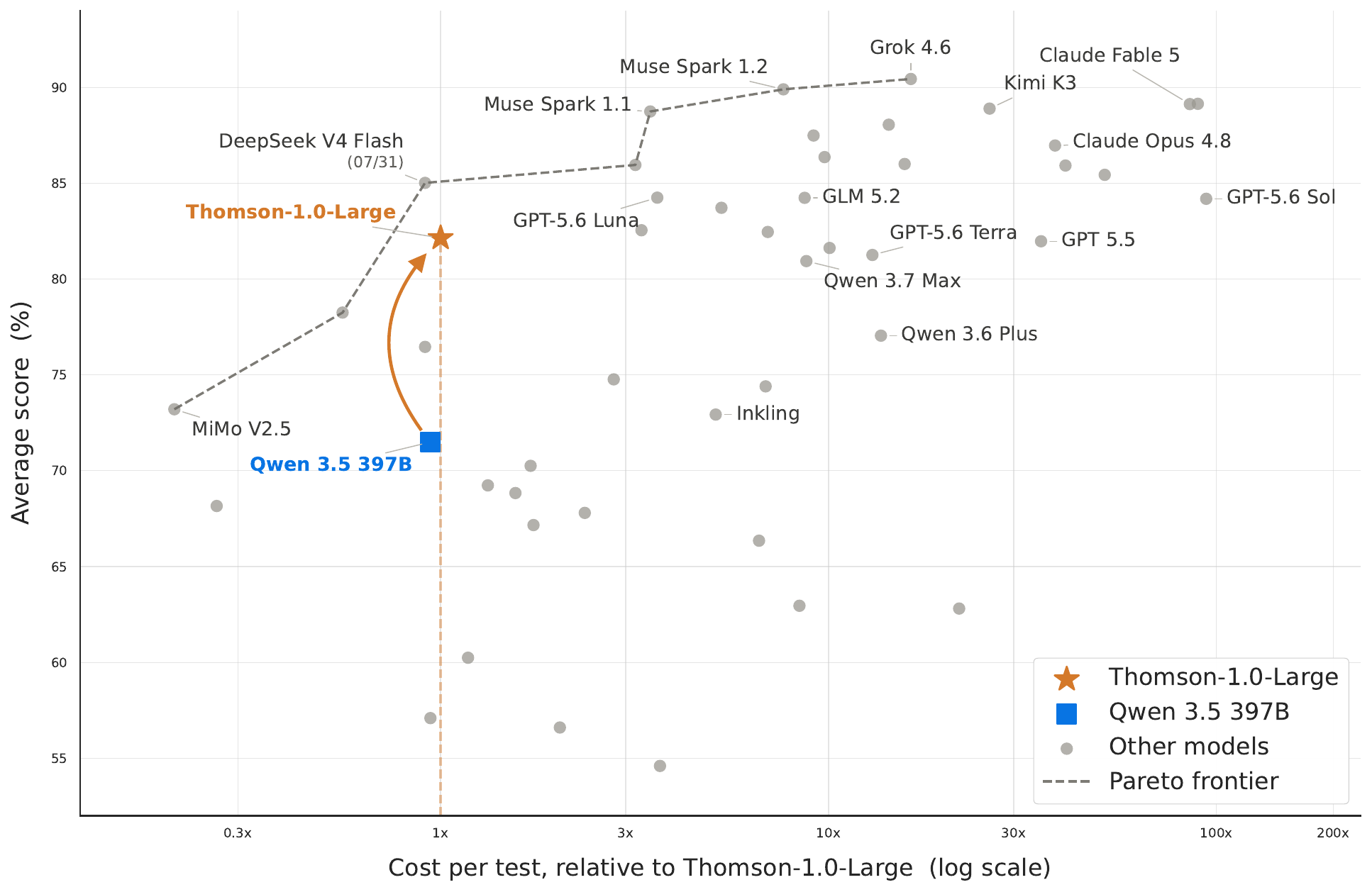}
        \caption{Cost/Performance trade-off on independent agentic suite.}
        \label{fig:headline_results_graph_a}
    \end{subfigure}
    \begin{subfigure}[b]{0.4\textwidth}
        \centering
        \includegraphics[width=0.95\linewidth]{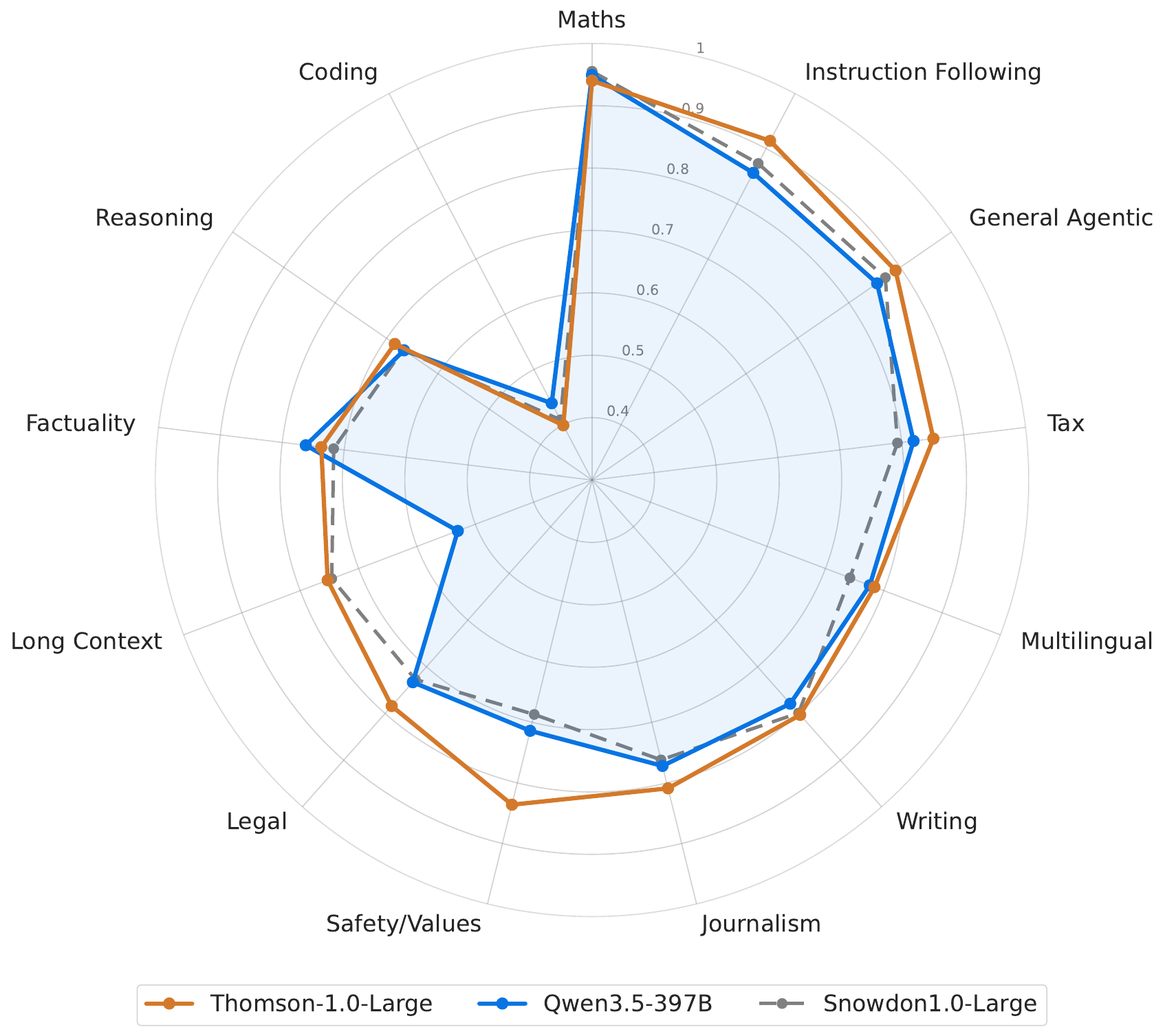}
        \caption{Per-category score improvements}
        \label{fig:headline_results_graph_b}
    \end{subfigure}

    \begin{subfigure}[b]{\textwidth}
        \centering
        \includegraphics[width=\linewidth]{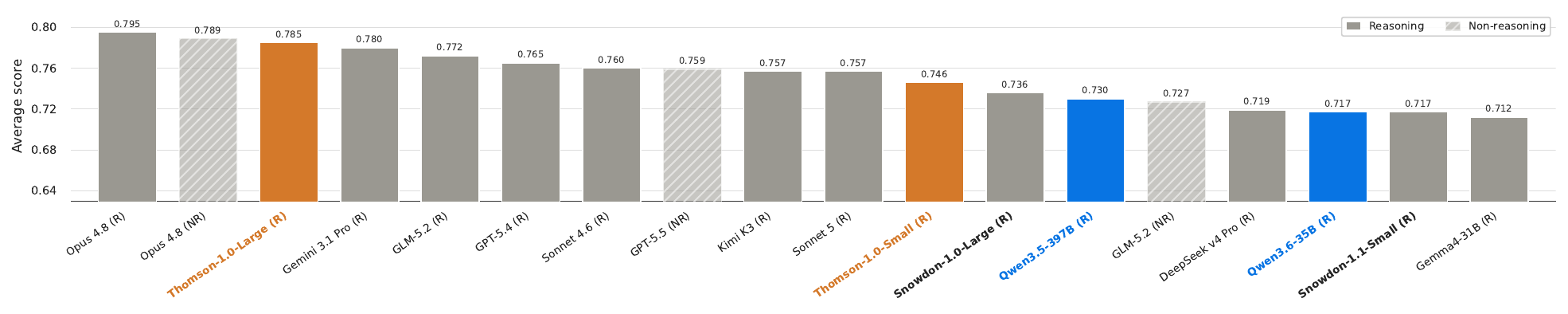}
        \caption{Aggregate score over our broader capability evaluation suite.}
        \label{fig:headline_results_graph_c}
    \end{subfigure}
    \caption{Combined \texttt{Thomson-1.0-Large} scores show frontier-level performance over a wide range of benchmarks. Improved scores are seen across most benchmark categories, with large gains in some areas. Please note that \ref{fig:headline_results_graph_a} shows the cost/performance trade-off on two target domain agentic tasks, while \ref{fig:headline_results_graph_c} is an aggregate score over a broader set of categories.}
    \label{fig:headline_results_graph}
\end{figure}

We introduce \texttt{Thomson}, a new family of frontier Foundation Models of high proficiency across a wide range of specialised and general-purpose domains as well as practical deployment settings. \texttt{Thomson} models are developed within a Continual Learning paradigm with the explicit goal of AI sovereignty, demonstrating that frontier performance is attainable by a much wider range of actors and institutions than commonly thought. Our model development moves beyond the superficiality of narrow fine-tuning or customisation exercises and avoids the significant economic inefficiency of training from scratch. The \texttt{Thomson} family was developed by a technical team not exceeding three dozen engineers and scientists on a modest compute cluster with no more than 368 B200 GPUs available at any stage of experimentation. Instead of working on either end of a spectrum ranging from pre-training from scratch to limited customisation exercises, we repurpose Open-weight models (specifically the \texttt{Qwen3.5-397B} and \texttt{Qwen3.6-35B} models) and substantially improve them on a wide range of performance domains, giving rise to a distinctive $\pi$-shaped performance improvement pattern. The total model development timeline for \texttt{Thomson-1.0-Small} and \texttt{Thomson-1.0-Large} (measured from the date of the first experiments with \texttt{Qwen3.5-397B}) was three months, covering large-scale data curation, Mid- \& Post-Training, several innovations on reward design (see Section~\ref{sec:training}), a mature evaluation protocol, human testing and adversarial safety studies. Once this pipeline is established, we believe that it can run in substantially shorter time frames. The cost of the final training run for \texttt{Thomson-1.0-Large} (measured in GPU costs over three weeks of training) is conservatively estimated to be under USD 450,000. The total cost of development (including staff, compute costs, domain expert compensation, and vendor partnerships) is estimated at approximately USD 40M, with most of it dedicated to reusable research, infrastructure engineering \& experimentation over a meaningfully longer time period than the aforementioned three months. We make full-weight updates (rather than relying on parameter-efficient methods) and do not engage in large-scale distillation, which both assumes the existence of a stronger teacher model and is typically unable to surpass it. In addition, we apply rigorous scrutiny to ensure that all of our training data and AI use are permissively licensed. Where we utilise open-weight models during development (e.g., as LLM judges), we show that they are eventually surpassed in performance and may be replaced by our own models in future iterations. Taken together, we believe that our findings generalise beyond the specific choices of open-weight checkpoint, model size, and target domain, giving rise to a new paradigm for frontier model development.

\texttt{Thomson} models are trained through three complementary modules, each yielding a fully developed Foundation Model. The implications of this finding are substantially more far-reaching than the models themselves: We argue that our development process can be seen as a blueprint for a wide range of institutions to develop private yet competitive Foundation Models at a fraction of the cost previously imagined. Specifically, we demonstrate that Continual Learning allows improvement of an open,  \textit{instruction-tuned} reasoning model broadly comparable to frontier performance in November 2025\footnote{Gemini 3 Pro: November 18, 2025; \texttt{GPT-5.1}: November 12, 2025; \texttt{Claude Opus 4.5}: November 24, 2025} to the point of surpassing recent flagship releases ranging from \texttt{Sonnet 5} \& \texttt{GLM-5.2} (June 2026), \texttt{GPT-5.5} \& \texttt{DeepSeek-V4 Pro} (April 2026) to \texttt{Gemini 3.1 Pro} (February, 2026) on a wide range of tasks (see Figure~\ref{fig:headline_results_graph_a}). The results \textit{bridge up to seven months of model improvements by the world's most generously funded AI companies} on a training compute budget typically reserved for smaller-scale ablation studies. Figure~\ref{fig:headline_results_graph_b} shows that, rather than making only shallow changes to a base model, our modifications improve on existing open-weight models in a consistent and meaningful way across a wide range of domains.

\definecolor{tromnge}{RGB}{250,100,0}      
\newcommand{\trtint}{tromnge!12}          
\newcolumntype{T}{>{\columncolor{\trtint}}c}
\begin{table}[t]
\centering
\footnotesize
\setlength{\tabcolsep}{3.5pt}
\renewcommand{\arraystretch}{1.05}
\resizebox{\textwidth}{!}{%
\begin{tabular}{l l c T c c c c c c c c }
\toprule
\textbf{Domain} & \textbf{Benchmark} &
\begin{tabular}[b]{@{}c@{}}\textbf{Opus}\\\textbf{4.8}\end{tabular} &
\begin{tabular}[b]{@{}c@{}}\textbf{Thomson}\\\textbf{1.0-Large}\end{tabular} &
\begin{tabular}[b]{@{}c@{}}\textbf{Gemini}\\\textbf{3.1 Pro}\end{tabular} &
\begin{tabular}[b]{@{}c@{}}\textbf{GLM}\\\textbf{5.2}\end{tabular} &
\begin{tabular}[b]{@{}c@{}}\textbf{GPT}\\\textbf{5.4}\end{tabular} &
\begin{tabular}[b]{@{}c@{}}\textbf{Kimi}\\\textbf{K3}\end{tabular} &
\begin{tabular}[b]{@{}c@{}}\textbf{Sonnet}\\\textbf{5}\end{tabular} &
\begin{tabular}[b]{@{}c@{}}\textbf{Snowdon}\\\textbf{1.0-Large}\end{tabular} &
\begin{tabular}[b]{@{}c@{}}\textbf{Qwen3.5}\\\textbf{397B}\end{tabular} &
\begin{tabular}[b]{@{}c@{}}\textbf{DeepSeek}\\\textbf{v4 Pro}\end{tabular} \\
\midrule
& \textbf{Overall Avg.}                & \textbf{79.5}  & 78.5           & 78.0           & 77.2           & 76.5           & 75.7           & 75.7           & 73.6           & 73.0           & 71.9 \\
\midrule
\multirow{11}{*}{\textbf{Legal}}
& Stanford LegalBench       & 81.8           & 82.3           & \textbf{84.3}  & 82.9           & 82.3           & 83.3           & 81.4           & 82.8           & 78.8           & 76.8 \\
& Info. Retrieval           & 51.2           & 53.6           & \textbf{53.8}  & 53.6           & \textbf{53.8}  & 53.0           & 47.9           & 51.2           & 49.0           & 51.5 \\
& Reasoning                 & 75.2           & 73.2           & \textbf{77.8}  & 73.1           & 68.4           & 74.8           & 73.1           & 70.8           & 66.5           & 70.1 \\
& Classification            & 70.5           & 70.9           & \textbf{74.2}  & 70.2           & 70.0           & 71.6           & 70.4           & 69.8           & 68.7           & 67.8 \\
& Doc. Processing \& RAG    & 78.9           & \textbf{79.7}  & 78.2           & 76.5           & 78.6           & 79.0           & 74.7           & 71.2           & 74.1           & 73.7 \\
& Summarisation             & 88.3           & \textbf{90.0}  & 89.4           & \textbf{90.0}  & 86.3           & 87.7           & 84.8           & 87.9           & 82.5           & 89.1 \\
& Contract Under.           & 74.4           & 71.0           & 75.9           & 73.6           & 74.9           & \textbf{77.2}  & 69.6           & 71.7           & 68.4           & 68.3 \\
& Human Queries             & 85.4           & \textbf{89.2}  & 86.4           & 87.3           & 88.6           & 61.4           & 84.3           & 87.1           & 86.7           & 86.4 \\
& Deep Research             & \textbf{90.8}  & 88.9           & 80.6           & 89.0           & 85.9           & 89.8           & 86.0           & 78.4           & 87.3           & 84.0 \\
& Harvey LAB                & \textbf{86.9}  & 85.7           & 55.5           & 84.6           & 76.1           & 83.7           & 80.9           & 56.3           & 70.6           & 83.1 \\
\cmidrule(l){2-12}
& \textit{Domain Avg.}      & 78.3           & \textbf{78.4}  & 75.6           & 78.1           & 76.5           & 76.1           & 75.3           & 72.7           & 73.3           & 75.1 \\
\midrule
\multirow{3}{*}{\textbf{Tax}}
& Deep Research             & \textbf{85.8}  & 82.4           & 76.0           & 83.5           & 80.9           & 85.1           & 82.0           & 70.4           & 78.7           & 80.7 \\
& Tax Q\&A                  & 86.9           & 87.9           & 84.5           & 85.7           & 86.6           & 88.3           & \textbf{88.7}  & 88.2           & 85.1           & 82.8 \\
\cmidrule(l){2-12}
& \textit{Domain Avg.}      & 86.3           & 85.1           & 80.2           & 84.6           & 83.8           & \textbf{86.7}  & 85.4           & 79.3           & 81.9           & 81.8 \\
\midrule
\textbf{Journalism}
& Deep Research             & 82.3           & 80.9           & 83.0           & 79.0           & 66.3           & \textbf{84.5}  & 74.1           & 76.2           & 77.2           & 78.6 \\
\midrule
\multirow{10}{*}{\textbf{General}}
& Factuality                & 71.4           & 73.7           & \textbf{82.6}  & 66.2           & 68.3           & 69.5           & 59.8           & 71.7           & 76.2           & 69.1 \\
& Long Context              & 75.2           & 75.3           & 75.0           & \textbf{75.9}  & 70.0           & 73.5           & 70.7           & 74.6           & 53.0           & 69.5 \\
& Multilingualism           & 83.2           & 78.4           & \textbf{85.7}  & 81.7           & 82.7           & 84.9           & 79.5           & 74.2           & 77.6           & 78.4 \\
& Instr. Following          & 86.1           & \textbf{91.4}  & 84.8           & 89.4           & 89.0           & 85.8           & 86.8           & 87.3           & 85.6           & 89.2 \\
& Writing                   & 79.3           & 80.3           & 78.5           & 78.1           & 79.1           & 78.2           & 78.7           & 79.9           & 77.9           & \textbf{80.7} \\
& Reasoning                 & 73.7           & 68.4           & 74.8           & 67.3           & 71.6           & \textbf{76.2}  & 66.8           & 66.8           & 66.6           & 65.0 \\
& General Agent             & 83.4           & \textbf{89.1}  & 84.4           & 77.5           & 75.6           & 61.0           & 74.1           & 87.1           & 85.5           & 72.0 \\
& Coding                    & 57.4           & 39.9           & 50.0           & 56.0           & 50.8           & \textbf{66.8}  & 57.4           & 40.9           & 43.9           & 45.6 \\
& Maths                     & \textbf{98.7}  & 94.0           & 97.4           & 91.2           & 97.7           & 96.2           & 85.9           & 95.5           & 94.9           & 94.9 \\
\cmidrule(l){2-12}
& \textit{Domain Avg.}      & 78.7           & 76.7           & \textbf{79.2}  & 75.9           & 76.1           & 76.9           & 73.3           & 75.3           & 73.5           & 73.8 \\
\midrule
\multirow{4}{*}{\begin{tabular}[c]{@{}l@{}}\textbf{Safety /}\\\textbf{Values}\end{tabular}}
& Political Neutrality      & 82.8           & \textbf{97.3}  & 93.3           & 91.0           & 83.8           & 33.5           & 82.3           & 85.3           & 51.5           & 31.3 \\
& Robustness                & \textbf{78.2}  & 60.3           & 65.3           & 50.4           & 68.3           & 72.7           & 77.1           & 42.2           & 66.7           & 38.1 \\
& Adversarial Testing       & --             & 93.4           & --             & 93.6           & --             & 87.3           & --             & 78.8           & \textbf{95.9}  & 81.1 \\
\cmidrule(l){2-12}
& \textit{Domain Avg.}      & --             & \textbf{83.6}  & --             & 78.3           & --             & 64.5             & --             & 68.7           & 71.4           & 50.2 \\
\bottomrule
\end{tabular}}
\caption{Cross-domain benchmark results for reasoning-mode models. Scores are percentages; the best score in each row is shown in \textbf{bold}. Dashes (--) mark benchmarks that were not run for a given model. Note: Fair Adversarial testing against proprietary models was not possible due to the likely presence of unknown guardrail mechanisms.  Overall Avg. is the unweighted mean over all individual benchmarks, excluding Adversarial Testing. Harvey LAB: Harvey Legal Agent Benchmark.}
\label{tab:benchmarks-large}
\end{table}

\definecolor{tromnge}{RGB}{250,100,0}      
\newcolumntype{T}{>{\columncolor{\trtint}}c}
\begin{table}[t]
\centering
\footnotesize
\setlength{\tabcolsep}{3.5pt}
\renewcommand{\arraystretch}{1.05}
\resizebox{\textwidth}{!}{%
\begin{tabular}{l l T c c c c}
\toprule
\textbf{Domain} & \textbf{Benchmark} &
\begin{tabular}[b]{@{}c@{}}\textbf{Thomson}\\\textbf{1.0-Small}\end{tabular} &
\begin{tabular}[b]{@{}c@{}}\textbf{Snowdon}\\\textbf{1.1-Small}\end{tabular} &
\begin{tabular}[b]{@{}c@{}}\textbf{Gemma4}\\\textbf{31B}\end{tabular} &
\begin{tabular}[b]{@{}c@{}}\textbf{Qwen3.6}\\\textbf{35B}\end{tabular} &
\begin{tabular}[b]{@{}c@{}}\textbf{Haiku}\\\textbf{4.5}\end{tabular} \\
\midrule
& \textbf{Overall Avg.}                & \textbf{74.6} & 71.7          & 71.2          & 71.7          & 68.2 \\
\midrule
\multirow{11}{*}{\textbf{Legal}}
& Stanford LegalBench                  & 79.9          & 80.9          & \textbf{83.1} & 80.3          & 80.7 \\
& Info. Retrieval                      & 49.6          & 48.6          & \textbf{51.9} & 49.4          & 49.2 \\
& Reasoning                            & 68.2          & 67.3          & \textbf{71.9} & 64.7          & 65.5 \\
& Classification                       & 70.0          & 70.1          & 69.4          & \textbf{70.4} & 68.1 \\
& Doc. Processing \& RAG            & \textbf{78.8} & 71.2          & 76.6          & 74.7          & 43.8 \\
& Summarisation                        & \textbf{89.4} & 88.1          & \textbf{89.4} & 89.0          & 84.3 \\
& Contract Under.                      & 67.3          & 64.8          & \textbf{73.2} & 63.7          & 70.1 \\
& Human Queries                        & \textbf{90.2} & 82.2          & 81.2          & 82.6          & 74.9 \\
& Deep Research                        & \textbf{85.0} & 80.0          & 74.0          & 82.0          & 80.0 \\
& Harvey Legal Agent Bench.            & \textbf{73.4} & 71.5          & 34.2          & 69.5          & 60.5 \\
\cmidrule(l){2-7}
& \textit{Domain Avg.}                 & \textbf{75.2} & 72.4          & 70.5          & 72.7          & 67.7 \\
\midrule
\multirow{3}{*}{\textbf{Tax}}
& Deep Research                        & \textbf{78.6} & 68.0          & 75.0          & 68.0          & 62.0 \\
& Tax Q\&A                          & \textbf{86.6} & 85.2          & 84.5          & 86.2          & 79.4 \\
\cmidrule(l){2-7}
& \textit{Domain Avg.}                 & \textbf{82.6} & 76.5          & 79.6          & 77.3          & 70.7 \\
\midrule
\textbf{Journalism}
& Deep Research                        & 74.2          & 67.5          & 74.7          & 73.0          & \textbf{81.0} \\
\midrule
\multirow{10}{*}{\textbf{General}}
& Factuality                           & \textbf{61.1} & 59.3          & 57.6          & 58.8          & 56.8 \\
& Long Context                         & \textbf{74.1} & 73.8          & 69.4          & 73.8          & 67.4 \\
& Multilingualism                      & 71.9          & 72.8          & 79.4          & 73.1          & \textbf{85.8} \\
& Instruction Following                & 86.1          & 85.5          & \textbf{89.3} & 85.6          & 78.2 \\
& Writing                              & \textbf{81.0} & 79.3          & 75.5          & 79.5          & 77.9 \\
& Reasoning                            & 61.5          & 61.3          & \textbf{66.1} & 61.5          & 49.3 \\
& General Agent                        & \textbf{85.8} & 81.4          & 72.9          & 80.3          & 59.7 \\
& Coding                               & 37.4          & 35.6          & 34.6          & \textbf{39.8} & 32.9 \\
& Maths                                & 86.7          & 88.0          & \textbf{91.1} & 87.5          & 66.5 \\
\cmidrule(l){2-7}
& \textit{Domain Avg.}                 & \textbf{71.7} & 70.8          & 70.7          & 71.1          & 63.8 \\
\midrule
\multirow{4}{*}{\begin{tabular}[c]{@{}l@{}}\textbf{Safety /}\\\textbf{Values}\end{tabular}}
& Political Neutrality                 & \textbf{98.5} & 91.5          & 91.5          & 78.5          & 92.0 \\
& Robustness                           & 56.3          & 47.3          & 41.7          & 48.7          & \textbf{70.2} \\
& Adversarial Testing                  & 89.1          & 87.7          & 88.5          & \textbf{89.2} & -- \\
\cmidrule(l){2-7}
& \textit{Domain Avg.}                 & \textbf{81.3} & 75.5          & 73.9          & 72.1          & 81.1 \\
\bottomrule
\end{tabular}}
\caption{Cross-domain benchmark results for small models. The best score in each row is shown in \textbf{bold}. Dashes (--) mark benchmarks that were not run for a given model. All models were run with medium reasoning effort. Overall Avg. is the unweighted mean over all individual benchmarks, excluding Adversarial Testing.}
\label{tab:benchmarks-small}
\end{table}

More important than the concrete results shown in this report are the principles underlying our demonstrated improvements, which we believe are more general and can be readily applied to other open-weight models, allowing a capable team to further reduce iteration time. Thus, we believe the reader is best advised to think of this report as a set of building blocks for a model factory applicable more broadly to a wide range of open-weight models, with the \texttt{Thomson} models being merely an initial proving ground.

For the purpose of this report, we consider the following essential principles of SovereignAI:
\begin{enumerate}[label=\S\arabic*., ref=\S\arabic*]
    \item\label{item:sovereignty-model} Training \& Model Sovereignty: See Section~\ref{sec:training} \& broader report. Largely achievable through work with open-weight models (for discussion, see Section~\ref{ssec:discussion_limitations}).
    \item\label{item:sovereignty-data} Data \& Tool Sovereignty: See Sections~\ref{sec:preliminaries},  \ref{sec:training}. Fully achievable for private data. No control over data used to train open-weight model.
    \item\label{item:sovereignty-values} Governance \& Ethical Alignment: See Section~\ref{ssec:values}. Substantial progress through alignment with custom constitutions at various training stages. Subject to current limitations of AI alignment more broadly.
    \item\label{item:sovereignty-infra} Infrastructure Sovereignty: See Section~\ref{sec:infra}. Largely achievable by building on increasingly mature open-source stacks for training, model serving \& tool infrastructure. Remaining dependence on critical hardware infrastructure.
    \item\label{item:sovereignty-econ} Control Over Economic Factors: Substantially increased by operating models at cost while maintaining control over release/update cycles. Remaining uncertainty based on cost/access to serving hardware.
\end{enumerate}

The pillars of our approach to model development are:

    \pillaritem{Continual Learning}\label{item:develop-pillar-cl}: Taking care to preserve capabilities not directly affected in each phase of model improvement (stability) while maintaining the ability to master new skills and absorb new knowledge (plasticity). This is achieved through the careful design of our model improvement pipeline, including critical choices and modifications of learning algorithms both effective and compute-efficient within an accessible budget. In addition, we curate a representative and low-cost evaluation suite for knowledge retention, providing a reliable measure of skill retention without overtly optimising on benchmarks.

\pillaritem{Data-centric Machine Learning}\label{item:develop-pillar-dcml}: The highest standards of data quality control and design targeted at learning efficiency and skill improvement in vital domains. To name a few examples, we construct contrasting data sources to re-align a given model with an institution's values; re-phrase, filter, de-duplicate, and enhance mid-training data to enable greater post-training efficiency; calibrate data mixtures for post-training through Bayesian Optimisation; guide invaluable human subject-matter data collection towards a semantic space with low density in the existing training data distribution; and carefully configure additional automated evaluation benchmarks to be as closely aligned with human judgements as possible.

\pillaritem{Agentic training \& tool use}\label{item:agents}: Recognising the near-universal use of tool augmentation as a common easy-access strategy to introduce private data at inference, we develop data generation, training,  inference and evaluation strategies to enable the tailoring of models to both general-purpose and privately implemented tool systems. This includes the design of a broad set of tools appropriate for the serving context, a full Deep Research harness (Section~\ref{sec:preliminaries}), and training stages ranging from localised error correction to end-to-end reinforcement learning (RL) for Deep Research. In addition, we provide a detailed description of thoroughly designed reward structures to incentivise faithful use and accurate citation patterns, vital to reducing hallucinations in high-stakes settings.

After discussing the specific modelling goals and broader impact of this work, the remainder of the report carefully lays out the different phases of model development, data synthesis, and evaluation, while also addressing broader practical infrastructure and serving requirements.

\subsection{Modelling Goals}
\label{ssec:prof_work_context}

\texttt{Thomson} was developed with a deliberate focus on economically impactful, high-stakes professional work across legal, tax, and journalism domains. This scope reflects a realistic and compelling use case for an institution with strong incentives to pursue SovereignAI. Accordingly, we address modelling decisions suited to domains that combine the formality, logic, and rigorous reasoning characteristic of technical fields with the nuance, interpretation, and tolerance for uncertainty of the humanities. Several of these principles -- particularly the emphasis on citation quality and conditional relevance (whereby precedents of considerable age may remain binding, while others are superseded by subsequent rulings or evidence) -- bear a natural affinity to scientific practice. We thus believe they are of broader methodological interest. Where relevant to readers with a specific interest in the legal domain, results of particular significance to that area are highlighted throughout.

A notable insight that emerged during development was the behaviour of our Continual Learning pipeline: rather than merely preserving performance on tasks outside the primary focus, the pipeline demonstrably improved performance on them. This result was initially unexpected and represents a highly positive finding for the claims made in this report. It provides strong grounds for believing that the modelling principles described here are domain-agnostic and transferable across a wide range of settings and institutional contexts.

\subsection{Headline Results}
\label{ssec:headline_results}

The following sections show some of the headline results for \texttt{Thomson}, both when run under identical conditions against frontier models as well as in a broader system comparison. Extended results, including detailed descriptions, ablation studies and analysis, can be found in Section~\ref{sec:evaluation}.

\subsubsection{Model Results}
\label{ssec:headline_results_model}

Figure~\ref{fig:headline_results_graph} shows \texttt{Thomson-1.0-Large} in comparison to a wide range of competitors as well as its base model (\texttt{Qwen3.5-397B-A17B}) in three settings. Figure~\ref{fig:headline_results_graph_a} shows an average cost vs performance trade-off on two agentic tasks in the legal target domain, normalised to the cost of running \texttt{Qwen3.5-397B} architectures.\footnote{Averaged over (1) \href{https://www.vals.ai/benchmarks/hlab}{Harvey Legal Agent Benchmark} and (2) \href{https://www.vals.ai/benchmarks/legal\_research}{Legal Research Bench}. Third-party model costs and results are taken from publicly available \href{https://www.vals.ai/}{vals.ai} leaderboards. \texttt{Thomson-1.0-Large} and \texttt{Qwen3.5-397B} costs are estimated using publicly available pricing data from Alibaba (both accessed in August 2026). All costs take token usage into account.} We show near-Pareto-optimal performance, with substantial improvements over \texttt{Qwen3.5-397B}, rivalling a wide range of competitive frontier models. The cost difference between subsequent versions of models from the same provider also strengthens our argument for economic sovereignty (\ref{item:sovereignty-econ}). Figure~\ref{fig:headline_results_graph_b} is a clear demonstration of $\pi$-shaped Continual Learning, showing material performance gains on several dimensions, while holding the baseline performance on most other domains. Figure~\ref{fig:headline_results_graph_c} is an aggregate overall performance score computed over target, general domains and safety scores (see Table~\ref{tab:benchmarks-large}). All results paint a clear picture of sustained frontier performance and substantial improvements through Continual Learning.

\begin{figure}
    \centering
    \includegraphics[width=0.45\linewidth]{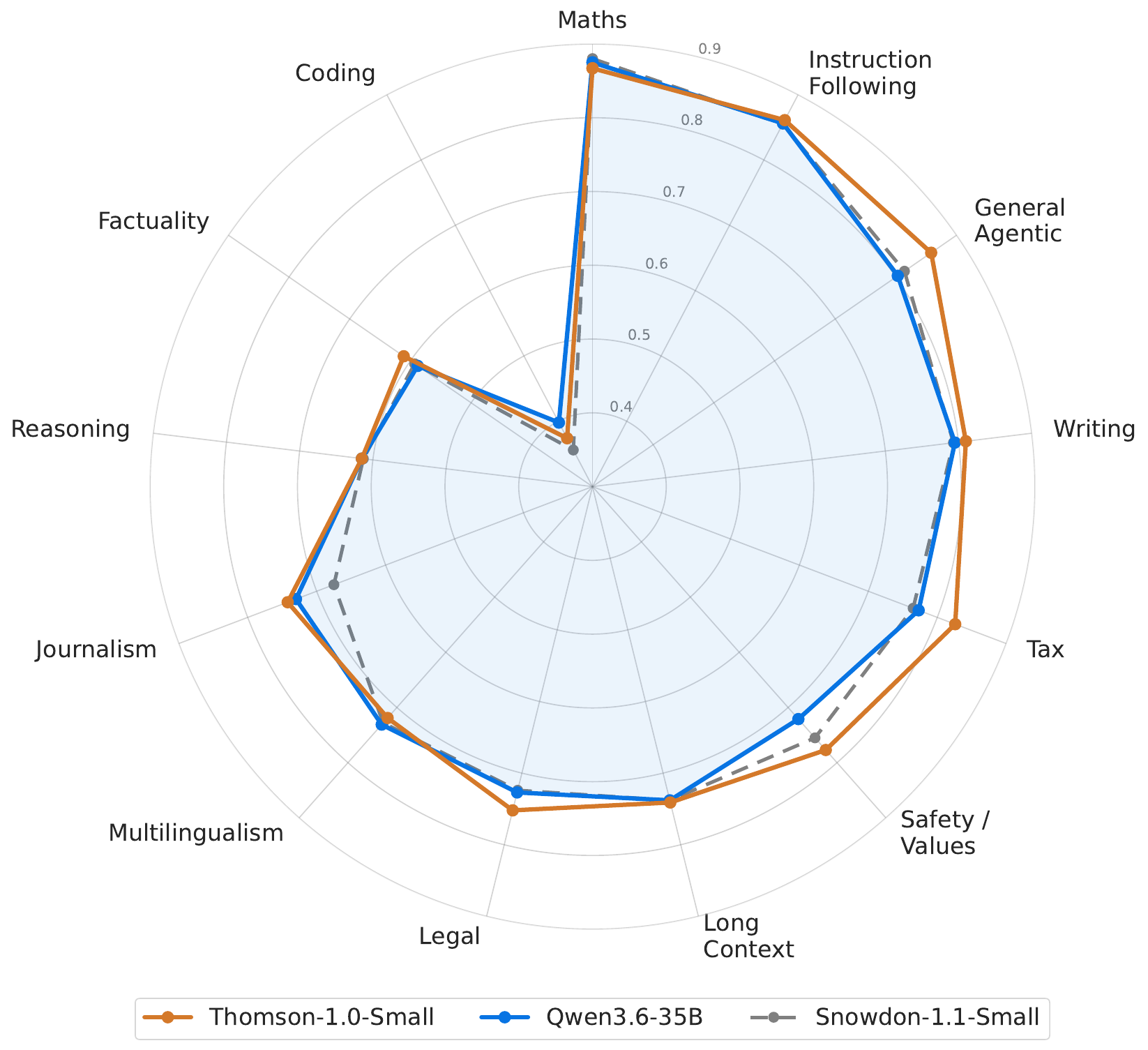}
    \caption{Per-Category score improvements of \texttt{Thomson-1.0-Small}.}
    \label{fig:headline_results_small_radar}
\end{figure}

For completeness, we also share results for \texttt{Snowdon-1.0-Large} \& \texttt{Snowdon-1.1-Small}, a value-realigned version of \texttt{Qwen3.5-397B/Qwen3.6-35B} (see Figures~\ref{fig:headline_results_graph_b}, \ref{fig:headline_results_small_radar}), developed through the \href{https://www.imperial.ac.uk/frontier-ai/}{Frontier AI Research Lab}, a joint research lab established by Thomson Reuters and Imperial College London. The \texttt{Snowdon-1.0-Large} development process is described in Section~\ref{ssec:values} and in full detail in \citep{imperial2026realignment}.

Table~\ref{tab:benchmarks-large} shows detailed results in a wide range of settings, from high-volume automated tasks (such as document processing) to answer quality on real-world human interactions, high-budget Deep Research tasks, a wide range of general domain evaluations, as well as safety and value evaluations judged through adversarial testing. Table~\ref{tab:benchmarks-small} reports the same breakdown for the small models, alongside Figure~\ref{fig:headline_results_small_radar}, which shows their per-category gains over \texttt{Qwen3.6-35B}. Every model is evaluated through the same harness, on identical prompts, comparable inference parameters (such as reasoning effort) and grading pipelines designed to faithfully measure performance rather than insignificant artefacts. No model is granted retrieval or tool access unless the benchmark itself defines it, in which case all models receive the same tools. However, closed models are accessible only as served endpoints, and we cannot rule out provider-side routing, system prompt augmentation or other scaffolding. We consider these unavoidable rather than disqualifying. In short, we take care to ensure the protocol is as fair and unbiased as possible.

\texttt{Thomson-1.0-Large} places close to the strongest proprietary model tested (which is likely of significantly larger size) in the comparison on the overall average, ahead of heavyweights such as \texttt{Gemini 3.1 Pro}, \texttt{Sonnet 5}, \texttt{GLM-5.2}, and importantly, dominating performance against its base model by a significant margin. \texttt{Thomson}'s strength is particularly pronounced across several of the domains we explicitly target, such as natural human interactions, document-heavy workflows or Deep Research. On safety and values, the effect of implementing value sovereignty (\ref{item:sovereignty-values}) is clearly visible, alleviating a regular concern of open-weight model use; we note that fair adversarial comparison against proprietary systems is not possible, since undisclosed guardrail layers sit between the served endpoint and the model. We thus omit such entries rather than reporting misleading numbers.

In terms of limitations, we find coding -- the only domain showing mild forgetting relative to the base model -- to fall clearly below frontier performance. While this is deliberately not a target domain (and arguably already well-targeted by the vast majority of AI developers), it is plausible that coding may affect the downstream performance of other domains, especially given the increased prevalence of agentic tasks requiring general computer-use skills. General mathematical and abstract reasoning trail the strongest proprietary models but are within the performance level of \texttt{Qwen}, showing successful protection against forgetting.

\subsubsection{System Results}
\label{ssec:headline_results_system}

\begin{figure}
    \centering
    \includegraphics[width=\linewidth]{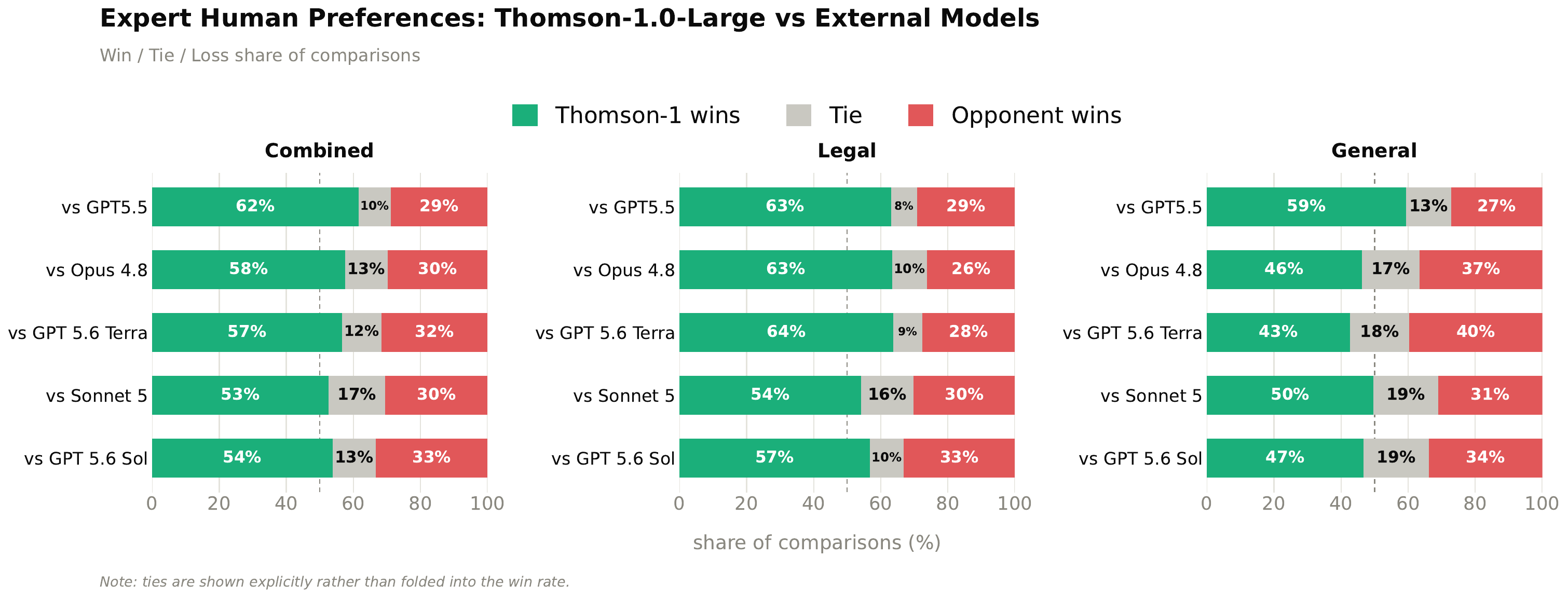}
    \caption{Blind human evaluation study of the \texttt{Thomson} system against systems provided by major frontier labs. \texttt{Thomson-1.0-Large} is given access to recent news (Reuters feed) as well as access to legal databases. External OpenAI \& Anthropic models are given broad web access. Scores computed on an aggregate of over 3,000 preference-rated conversations.}
    \label{fig:human_comparison}
\end{figure}

The preceding section deliberately omits various common comparative advantages institutions would deploy in practice. In keeping with our arguments for sovereignty (especially \ref{item:sovereignty-data}), this section answers a complementary question of high practical relevance: \emph{what are the best results competing institutions can currently produce?} To approximate the question, we provide a downstream system demonstration rather than aiming to isolate the model's contribution alone. To do so, we test models from leading AI providers OpenAI and Anthropic, given web access, against \texttt{Thomson-1.0-Large} when given access to some of the world's most authoritative licensed databases on legal and news information, testing both systems on general and domain-specific queries in a blind comparison. A priori, it may be expected that broad web access should provide a major advantage in general tasks while still providing substantial benefits in the target domain. In addition, both providers are likely to have specialised their models for web use, given the ubiquity of this use case.

Figure~\ref{fig:human_comparison} reports a blind preference study over more than 3,000 expert-rated conversations, split into legal and general sets, with subject-matter experts unaware of which system produced which response. Instructions were deliberately limited (with the exception of providing realistic limits to what the systems could produce), testing systems under real-world conversational use. Aggregated across both sets, the \texttt{Thomson} system is preferred over each of the five external systems. The structure of the result is more informative than the headline: on legal conversations, where sovereign data access is most consequential, preferences are consistently and clearly favourable, in expectation of the privileged position. However, on open-domain general conversations, the picture is mixed and considerably tighter, a promising result given that (a) both OpenAI and Anthropic will likely possess historical user interaction data on general queries at unparalleled scale, and (b) web access is arguably broader than only news access. Further details (including ablation study results removing system components) are provided in Section~\ref{sec:evaluation}.

Taken together with Section~\ref{ssec:headline_results_model}, the two sets of results describe a model factory: a reproducible pipeline for turning an open-weight checkpoint and an institution's proprietary data into a deployable, competitive, and governable system.

\subsection{Discussion \& Limitations}
\label{ssec:discussion_limitations}

A critical reader may pause and consider the reliance on an initial open-weight starting point as incompatible with the principles of SovereignAI. While ownership of the full training stack arguably goes further, the implied economic and computational demands preclude most interested actors from making meaningful progress towards independence. In this report, we are interested in moving beyond a bipolar view of sovereignty (i.e., conditioning any notion of sovereignty on full ownership) by instead advancing along what we consider a sovereignty spectrum. To that end, we present substantial progress on the model -- the core of any such strategy -- as well as beyond it, through tool design, evaluation guidelines, serving infrastructure, and harness development.

In choosing this approach, we explicitly aim to capitalise on the outstanding opportunity of a competitive open-weight landscape that has flourished despite the substantial early performance margin enjoyed by closed-model developers. Indeed, it is an increasingly common view that the gap between open-weight models and the best closed models has collapsed from multiple months \citep{aisi2026openweight} to a much shorter time frame, suggesting that the additional training recommended in this report can not only close any remaining gap, but potentially surpass some of the world's most competitive models. Further arguments in favour of this approach are the general consensus that additional large-scale pre-training is no longer giving rise to large model improvements (drastically decreasing dependence on the most costly stage of training), as well as a noticeably larger collective body of knowledge on how to develop frontier models \& systems (with LLM reasoning being the latest such breakthrough quickly made accessible by \texttt{DeepSeek-R1} \citep{guo2025deepseek}).

It is also important to note that once a first generation of models has been built using the principles laid out in this report, we observe that any subsequent model generation can be increasingly based on the resulting checkpoints from previous training rounds (similarly to how some model builders aim to rely only on internally developed models). It is our view that, after a few such model iterations, it becomes increasingly tenuous and of little relevance that a model builder relies on an open-weight model at the start of the iteration, no matter how distant the original dependence might be.

While valid arguments persist regarding architectural control (which is more difficult to alter in later stages of model training), a rich set of models spanning a wide range of sizes and architectures is readily available. Finally, recent work \citep[e.g.,][]{o2026deep} has begun to show that some pre-training properties may be resilient to alteration in later stages. So far, this has been exploratory research motivated by highly desirable properties (tamper resistance of safety features), and there is little incentive for an open-weight developer to remove benign properties in a similar fashion. In addition, by demonstrating that Mid-Training can still be performed on \emph{instruction-tuned} models (a highly desirable yet unconventional approach; Section~\ref{ssec:cpt}), we present a practical solution for injecting additional knowledge present in private data. Thus, we believe that for the majority of potential beneficiaries of AI, our arguments in favour of practical sovereignty through Continual Learning carry more weight.

\tableofcontents

\newpage
\newcommand{\hlt}[1]{\hl{#1}}

\section{Preliminaries}
\label{sec:preliminaries}

\subsection{Constitutional AI}
\label{ssec:constituional_ai}

Recognising the Sovereignty goal \ref{item:sovereignty-values} (Governance and Ethical Alignment), we draw inspiration from initial attempts to align AI with model constitutions \citep{bai2022constitutional} and show throughout the report how this can be achieved through complementary training processes designed for actors with both modest and more generous computational budgets. Constitutional AI sits at the core of how we approach Sovereignty in model development. A sovereign model must be governed by principles that are transparent, publicly accountable, and open to scrutiny, rather than by proprietary value systems that are subject to change. For this reason, we deliberately depart from existing commercial constitutions and instead ground our alignment process in the Public AI constitutional project \citep{patriniche2026publicai}, an open project that emphasises public contribution and debate. As this constitution allows free use and modification, we believe \ref{item:sovereignty-values} can be readily implemented by making appropriate modifications and following our development methodology. For the development of \texttt{Thomson}, this choice reflects our conviction that the normative foundations of a model should themselves be a shared public resource, developed in the open and subject to community input. Throughout development, this constitution serves as the reference framework against which model behaviour is shaped.

We stress that alignment to this constitution remains an aspiration rather than a solved problem. While we make every effort to faithfully reflect its principles in the resulting model, we regard this as ongoing work, and we make no claim of complete or guaranteed adherence. Our hope is that the approach described here, along with the choices and trade-offs it entails, will prove useful to the wider community and help inspire future work on grounding sovereign models in openly developed constitutional principles.

\subsection{Mitigating \& Measuring Forgetting}
\label{ssec:monitoring_forgetting}

Continual Learning has its roots in the study of lifelong learning in neural networks, with the phenomenon of \emph{catastrophic forgetting} first characterised in the late 1980s \citep{mccloskey1989catastrophic, ratcliff1990connectionist} and subsequently formalised, along with a number of proposals for algorithmic remedies developed over the following three decades \citep{parisi2019continual, wang2024comprehensive}. In the era of large Foundation Models, however, forgetting rarely manifests as the abrupt, catastrophic collapse observed in earlier, smaller-scale settings; instead, it tends to be subtle and gradual, surfacing as a slow erosion of specific capabilities that is considerably harder to detect and to measure reliably. Throughout the development of this approach, we found that the careful, disciplined iteration and execution of well-established ideas -- Data Rehearsal, Architectural Regularisation, and Functional Regularisation -- applied selectively but consistently throughout every stage of the pipeline constitutes a substantially more impactful and practical strategy than isolated modifications to any single training algorithm.

A further distinguishing feature of our approach is that we operate throughout on \textit{instruction-tuned}, rather than simply pre-trained, open-weight models: this is vital in order to retain and build upon the substantial capabilities already instilled during prior post-training, but requires careful experimentation and validation at critical stages of the pipeline (most notably Mid-Training; see Section~\ref{ssec:cpt}) where the assumptions underlying standard training recipes for base model adaptation no longer straightforwardly hold.

\begin{table}
\footnotesize
\begin{tabular}{@{} >{\bfseries}l p{6.4cm} p{6.4cm} @{}}
\toprule
\normalfont\bfseries & \normalfont\bfseries Development Evals (CapTrack) & \normalfont\bfseries Test Evals \\
\normalfont\bfseries & \scriptsize{(only $\approx10-15\%$ of examples used)} & \normalfont \scriptsize{(All examples used)} \\
\midrule

\rowcolor{cToolBg}Factuality& RAGTruth, TruthfulQA, PopQA       & SimpleQA, Faitheval \\
\midrule

Multilingualism & XTREME, MGSM & XTREME$^*$, MGSM, MMMLU \\
\midrule

\rowcolor{cToolBg}Instruction Follow. & IFEval, FollowBench     & IFEval, FollowBench$^*$ \\
\midrule

Writing & MT-Bench\note{1st turn}, OASST1, ELI5 & WritingBench \\
\midrule

\rowcolor{cToolBg}Coding & HumanEval, MBPP, BFCL, MNMS & SWE-Bench Pro, TerminalBench 2.1 \\
\midrule

Maths & GSM-8K, MATH, LiveMathBench & AIME 2025-2026, GSM-8K, MATH \\
\midrule

\rowcolor{cToolBg}Reasoning & MMLU-Pro, SuperGPQA   & GPQA-Diamond, Humanity's Last Exam, MMLU-Pro \\
\midrule

General Agent & MT-Bench\note{2nd turn}, StructFlowBench & GDPVal, Tau2 \\
\midrule

\rowcolor{cToolBg}Long Context & HotpotQA, QASPER, RULER-32K, LongBench-V2     & HotpotQA, MuSiQue, NovelQA, NQ, QAMPARI, Quest, $\infty$Bench \\

\midrule

Safety \& Values & HarmBench (unsafe), GSM-8k (benign), RULER-4k\note{incomplete}, WinoGrande, HellaSwag                    & Adversarial Testing, Political Neutrality, Robustness \\
\midrule

\rowcolor{cToolBg}Misc & BoolQ, Schema-wrapped MMLU-Pro, Rephrased MMLU/GSM8K &  \\

\bottomrule
\end{tabular}
\caption{General-purpose evaluations used for \texttt{Thomson}. Aggregate scores for Development Evals are monitored during the model training process to measure and control forgetting, while Test Evals are reported as a genuine performance evaluation. None of the evaluations has dedicated training sets ($^*$uses CapTrack examples).}
\label{tab:captrack_table}
\end{table}

A vital tool throughout model development is the monitoring of capability preservation throughout all training stages. In order to effectively achieve this, we rely on a capability-centric meta-benchmark for measuring behavioural drift in large language models (CapTrack; for full details see \citep{thede2026captrack}). Rather than proposing another standalone benchmark, CapTrack intentionally organises established community benchmarks into a structured capability taxonomy spanning three complementary dimensions: \textsc{CAN} (latent competence), \textsc{WILL} (default behavioural preferences), and \textsc{HOW} (protocol compliance and execution). This organisation enables a multifaceted view of model quality beyond aggregate accuracy, capturing changes in a semantically meaningful manner. Throughout this work, CapTrack serves as the primary framework for tracking capability evolution and identifying potential model drift across the training pipeline.

Each capability is assessed through multiple complementary evaluations and capability-specific metrics, providing robust capability estimates while reducing reliance on any individual benchmark. Rather than focusing on absolute benchmark scores, CapTrack quantifies relative changes in capability between model checkpoints and their corresponding base models, enabling consistent tracking of behavioural drift throughout training.

In the design of Captrack, we intentionally combine our aim for efficiency as well as pillars (i) and (ii) of our model development by applying data-pruning techniques to constituent benchmarks. Concretely, we categorise each evaluation example across 16 cognitive dimensions \citep{zhou2025general}, allowing us to apply semantic de-duplication / data pruning to systematically subsample and drastically reduce the size of the remaining data while preserving as clear a signal as possible, an approach we term Scales++ (see \citep{bean2025scales++} for full details). As a result, this provides broad capability coverage while remaining feasible to run repeatedly across multiple checkpoints and training stages on modest compute budgets.

To maintain scientific integrity, we contrast CapTrack with our general capability evaluation suite in Table~\ref{tab:captrack_table}, adding additional benchmarks for most categories, motivated by the development/unseen evaluation suite split recommended in \citep{lambert2024tulu}. We also note the additional benefit of using Scales++, which reduces sample overlap even when benchmarks are re-used, as the CapTrack benchmark versions only contain about 10-15\% of the samples in the original data. An exception is Instruction Following, which we believe is indirectly measured through most other evaluations (automated scoring highly penalises failures to comply with format instructions). 

\subsection{Agentic Deep Research} \label{subsec:AgenticDeepResearch}

Tool use recurs throughout this report as an essential model capability targeted through data design and training, with its infrastructure being a critical component best owned independently and usable with any model (\ref{item:sovereignty-data}), thereby also enabling fair evaluation. Its function is to augment parametric knowledge (memorised by a model during training) with access to external resources at test time, enabling the utilisation of private data or knowledge recency. This section describes the preliminaries, aiding the reader with the technical details in Section~\ref{sec:training}.

The most powerful instantiation of tool calling is Deep Research, where a model uses a suite of tools to independently retrieve and synthesise evidence over a long horizon, typically for high-impact tasks that prioritise output quality over model response latency. It is thus one of the clearest tests of sustained agentic capability \citep{du2026deepresearch,huang2025deep}: Successful research requires planning over long horizons, efficacious utilisation of search tools, faithful interpretation of long source documents, meaningful synthesis, as well as appropriate and highly accurate citations. While capabilities like tool-calling \citep{schick2023toolformer,yao2024tau}, long context interpretation \citep{bai2024longbench,hsieh2024ruler}, and reasoning \citep{wei2022chain} are tested separately within our evaluation suite (Section~\ref{sec:evaluation}), we look to agentic research as a test of integrating these capabilities in meaningful settings.

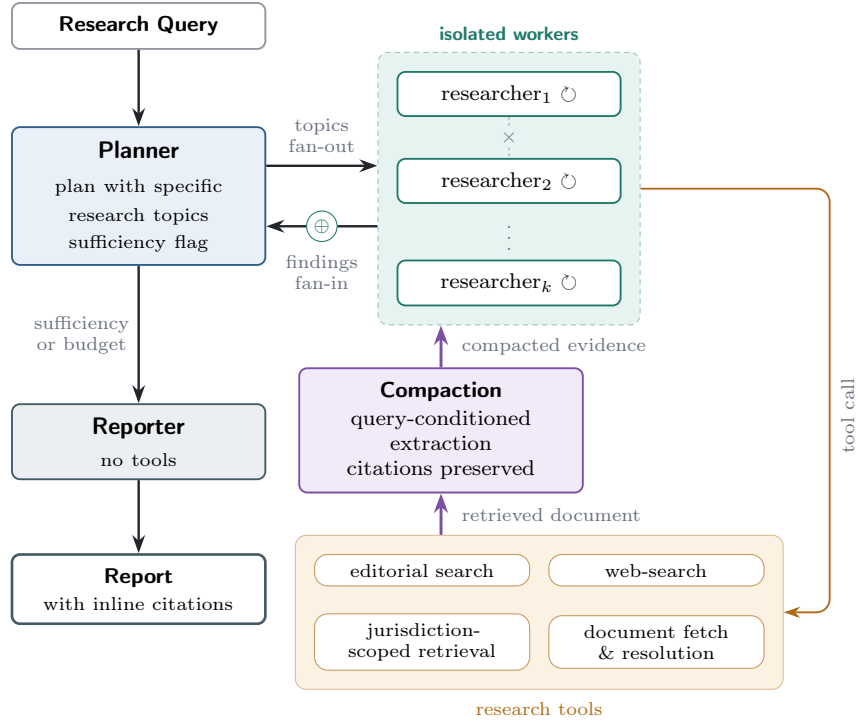
\begin{figure}[t]
\centering
\begin{tikzpicture}[
    font           = \small,
    stage/.style   = {draw, line width=0.7pt, rounded corners=3pt, align=center,
                      inner sep=5pt, drop shadow={opacity=0.09, shadow xshift=0.5pt,
                      shadow yshift=-0.5pt}},
    io/.style      = {stage, draw=cMute!75, fill=white,   text width=30mm, font=\footnotesize},
    outblk/.style  = {stage, draw=cRep, line width=1.0pt, fill=white,
                      text width=30mm, font=\footnotesize},
    planner/.style = {stage, draw=cPlan, fill=cPlanBg, text width=30mm, font=\small},
    worker/.style  = {stage, draw=cRes,  fill=white,   text width=26mm, font=\footnotesize},
    comp/.style    = {stage, draw=cComp, fill=cCompBg, text width=34mm, font=\footnotesize},
    rep/.style     = {stage, draw=cRep,  fill=cRepBg,  text width=30mm, font=\small},
    pill/.style    = {draw=cTool!70, fill=white, rounded corners=5pt, inner sep=3.5pt,
                      align=center, font=\scriptsize, text width=26mm},
    reducer/.style = {draw=cRes, fill=white, circle, inner sep=1.2pt,
                      font=\scriptsize, text=cRes},
    flow/.style    = {-{Stealth[length=2.4mm,width=1.7mm]}, line width=0.9pt, draw=cInk},
    stub/.style    = {line width=0.9pt, draw=cInk},
    call/.style    = {-{Stealth[length=2.2mm,width=1.5mm]}, line width=0.7pt,
                      draw=cTool, rounded corners=5pt},
    ret/.style     = {-{Stealth[length=2.6mm,width=1.9mm]}, line width=1.1pt, draw=cComp},
    lbl/.style     = {font=\scriptsize, text=cMute, align=center, inner sep=2pt}
  ]

  \node[io]      (q)       at (0, 2.90) {\textbf{Research Query}\\[1pt]};
  \node[planner] (planner) at (0, 0.65) {\textbf{Planner}\\[2pt]
                                         {\scriptsize plan with specific research topics}\\[-1pt]
                                         {\scriptsize sufficiency flag}};
  \node[rep]     (rep)     at (0,-2.60) {\textbf{Reporter}\\[1pt]{\scriptsize no tools}};
  \node[outblk]  (out)     at (0,-4.55) {\textbf{Report}\\[1pt]
                                         {\scriptsize with inline citations}};

  \node[worker] (r1)  at (4.90, 2.00) {researcher$_1$~$\circlearrowright$};
  \node[worker] (r2)  at (4.90, 0.85) {researcher$_2$~$\circlearrowright$};
  \node[font=\scriptsize, text=cMute] (dts) at (4.90, 0.20) {$\vdots$};
  \node[worker] (rk)  at (4.90,-0.50) {researcher$_k$~$\circlearrowright$};

  \begin{scope}[on background layer]
    \node[draw=cRes!45, dash pattern=on 2pt off 1.6pt, rounded corners=4pt,
          fill=cResBg, inner sep=7pt, fit=(r1)(r2)(dts)(rk),
          label={[font=\scriptsize, text=cRes, yshift=1pt]above:%
                 {\textbf{isolated workers}}}]
         (band) {};
  \end{scope}

  \draw[dash pattern=on 1pt off 1.3pt, draw=cMute!70] (r1.south) -- (r2.north)
        node[midway, fill=cResBg, inner sep=1pt, text=cMute!90, font=\scriptsize]
        {$\times$};

  \node[comp] (cmp) at (4.00,-2.45)
       {\textbf{Compaction}\\[1pt]
        query-conditioned extraction\\
        citations preserved};

  \node[pill] (t1) at (3.75,-4.30) {editorial search};
  \node[pill] (t2) at (6.85,-4.30) {web-search};
  \node[pill] (t3) at (3.75,-5.25) {jurisdiction-scoped retrieval};
  \node[pill] (t4) at (6.85,-5.25) {document fetch \& resolution};

  \begin{scope}[on background layer]
    \node[draw=cTool!45, rounded corners=4pt, fill=cToolBg, inner sep=7pt,
          fit=(t1)(t2)(t3)(t4),
          label={[font=\scriptsize, text=cTool, yshift=-1pt]below:%
                 {research tools}}]
         (tools) {};
  \end{scope}

  \draw[flow] (q) -- (planner);
  \draw[flow] (planner) -- node[lbl, left, xshift=-1pt]
       {sufficiency\\ or budget} (rep);
  \draw[flow] (rep) -- (out);

  \coordinate (outL) at ([yshift=4mm]planner.east);
  \coordinate (outR) at (outL -| band.west);
  \coordinate (inL)  at ([yshift=-4mm]planner.east);
  \coordinate (inR)  at (inL -| band.west);

  \draw[flow] (outL) -- (outR);
  \node[lbl, anchor=south] at ($(outL)!0.5!(outR) + (0,0.10)$) {topics\\ fan-out};

  \node[reducer] (red) at ($(inL)!0.5!(inR)$) {$\oplus$};
  \draw[stub] (inR) -- (red);
  \draw[flow] (red) -- (inL);
  \node[lbl, anchor=north] at ($(inL)!0.5!(inR) - (0,0.30)$) {findings\\ fan-in};

  \draw[call] (band.east) -- ([xshift=25mm]band.east) |- (tools.east);
  \node[lbl, rotate=90, anchor=south] at ([xshift=29mm,yshift=-30mm]band.east)
       {tool call};

  \draw[ret] (tools.north -| cmp.south) -- (cmp.south);
  \node[lbl, anchor=west] at ($(tools.north -| cmp.south)!0.5!(cmp.south) + (0.20,0)$)
       {retrieved document};

  \draw[ret] (cmp.north) -- (band.south -| cmp.north);
  \node[lbl, anchor=west] at ($(cmp.north)!0.5!(band.south -| cmp.north) + (0.20,0)$)
       {compacted evidence};

\end{tikzpicture}
\caption{Deep Research agent harness. A planner decomposes the research query into self-contained research topics and dispatches them to isolated workers, each a ReAct loop with a bounded tool-call budget; workers share no context, so coverage is a property of the plan rather than of lateral coordination. Tool results return through query-conditioned compaction, which discards prose but preserves citations, keeping the evidence a worker accumulates small while the addressable corpus stays large. The planner re-plans against remaining gaps or hands off to a reporter that has no tools, which is then tasked with producing a Deep Research report with in-line citations.}
\label{fig:harness:flow}
\end{figure}

In many domains, Deep Research is also one of the most economically consequential applications of AI \citep{xu2025comprehensive}, spanning scientific research, finance, compliance, journalism and dozens of other domains combining source identification, interpretation, and synthesis. In biomedical research, determining whether a proposed mechanism, drug interaction, or trial design is consistent with the existing literature requires reviewing primary studies, dosage and safety data, and regulatory guidance that is frequently revised as new trials report out. While our research system is not intended to have the sophistication of a full research product, close alignment to real-world tasks also bolsters its value as a training and evaluation environment (Section~\ref{sec:evaluation}), demonstrating several of the SovereignAI principles introduced in Section~\ref{ssec:sovereign_ai_through_continual_learning} put into practice most concretely.

This makes Deep Research a demanding evaluation target, requiring a harness/scaffold that can run a complete research episode end-to-end, identically for every model under test. Furthermore, complex research reports are not deterministically verifiable, and valuable information for complex economic domains is not easily accessible on the open web. While open-web research discussed in other reports \citep[e.g.][]{shao2025dr} shares a similar agentic loop, we tackle the following unique challenges motivated by the context of the report (Section~\ref{ssec:prof_work_context}):

\textbf{\emph{(C1) The evidence is not on the open web.}}
Authoritative evidence often sits behind private or licensed databases and paywalled corpora rather than the indexed open web (e.g., the Reuters News archive for reliable global news spanning decades, restricted-access outlets and journals for scientific publishing and financial reporting, and licensed professional research databases for primary source material). Consequently, a model cannot (and should not) fall back on memorised web text or general search coverage, so performance has to come from inference-time search and retrieval against these gated sources. On the positive side, this also makes contamination a much smaller concern than for open-web browsing benchmarks such as GAIA \citep{mialon2024gaia} or BrowseComp \citep{wei2025browsecomp}.

\textbf{\emph{(C2) Documents are orders of magnitude longer.}} A typical web page is a few thousand tokens, while this report deals with documents that routinely exceed 50k (e.g., a regulatory filing, a clinical trial protocol, or a technical standard). The bottleneck moves from \emph{finding} a document to \emph{absorbing} it,  i.e., retrieval is cheap to invoke and expensive to consume, and an agent that fetches indiscriminately will exhaust its context before it can synthesise and, as a result, will affect measured quality. We note that a larger window does not dissolve the problem either, since long context degradation appears well before a window is nominally full \citep{hsieh2024ruler}.

\textbf{\emph{(C3) Correctness is query-dependent, not intrinsic to the document.}} A document can be internally valid and still be the wrong answer to a given query, because its validity depends on facts external to the document itself -- facts that change over time and are not visible from the text alone (e.g., a guideline or trial result on medicine later superseded by new evidence). A model can therefore correctly use information in the document, but incorrectly for the query at a given timepoint, resulting in a fluid, well-cited report that is nonetheless incorrectly conditioned on the query. There are also hard constraints on applicability -- analogous to the jurisdiction in law -- that shape the search process itself.

Figure~\ref{fig:harness:flow} shows the design of a Deep Research agent harness used throughout this report. Specifically, we adapt DeerFlow \citep{deerflow2025}, an open-source LangGraph \citep{langgraph2024} research agent implementing the orchestrator-worker pattern common to multi-agent research systems \citep{anthropic2025multiagent}. We retain the harness topology, with significant adaptations to the evaluation and training to address unique challenges in non-verifiable domains.  The harness has the following architecture:

\textbf{Planning.}
In response to a research query, the planner emits a typed plan typically of 3-5 steps. Each step is a self-contained retrieval topic specifying what evidence to collect. Planning runs as direct structured generation rather than inside a tool-calling loop.

\textbf{Research.}
Each researcher is a ReAct agent \citep{yao2022react} with a bounded tool-call budget, working to complete a research topic using the tool suite. It sees the broader query and its own topic, not the other topics and findings of other researchers. This isolation is the main structural control on context growth.

\textbf{Fan-in and Re-planning.}
Findings of individual researchers are merged back via a fan-in to the planner. The planner re-examines the accumulated findings and either issues another research round targeting gaps or declares sufficiency; rounds are capped.

\textbf{Reporter.} The reporter receives the finalised plan and all findings, and has no tools. It is tasked with writing a research report with an explicit instruction to include inline citations to the specific document for every proposition/claim.

\textbf{Tool suite.} Addressing (C1) and the use of authoritative corpora, the harness dynamically wraps a set of custom research tools, fully controlled by the entity, thus implementing SovereignAI goal \ref{item:sovereignty-data}.

\textbf{\emph{Context management.}}
Addressing (C2), the challenge of long documents, we adapt the harness to use \emph{query-conditioned evidence extraction}. Any tool result above a length threshold is compacted by a model conditioned on the research topic and the current findings.

\newpage
%


\section{Model Development}
\label{sec:training}

\subsection{Overview}
\label{ssec:training_overview}

\begin{figure}[h]
\centering
\includegraphics[width=\linewidth]{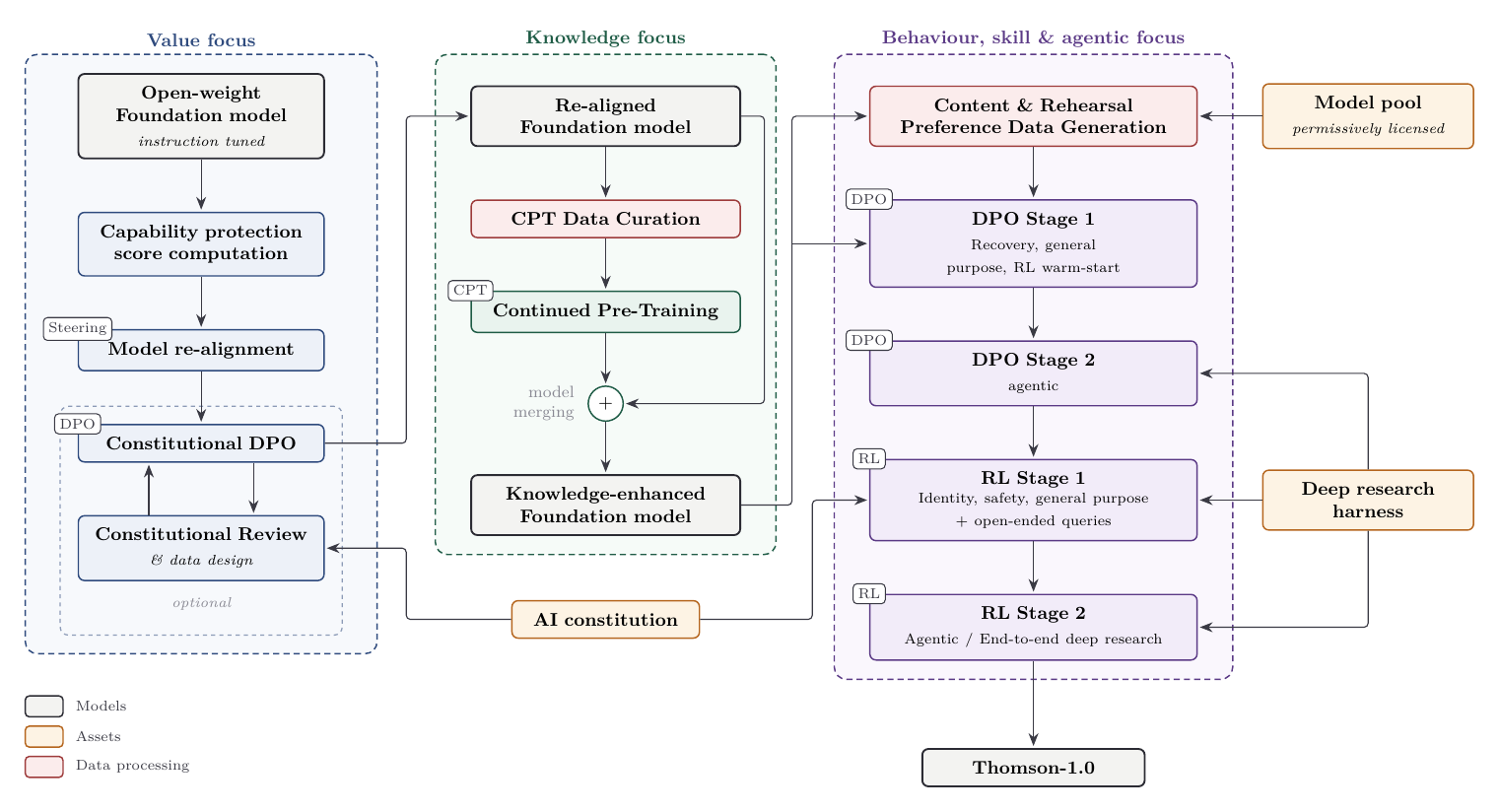}
\caption{Model development pipeline for \texttt{Thomson~1}. The pipeline is organised into three sequential modules with distinct development foci. \textbf{Value focus}: Starting from an open-weight instruction-tuned Foundation Model, we perform value re-alignment through constitutional DPO, potentially augmented with activation steering. \textbf{Knowledge focus}: Data-centric continual pre-training (CPT) ingests proprietary domain data, with model merging protecting generic capabilities while absorbing domain knowledge. \textbf{Behaviour, skill \& tool focus}: Two-stage DPO (broad preference alignment, then agentic specialisation) warm-starts two-stage RL (short-context with diverse queries, then long-context end-to-end Deep Research). Agentic data from the Deep Research harness feeds DPO Stage~2 and both RL stages. This modular design allows practitioners to adjust computational investment to match sovereignty requirements.}
\label{fig:training_stages}
\end{figure}

Figure~\ref{fig:training_stages} shows our overall model development diagram, separated into three modules with distinct development foci (values, knowledge, behaviour/skill/tool). Upon completion of a series of stages designed to achieve the respective development goal, we aim to arrive at a fully capable Foundation Model which may either be used directly or serve as input to the next module to further improve its capabilities. This allows model builders to further adjust computational budgets to sovereignty goals. As a practical choice, we found it valuable to first realign existing \textit{instruction-tuned} Foundation Models with our own constitution, hypothesising that current alignment occurs primarily during post-training and may thus be altered with relatively cheap methods (such as activation steering).

Privately held unstructured data is best utilised through knowledge injection in the secondary stage, allowing further learning with a pre-training objective. To enable continued pre-training on an \emph{instruction-tuned model}, we develop a data-centric pipeline focused on the automated enhancement and restructuring of raw data. This has the dual benefit of reducing distribution mismatch with the typical shape of post-training data and serving as a soft warm start for the final module. In addition, a vital model-merging step prevents otherwise detrimental degradation of model capabilities. This paradigm is arguably closer to what is often referred to as Mid-Training.

Our final module combines typical post-training stages to unlock newly acquired knowledge for practical tasks. Direct preference optimisation (DPO) \citep{rafailov2023direct} stages serve the dual purpose of warm-starting a model for online reinforcement learning \citep{shao2024deepseekmath} while allowing fine-grained edits to any remaining degradation not fully recovered by model merging. Among all model modifications, we found DPO and RL to be the most robust to model degradation, provided sufficient care is taken to design high-quality preference data and training environments. Note that we explicitly forgo any further supervised fine-tuning (SFT) as an isolated post-training stage due to its tendency to aggressively alter model behaviour, leading to rapid forgetting even across broad data distributions \citep{thede2026captrack}.

\subsection{Value Re-alignment}

\textit{(Co-authored by Imperial College London)}

\label{ssec:values}
We re-align the values of open-weight models following the two-stage procedure of~\citep{imperial2026realignment}, which brings the model's expressed values into line with the Public AI Constitution~\citep{patriniche2026publicai} rather than with those of its original training distribution. The first stage is Fisher-protected directional ablation (Section~\ref{subsec:fisher}): a training-free weight edit that identifies the direction in activation space separating value-laden responses from neutral ones and projects it out of the model's weights. The second stage is Constitutional DPO (Section~\ref{subsec:constiutional-dpo}) over paired responses drawn from both expert-curated and synthetic data, which consolidates the edit. The binding constraint throughout is that the model's inherent capability must survive: a realignment that trades away general competence is of no practical use to us, so we treat capability retention as an objective rather than as a diagnostic.

\subsubsection{The Alignment--Capability Frontier}

\begin{figure}[h]
    \centering
    \includegraphics[width=0.85\linewidth]{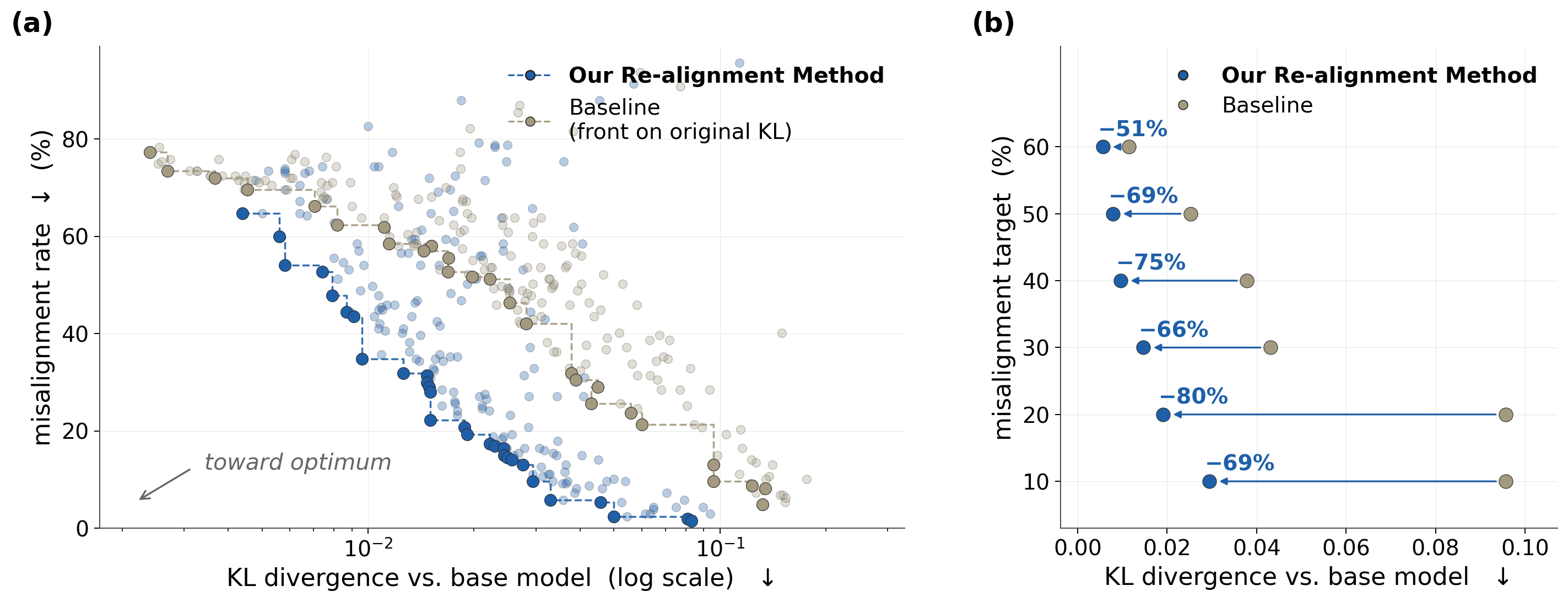}
    \caption{%
Misalignment-rate vs. capability preservation. Our realignment traces a strictly better frontier than the plain-abliteration baseline, reaching any given misalignment level at substantially lower KL divergence from the base model.
  \textbf{(a)} Misalignment rate on the validation prompt set vs.\ KL divergence from the base model; faded markers are individual trials, solid markers joined by dashed
  lines are the Pareto fronts. Note: Baseline KL scores are replayed under our KL score definition for comparability, with their front selected on the original KL scores.
  \textbf{(b)} KL divergence cost to reach each misalignment rate; our realignment method incurs $2.0\times$--$5.0\times$ less KL divergence scores than the baseline, suggesting better model capability preservation during ablations.%
}
\label{fig:realignment-pareto-frontier}
\end{figure}

Because alignment and capability can pull against each other, we do not reduce the search to a single scalar objective; we characterise each method by its Pareto front over two measurements. The first is the \emph{misalignment rate}: the fraction of a held-out validation set of value-probing prompts on which the edited model fails to respond in line with the desired value. The second is the capability cost, measured as the \emph{multi-token Kullback-Leibler (KL) divergence} between the edited model and its corresponding base model over a curated prompt set (KL set). That set is deliberately broad, including benign everyday prompts, capability-specific and safety probing prompts, so that a low KL score indicates less behavioural drift, rather than merely preserving behaviour on the alignment distribution itself.

Figure~\ref{fig:realignment-pareto-frontier} reports this front. Our realignment dominates the baseline abliteration over the entire range: for every misalignment target we consider, it reaches that target at $51$--$80\%$ lower KL divergence cost ($2.0\times$--$5.0\times$ cheaper), as quantified in panel~(b). The advantage is widest at aggressive targets, where the baseline can only continue reducing misalignment by accepting capability damage. The baseline method illustrated here shares our direction-extraction data and validation set, but during its trial search it tracks a single-token KL divergence over a limited prompt set. To ensure that our advantage does not merely reflect a change of metric, we recompute the KL score for every baseline trial on our KL set under the identical multi-token definition. Its front is still selected using the KL scores it actually optimised against, and only then re-plotted at the recomputed coordinates; we do not re-optimise the baseline after recomputation. Both searches are given the same budget of $201$ trials.

Optimising this frontier yields a family of ablated checkpoints rather than a single model. From it, we select a small set of operating points and consolidate each with a preference optimisation step, searching jointly over the choice of front and the DPO hyperparameters. Crucially, the final \texttt{Snowdon} checkpoints are \emph{not} chosen on the search-time proxies of
Figure~\ref{fig:realignment-pareto-frontier} but on the downstream quantities: a realignment score measured with Perspective Bench and a general-capability score aggregated over $18$ benchmarks in CapTrack as well as an unsafe prompt refusal score. The evaluation that follows (Section~\ref{sec:value-evaluation}) asks whether the frontier's proxy advantage survives this change of measurement, on a held-out test set the search never saw.

\begin{figure}
    \centering
    \includegraphics[width=0.65\linewidth]{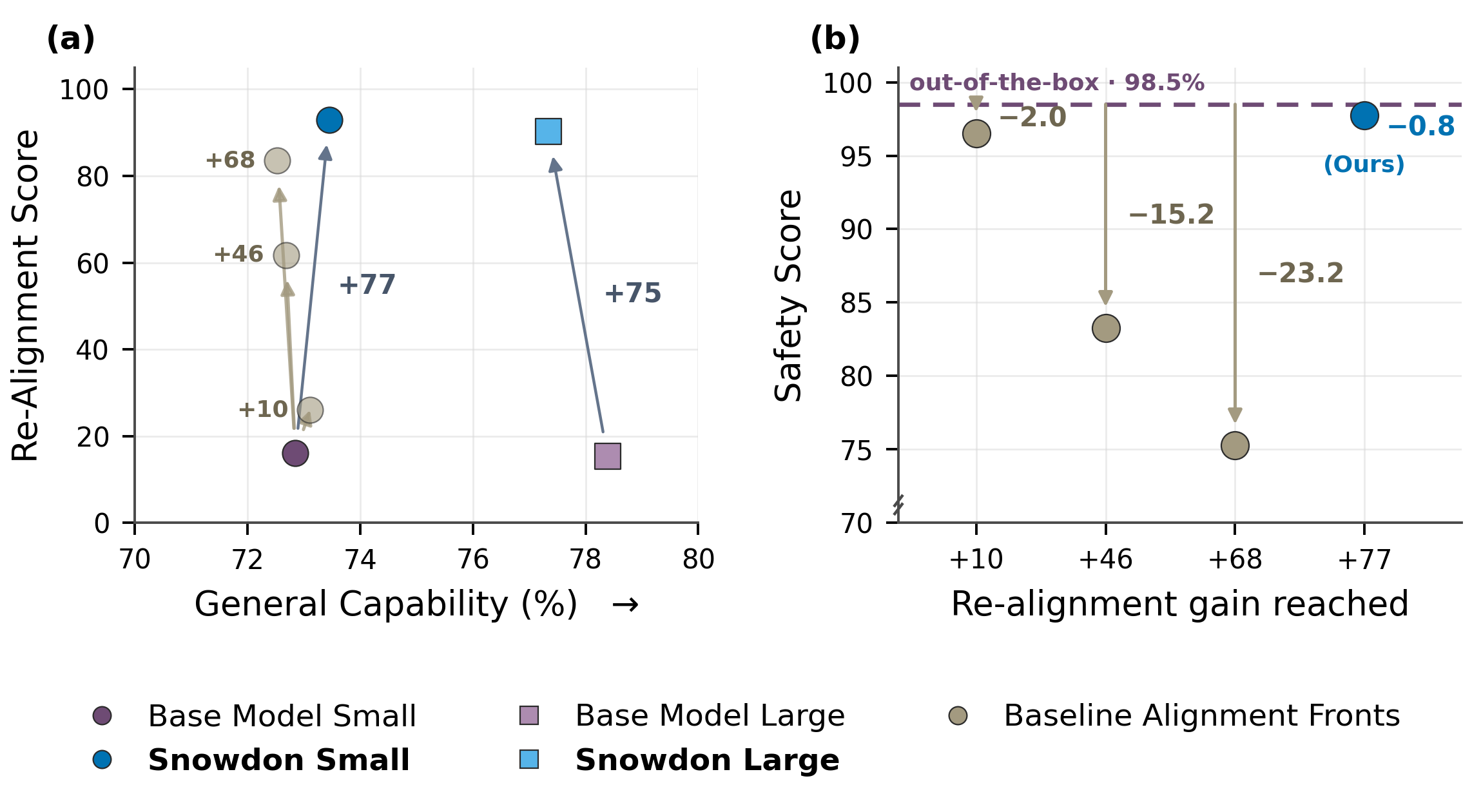}
\caption{Realignment on the small and large models. Each panel contrasts the out-of-the-box base models (\emph{Base Model Small/Large}) with our re-aligned checkpoints (\emph{Snowdon Small/Large}); the
\emph{Baseline Alignment Fronts} are plain ablated versions of the small model.
\textbf{(a)} Realignment score (Perspective Bench) against general capability (CapTrack): both sizes gain large points of realignment with essentially no capability loss, while baselines reach only partial re-alignment.
\textbf{(b)} Safety, as the unsafe-request refusal rate's drop from the
out-of-the-box line, plotted against the realignment gain each method reaches. Baselines improve re-alignment only by giving up safety score, and still plateau below the re-alignment level \texttt{Snowdon} reaches; \texttt{Snowdon} attains the largest gain at a cost of 0.8\% refusal. Note: models are tested in non-reasoning mode for this experiment.}
\label{fig:realignment-overview}
\end{figure}

\subsubsection{Fisher-Routed Directional Ablation}
\label{subsec:fisher}

We suppress the target misaligned behaviour with a rank-one weight edit. The method separates two choices. A difference of means identifies which residual-stream direction to steer. An activation-space Fisher estimate determines where to
place the resulting correction.

\paragraph{Estimating the Direction.}
We compare the model's state on two prompt populations. The sensitive set $S$ elicits the target behaviour, while the neutral set $N$ does not. We generate both sets independently from the same topic taxonomy.

\begin{figure}[htbp]
    \centering
    \includegraphics[width=\linewidth]
    {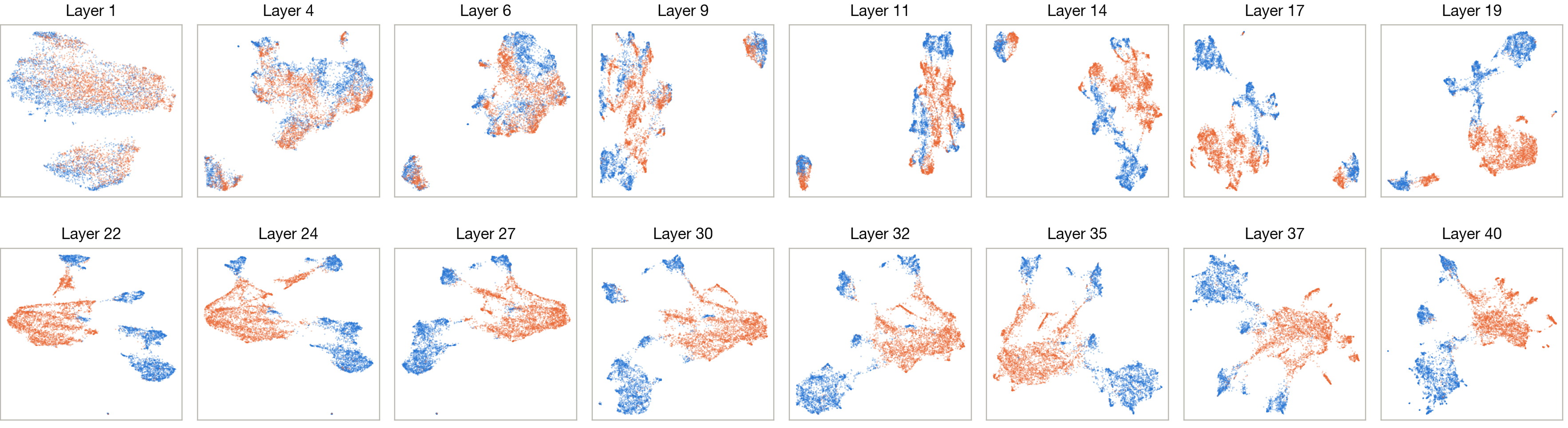}
    \caption{The misaligned behaviour has a separable representation. UMAP projections of final-prompt-token residual stream activations, computed independently at each layer, for the sensitive set S (elicits the misaligned behaviour) vs. the neutral set N (does not). Points from the two prompt sets are interleaved in early layers and become separated by mid-stack. Each panel is an independent fit, so panels show separability at a depth, not trajectories across depth.}
    \label{fig:set-a-vs-set-b}
\end{figure}

Let $\mu^S_\ell$ and $\mu^N_\ell$ denote their mean residual-stream activations
at layer $\ell$. We record each activation at the final prompt token, immediately
before the model begins its response. Their normalised difference defines the
candidate direction at that layer
\citep{arditi2024refusal,belrose2023diffinmeans}:
\begin{equation}
  \label{eq:direction-s}
  v_\ell
  =
  \frac{\mu^S_\ell-\mu^N_\ell}
       {\|\mu^S_\ell-\mu^N_\ell\|_2}.
\end{equation}
Averaging suppresses prompt-specific variation and retains the systematic difference between the populations. Figure~\ref{fig:set-a-vs-set-b} illustrates the resulting layer-wise separation between $S$ and $N$ using two-dimensional UMAP projections \citep{mcinnes2018umap}. Each search trial chooses a continuous
source-layer index and interpolates the adjacent directions. We normalise the result and use the resulting direction $v$ at every edited layer.

\paragraph{Degrees of Freedom in the Writeback.}
Let $W$ be the weight matrix of a component whose output enters the residual
stream, and let $b$ be its writeback vector. We write the edit as
\begin{equation}
  \label{eq:edit-s}
  \Delta W=b(v^\top W),
  \qquad
  v^\top b=-\lambda
  \quad\Longrightarrow\quad
  v^\top(W+\Delta W)=(1-\lambda)v^\top W.
\end{equation}
The constraint fixes the removal strength $\lambda$ while leaving $d-1$
orthogonal degrees of freedom in $b$. Standard directional ablation sets these
to zero, giving $b=-\lambda v$. We instead choose
$b=-\lambda v+b_\perp$, where $v^\top b_\perp=0$, to minimise the edit's
predicted effect on other model behaviour. During search, we implement each
rank-one edit as a LoRA adapter \citep{hu2021lora}.

\paragraph{Measuring Output Sensitivity.}
For each edited component, we estimate the diagonal Fisher information of the
model's predictive distribution in the component's output space
\citep{kirkpatrick2017overcoming}. Omitting layer, component, and token indices,
\begin{equation}
  \label{eq:fisher-s}
  F_i
  =
  \mathbb{E}_{x\sim\mathcal D}
  \mathbb{E}_{\hat a\sim p_\theta(\cdot\mid x)}
  \left[
    \left(
      \frac{\partial\log p_\theta(\hat a\mid x)}
           {\partial z_i}
    \right)^2
  \right],
\end{equation}
where $z_i$ is output coordinate $i$ and $\mathcal D$ is the reference
distribution. Large $F_i$ indicates that changing coordinate $i$ has a larger
local effect on the model's predictive distribution; small $F_i$ indicates
lower estimated sensitivity.

We estimate $F$ on a reference set spanning safety, tool use, instruction
following, multilingual behaviour, and general capability, disjoint from the
data used to score search trials.

\paragraph{Fisher-Routed Writeback.}
Let $\overline F$ denote the mean of the coordinates of $F$. We define
\begin{equation}
  \label{eq:writeback-s}
  h_i = F_i+\rho\overline F,
  \qquad
  g_i = \frac{v_i}{h_i},
  \qquad
  s = v^\top g,
  \qquad
  b = -\frac{\lambda}{s}g.
\end{equation}
We fix the ridge fraction at $\rho=0.05$.

Each step has one role. The first adds a floor to the Fisher values, preventing
coordinates with $F_i\approx 0$ from receiving an excessive correction. The
second reduces the contribution of coordinates with high predictive
sensitivity. The third measures the remaining projection of the routed vector
onto $v$. The final step sets the scale required to remove the chosen fraction
$\lambda$.

Because $s=v^\top g$, the normalisation preserves $v^\top b=-\lambda$
exactly. Fisher routing therefore changes where the correction is written, not
how much of the target projection is removed.

This writeback minimises the local diagonal-Fisher cost among all vectors that
satisfy $v^\top b=-\lambda$ \citep{amari1998natural}. If $F$ is uniform, then
$g$ is proportional to $v$, and the solution reduces to $b=-\lambda v$.
Standard directional ablation is therefore the flat-Fisher case.

\begin{figure}[h]
    \centering
    \includegraphics[width=0.6\linewidth]
    {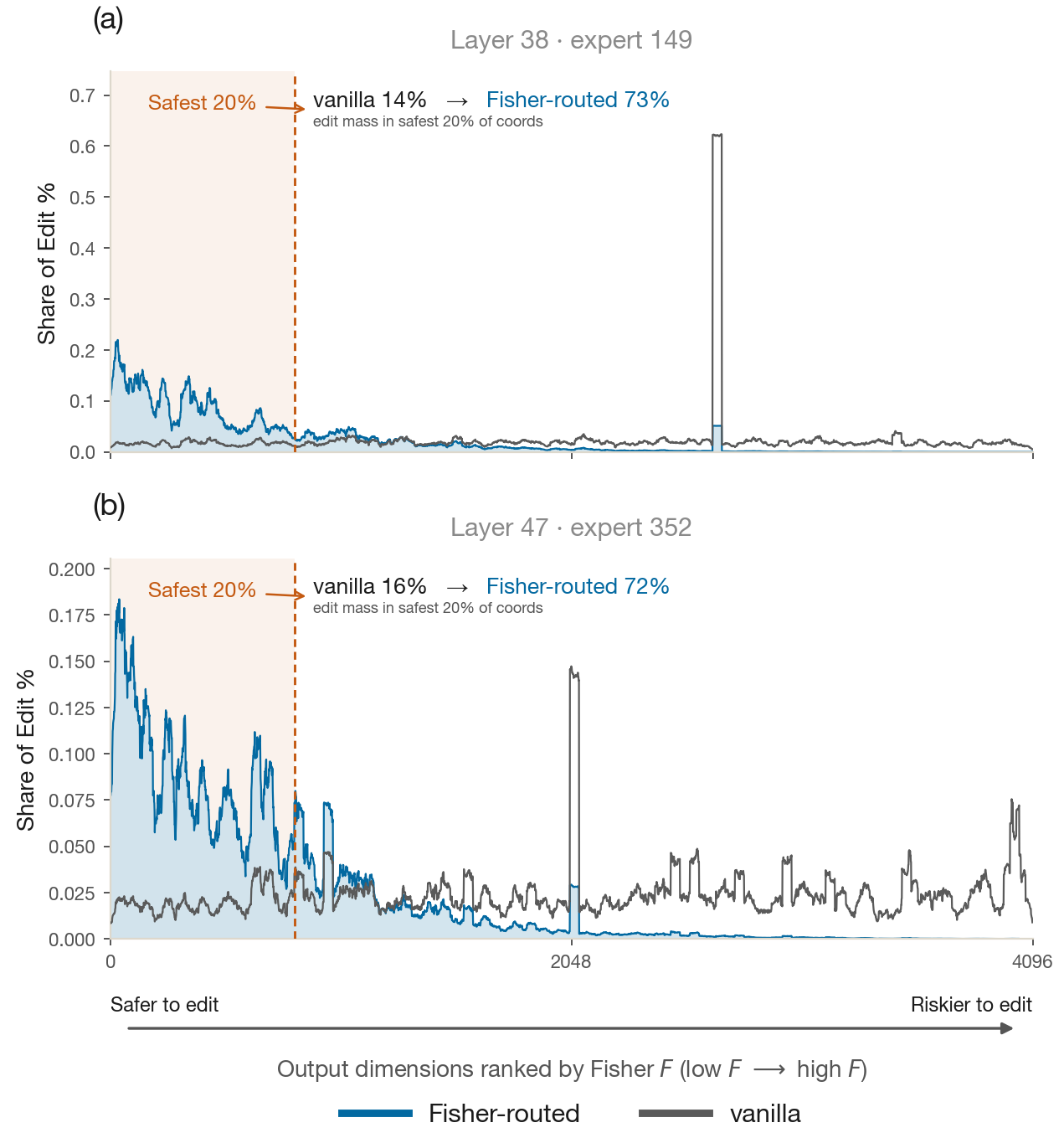}
    \caption{%
    Fisher routing shifts the writeback towards output coordinates with lower
    estimated sensitivity. Coordinates are ordered from low to high Fisher
    value for the selected experts at layers 38 and 47. Grey shows the
    standard writeback $b=-\lambda v$; blue shows the Fisher-routed writeback (ours)
    from Equation~\eqref{eq:writeback-s}. The vertical line marks the lowest-
    Fisher fifth of the output coordinates. Labels report the share of
    squared writeback mass within that fifth. 
    }
    \label{fig:fisher-protection-experts}
\end{figure}

Figure~\ref{fig:fisher-protection-experts} shows the routing effect directly.
Both edits satisfy the same removal constraint. The Fisher-routed edit places
more of its writeback on coordinates with lower estimated predictive
sensitivity.

\paragraph{Search Procedure}
\label{subsec:tpe-search}
Our realignment procedure is an automated trial-search loop. Each trial applies a rank-one update to each targeted component, including combination of direction, location and strength, and the resulting model is scored on in-loop proxies for misalignment rate and capability drift. Trials are proposed by Tree-structured Parzen Estimator (TPE) in Optuna \citep{bergstra2011algorithms,akiba2019optuna}, which concentrates the search on promising regions of the edit space.

We adapt this framework by replacing the standard writeback
$b=-\lambda v$ with the Fisher-routed writeback from
Section~\ref{subsec:fisher}, and by measuring model drift with multi-token KL
over a broader anchor set. Each trial is evaluated on held-out misalignment
rate and KL divergence; the non-dominated trials form the
alignment--capability frontier in
Figure~\ref{fig:realignment-pareto-frontier}. The plain-abliteration baseline
uses the standard writeback and its original single-token KL objective. Both
searches use the same direction data, validation set, and trial budget.
Selected points on our frontier are passed to constitutional DPO.

\subsubsection{Constitutional DPO} \label{subsec:constiutional-dpo}
We consolidate selected operating points from the alignment--capability frontier (Figure~\ref{fig:realignment-pareto-frontier}) with a single epoch of length-normalised DPO, searching jointly over the choice of front and the DPO hyperparameters. We call this stage Constitutional DPO: the preference pairs combine human expert-curated examples on a heterogeneous topic mix with synthetic examples on the sensitive topics we target, and in each pair the chosen response is the one that reflects the Public AI Constitution.

\paragraph{Human Expert Pairs.}
Subject-matter experts (SMEs) prompted an early-stage exploration model across a broad topic taxonomy. General domains included coding, culture, history, languages, law, mathematics, medicine, philosophy, reasoning, science, and technology. Sensitive domains included political, historical, and social topics where credible narratives diverge or states maintain institutional positions. Prompts were designed to elicit value-relevant behaviour or test a constitutional constraint. Guided by the Public AI Constitution, SMEs made targeted edits where minor correction was sufficient and rewrote responses requiring substantive changes to their conclusions or framing. Each revision was checked against the Constitution and used as the chosen completion, with the original model response serving as the rejected completion.

\paragraph{Synthetic Pairs.} To scale beyond the human-curated set, each base model answer is rewritten by an ensemble of three open-source models conditioned on the constitution. An LLM judge, using criteria drawn from the constitution, discards revisions that are out of date or that refuse to answer, deflect or return a vague non-answer. Surviving revisions become chosen responses, paired against the OOB rejected response.

\paragraph{Hyperparameter Search.} We grid-search KL strength $\beta \in \{5, 10\}$, learning rate $\in \{1\times10^{-7}, 2\times10^{-7}\}$, NLL weight $\lambda \in \{0.9, 1.0\}$, objective (DPO vs.\ length-normalised DPO), and data mixture (equal 1:1:1:1 vs.\ re-weighted across expert and synthetic sources), selecting the configuration maximising realignment score subject to retaining baseline general-capability performance. The resulting \texttt{Snowdon} checkpoints (Figure~\ref{fig:realignment-overview}(b), blue points) push realignment further while preserving the capability gains from the Fisher-protected edit.

\subsubsection{Realignment Evaluation}
\label{sec:value-evaluation}

The frontier (Figure~\ref{fig:realignment-pareto-frontier}) was traced with search-time proxies, misalignment rate and KL divergence, measured on data that drove the search. We then evaluate the final checkpoints on two independent, downstream measurements and ask whether that advantage survives the change of metric.

\paragraph{Realignment.} We measure realignment with Perspective Bench~\citep{imperial2026realignment}: prompts on geopolitically contested topics, each response graded by a panel of five LLM judges across six rubric dimensions and assigned an overall ideological lean by majority vote. We summarise the lean distribution as a single \emph{realignment score} ($0$--$100$, higher is more aligned).

\paragraph{Model Capability.} We track general capability with CapTrack~\citep{thede2026captrack} (Section~\ref{ssec:monitoring_forgetting}), reporting its latent-competence (CAN) group: what a model can do under ideal prompting across knowledge, reasoning, comprehension, faithfulness, and robustness. Our score is the mean accuracy over 18 benchmarks; we avoid CapTrack's headline average, which mixes in behavioural metrics of differing polarity. This is mainly to keep track of capability preservation of our realignment process. Additionally, we measure \emph{Safety Score} as CapTrack's W1 unsafe-request refusal rate (higher values indicate that the model refuses a larger proportion of unsafe requests).

\paragraph{Results.} Realignment succeeds at both model scales while largely preserving capability (Figure~\ref{fig:realignment-overview}). Both base models start relatively low on the realignment axis and, after the full pipeline, gain roughly $75$ points while general capability holds within a point of its starting value (panel~a). Panel~(b) shows the same holds for safety: every baseline achieves re-alignment by giving up unsafe-request refusal and still plateaus below \texttt{Snowdon}'s re-alignment level, whereas \texttt{Snowdon} reaches the largest gain on realignment only with a 0.8 percentage-point drop in safety refusals. Both results are measured on benchmarks the search never scored against.

\subsection{Continual Pre-training}
\label{ssec:cpt}

\textit{(Co-authored by DatologyAI)}

\subsubsection{Data-Centric Subselection and Enhancement}

\begin{figure}[htbp]
    \centering
    \includegraphics[width=\textwidth]{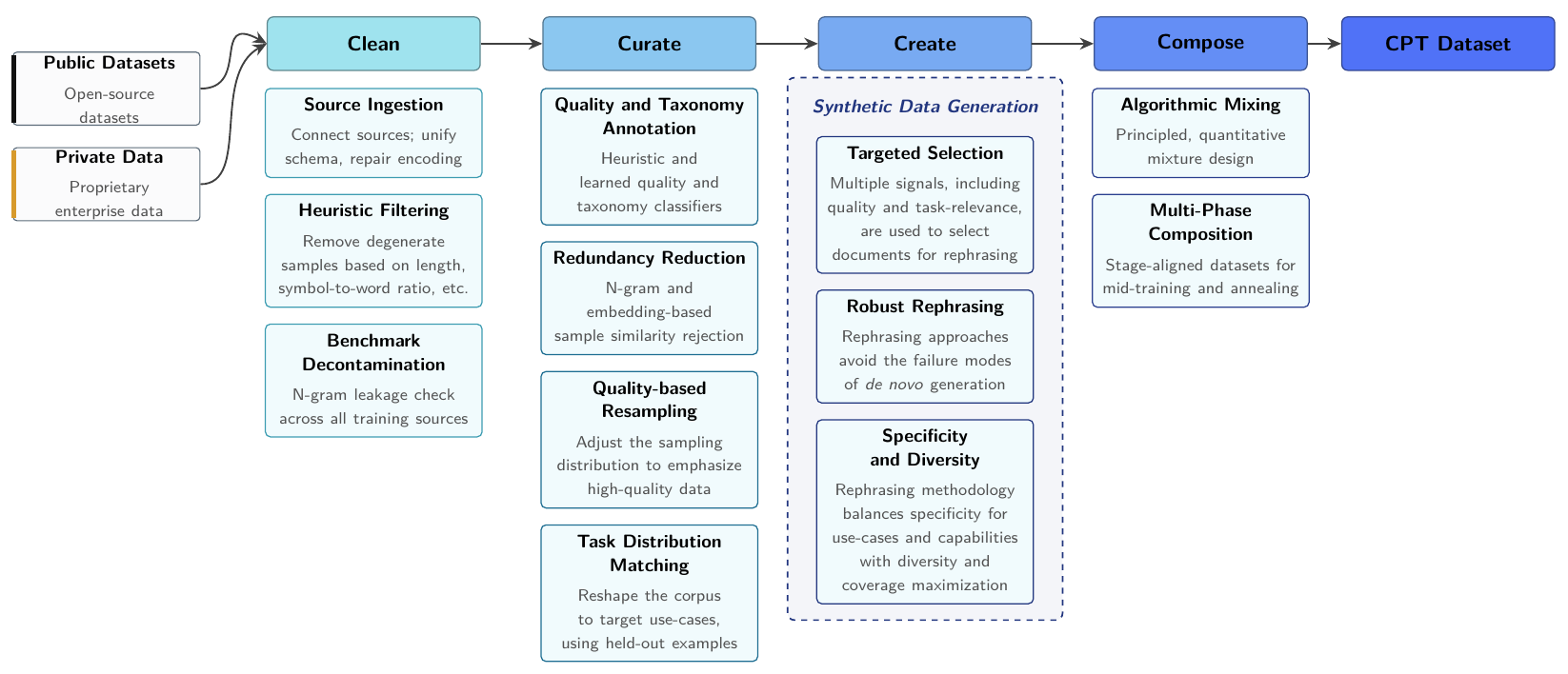}
    \caption{Data curation pipeline for continued pre-training}
    \label{fig:mid_training_curation}
\end{figure}

Data curation is at the heart of our Continual Learning strategy for the knowledge focus module (Figure~\ref{fig:training_stages}) of the \texttt{Thomson} model development pipeline. To do so, we partnered with DatologyAI to curate a high-quality mid-training dataset of 200B tokens from a corpus of permissively public and high-quality proprietary data of over 19T tokens. The key tension in ``Continual Pre-training'' (CPT) for \textit{instruction-tuned} models is the need to elicit deep domain expertise without eroding the model's existing capabilities. This is the standard ``learn without forgetting'' tension that all Continual Learning must navigate, but with the important twist that we perform CPT on top of an \textit{instruction-tuned} (and later value-realigned) checkpoint. We thus make an important terminological distinction and consider this a form of ``mid-training'', hinting that data is carefully curated to sit at an in-between point between raw, unstructured pre-training and formats typically seen in Post-Training. Standard Continual Pre-training on such post-trained models disproportionately degrades the vital capabilities previously developed during post-training \citep{jindal2024balancing, padmanabhan2026updating} and thus presents a critical challenge for Continual Learning. This raises the stakes of forgetting mitigation and, with it, the value of curated replay data.

Our approach to mitigating this is a combination of classic rehearsal-based Continual Learning \citep[e.g.][]{robins1995catastrophic, shin2017continual, rolnick2019experience} along with a modern approach to data-centric techniques and synthetic data \citep[e.g.][]{shin2017continual, titsias2019functional, li2024datacomp, maini2025beyondweb}. Our mid-training data mix thus comprises three roughly equal parts: curated proprietary documents, synthetic data generated from those documents, and curated general-capability replay data.

\begin{figure}[h]
    \centering
    \includegraphics[width=\textwidth]{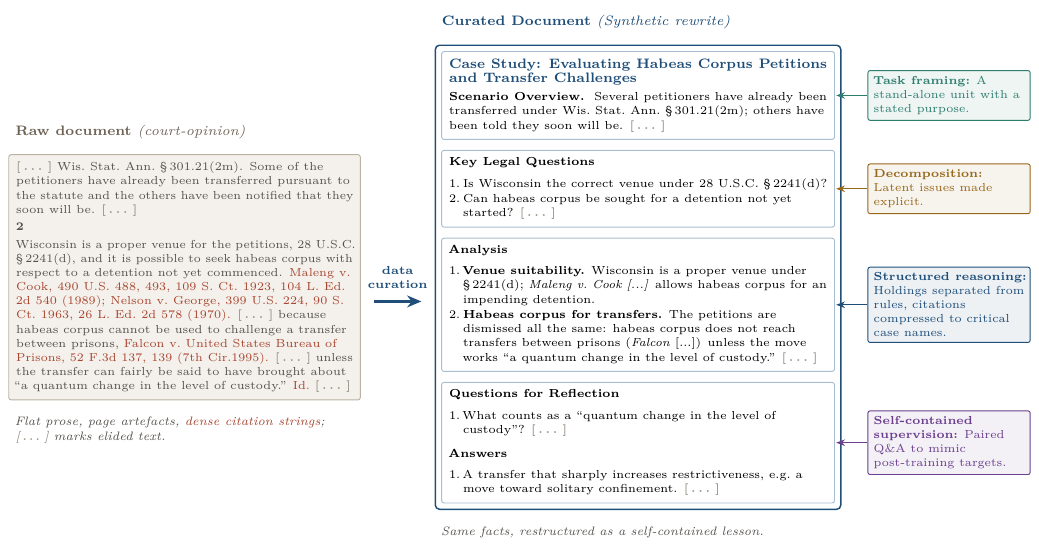}
    \caption{Example curation of a dense document}
    \label{fig:data_curation_example}
\end{figure}

The knowledge of the raw proprietary corpus (containing decades of news, contracts, long and numerically dense regulatory filings, US and international case law, statutes, regulations, practitioner guidance, etc.) is distilled through a series of processing stages (Figure~\ref{fig:mid_training_curation}). The much denser corpus yielded is representative of a corpus of data often held privately by institutions and thus missing in the pre-training data of third-party models. This allows expanding the embedded knowledge through model training while preserving data sovereignty (\ref{item:sovereignty-data}).  The deployed pipeline runs as distributed jobs on Kubernetes and processes each source.

\begin{itemize}
    \item \textsc{Ingestion and Normalisation:} Per-source parsing of heterogeneous formats into a standardised markdown document schema.
    \item \textsc{Exact Deduplication:} applied per-source where duplication analysis warranted it.
    \item \textsc{Length Filtering and Chunking:} with documents capped at 256k tokens.
    \item \textsc{Decontamination:} against all evaluation targets via 13-gram overlap (\`a la \citep{brown2020language}), applied before any selection stage so that no later stage can include evaluation data in training.
    \item \textsc{Quality Filtering:} using heuristic and model-based quality and repetition filters. These filters serve multiple goals. First, removing the bad tail of degenerate documents, such as excessively short, repetitive, poorly-formatted, and insufficiently linguistic documents. And second, identifying fluent, information-dense, and/or relevant documents.
    \item \textsc{Task and Use-Case Distribution Matching:} We upsample documents from the corpus to better align with the distribution of tasks the model will serve in the target domain. We emphasise that evaluation items have already been removed (Stage 4), and that all curation research was conducted without the use of proprietary data or evaluations (as described below).
    \item \textsc{Synthetic Generation:} (see below).
    \item \textsc{Mixing, Shuffling, and Export:} with mixture proportions determined systematically and enforced by token count.
\end{itemize}

Example~\ref{fig:data_curation_example} shows a rephrased example data point, highlighting the key distinction between large-scale pre-training on raw, uncurated content and targeted mid-training with high-quality data designed to teach new knowledge while reducing degradation.

\paragraph{Evaluation.}
To maintain evaluation integrity, our full curation recipe -- including mixture proportions, filter thresholds, and the synthetic data approach -- was developed through mid-training ablations on small open models (specifically \texttt{Qwen3-8B} \citep{yang2025qwen3} and \texttt{Llama-3.1-8B} \citep{grattafiori2024llama}). We used public corpora as stand-ins for the full proprietary corpus, and open benchmarks re-implemented in a common evaluation harness \citep{gu2406olmes}. General-purpose evaluations encompassed world knowledge, reading comprehension, language understanding, general reasoning, mathematics, coding, and instruction following (Section~\ref{ssec:monitoring_forgetting}). Target domain evaluations were composed of the LSAT subsets of AGIEval \citep{zhong2024agieval}, law-related subsets of MMLU \citep{hendrycks2020measuring} as well as a subset of tasks from LegalBench \citep{guha2023legalbench} which we adapted to maximise the signal-to-noise ratio using data similar to \citep{heineman2026signal}.

In cases where the default evaluation protocol scores a generated answer by exact match (conflating capability with format compliance), we implemented subsets in multiple scoring formats: exact match, perplexity-based option selection, and chain-of-thought with answer extraction. We found that perplexity-based selection yielded the lowest noise, particularly on models with diminished instruction-following capability. To reduce the cost of evaluation and maximise signal, we evaluated twelve intermediate checkpoints of OLMo-2-7B \citep{olmo20242} spanning 1B to 3.9T training tokens on a subset of tasks, computed the rank correlation between tokens seen and score, and retained the tasks whose scores improved reliably with training. Fewer than a third exceeded a rank correlation of 0.7.

\paragraph{Synthetic \& Replay Data.}
Our replay data is DatologyAI's general-capability mid-training mixture. This constitutes curated web text, mathematics, code, multilingual data, and a small amount of instruction-style data. Replay is standard practice in domain-adaptive pre-training \citep{ibrahim2024simple}, but the effects of data curation on replay data utility are poorly understood. In ablation studies, we observe that replacing publicly-available replay data derived from Dolmino \citep{olmo20242} with the curated replay data has the potential to substantially improve code (+5.4pp average across HumanEval++, MBPP, and MBPP++) and reading comprehension (PubMedQA +7.2pp, BoolQ +4.5pp), indicating that replay data curation has meaningful, capability-specific downstream effects.

Taking inspiration from Deep Generative replay \citep[e.g.][]{shin2017continual}, we further develop synthetic data. This consists of rephrasings of algorithmically-selected source material, generated using an adaptation of DatologyAI's BeyondWeb method \citep{maini2025beyondweb}. BeyondWeb's efficacy comes from two complementary mechanisms. Diversity, which is achieved by rephrasing source documents into a breadth of formats and registers, and specificity, which is achieved through careful selection of documents for rephrasing using various quality and relevance signals.

While naively applying our standard BeyondWeb-style rephrasing pipeline to domain-specific documents demonstrated clear benefits in internal ablations, we made numerous adaptations and improvements to our synthetic data curation in order to promote strong capabilities in \texttt{Thomson}. After examining benchmarks that made it past our signal and noise analysis, we  identified six core capabilities that we thought were required by these benchmarks: foundational knowledge and language; (verbal) reasoning; reading comprehension and summarisation over supplied documents; long-context handling; specialised knowledge; and numerical reasoning over formal rules. We then expanded the synthetic rephrasing format inventory from general-purpose patterns to also target these capabilities. Before our final run, we validated candidate generator models specifically on their fitness for synthetic data applications. We used a mix of heuristic and model-based assessments, as well as extensive human evaluation. We emphasise that this is for generator qualification, not curation of individual documents. Figure~\ref{fig:data_curation_example} shows a re-phrased example.

Furthermore, we changed the way in which rephrased documents are assembled into contexts. Typically, individual documents are sampled IID from the full pre-training corpus. Instead, we found that leveraging the relationship between the source(s) and rephrased document(s) can meaningfully impact downstream model quality. Finally, we extended the input length for seed data input and output to accommodate the need for long-context reasoning over long and/or many documents.

\subsubsection*{Lessons Learned and Future Work}

Our experiments provided the following insights:
\begin{itemize}
     \item Selection of optimal mid-training learning rate (LR). This depends on the base model and Mid-Training dataset \citep{gupta2023continual, ibrahim2024simple, parmar2024reuse}, hence it is standard practice to run an LR sweep using short training runs before committing to a full mid-training run. We found that the LR minimising training loss continued to deliver the strongest checkpoint after model merging with an appropriate coefficient.
    \item Recipes developed on base models transfer to post-trained models after model merging. We conducted our mid-training curation research using both base and post-trained models, and found that the effects of data curation interventions were qualitatively similar across model types after model merging.
\end{itemize}

Our future work aims to explore the following directions:

\begin{itemize}
    \item Increase data pool. To maintain economically efficient Continual Learning, we restricted our CPT budget to a modest 200B tokens, pruning over 98\% of the original data. While this represents the highest quality data and  we believe this represents a realistic setting for many potential actors interested in SovereignAI, scaling up the data pool through this pipeline is expected to unlock further improvement of \texttt{Thomson}.
    \item Curation to reduce hallucination. Hallucinations are a known weakness of LLMs in high-stakes domains \citep[e.g.][]{dahl2024large}, and are contained in synthetic data \citep{gudibande2023false, zhang2026survey}. Our next cycle of data curation will target reduction of hallucination in synthetic training data and in downstream models.
    \item Curation for multi-document reasoning. Complex tasks require reasoning across many related documents, and this is where current models are weakest. Grouping related documents during curation to teach models how to reason across documents, rather than curating each in isolation, is a promising direction to improve reasoning in complex settings.
\end{itemize}

\subsubsection{Training \& Capability Recovery}
\subsubsection*{Training runs}

\begin{figure}[htbp]
\centering
\includegraphics[width=\linewidth]{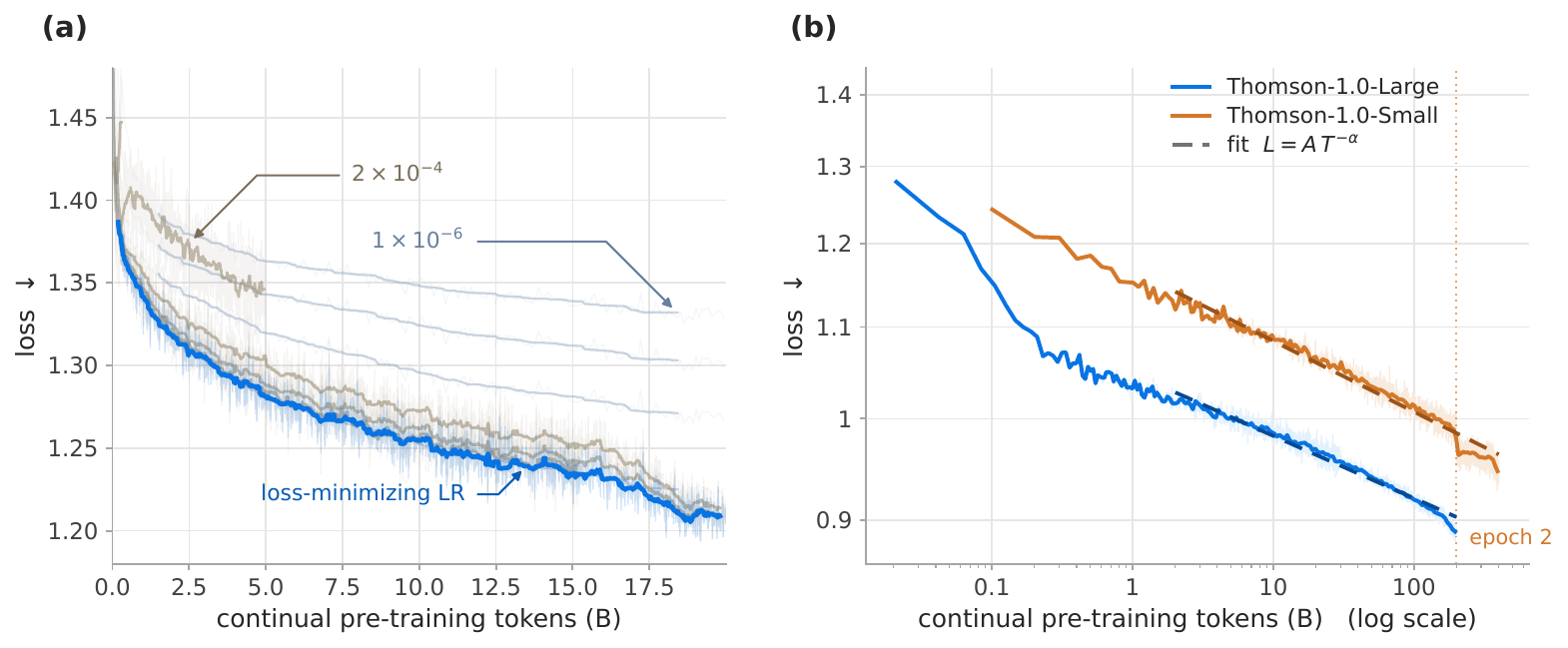}
\caption{Learning rate selection and loss scaling during continual pre-training.
(a) Training cross entropy across candidate learning rates under a fixed pilot
budget, with the selected learning rate minimising stable training loss. (b) Loss
trajectories for the Small and Large models as a function of consumed continual
pretraining tokens, with dashed lines denoting power-law fits.}
\label{fig:cpt-train-loss}
\end{figure}

Before committing to a full CPT run, we select the peak learning rate independently at each model scale using short pilot runs that hold the remaining training configuration fixed. Figure~\ref{fig:cpt-train-loss} shows that learning rates at the upper extreme produce optimisation instability, whereas those at the lower extreme reduce cross entropy too slowly under the same token budget. The intermediate rate that achieves the lowest stable training loss within the pilot budget is then used for the full continual pre-training run.\par 

Each model is trained with sequences of \(8{,}192\) tokens and a global batch size of \(512\), corresponding to approximately \(4.2\) million tokens per optimisation step. Following LR warmup, we apply inverse square root decay, under which the LR at step \(t\) is proportional to \(t^{-1/2}\). Unlike schedules such as cosine decay, in which the LR is parametrised by the total training budget, inverse square root decay is independent of the final token budget. This allows the stable training phase to be extended without redefining the schedule or introducing a discontinuity in the LR. A final annealing stage promotes controlled convergence by linearly decreasing the LR to its minimum value over the last \(20\%\) of the continual pre-training budget. In early-stage experiments, we also investigated an extended long context Mid-Training stage. Although this additional training recovered long context capabilities lost during continual pre-training, model merging followed by long context DPO and agentic RL proved largely sufficient, and the dedicated extension was therefore omitted from the final training recipe.

\subsubsection*{Model Merging}
\label{ssec:model-merging}

Model merging is a central component of our mid-training procedure when starting from a model that has already undergone post training. Although merging can mitigate forgetting during continued training more generally~\citep{alexandrov2024mitigating,wortsman2022model}, it is particularly important in this setting since mid-training disproportionately erodes capabilities acquired during post training. \par

During early development, we sought a compute efficient procedure that balanced domain adaptation against forgetting under a fixed budget for continual pre- and post-training. Alongside model merging, we evaluated several regularisation techniques using the same curated CPT mixture, including low rank adaptation, lower continual pre-training learning rates, and shorter training schedules. Each of these regularisation techniques reduced the loss of general capability but weakened domain adaptation to a similar extent. After post-training, none of the resulting checkpoints improved on either axis relative to a model post-trained without continual pre-training.\par

Consistent with prior work on domain adaptation of post trained models~\citep{labrak2024biomistral,siriwardhana2024domain}, these results led us to merge the continually pretrained checkpoint with the checkpoint preceding mid-training, which we found necessary to restore the lost capabilities. Additional post-training on generic data provides another route to recovery, but its computational cost grows with the severity of forgetting and represents overhead relative to the domain adaptation objective, potentially placing it beyond the budget of many institutions.\par

\paragraph{Model Merging Benefits from Stronger Continual Pre-Training.} 
Model merging, in its simplest form, linearly interpolates in weight space between the continually pre-trained checkpoint and the model initialisation preceding continual pre-training, with the relative contribution of each checkpoint governed by a single merging coefficient~\citep{ilharco2022editing,wortsman2022model}. 
We found that an appropriate merge coefficient can recover most of the general capability lost during continual pre-training. More surprisingly, after post-training under the same budget, a merged checkpoint derived from sufficiently strong continual pre-training outperformed its counterpart trained without continual pre-training on both the domain and general capability axes.\par

\begin{figure}[htbp]
\centering
\includegraphics[width=\linewidth]{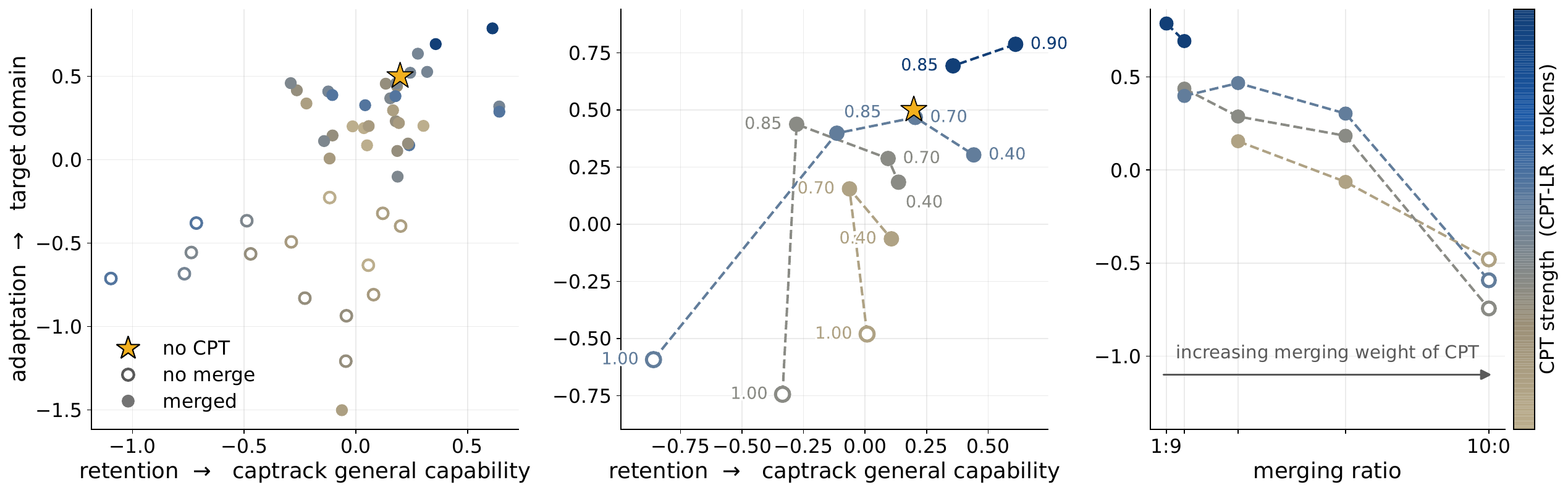}
\caption{Effect of continual pre-training strength and merge coefficient on target domain performance and general capability retention under a fixed post-training budget. Colour denotes CPT strength, while each numerical annotation gives the merge coefficient assigned to the CPT checkpoint. Axes report mean standardised performance over the target and general domains.}
\label{fig:merge-by-strength}
\end{figure}

To characterise how continual pre-training, merging and post-training jointly
determine the trade-off between domain adaptation and forgetting, we sweep the continual pre-training learning rate over $\{2\!\times\!10^{-6},
5\!\times\!10^{-6}, 1\!\times\!10^{-5}, 2\!\times\!10^{-5}\}$ and the token budget over $\{20, 40,
100, 200\}$B tokens, and interpolate each resulting checkpoint towards the model preceding CPT at merge ratios from $1{:}9$ to $10{:}0$, holding the post-training budget
fixed. We define \emph{CPT strength} as the product of the learning rate and
the token budget, a single index of how far training displaces the weights from
their initialisation. Figure~\ref{fig:merge-by-strength} shows that aggressive continual pre-training followed by a conservative merge ratio, by which we mean one that retains less of the continually pre-trained weights, consistently outperforms the regularised alternatives on both axes at once, and that the Pareto front is occupied by high-strength merged checkpoints rather than by restrained ones with a lower CPT LR or a smaller training budget.\par

\subsection{Direct Preference Optimisation}
\label{ssec:dpo}

Post-training begins from the merged CPT checkpoint, converting
newly injected knowledge into desired, task-specific behaviour.
On-policy RL is the most direct instrument for this
conversion, but it is expensive, and several of the behaviours required
for subsequent exploration -- e.g., model identity, output-format
conventions, and related inductive biases -- are specified more
economically by off-policy preference data than they are acquired
through fully on-policy search. We therefore precede reinforcement
learning with DPO \citep{rafailov2023direct}, which warm-starts the policy at substantially lower cost. \par

A common alternative for warm-starting RL is SFT; however, we found that naive SFT in our setting can rapidly
degrade generic capability, even when the supervised mixture is broad and includes generic replay data, while DPO tends to be much more robust to catastrophic forgetting \citep{thede2026captrack}. We forgo a standalone SFT stage for this reason, and instead
turn towards a DPO training objective with an SFT intervention to still
account for the instruction-following and output-formatting
supervision when needed for a particular task collection. The supervision
that such a stage would otherwise provide is already present in the
CPT mixture, both as a small instruction-style replay component and as synthetic rephrasings of proprietary documents into task-shaped and instruction-following formats (Section~\ref{ssec:cpt}), motivated by prior findings that absorbing such supervision under a pre-training objective, rather than as an isolated post-training SFT pass, reduces distribution mismatch with later stages and is a more effective way of using SFT-like data on an already-aligned model~\citep{cheng2024instruction,baek2026finetuner}.\par

We apply DPO in two sequential stages (see Figure~\ref{fig:training_stages}), separating broad alignment from
long-context, agentic specialisation:\label{sssec:dpo-stages}
\begin{itemize}
    \item \textsc{DPO Stage~1 (broad preference alignment):} Performs preference alignment over a large heterogeneous collection mixture at moderate context length. We employ multi-fidelity Bayesian optimisation to learn data mixture proportions that improve both performance and robustness (Section~\ref{sssec:dpo-mixing}).

    \item \textsc{DPO Stage~2 (agentic specialisation):} Initialises from the Stage~1 checkpoint and specialises the policy for agentic behaviour on a targeted set of synthetic Deep Research preference collections. This stage operates at extended context length (up to 64k tokens) to support long-horizon research and tool-use trajectories.
\end{itemize}

The remainder of this section covers, in order, the shared training
algorithm in Section~\ref{sssec:dpo-algorithm}, DPO Stage~1 preference data
and data mixture optimisation in Sections~\ref{sssec:dpo-stage1-data}
and~\ref{sssec:dpo-mixing}, DPO Stage~2 agentic preference data and
long-context specialisation in Sections~\ref{ssec:agentic-data}
and~\ref{sssec:dpo-stage2-training}, and, finally, how the resulting DPO
checkpoint provides a stronger initialisation for subsequent
reinforcement learning in Section~\ref{sssec:dpo-warmup-results}.

\subsubsection{Algorithmic Overview}
\label{sssec:dpo-algorithm}
The DPO training objective used throughout both stages is
\begin{equation}
  \mathcal{L}(\theta)
  =
  \mathbb{E}_{(x,y_{c,w},y_{c,l},c)}
  \left[
    -\log\sigma\!\left(
      \beta\left[
        \bar{h}_\theta(x,y_{c,w})-\bar{h}_\theta(x,y_{c,l})
      \right]
    \right)
    +
    \alpha_c\,\mathcal{L}_{\mathrm{SFT}}(x,y_{c,w})
  \right].
  \label{eq:dpo-combined}
\end{equation}
Here, $c$ indexes a training collection, or dataset, and $y_{c,w}$ and $y_{c,l}$ are its chosen and rejected responses for prompt $x$. Equation~\eqref{eq:dpo-combined} differs from standard DPO in that we employ the length-normalised implicit reward $\bar{h}_\theta(x,y)$ defined in Equation~\eqref{eq:dpo-ln-reward}, and augment the preference loss with an SFT term $\mathcal{L}_{\mathrm{SFT}}$ weighted by the collection specific coefficient $\alpha_c$.

\paragraph{Collection-Dependent SFT Anchoring.}
When tasks are often open-ended, what counts as a target is
task- and preference-dependent, and there is often no universal, single
gold response for a query. We introduce
a per-collection coefficient $\alpha_c$ that controls how much absolute imitation is
warranted on the chosen response. Setting $\alpha_c$ per collection
lets us encode domain preferences and requirements using expert
knowledge of each task. At one end of the scale, collections with
high-quality, strict gold-factuality and other format-specific
outputs -- receive a large $\alpha_c$. Where preference is instead
driven by more abstract aspects such as style, malicious query refusal, and
helpfulness, the gold response need not be unique. 
In these cases $\alpha_c$ is small or zero, and the preference
gap still presents a meaningful optimisation
direction~\citep{geng2025delta}. DPO Stage~1 uses the full spread of these
weights; DPO Stage~2, whose collections are all curated imitation targets
of comparable status, applies a single modest weight uniformly.

In accordance with prior work~\citep{grattafiori2024llama}, we find that
the SFT term also stabilises DPO training. Since standalone DPO optimises only the relative margin between the chosen and rejected responses, their
likelihoods may both decrease while the preference objective improves,
provided that the rejected response decreases more rapidly. The
auxiliary SFT term directly anchors the chosen response and thereby
counters this likelihood displacement.

\paragraph{Length Normalised Implicit Reward.}
Standard DPO defines the implicit reward as
$h_\theta(x,y)=\log\pi_\theta(y\mid x)-\log\pi_{\mathrm{ref}}(y\mid x)$,
which sums token level log ratios across the response. Its magnitude
therefore varies with sequence length, allowing length to influence the
preference margin independently of response quality. We reduce this
dependence by using the mean token log ratio, following
SimPO~\citep{meng2024simpo} and T\"ulu~3~\citep{lambert2024tulu}.
\begin{equation}
  \bar{h}_\theta(x,y)
  \;=\;
  \frac{1}{\lvert y\rvert}
  \sum_{t=1}^{\lvert y\rvert}
  \log
  \frac{\pi_\theta(y_t\mid x,y_{<t})}
       {\pi_{\mathrm{ref}}(y_t\mid x,y_{<t})}.
  \label{eq:dpo-ln-reward}
\end{equation}
We apply the same token averaging to the SFT term $\mathcal{L}_{\mathrm{SFT}}$, preventing longer responses from
contributing disproportionately due to response length alone.



\subsubsection{DPO Stage 1 -- Content \& Rehearsal Preference Data Generation}
\label{sssec:dpo-stage1-data}



\subsubsection*{Rehearsal DPO Preference Data} 
\label{sssec:pool-dpo}

The purpose of Rehearsal DPO is to strengthen previously acquired knowledge and abilities that were weakened over the course of model re-alignment and mid-training.
We base the Rehearsal DPO corpus on Math\footnote{https://huggingface.co/datasets/nvidia/Nemotron-SFT-Math-v3} \citep{du2025nemotron}, Safety\footnote{https://huggingface.co/datasets/nvidia/Nemotron-SFT-Safety-v1}, Instruction~Fine-tuning, Science, and Multilingual\footnote{https://huggingface.co/datasets/nvidia/Nemotron-Post-Training-Dataset-v2} subsets of the Nemotron Super dataset family
and augment them with new synthetic responses and reasoning traces, generated by permissively-licensed, open-weights models.

The DPO pair synthesis follows a process inspired by \citep{lambert2024tulu} and \citep{zhang2025ccr}. First, we define a pool of generators powered by selected open-weight LLMs. For each input query, each generator in the pool creates one candidate response. These responses are subsequently pairwise evaluated via LLM judge model equipped with a set of domain-specific criteria and the original answer as a reference, and ranked by the number of times they were judged to be the better option in the direct one-to-one comparison. To avoid a judge's potential self-preference bias, the answer-generation pool never includes any model from the same family as the judge. Finally, the highest-ranking response in the pool is selected as the preferred answer, and its rejected counterpart is either the second-best candidate in the pool, or, to keep the training examples on policy, the response given by the model under training, if the latter happened to rank third or lower in the pool.

\subsubsection*{Content-Driven Synthetic Preference Data}
\label{ssec:content-synth}

A strong argument for SovereignAI is the comparative advantage of private data repositories not seen by general-purpose models (\ref{item:sovereignty-data}). Mid-Training absorbs that material as knowledge, but knowledge alone is insufficient for such data to be surfaced in the appropriate manner during practical use cases. This section describes a practical approach to further utilise private data beyond the Mid-Training stage. 

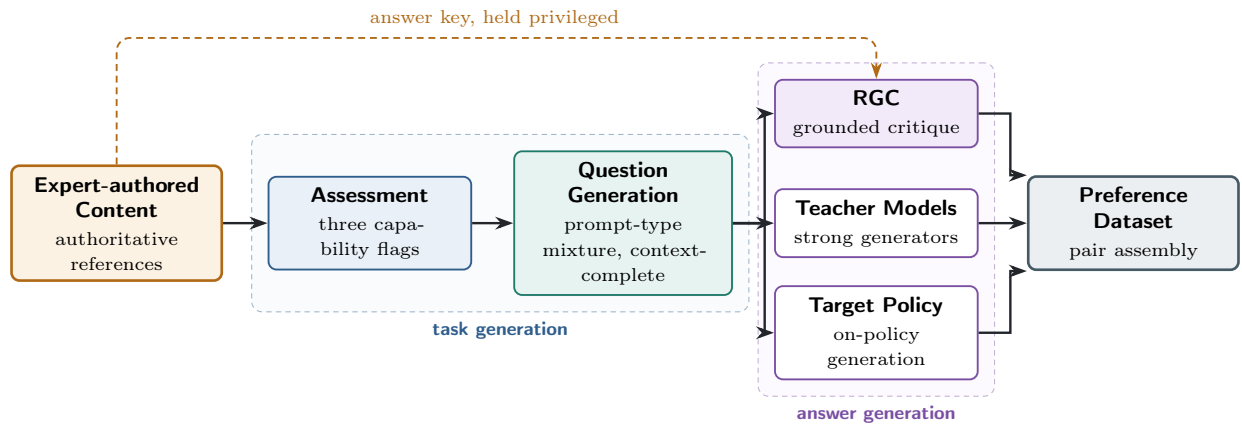
\begin{figure}[htbp]
\centering
\adjustbox{max width=\textwidth}{%
\begin{tikzpicture}[
    font           = \small,
    stage/.style   = {draw, line width=0.7pt, rounded corners=3pt, align=center,
                      inner sep=4pt, drop shadow={opacity=0.09, shadow xshift=0.5pt,
                      shadow yshift=-0.5pt}},
    content/.style = {stage, draw=cTool, line width=1.0pt, fill=cToolBg,
                      text width=25mm, font=\footnotesize},
    gate/.style    = {stage, draw=cPlan, fill=cPlanBg, text width=24mm, font=\footnotesize},
    gen/.style     = {stage, draw=cRes,  fill=cResBg,  text width=26mm, font=\footnotesize},
    roll/.style    = {stage, draw=cComp, fill=white,   text width=24mm, font=\footnotesize},
    rollg/.style   = {stage, draw=cComp, fill=cCompBg, text width=24mm, font=\footnotesize},
    asm/.style     = {stage, draw=cRep,  line width=1.0pt, fill=cRepBg,
                      text width=25mm, font=\footnotesize},
    flow/.style    = {-{Stealth[length=2.4mm,width=1.7mm]}, line width=0.9pt, draw=cInk},
    sup/.style     = {-{Stealth[length=2.2mm,width=1.5mm]}, line width=0.7pt,
                      draw=cTool, dash pattern=on 2.6pt off 1.8pt, rounded corners=5pt},
    lbl/.style     = {font=\scriptsize, text=cMute, align=center, inner sep=2pt}
  ]

  \node[content] (doc) at (0, 0)
       {\textbf{Expert-authored Content}\\[1pt]
        {\scriptsize authoritative references}};

  \node[gate] (assess) at (3.35, 0)
       {\textbf{Assessment}\\[1pt]
        {\scriptsize three capability flags}};
  \node[gen]  (qgen)   at (6.70, 0)
       {\textbf{Question Generation}\\[1pt]
        {\scriptsize prompt-type mixture, context-complete}};

  \begin{scope}[on background layer]
    \node[draw=cPlan!45, dash pattern=on 2pt off 1.6pt, rounded corners=4pt,
          fill=cPlanBg!25, inner sep=6pt, fit=(assess)(qgen),
          label={[font=\scriptsize, text=cPlan]below:%
                 {\textbf{task generation}}}] (taskgen) {};
  \end{scope}

  \node[rollg] (rgc)   at (10.05,  1.45) {\textbf{RGC}\\[1pt]
                                          {\scriptsize grounded critique}};
  \node[roll]  (teach) at (10.05,  0.00) {\textbf{Teacher Models}\\[1pt]
                                          {\scriptsize strong generators}};
  \node[roll]  (pol)   at (10.05, -1.45) {\textbf{Target Policy}\\[1pt]
                                          {\scriptsize on-policy generation}};

  \begin{scope}[on background layer]
    \node[draw=cComp!45, dash pattern=on 2pt off 1.6pt, rounded corners=4pt,
          fill=cCompBg!25, inner sep=6pt, fit=(rgc)(teach)(pol),
          label={[font=\scriptsize, text=cComp]below:%
                 {\textbf{answer generation}}}] (ansgen) {};
  \end{scope}

  \node[asm] (asm) at (13.45, 0) {\textbf{Preference Dataset}\\[1pt]
                                  {\scriptsize pair assembly}};

  \draw[flow] (doc)    -- (assess);
  \draw[flow] (assess) -- (qgen);
  \draw[flow] (qgen)   -- (teach);
  \draw[flow] (qgen.east) -- ++(0.42,0) |- (rgc.west);
  \draw[flow] (qgen.east) -- ++(0.42,0) |- (pol.west);
  \draw[flow] (rgc.east)   -| ++(0.42,0) |- (asm.north west);
  \draw[flow] (teach.east) -- (asm.west);
  \draw[flow] (pol.east)   -| ++(0.42,0) |- (asm.south west);

  \draw[sup] (doc.north) -- (0, 2.45) -- (10.05, 2.45) -- (rgc.north);
  \node[lbl, text=cTool, anchor=south] at (5.0, 2.52)
       {answer key, held privileged};

\end{tikzpicture}}
\caption{Expert-authored content as the supervision signal. It decides which documents can
carry a training task and which task to generate. Answers to those questions are then
generated by three routes: RGC, which uses the source content as a reference
inside its critique stage, and teacher models and the target policy, which do not.
The resulting candidates are assembled into preference pairs.}
\label{fig:content-supervision}
\end{figure}

For large document corpora, we reverse-engineer queries for which a document implicitly provides an authoritative answer. If appropriate questions are recoverable, the document that suggested the questions also serves as their answer key, and the only thing left to synthesise is the reasoning
connecting the two. In this sense, the content supervises every stage
(Figure~\ref{fig:content-supervision}): it decides which tasks it can
support, it generates the questions, and it guides the reasoning that
answers them. No practitioner is in the loop at generation time, yet every
target traces back to material a practitioner wrote.

\paragraph{Content Assessment.}
Not every passage of content can carry a training task, and the deciding
factor is rarely quality of writing, which is uniformly high. What varies is
whether a passage is substantive enough to be worth asking about. Each document
is therefore assessed by a judge and probed for three separate capabilities
(Table~\ref{tab:task-assessment}). A passage passes only if it contains substance of interest
rather than strings of citations, and is challenging enough that training on it
would meaningfully improve the model. Brief, superficial, or purely
introductory material is rejected.

\begin{table}[htbp]
\centering
\small
\begin{tabular}{@{}p{0.26\linewidth}p{0.64\linewidth}@{}}
\toprule
\textbf{Capability} & \textbf{What the assessor looks for} \\
\midrule
question-answer pairs &
factual content that converts into clear, definitive questions; answers
that are specific and authoritative rather than vague; enough detail to support
several distinct pairs \\[2pt]
reasoning traces &
step-by-step reasoning with explicit connectives such as ``because'',
``therefore'', ``since'' or ``given that''; clear cause and effect; identifiable
steps running from premises to conclusion \\[2pt]
concept definitions &
detailed explanations of concepts, with context and examples; specific rather
than general; self-contained enough to be understood without the surrounding
document \\
\bottomrule
\end{tabular}
\caption{The three capability criteria applied to every document, each returned
as an independent true/false judgement on top of a shared quality bar. The
datasets described here keep documents that satisfy all required criteria.}
\label{tab:task-assessment}
\end{table}

\paragraph{Task Generation.}
Questions are generated one document at a time, controlling: what kind of question gets asked, and whether it is self-contained.

To get a diverse set of questions we sample one of nine prompt types per document
from a weighted mixture (Table~\ref{tab:question-prompts}). Six vary the
cognitive work demanded, from general reasoning over the whole document to
counterfactuals, analogies, claim verification, quantitative analysis, and causal
or temporal chains. Three vary the professional stance instead.

The model that will answer these questions never sees the document they came from, so each question has to
carry everything needed to answer it: any authority
the question turns on quoted inside the question, all necessary facts present as
posed. Generated questions are then filtered on the same standard.

\begin{table}[htbp]
\centering
\small
\begin{tabular}{@{}llr@{}}
\toprule
\textbf{Prompt Type} & \textbf{What it asks for} & \textbf{Weight} \\
\midrule
general reasoning      & reasoning across the whole document      & 30\% \\
counterfactual         & how the analysis changes if facts change & 10\% \\
analogy                & reasoning by comparison to a related case & 10\% \\
claim verification     & whether a stated claim holds             & 10\% \\
quantitative           & numerical or computational analysis      & 10\% \\
temporal / causal      & sequence and cause-and-effect chains     & 10\% \\
stance, judge          & record building, ambiguities, scope limits & 10\% \\
stance, lawyer         & doctrine, compliance, client consequences & 5\% \\
stance, layperson      & the same material in plain language      & 5\% \\
\bottomrule
\end{tabular}
\caption{Question generation prompt types and their default sampling weights. One
prompt type is drawn per document, so the mixture controls the distribution of
task types independently of the topical distribution of the underlying content.}
\label{tab:question-prompts}
\end{table}

\paragraph{Answer Generation.} Task generation yields questions without targets. Each question, however, carries
the document it was derived from, and that document is an authoritative answer to
it by construction: the question was written to be answerable from that passage.
Questions and answer keys therefore come paired, at no annotation cost.

Candidate answers are produced by three routes (Figure~\ref{fig:content-supervision}),
which differ in whether the source content participates. Teacher models and the
target policy answer from the question alone. Only Reasoning with Grounded Critique (see below) uses the document, and only inside its critique stage. All three routes feed the pool from which preference pairs are assembled (Section~\ref{ssec:dpo}).

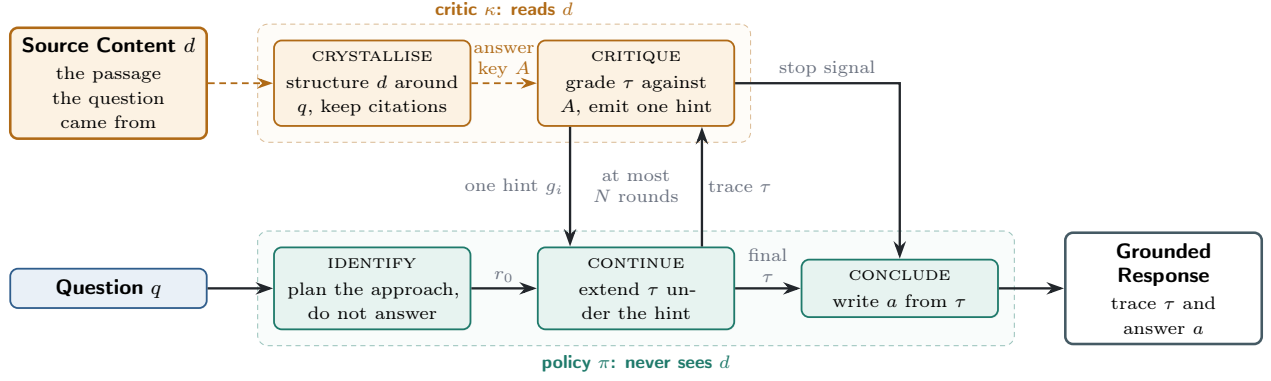
\begin{figure}[htbp]
\centering
\adjustbox{max width=\textwidth}{%
\begin{tikzpicture}[
    font           = \small,
    stage/.style   = {draw, line width=0.7pt, rounded corners=3pt, align=center,
                      inner sep=4pt, drop shadow={opacity=0.09, shadow xshift=0.5pt,
                      shadow yshift=-0.5pt}},
    content/.style = {stage, draw=cTool, line width=1.0pt, fill=cToolBg,
                      text width=24mm, font=\footnotesize},
    critic/.style  = {stage, draw=cTool, fill=cToolBg, text width=24mm, font=\footnotesize},
    pol/.style     = {stage, draw=cRes,  fill=cResBg,  text width=24mm, font=\footnotesize},
    qnode/.style   = {stage, draw=cPlan, fill=cPlanBg, text width=24mm, font=\footnotesize},
    outblk/.style  = {stage, draw=cRep, line width=1.0pt, fill=white,
                      text width=24mm, font=\footnotesize},
    flow/.style    = {-{Stealth[length=2.4mm,width=1.7mm]}, line width=0.9pt, draw=cInk},
    sup/.style     = {-{Stealth[length=2.2mm,width=1.5mm]}, line width=0.7pt,
                      draw=cTool, dash pattern=on 2.6pt off 1.8pt, rounded corners=5pt},
    lbl/.style     = {font=\scriptsize, text=cMute, align=center, inner sep=2pt}
  ]

  \node[content] (doc)  at (0.00, 2.80)
       {\textbf{Source Content $d$}\\[1pt]
        {\scriptsize the passage the question came from}};
  \node[critic]  (crys) at (3.60, 2.80)
       {\textsc{crystallise}\\[1pt]
        {\scriptsize structure $d$ around $q$, keep citations}};
  \node[critic]  (crit) at (7.20, 2.80)
       {\textsc{critique}\\[1pt]
        {\scriptsize grade $\tau$ against $A$, emit one hint}};

  \begin{scope}[on background layer]
    \node[draw=cTool!45, dash pattern=on 2pt off 1.6pt, rounded corners=4pt,
          fill=cToolBg!25, inner sep=6pt, fit=(crys)(crit),
          label={[font=\scriptsize, text=cTool]above:%
                 {\textbf{critic $\kappa$: reads $d$}}}] (kbox) {};
  \end{scope}

  \node[qnode]  (q)    at ( 0.00, 0) {\textbf{Question $q$}};
  \node[pol]    (iden) at ( 3.60, 0)
       {\textsc{identify}\\[1pt]
        {\scriptsize plan the approach, do not answer}};
  \node[pol]    (cont) at ( 7.20, 0)
       {\textsc{continue}\\[1pt]
        {\scriptsize extend $\tau$ under the hint}};
  \node[pol]    (conc) at (10.80, 0)
       {\textsc{conclude}\\[1pt]
        {\scriptsize write $a$ from $\tau$}};
  \node[outblk] (out)  at (14.40, 0)
       {\textbf{Grounded Response}\\[1pt]
        {\scriptsize trace $\tau$ and answer $a$}};

  \begin{scope}[on background layer]
    \node[draw=cRes!45, dash pattern=on 2pt off 1.6pt, rounded corners=4pt,
          fill=cResBg!25, inner sep=6pt, fit=(iden)(cont)(conc),
          label={[font=\scriptsize, text=cRes]below:%
                 {\textbf{policy $\pi$: never sees $d$}}}] (pbox) {};
  \end{scope}

  \draw[sup] (doc)  -- (crys);
  \draw[sup] (crys) -- node[lbl, text=cTool, above]
       {answer\\ key $A$} (crit);

  \draw[flow] (q)    -- (iden);
  \draw[flow] (iden) -- node[lbl, above] {$r_0$} (cont);
  \draw[flow] (cont) -- node[lbl, above] {final\\ $\tau$} (conc);
  \draw[flow] (conc) -- (out);

  \draw[flow] ([xshift=-9mm]crit.south) -- node[lbl, left]
       {one hint $g_i$} ([xshift=-9mm]cont.north);
  \draw[flow] ([xshift=9mm]cont.north) -- node[lbl, right]
       {trace $\tau$} ([xshift=9mm]crit.south);
  \node[lbl] at (7.20, 1.40) {at most\\ $N$ rounds};

  \draw[flow] (crit.east) -| node[lbl, above, pos=0.28]
       {stop signal} (conc.north);

\end{tikzpicture}}
\caption{Reasoning with Grounded Critique (Algorithm~\ref{alg:rgc}).}
\label{fig:rgc}
\end{figure}

\paragraph{Reasoning Trace Generation via Reasoning with Grounded Critique.}
\label{sssec:rgc}
Reasoning with Grounded Critique (RGC) uses the source content as an answer key. The document that generated the question also contains its answer -- only the reasoning that connects them is missing. RGC uses the grounding document in critic-mode to obtain guidance for the generation of the reasoning trace. The target model reasons from the question alone, under that guidance, and writes the final answer from its own trace. The output is a trace the policy itself generated, steered at every step by a document.
The procedure is described in Algorithm~\ref{alg:rgc} and illustrated in Figure~\ref{fig:rgc}. Data generation is divided into five stages:
\begin{itemize}
    \item \textsc{crystallise} reads the source content once and restates it as a structured
argument with its citations intact, organised around the question rather than as a
neutral summary.
    \item \textsc{identify} then starts the reasoning trace from the question alone. It is a planning step explicitly barred from answering: it identifies the sub-questions that will
have to be resolved and sketches a strategy.
    \item \textsc{critique} sees the question, the trace so far, and the crystallised document, and returns exactly one piece of guidance, tagged with its kind: stay on the current path
(\textsc{continue}), add the single most valuable missing insight
(\textsc{insight}), fix a specific flaw (\textsc{correct}), or stop because the
reasoning now suffices (\textsc{conclude}).
    \item \textsc{continue} receives the question, the trace so far and that guidance and extends the trace.
    \item \textsc{conclude} writes the
final answer from the question and accumulated trace alone.
\end{itemize}

The process alternates between \textsc{critique} and \textsc{continue} until a stop signal or exhaustion of the iteration budget ends the loop. Every generative stage (\textsc{identify}, \textsc{continue}, \textsc{conclude}) runs the target policy, so traces stay close to the distribution the model already produces rather than importing a teacher's voice. The loop spends several rounds per example, which is test-time compute paid once at data-creation time instead of at every inference.

The grounding therefore acts as an additional signal during construction of the
trace.

\begin{algorithm}[htbp]
\DontPrintSemicolon
\SetKwInOut{Input}{Input}
\SetKwInOut{Output}{Output}
\SetKwFunction{Crys}{Crystallise}
\SetKwFunction{Iden}{Identify}
\SetKwFunction{Crit}{Critique}
\SetKwFunction{Cont}{Continue}
\SetKwFunction{Conc}{Conclude}
\Input{question $q$, source content $d$, budget $N$, policy $\pi$, critic $\kappa$}
\Output{reasoning trace $\tau$, final answer $a$}
\BlankLine
$A \leftarrow$ \Crys{$q$, $d$} \tcp*[h]{$\star\ \kappa$: structure $d$ around $q$, keep citations}
$r_0 \leftarrow$ \Iden{$q$} \tcp*[h]{$\pi$: sees $q$ only}
$\tau \leftarrow \langle r_0 \rangle$\;
\For{$i \leftarrow 1$ \KwTo $N$}{
  $(g_i, c_i) \leftarrow$ \Crit{$q$, $\tau$, $A$} \tcp*[h]{$\star\ \kappa$: $g_i{=}$hint, $c_i{=}$its tag}
  \lIf{$c_i = \textsc{conclude}$}{\KwSty{break}}
  $r_i \leftarrow$ \Cont{$q$, $\tau$, $g_i$} \tcp*[h]{$\pi$: sees $g_i$, not $A$ or $d$}
  $\tau \leftarrow \tau \mathbin{\|} \langle r_i \rangle$ \tcp*[h]{append, never rewrite}
}
$a \leftarrow$ \Conc{$q$, $\tau$} \tcp*[h]{$\pi$: sees $q$ and $\tau$ only}
\Return $(\tau, a)$\;
\caption{Reasoning with Grounded Critique. At each round the critic returns a
hint $g_i$ together with its tag $c_i \in \{\textsc{continue}, \textsc{insight},
\textsc{correct}, \textsc{conclude}\}$. The lines marked $\star$ are the only two at
which the source document $d$ is read, both by the critic $\kappa$.}
\label{alg:rgc}
\end{algorithm}

\paragraph{Preference Pair Assembly.} Each question yields a pool of candidate rollouts drawn from the answer
generation routes described above (Figure~\ref{fig:content-supervision}). A judge then compares these candidates against one another to select a chosen and a rejected response, and critically, this comparison is made with privileged access to the expert-authored source content, so the comparison is anchored to what the source content actually supports rather than to surface qualities of the writing. This ensures high quality, diverse preference pairs: grounding in the source content lets the judge catch a strong-sounding but subtly incorrect candidate and reject it in favour of an answer faithful to the document. At the same time, because the candidate pool draws on multiple answer generation routes rather than a single one, the chosen and rejected responses in a given pair need not come from the same route, keeping the dataset diverse across reasoning styles and generation strategies.

\subsubsection*{Ontology-Driven Synthetic Preference Data}
\paragraph{Domain Ontologies.} The previous section describes how we derive supervision from content by recovering the questions a document already answers. Inspired by \citep{song2026knowledge}, we take advantage of a second route that opens whenever the content implicitly adheres to an
\emph{ontology}: a specification of the key entity types and the relations among
them. In specialised domains such a structure almost always exists and is almost
never written down, precisely because the practitioners writing the document
already share it; yet it is what the prose is organised around. A clinical case
report, for instance, is organised around entities such as symptoms, test
results, comorbidities, diagnosis, treatment and outcome, and around the
relations that connect them: the diagnosis together with the patient's
comorbidities guides the treatment, which itself drives the outcome. No sentence
in the report states this schema, but every sentence is placed within it.
Consequently, the document admits a faithful representation as a knowledge graph
whose schema is the ontology itself. Two properties follow. First, tasks that are
meaningful to practitioners can typically be expressed in terms of the ontology,
so that inputs and targets may be obtained by traversing the graph rather than
authored by hand. Recovering the diagnosis from the symptoms and test results
attached to a case is one such task: it amounts to anchoring on a diagnosis,
collecting the symptom and test-result nodes adjacent to it, and withholding the
diagnosis itself. Second, because the schema is fixed, the same traversal can be
applied to every knowledge graph derived from documents in the corpus. Dataset
construction then becomes mechanical.

\paragraph{From Graph to Pairs.}
Let $\mathcal{O}$ be an ontology and $D$ a
document conforming to it. A \emph{traversal} $\mathcal{T}$ is a fixed procedure
that starts from a chosen \emph{anchor} node $a$, the entity a sample is built
around, such as a single diagnosis, and walks the graph to
select an input set $I$ and a target set $Y$, returning the pair $(I, Y)$.
Training samples are then obtained in three steps:

\begin{enumerate}
    \item \textsc{Instantiate:} Extract a knowledge graph $G_D$ from $D$ using $\mathcal{O}$ as schema, so that every node and edge carries a type drawn from it.
    \item \textsc{Collect:} Sample an anchor $a$ and evaluate $\mathcal{T}(G_D, a)$. Each anchor yields one sample, so a document contributes as many samples as it admits valid anchors.
    \item \textsc{Render:} Verbalise the task as an instruction, the entities of $I$ as the input, and the entities of $Y$ as the reference answer.
\end{enumerate}
Finally, when the objective is a preference dataset rather than a single
reference answer, negatives come for free: the entities of the target type that occur in the same graph but were not collected are topically close and, up to extraction error, wrong by construction.

\paragraph{Data construction.} We applied the pipeline to ontologies unique to reasoning in our target domains. We use an LLM to instantiate the knowledge graph for a given document and create preference data as described above. Example~\ref{ex:ontology} shows the process for a data point.

\tikzset{
  font           = \small,
  stage/.style   = {draw, line width=0.7pt, rounded corners=3pt, align=center,
                    inner sep=4pt, drop shadow={opacity=0.09, shadow xshift=0.5pt,
                    shadow yshift=-0.5pt}, text width=30mm, font=\scriptsize},
  issuenode/.style = {stage, draw=cRep,  line width=1.0pt, fill=cRep!8},
  factnode/.style  = {stage, draw=cPlan, fill=cPlanBg},
  rulenode/.style  = {stage, draw=cRes,  fill=cResBg},
  rulenodeout/.style = {stage, draw=cRes!75, fill=white},
  direct/.style  = {-{Stealth[length=2.4mm,width=1.7mm]}, line width=0.9pt, draw=cRes},
  indirect/.style= {-{Stealth[length=2.2mm,width=1.5mm]}, line width=0.7pt, draw=cRes,
                    dash pattern=on 2.6pt off 1.8pt, rounded corners=5pt},
  arises/.style  = {-{Stealth[length=2.2mm,width=1.5mm]}, line width=0.7pt, draw=cPlan},
  lbl/.style     = {font=\scriptsize, text=cMute, align=center, inner sep=2pt},
  tlbl/.style    = {font=\scriptsize\ttfamily, align=center, inner sep=2pt},
}

\begin{agexample}{Preference pair from \texttt{rule} traversal}
\centering
\adjustbox{max width=\linewidth}{%
\begin{tikzpicture}

  \node[rulenode] (rl2) at (0.00, 2.18)
       {qualified immunity: no violation of clearly established law, or
        objectively reasonable};
  \node[rulenode] (rl5) at (0.00, 0.34)
       {motion to dismiss: complaint must state a plausible claim};
  \node[rulenode] (rl1) at (0.00,-2.41)
       {absolute immunity: judicial-phase acts only, not investigatory
        functions};

  \begin{scope}[on background layer]
    \node[draw=cRes!55, line width=0.7pt, rounded corners=5pt,
          fill=cResBg!20, inner xsep=8pt, inner ysep=21pt, fit=(rl2)(rl5)(rl1),
          label={[tlbl, text=cRes]above:{Rule}},
          label={[lbl, text=cRes, rotate=90, anchor=south, inner sep=3pt]left:%
                 {\textbf{chosen set $Y$ = direct $\cup$ indirect}}}] (ybox) {};
    \node[draw=cRes!45, dash pattern=on 2pt off 1.6pt, rounded corners=4pt,
          fill=cResBg!35, inner sep=5pt, fit=(rl2)(rl5),
          label={[lbl, text=cRes, inner sep=1pt]above:%
                 {\textbf{direct (\textsc{address})}}}] (dbox) {};
    \node[draw=cRes!45, dash pattern=on 2pt off 1.6pt, rounded corners=4pt,
          fill=cResBg!35, inner sep=5pt, fit=(rl1),
          label={[lbl, text=cRes, inner sep=1pt]below:%
                 {\textbf{indirect (\textsc{apply})}}}] (ibox) {};
  \end{scope}

  \node[issuenode] (li2) at (5.30, 1.46)
       {\textbf{Anchor $\ell$}\\[1pt]
        {\scriptsize qualified immunity for alleged investigative acts}};
  \node[tlbl, text=cRep, above=1.5mm of li2] {Issue};

  \node[factnode] (mf2) at (5.30,-1.23)
       {prosecutors directed the police investigation before indictment};
  \node[factnode] (mf3) at (5.30,-3.19)
       {prosecutors impeded the cold-case investigation};

  \begin{scope}[on background layer]
    \node[draw=cPlan!45, dash pattern=on 2pt off 1.6pt, rounded corners=4pt,
          fill=cPlanBg!25, inner sep=6pt, fit=(mf2)(mf3),
          label={[tlbl, text=cPlan]below:{Fact}}] (fbox) {};
  \end{scope}

  \node[rulenodeout] (rl3) at (10.70, 0.50)
       {\emph{Monell}: municipal liability needs a policy or custom};
  \node[rulenodeout] (rl4) at (10.70,-1.85)
       {Eleventh Amendment immunity for official-capacity claims};

  \begin{scope}[on background layer]
    \node[draw=cMute!45, dash pattern=on 2pt off 1.6pt, rounded corners=4pt,
          fill=cMute!5, inner sep=6pt, fit=(rl3)(rl4),
          label={[tlbl, text=cRes]above:{Rule}},
          label={[lbl, align=center]below:%
                 {\textbf{rejected:}\\ reached by neither path}}] (rbox) {};
  \end{scope}

  \draw[direct] (rl2.east) to[out=0,in=155]
        node[lbl, text=cRes, above, pos=0.5] {address} (li2.north west);
  \draw[direct] (rl5.east) -- node[lbl, text=cRes, above, pos=0.45] {address} (li2.west);

  \draw[indirect] (rl1.east) to[out=0,in=200]
        node[lbl, text=cRes, above, pos=0.6] {apply} (mf2.west);
  \draw[indirect] (rl1.south east) to[out=-20,in=180]
        node[lbl, text=cRes, below, pos=0.78] {apply} (mf3.west);

  \draw[arises] (li2.south) -- node[lbl, text=cPlan, right] {arise} (mf2.north);
  \draw[arises] (li2.south east) to[out=-75,in=25]
        node[lbl, text=cPlan, right, pos=0.55] {arise} (mf3.east);
\end{tikzpicture}}

\smallskip
{\footnotesize The rule traversal on an extracted graph.
Node labels are abbreviated -- the pair appears verbatim below.\par}

\medskip
\raggedright
\footnotesize
\textbf{Input} -- the anchor $\ell$ and the facts it arises from
\begin{quote}
\texttt{instruction}: ``Analyse the provided facts and issue, and
identify the most important rules that are applicable to address the
issue.''

\smallskip
\texttt{facts}:
\begin{itemize}[leftmargin=1.2em,itemsep=1pt,topsep=1pt]
  \item Prosecutors allegedly directed police investigation prior to indictment,
        including ordering incomplete polygraph, directing DNA not be tested, and
        relying on non-credible witness
  \item Prosecutors allegedly impeded cold case investigation by coercing
        witnesses and threatening police officials
\end{itemize}

\smallskip
\texttt{issue}: Whether prosecutors are entitled to qualified immunity for
alleged investigative acts.
\end{quote}

\medskip
\textbf{\textcolor{red!55!black}{Rejected}} -- rules of the same opinion reached
by neither path
\begin{quote}\itshape
\begin{itemize}[leftmargin=1.2em,itemsep=1pt,topsep=1pt]
  \item A municipality may be held liable under \S~1983 when execution of a
        government policy or custom inflicts the injury, but not under
        respondeat superior.
  \item District attorneys acting in a quasi-judicial capacity represent the state
        and are entitled to Eleventh Amendment immunity for official capacity
        suits.
\end{itemize}
\end{quote}

\medskip
\textbf{\textcolor{green!40!black}{Chosen}} -- the union of the two paths
\begin{quote}\itshape
\begin{itemize}[leftmargin=1.2em,itemsep=1pt,topsep=1pt]
  \item \upshape[\textsc{direct}]\itshape\ Qualified immunity protects officials
        if their actions did not violate clearly established law or were
        objectively reasonable.
  \item \upshape[\textsc{direct}]\itshape\ To survive a motion to dismiss, a
        complaint must contain sufficient factual matter to state a plausible
        claim for relief.
  \item \upshape[\textsc{indirect}]\itshape\ Prosecutors are absolutely immune for
        acts intimately associated with the judicial phase of the criminal
        process, but not for administrative or investigatory functions unrelated
        to preparation for judicial proceedings.
\end{itemize}
\end{quote}

\medskip
\hrule
\smallskip
\textbf{Why these three, in this order.}
\begin{enumerate}[leftmargin=1.4em,itemsep=1pt,topsep=2pt]
  \item Absolute immunity reaches only judicial-phase acts -- not these, so
        \emph{qualified} immunity is what is in issue.
  \item Qualified immunity then turns on clearly established law or objective
        reasonableness.
  \item At the pleading stage, that question is asked of the complaint rather than
        of proven facts.
\end{enumerate}
\smallskip
$\rightarrow$ The opinion concludes: no qualified immunity at the motion to
dismiss stage.

\smallskip
\textbf{What the indirect path buys.} Rule 1 carries no \textsc{address} edge to
$\ell$; in the graph it addresses the neighbouring absolute-immunity issue. It
enters the target set only because the opinion \textsc{applies} it to both of
$\ell$'s facts. The direct path alone would drop the rule that frames the
question.

\smallskip
\textbf{Why the negatives are hard.} Both concern who else is answerable -- the
county under \emph{Monell}, the district attorneys in their official capacity --
not whether these prosecutors are shielded for this conduct.

\label{ex:ontology}
\end{agexample}

\subsubsection{DPO Stage 1 -- Data-Mixture Optimisation}
\label{sssec:dpo-mixing}

The proportions assigned to training collections can materially affect the outcome of post-training under a fixed compute budget~\citep{djuhera2026data,thakkar2024deep}. Selecting these proportions is a joint optimisation problem because the value of any collection depends on the weights assigned to the others, as well as on the base model and training horizon. As the number of collections grows, the candidate mixtures form an increasingly high-dimensional probability simplex, making manual search expensive and unlikely to capture these interactions.\par

This motivates learning the data-mixture. A line of work trains proxy models on sampled
mixtures and uses their measured quality to choose weights for the target model: RegMix~\citep{liu2025regmix} regresses validation loss on the mixture weights, while DoReMi~\citep{xie2023doremi} reweights collections directly on a proxy model through group distributionally robust optimisation. These approaches reduce dependence on manual tuning but leave two difficulties in our setting.
 
\begin{itemize}
  \item \textbf{Cost of training and evaluation.} Every observation costs one full round of training followed by evaluation across a wide and diverse range of downstream tasks, which for a post-trained model demands generation and model-based grading rather than the negative log-likelihood measurements that RegMix and DoReMi rely on for pre-trained models. A dense search at the target scale is therefore prohibitively expensive.
  \item \textbf{Model Dependent Optima.} The optimal mixture depends on both the capabilities and scale of the base model. Transferring a mixture selected on a small proxy model directly to the target model consequently assumes that mixture rankings remain stable across scale, which need not hold.
\end{itemize}

To address these challenges, we adapt ADMIRE-BayesOpt~\citep{chen2025admire} and formulate data mixture selection as a sequential multi-fidelity Bayesian optimisation problem. Let $\boldsymbol{\pi}=(\pi_1,\ldots,\pi_D)$ denote the proportions assigned to $D$ collections, where $\pi_d\geq0$ and $\sum_{d=1}^{D}\pi_d=1$, and let $m$ denote model scale. Each training and evaluation run provides a noisy observation of the quality associated with one mixture and model pair. A multi-fidelity Gaussian process (GP) serves as a surrogate over this joint space, predicting the quality of untested configurations, quantifying uncertainty in those predictions, and learning how evidence transfers across model scales. Optimisation proceeds sequentially by updating the surrogate after each batch of observations (data mixtures, model scale, and associated evaluation results) and using an acquisition function to select the next mixtures and model scales from their predicted quality, uncertainty, and experimental cost. This process naturally balances exploration of uncertain regions against exploitation of promising mixtures, while allocating inexpensive proxy runs when they remain informative and reserving target scale evaluations for refinement of the final optimum.

\paragraph{Optimisation Objective.}
For $Q$ in-house evaluation tasks, let $s_q(\boldsymbol{\pi},m)>0$ denote the
standardised score on task $q$ after training model $m$ with mixture
$\boldsymbol{\pi}$, shifted by a constant to remain positive across the
mixtures we evaluate. We optimise their geometric mean
\begin{equation}
\label{eq:datamix-geometric-objective}
f(\boldsymbol{\pi},m)=\left(\prod_{q=1}^{Q}s_q(\boldsymbol{\pi},m)\right)^{1/Q}.
\end{equation}
This objective assigns a larger marginal benefit to the same absolute improvement on a lagging task than on an already strong task. The preference is aligned with the role of mid-training (including DPO stages) in our training pipeline, which must provide a broadly capable initialisation for the subsequent RL stage rather than maximise retrospective performance on a narrow subset of evaluations. Since RL improvements depend on capabilities already present in the initial policy~\citep{yue2025rlvr,wang2025oneshot}, a mixture that leaves one area substantially behind can constrain later exploration even when its arithmetic mean is high.

\paragraph{Multi-Fidelity Optimisation.}
We conduct the sequential search with batched queries in three phases, using the smaller model as an inexpensive source of information and the target model as the high fidelity objective.

\begin{itemize}
  \item \textsc{Phase 1 -- Parallel Exploration.} The search is initialised with
    mixtures drawn from a Dirichlet distribution centred on a manually specified
    prior, with rejection sampling on cosine distance to maintain coverage of the
    simplex. Each mixture is trained and evaluated on the proxy model in parallel,
    and the resulting observations provide the initial fit of the GP surrogate at
    low cost.
  \item \textsc{Phase 2 -- Batched Multi-Fidelity Querying.} Each iteration fits
    the multi-fidelity GP to all observations collected so far, maximises the
    acquisition function jointly over the mixture and the model scale, and selects
    a batch of candidate experiments subject to the same diversity constraint. The
    selected configurations are trained and evaluated, their results are appended
    to the observation set, and the procedure repeats. The multi-fidelity GP and the Bayesian optimisation loop together balance exploration against exploitation, with transferability across model scale modelled explicitly through the cross-scale covariance, which makes the search over data mixtures more efficient.
  \item \textsc{Phase 3 -- Target-Scale Refinement.} We restrict the remaining queries to the target model and use the transferred GP surrogate to refine its model-specific optimum. This phase concentrates the high-fidelity budget on resolving the most promising and uncertain target scale mixtures, yielding the final mixture for each target model.
\end{itemize}

\begin{figure}[htbp]
\centering
\includegraphics[width=1.0\linewidth]{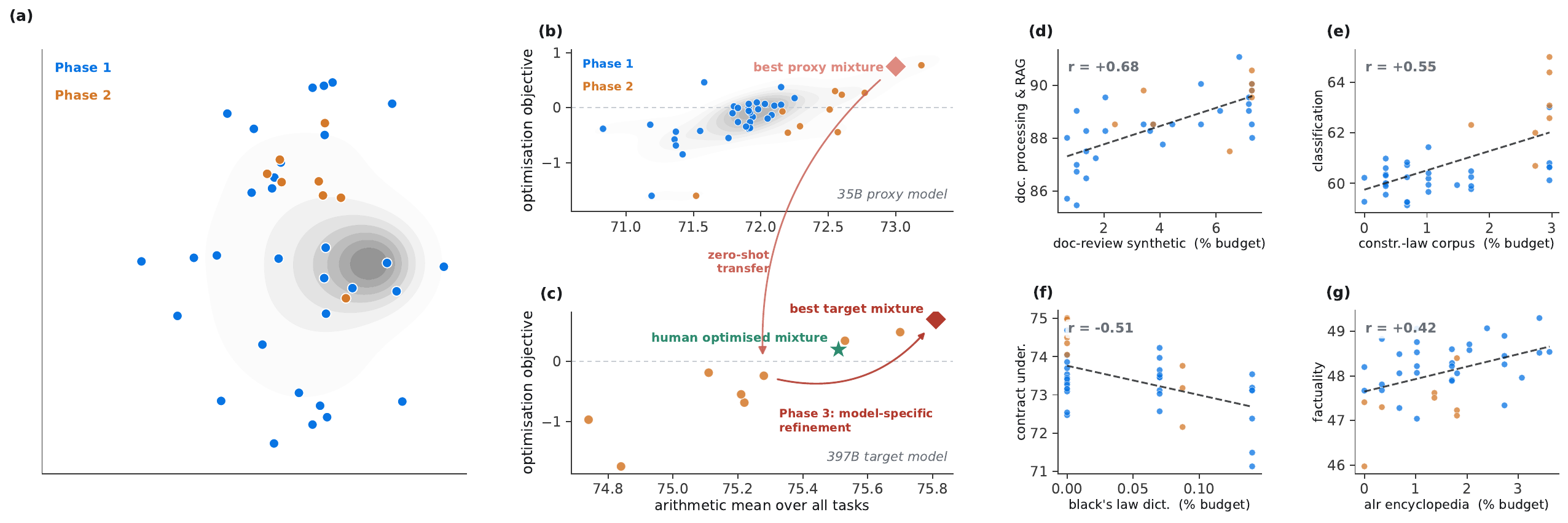}
\caption{Multi-fidelity data-mixture optimisation for the DPO Stage 1 training. Panel (a) shows the distribution of candidate data mixtures explored during the search, projected onto the first two principal axes of the mixture simplex. Panels (b) and (c) show the optimisation progress at the proxy and target model scales, plotting the geometric-mean optimisation objective against the arithmetic mean over all evaluation tasks. Arrows trace the best proxy mixture as it is transferred zero-shot to the target scale, where it is no longer optimal, and the subsequent Phase 3 refinement that reaches the best target mixture. Panels (d)--(g) show the association between individual evaluation scores and the budget share allocated to the collection most correlated with each task, with the Pearson correlation inset.}
\label{fig:datamix_optimisation}
\end{figure}

As shown in Figure~\ref{fig:datamix_optimisation}, relative to the human-optimised mixture, the final DPO Stage 1 training run on the learned best mixture increases the geometric mean objective from $0.20$ to $0.70$ and simultaneously achieves the highest arithmetic mean across all validation tasks, including tasks held out from the optimisation objective. We note that the best proxy mixture is suboptimal on the target model and would perform below the manually selected mixture if transferred without further search. The multi-fidelity procedure therefore reduces the cost of exploration without treating proxy performance as a scale invariant ranking.

\begin{figure}[htbp]
    \centering
    \includegraphics[width=\linewidth]{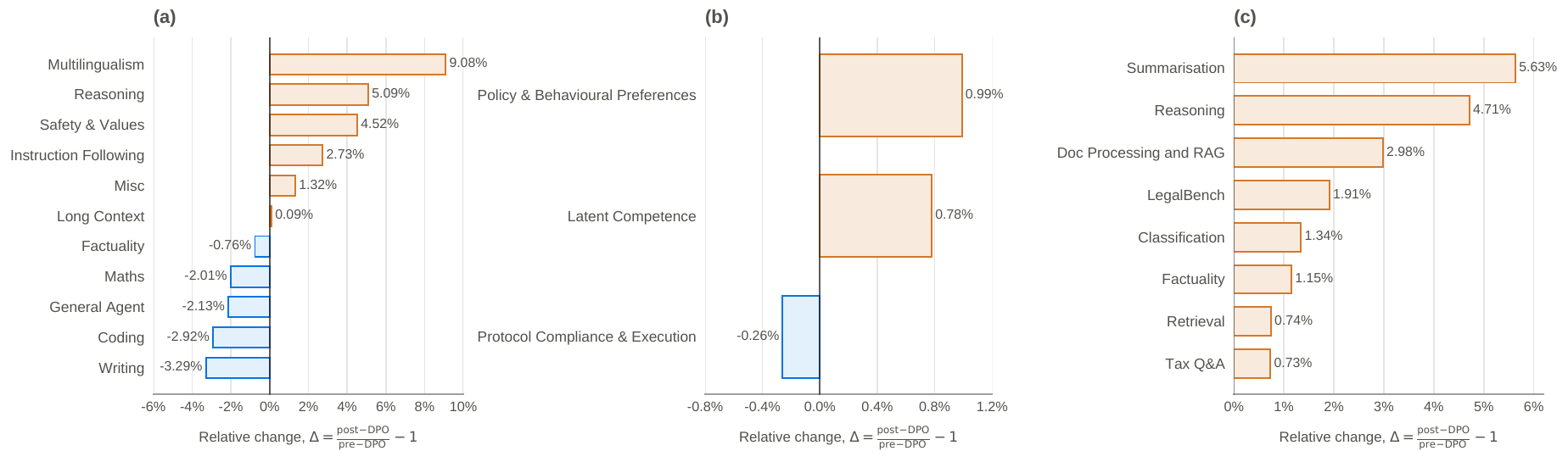}
    \caption{Relative Change in Performance after DPO Stage 1. (a) CapTrack -- Knowledge: largest gains in Multilingualism, Reasoning, and Safety \& Values, but small regressions in Coding and Writing. (b) CapTrack -- Ability: modest gains in Policy \& Behavioural Preferences with slight drop in Protocol Compliance \& Execution. (c) Target Domain: broad improvements led by Summarisation and Reasoning. Orange = improvement, blue = regression.}
    \label{fig:dpo-benchmark-differences}
\end{figure}

The effect of training the model on a selected DPO data mixture is shown in Figure~\ref{fig:dpo-benchmark-differences}. Alignment-adjacent capabilities improve while more technical, execution-heavy capabilities slightly regress. The largest gains appear on CapTrack's Knowledge axis, in Multilingualism (+9.08\%), Reasoning (+5.09\%), and Safety \& Values (+4.52\%), reflecting the direct targets of preference optimisation, with a small trade-off in Coding (-2.92\%) and Writing (-3.29\%). Changes on the Ability axis are smaller, with gains in Policy \& Behavioural Preferences (+0.99\%) and Latent Competence (+0.78\%) offset by a modest drop in Protocol Compliance \& Execution (-0.26\%). Encouragingly, this trade-off does not appear to damage downstream performance in the target domain: benchmarks show broad-based improvement, led by Summarisation (+5.63\%) and Reasoning (+4.71\%). Together, these results suggest that thoughtful DPO data selection can strengthen the model's alignment to a target-domain without meaningfully undermining general capabilities.

\subsubsection{DPO Stage 2 -- Agentic Preference Data Generation}
\label{ssec:agentic-data}

The agentic preference dataset supports an off-policy warm start for subsequent RL by familiarising the policy with the Deep Research harness and retrieval tools engineered to provide models with up-to-date information (\ref{item:sovereignty-data}). It also targets weaknesses observed in the current model, including failure to invoke available tools and fabrication of sources. Deep Research requires the model to plan an open-ended investigation, retrieve and interpret relevant evidence, and synthesise a grounded report. The task and harness are described in Section~\ref{subsec:AgenticDeepResearch}, while the corresponding evaluation protocol is presented in Section~\ref{sec:evaluation}. This section describes how completed trajectories are decomposed into pivotal decision points and converted into preference data.

Each preference example contains the off-policy prefix leading to a selected node and alternative chosen and rejected completions at that node. The node is a pivotal decision point because its output either determines the subsequent research state or contributes directly to final report quality. Pairing a lower-quality or behaviourally undesirable completion with a preferred alternative assigns supervision directly to the responsible decision. Depending on the selection criterion, this preference may encode a targeted behavioural correction, such as invoking retrieval tools before answering, or a relative quality improvement, such as stronger factual grounding during compaction and report synthesis.

\subsubsection*{Trajectory Collection and Pivotal Node Extraction}
\label{sssec:agentic-foundation}

We first collect complete trajectories from which prefixes and pivotal nodes can be extracted. For each item, human experts author a research \emph{query}, annotated with metadata such as domain and complexity, together with a two-tier \emph{rubric} specifying the essential and supplementary content of a competent response. The query initiates the harness execution, while the rubric provides the standard against which model response quality is assessed.

The queries are executed through \emph{fr-agents}, the Deep Research harness shown in Figure~\ref{fig:harness:flow}. Its graph contains four node types, with each node execution corresponding to one model call.

\begin{itemize}
  \item The \textsc{Planner} decomposes the research question into a plan.
  \item The \textsc{Researcher} decides at each turn whether and how to invoke a retrieval tool.
  \item The \textsc{Compaction} node condenses retrieved documents while preserving relevant evidence and citations.
  \item The \textsc{Reporter} synthesises the accumulated evidence into the final report.
\end{itemize}

Each node execution is logged with its trajectory prefix, completion, and available tool schemas. A single trajectory therefore yields multiple candidate decision points. The number of calls varies substantially by node because the researcher executes repeatedly whereas the reporter executes once. As summarised in Table~\ref{tab:agentic-nodes}, this asymmetry produces approximately two orders of magnitude more researcher examples than reporter examples.

\begin{table}[t]
\centering
\small
\begin{tabular}{@{}llll@{}}
\toprule
\textbf{Node} & \textbf{Calls / trajectory} & \textbf{Rejected-side limitation} & \textbf{Chosen-side improvement} \\
\midrule
Planner    & 1-2        & shallow or mis-scoped plan             & stronger planning behaviour \\
Researcher & $\sim$17 & answers without retrieval              & tool-grounded research \\
Compaction & many      & drops or distorts evidence              & factual evidence retention \\
Reporter   & 1         & unsupported or incomplete report       & grounded report synthesis \\
\bottomrule
\end{tabular}
\caption{Pivotal nodes extracted from Deep Research trajectories, including
their typical frequency, the limitation represented by the rejected completion,
and the behavioural or quality improvement represented by the chosen completion.}
\label{tab:agentic-nodes}
\end{table}

\paragraph{Counterfactual Completions from a Shared Prefix.}
A recorded node supplies an off-policy prefix containing the research state,
message history, and available tool schemas at a pivotal decision point. We
re-run the node under multiple open-weight models while holding this prefix fixed, producing
alternative completions that differ only in the decision being supervised.
Comparing completions from a shared prefix isolates the effect of that decision
without the trajectory drift that would arise from comparing independently
generated rollouts.

\subsubsection*{Preference Construction at Pivotal Nodes}
\label{sssec:agentic-dpo}

We convert the alternative completions at each pivotal node into preference
pairs using three complementary selection signals. Teacher identity transfers
behaviours that cannot yet be scored reliably, a programmatic tool-use rule
corrects an observed agentic failure, and factuality ranking selects the
higher-quality completion where groundedness can be measured. The rejected side
may therefore exhibit a specific undesirable behaviour or simply lower measured
quality, while the chosen side provides the corresponding preferred alternative.

Constructing these pairs offline permits stronger supervision than would be
practical within an online training loop. The chosen completion may be generated
by a stronger model with access to the SME rubric, and pair selection may use a
judge whose cost would be prohibitive as an online reward. The same prefix pool
can also be re-evaluated as the selection criteria improve. These preference
pairs are used in DPO Stage~2, which provides the off-policy warm start for RL.
A disjoint set of recorded prefixes at pivotal nodes is subsequently reused as
training queries for RL, allowing the same decision types to be optimised
on-policy without overlap with the DPO preference data.

\paragraph{Teacher-Guided Behavioural Repair.}
Teacher-guided pairs transfer planning and research behaviours for which a
reliable automatic quality metric is not available. For each shared prefix at a
\texttt{planner} or \texttt{researcher} node, a completion from a stronger
agentic model is selected as chosen and one from a weaker model as rejected,
provided both are valid. Relative model strength is established through
consistently higher performance on our Deep Research evaluation. This uses model
identity rather than a per-example judge to define the preference, capturing
behaviours such as decomposing the query into appropriately scoped research
tasks and continuing retrieval when the available evidence remains insufficient.

\begin{figure}[htbp]
\centering
\includegraphics[width=\linewidth]{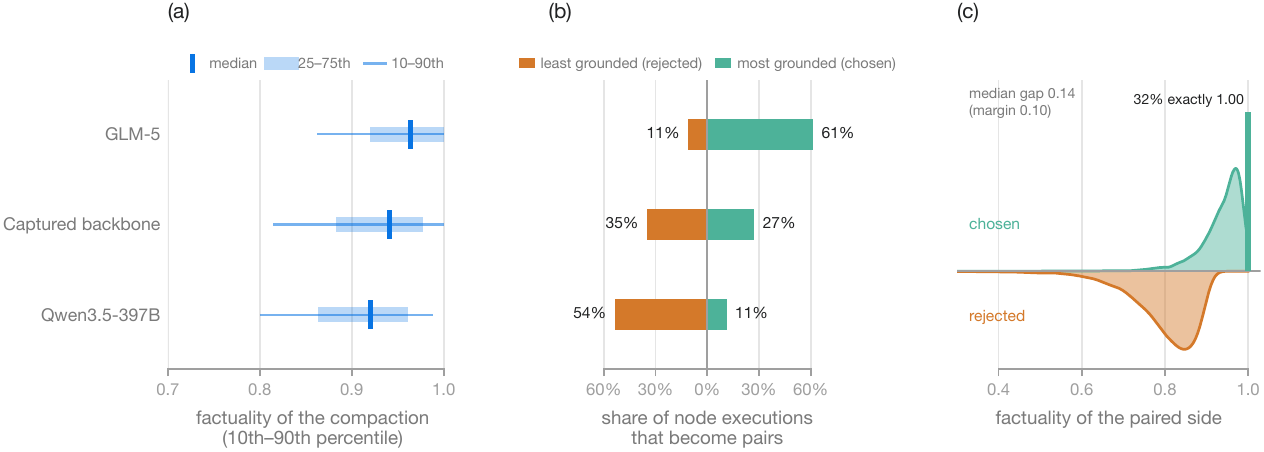}
\caption{Factuality-based selection for compaction preference pairs.
\textbf{(a)} Candidate factuality distributions overlap substantially, showing
that model identity does not reliably determine preference. \textbf{(b)}
Frequency with which each model supplies the chosen and rejected completions.
Every model appears on both sides of the preference pairs. \textbf{(c)}
Factuality distributions of the resulting pairs, with a median chosen-rejected
difference of 0.14.}
\label{fig:agentic-factuality-pairing}
\end{figure}

\paragraph{Factuality-Based Quality Selection.}
Factuality-based pairs improve the quality of \texttt{compaction} and
\texttt{reporter} outputs by preferring completions that remain grounded in the
evidence available in the shared prefix. Model identity is not a reliable
selection rule for these nodes because no candidate model is consistently the
most factual across examples, as shown in
Figure~\ref{fig:agentic-factuality-pairing}.

We score every candidate completion using the factuality judge described in
Section~\ref{ssec:research-evaluation}, which extracts claims and assesses their
support in the available source material. The highest-scoring completion forms
the chosen side and the lowest-scoring completion forms the rejected side.
Pairs with a score difference below 0.10 are removed to reduce ambiguous
preferences arising from judge variability. This criterion directly targets
the responsibilities of the two nodes. Compaction must preserve relevant
evidence without introducing unsupported content, while reporting must
synthesise the accumulated evidence into a grounded final response.

\begin{figure}[htbp]
\centering
\includegraphics[width=\linewidth]{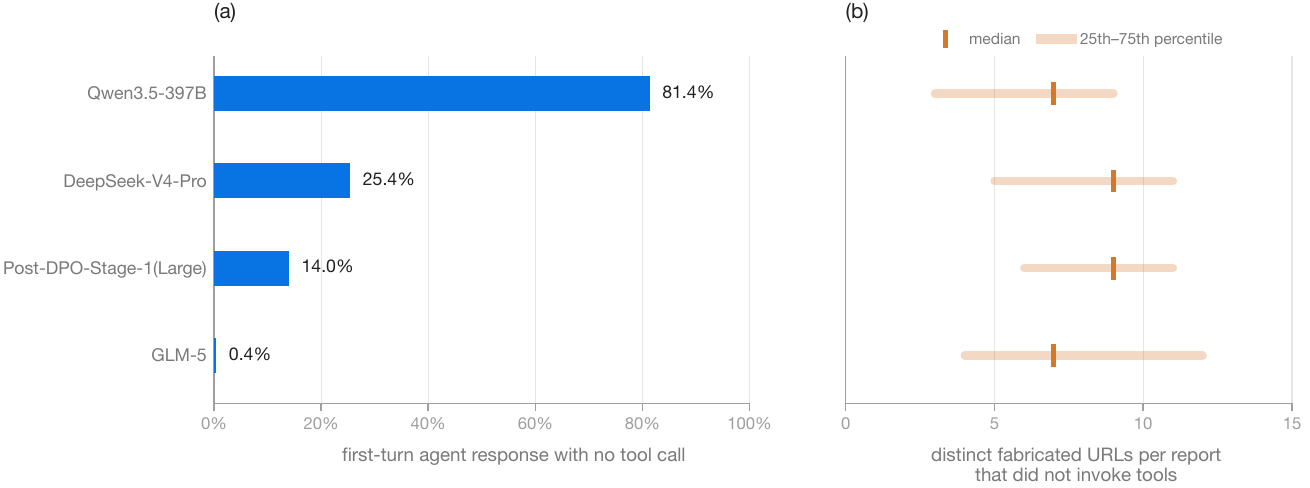}
\caption{Tool-use failure at the first \texttt{researcher} node in the Reuters
query pool, where the trajectory prefix contains no retrieved evidence.
\textbf{(a)} Share of first turns each backbone answered with zero tool calls
despite being handed the full retrieval tool set. The behaviour is present in
every backbone we ran, spanning two orders of magnitude.
\textbf{(b)} Distinct URLs cited in exactly those answers. Since the turn issued
no tool call, nothing was fetched, so every one of these URLs is invented; a
typical fabricated report carries seven of them. That contrast is what identifies
the failure as fabrication rather than a refusal or an empty turn, and it is the
reason we pair against it.}
\label{fig:agentic-no-toolcall}
\end{figure}

\paragraph{Programmatic Tool-Use Correction.}
Tool-use pairs correct a recurrent researcher failure in which the model answers
from parametric memory despite being given retrieval tools. This behaviour can
produce apparently researched responses containing fabricated citations because
no supporting document was retrieved. It occurs across all evaluated backbone
models, with substantially different frequencies, as shown in
Figure~\ref{fig:agentic-no-toolcall}.

Pair selection uses a programmatic predicate rather than a model judge. At a
shared prefix where retrieval is the required next action, the rejected
completion returns a substantive answer without invoking any tool, while the
chosen completion invokes at least one retrieval tool. The preference therefore
provides a targeted behavioural repair signal that teaches the policy to gather
evidence before composing an answer. It does not attempt to rank the substantive
quality of two completed reports.

\subsubsection*{Illustrative Preference Pairs}

The following examples show how a shared off-policy prefix is paired with
alternative rejected and chosen completions. The first expresses a targeted
behavioural correction, while the second encodes a measured difference in
factual quality.

\paragraph{Tool-Use Behaviour Correction.}
The shared prefix presents the same research state and retrieval tools to both
candidates. The rejected completion answers without gathering evidence, while
the chosen completion initiates retrieval.

\begin{agexample}{Tool-use preference pair at a \texttt{researcher} node}
\footnotesize
\textbf{Shared prefix}
\begin{quote}
``Retrieve direct quotes and policy positions from Nordic leaders and Reuters
analysis on the long-term viability of the Arctic Council as a governance body,
including any publicised Nordic proposals for reforming observer rules, altering
consensus mechanisms, or creating parallel frameworks to preserve governance
functions if the Council remains deadlocked at the ministerial level.''
\end{quote}

\medskip
\textbf{\textcolor{red!55!black}{Rejected}}
\begin{quote}
\upshape[\ldots]\itshape\
Following Russia's invasion of Ukraine in February 2022, the seven other Arctic
Council member states (Canada, Denmark, Finland, Iceland, Norway, Sweden, and the
United States) issued a joint statement on March 3, 2022, condemning the invasion
and announcing a pause in participation in Council meetings. [Joint statement by
the Arctic Council states on Russia's actions in
Ukraine](\hlurl{https://www.reuters.com/world/europe/arctic-council-states-pause-participation-meetings-due-ukraine-crisis-2022-03-03/})
This initial pause effectively halted all high-level governance functions, as the
Council operates on a consensus basis requiring the participation of all eight
members, including Russia. \upshape[\ldots]\itshape\ Norwegian Foreign Minister
Anniken Huitfeldt stated in March 2022 that while cooperation with Russia was
suspended, the ``Arctic Council remains the key forum for Arctic cooperation'' and
that the pause was intended to be temporary pending a change in circumstances.
[Norway's Huitfeldt on Arctic Council pause and future
cooperation](\hlurl{https://www.reuters.com/world/europe/norway-foreign-minister-says-arctic-council-cooperation-suspended-2022-03-04/})
\upshape\ [\ldots]
\end{quote}

\medskip
\textbf{\textcolor{green!40!black}{Chosen}}
\begin{quote}
\ttfamily\scriptsize\raggedright
Tool calls\\[3pt]
reuters\_search\\
\{\\
\hspace*{1.2em}"query": "Nordic Arctic Council governance reform consensus observer rules",\\
\hspace*{1.2em}"start\_date": "2022-02-01",\\
\hspace*{1.2em}"end\_date": "2026-06-11"\\
\}\\[3pt]
reuters\_search\\
\{\\
\hspace*{1.2em}"query": "Arctic Council ministerial deadlock Russia Nordic",\\
\hspace*{1.2em}"start\_date": "2022-02-01",\\
\hspace*{1.2em}"end\_date": "2026-06-11"\\
\}\\[3pt]
reuters\_search\\
\{\\
\hspace*{1.2em}"query": "Norway Sweden Finland Denmark Iceland Arctic Council future",\\
\hspace*{1.2em}"start\_date": "2022-02-01",\\
\hspace*{1.2em}"end\_date": "2026-06-11"\\
\}
\end{quote}
\end{agexample}

\paragraph{Factuality-Based Quality Selection.}
The shared prefix contains the same retrieved document for both candidates.
Both identify the central jurisdictional limitation, but the rejected completion
adds an unsupported threshold and a case citation absent from the source.
The highlighted spans show the resulting factuality failure.

\begin{agexample}{Factuality preference pair at a \texttt{compaction} node}
\footnotesize
\textbf{Shared prefix}
\begin{quote}
[\ldots] ``You are a research assistant analysing documents retrieved by
\texttt{document-fetch-tool} to answer a specific research question.
Overall research topic: Scottish sheriff court interdict enforcement research
plan. Current research step: lay representatives and McKenzie friends in Scottish
courts. Step description: research the rules governing party litigants' rights to
assistance from lay representatives, specifically (a) the common law right to a
McKenzie friend established in \emph{McKenzie v McKenzie}; [\ldots] (g) the
position under Scottish Civil Justice Council rules or practice directions.
[\ldots] Analyse the document(s) below and produce a targeted, context-aware
summary that:'' [\ldots] \emph{Practice Note (Sen Cts: McKenzie Friends: Civil and
Family Courts)}, 2010 WL 3017993 [\ldots]
\end{quote}

\medskip
\textbf{\textcolor{red!55!black}{Rejected}}: (factuality: 0.71)
\begin{quote}
\upshape[\ldots]\itshape\ Key claims: McKenzie friends have no right of audience
and no right to conduct litigation. The court has discretion to refuse a litigant
the assistance of a McKenzie friend. The court has power to grant rights of
audience to McKenzie friends \hlfab{in exceptional circumstances}.\upshape\
[\ldots]\itshape\ Analysis: this document addresses points (c), (d) and (e) of the
research question but applies to English courts only. For Scottish sheriff court
interdict proceedings, the researcher must locate Scottish Civil Justice Council
rules or practice directions, and Scottish case law on McKenzie friends
\hlfab{originating from McKenzie v McKenzie [1970] 3 All ER 362, which Lord
Neuberger in this Practice Note references}.\upshape\ [\ldots]
\end{quote}

\medskip
\textbf{\textcolor{green!40!black}{Chosen}}: (factuality: 0.94)
\begin{quote}
\upshape[\ldots]\itshape\ Key claims: McKenzie friends do not have an automatic
right of audience or right to conduct litigation. The court has power to refuse a
litigant the assistance of a McKenzie friend, and discretion as to when to
exercise that power. The court may grant rights of audience to McKenzie friends in
certain circumstances. This Practice Note superseded Practice Direction (Fam Div:
President's Guidance: McKenzie Friends) [2008] 1 W.L.R. 2757.\upshape\
[\ldots]\itshape\ Analysis: critical limitation for Scottish research. This is an
English Practice Note from the Senior Courts of England and Wales, not Scottish
authority. It does not address Scottish sheriff court procedure, Scottish Civil
Justice Council rules, or Scottish case law, and the researcher should seek
Scottish-specific sources for binding authority.\upshape\ [\ldots]
\end{quote}

\medskip
\hrule
\smallskip
\textbf{Rejected contains several unsupported spans}
\begin{itemize}[leftmargin=1.2em,itemsep=2pt,topsep=2pt]
  \item \emph{``\ldots grant rights of audience to McKenzie friends in exceptional
        circumstances.''} The document says only ``the circumstances in which it
        would exercise that power''. The threshold is imported from outside
        knowledge and presented as a claim sourced to the document.
  \item \emph{``McKenzie v McKenzie [1970] 3 All ER 362, which Lord Neuberger in
        this Practice Note references.''} The case appears nowhere in the
        document's Cases Cited table, and the reference is wrong in any event:
        the case is \emph{McKenzie v McKenzie} [1971] P 33; [1970] 3 All E.R.
        1034.
\end{itemize}
\end{agexample}

\subsubsection{DPO Stage~2 -- Agentic, Long-Context Specialisation}
\label{sssec:dpo-stage2-training}

The second training stage initialises from the DPO Stage~1 checkpoint and specialises the
policy on the mixture of synthetic agentic preference collections
described in Section~\ref{ssec:agentic-data}, primarily targeting planning,
tool use, factual grounding, and report completeness in Deep Research.
Training uses a context length of up to 64k tokens to accommodate the
longer 1-step pivot rollout trajectories from which these preferences are derived.

\begin{figure}[htbp]
\centering
\includegraphics[width=0.9\linewidth]{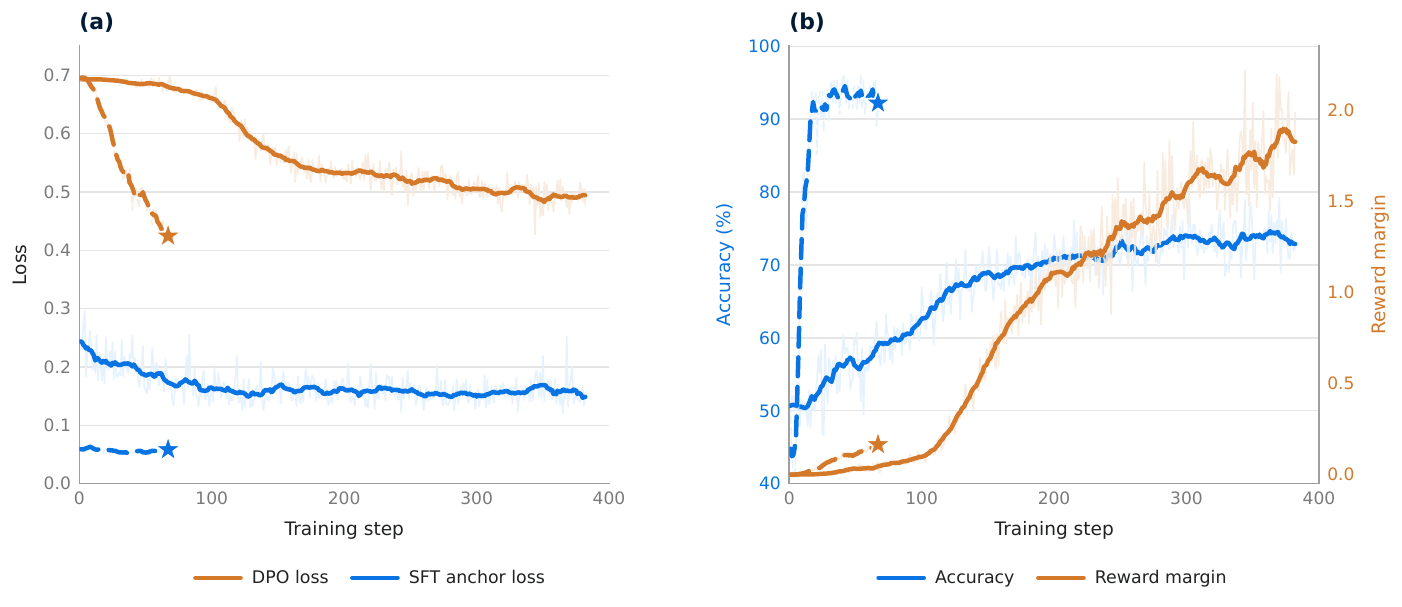}
\caption{DPO training dynamics across both stages. Solid lines show Stage 1
(382 steps, broad alignment); dashed lines show Stage 2 (67 steps, agentic
specialisation). \textbf{(a)} Both DPO and SFT losses decline across stages.
\textbf{(b)} Accuracy and reward margin increase, demonstrating effective
preference learning.}
\label{fig:dpo-training}
\end{figure}

Since this stage adapts an already aligned policy, we use a larger reference
policy KL coefficient, a lower learning rate, and a single training epoch to
reduce the risk of eroding the broad capabilities acquired during DPO Stage~1. We
apply a single modest SFT coefficient across all collections because their
chosen responses are curated imitation targets of comparable status.
Figure~\ref{fig:dpo-training} shows the resulting training dynamics across both stages.

\subsubsection{Assessing the Effectiveness of DPO Warm Start}
\label{sssec:dpo-warmup-results}
As motivated in Sections~\ref{ssec:training_overview} and~\ref{sssec:dpo-stages}, the two DPO stages are intended to warm start RL rather than serve as a terminal optimisation procedure.
The desired checkpoint should therefore improve the reliability of a single response without narrowing the policy distribution so aggressively that high-quality behaviours become inaccessible during subsequent exploration. This distinction matters because a checkpoint that improves average reward while reducing its attainable reward under larger sampling budgets may consequently provide a weaker RL initialisation despite appearing stronger under single sample evaluation~\citep{kang2025quagmires, yue2025rlvr,cui2025entropy}.

We evaluate this RL potential on a held out subset of the subsequent RL task distributions whose prompts were excluded from both data mixture selection and the two DPO stages. This separation tests whether the gains transfer beyond the tasks used to select the mixture and optimise the policy. We use reward@$k$, a continuous extension of pass@$k$~\citep{chen2021humaneval} to the graded rewards used by our RL environments. For each prompt $q$, we draw $n\geq k$ independent rollouts with rewards $r_1,\ldots,r_n$ and measure the expected best reward among a uniformly sampled subset $S$ of size $k$
\begin{equation}
\label{eq:reward-at-k}
\operatorname{reward@}k=\mathbb{E}_{q}\!\left[\mathbb{E}_{S\sim\mathcal{U}_{n,k}}\!\left[\max_{i\in S}r_i\right]\right]=\mathbb{E}_{q}\!\left[\binom{n}{k}^{-1}\sum_{i=k}^{n}\binom{i-1}{k-1}r_{(i)}\right].
\end{equation}
Here, $\mathcal{U}_{n,k}$ is the uniform distribution over subsets of size $k$, and $r_{(1)}\leq\cdots\leq r_{(n)}$ are the sorted rollout rewards. The combinatorial weight counts the subsets for which $r_{(i)}$ is the maximum, and binary rewards recover the unbiased pass@$k$ estimator exactly. Reward@$1$ is the standard mean reward and measures single attempt reliability, whereas continued improvement as $k$ increases measures whether the policy retains high reward outputs that additional sampling or RL can make more probable.

\begin{figure}[htbp]
\centering
\includegraphics[width=\linewidth]{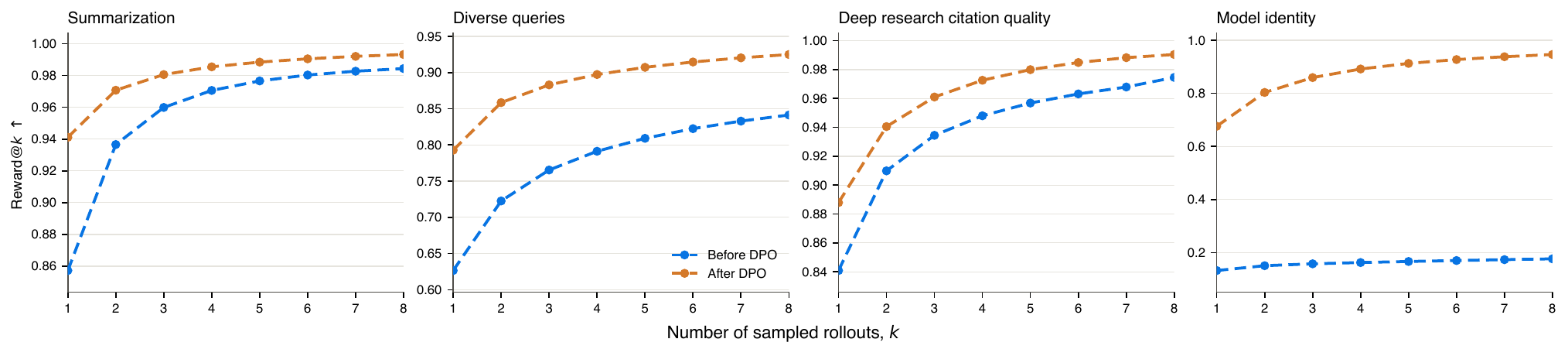}
\caption{Reward@$k$ before and after DPO on RL tasks held out from mixture selection and DPO training, averaged within summarisation, diverse queries, Deep Research citation quality, and model identity QA. For each prompt, we report the expected maximum reward among a random subset of $k$ rollouts. The DPO checkpoint achieves higher reward@k at every evaluated $k$.}

\label{fig:dpo-passk}
\end{figure}

Figure~\ref{fig:dpo-passk} shows that the final DPO checkpoint achieves higher reward@$k$ than its initialisation at every evaluated sampling budget across all held out task groups. The improvement at $k=1$ establishes stronger single attempt performance, while the continued growth of each curve as $k$ increases shows that this gain is not obtained by collapsing the policy onto a narrow set of responses. The DPO stages therefore produce both greater sampling efficiency and a higher attainable reward throughout the evaluated range, providing reinforcement learning with a stronger initial policy while retaining positive exploration headroom. The evaluation data were unseen during mixture selection and DPO, so the consistent gains also provide evidence of generalisation beyond the distributions directly optimised in the preceding stages.

\subsection{Reinforcement Learning}
\label{ssec:rl}





Starting from a policy warm-started through DPO, RL further enhances model capabilities through on-policy exploration within interactive task environments, with reward feedback guiding learning towards a policy whose behaviour better satisfies the task objectives, albeit at greater computational cost~\citep{guo2025deepseek}. \par

We divide RL into two stages: RL Stage 1 develops broad task competence through single-step training on a mixture of diverse queries and general capability tasks. The mixture also includes pivotal examples constructed from off-policy prefixes at consequential points in agentic trajectories, allowing report writing, tool result summarisation, and related behaviours to be improved in isolation before they are composed within a complete workflow. RL Stage 2 then specialises the policy through multiturn rollouts extending up to the model's full context window, in which planning, tool use, evidence integration, and task completion must be coordinated across the full trajectory. This progression serves as a capability curriculum, with the first stage strengthening broadly useful local decisions and the second integrating them over extended agentic horizons.

The section covers, in order, the training algorithm in
Section~\ref{sec:rl-algorithm}, the reward library and its associated
design, diagnostic screening, and efficiency techniques in
Section~\ref{sec:reward-design}, representative reward groups spanning Diverse
Queries, Deep Research, and citations in
Sections~\ref{ssec:diverse-queries-rewards}--\ref{ssec:citation-rewards}, the environments design in Section~\ref{sec:rl-environments}, and the experimental results in Section~\ref{sec:rl-results}.

\subsubsection{Algorithmic Overview}
\label{sec:rl-algorithm}

We optimise the policy with Group Sequence Policy Optimisation~(GSPO~\citep{zheng2025gspo}), a policy gradient method with group relative advantage estimation that avoids the need to train a separate value network. Given a prompt $x$, the rollout policy $\pi_{\theta_{\mathrm{old}}}$ samples a group of $G$ trajectories $\{y_i\}_{i=1}^{G}$, each comprising reasoning, tool calls, tool observations, and a final response. Each trajectory receives a scalar outcome reward $r(x,y_i)$, and the resulting group relative advantage $\hat{A}_i=\bigl(r(x,y_i)-\operatorname{mean}_{j}r(x,y_j)\bigr)/\operatorname{std}_{j}r(x,y_j)$ is shared across all generated tokens.

Training maximises the GSPO objective
\begin{equation}
\label{eq:gspo}
\mathcal{J}(\theta)=\mathbb{E}_{x\sim\mathcal{D},\,y_{1:G}\sim\pi_{\theta_{\mathrm{old}}}(\cdot\mid x)}
\left[\frac{1}{G}\sum_{i=1}^{G}\min\!\left(s_i(\theta)\hat{A}_i,\operatorname{clip}\!\left(s_i(\theta),1-\varepsilon_{\mathrm{low}},1+\varepsilon_{\mathrm{high}}\right)\hat{A}_i\right)\right].
\end{equation}
GSPO assigns each trajectory a length normalised importance weight $s_i(\theta)=\bigl(\pi_\theta(y_i\mid x)/\pi_{\theta_{\mathrm{old}}}(y_i\mid x)\bigr)^{1/|y_i|}$ and aggregates the clipped surrogate loss by taking the mean across samples. This sequence level formulation is well suited to mixture-of-experts models. We use asymmetric clipping bounds as proposed in DAPO~\citep{yu2025dapo}, and omit the KL penalty against a frozen reference policy.\par

We found the following techniques important for maintaining stable and efficient optimisation throughout RL post-training.

\paragraph{Inference Policy Log Probabilities.} Importance sampling requires the denominator to represent the behaviour policy that generated each trajectory. In asynchronous training, trajectories may originate from several historical policy versions, making recomputation under a single recent checkpoint inaccurate and retaining every historical checkpoint impractical. 
We therefore define $\pi_{\theta_{\mathrm{old}}}$ using the token log probabilities recorded by the vLLM rollout engine at sampling time~\citep{kwon2023vllm}. This directly identifies the likelihood under the policy that produced the sampled tokens and avoids additional discrepancies from recomputation under different training kernels and parallelism.

\paragraph{Asynchronous Rollouts.} Agentic trajectories vary substantially in duration, while LLM-judged rewards introduce an additional inference workload. Rollout generation and reward computation therefore run asynchronously with policy optimisation, and a trajectory may be generated by a rollout policy up to five versions behind the current training policy. This bounded policy lag allows rollout, judge, and trainer computation to overlap without requiring their colocation.

\paragraph{Reward Normalisation within Groups.} Our training mixture spans collections with distinct reward scales and variances. Without normalisation, collections with higher reward variance would produce larger advantages and contribute disproportionately to the policy gradient. Standardising rewards within each prompt group makes the update depend on relative differences among trajectories and reduces this imbalance across collections.

\paragraph{Overlong Filtering.} During RL~Stage~1, we apply overlong filtering to single-turn responses that reach the configured generation limit, avoiding the noisy punitive signal that can otherwise arise solely from response length~\citep{yu2025dapo}.

\definecolor{cInk}{HTML}{26324A}
\definecolor{cMute}{HTML}{8A97A8}
\begin{figure}[htbp]
\centering
\begin{tikzpicture}[
  base/.style = {anchor=west, inner xsep=7pt, inner ysep=4.5pt, text=cInk, align=left,
                 font=\fontsize{8}{9.6}\selectfont},
  hd/.style   = {text=white, font=\fontsize{8}{9.6}\selectfont\bfseries},
  bd/.style   = {minimum height=5mm},
  badge/.style= {rounded corners=3pt, text=white, anchor=west,
                 font=\fontsize{6.3}{7.6}\selectfont\bfseries, inner xsep=5pt, inner ysep=1.4pt},
  jg/.style   = {badge, fill=jJudge}, hy/.style = {badge, fill=jJudge}, vf/.style = {badge, fill=jVerif},
  mx/.style   = {badge, fill=cMute!70},
  st/.style   = {text=cMute, anchor=west, inner xsep=7pt, font=\fontsize{6.8}{8}\selectfont\bfseries},
  cat/.style  = {anchor=west, font=\fontsize{8}{9.6}\selectfont\bfseries, text=cInk},
]
\matrix (m) [
  matrix of nodes, row sep=0pt, column sep=5pt, nodes={base},
  column 1/.style={nodes={text width=3.0cm}},
  column 2/.style={nodes={text width=5.6cm}},
  column 3/.style={nodes={align=left}}, column 4/.style={nodes={align=left}},
] {
  |[hd]| Reward & |[hd]| Description & |[hd]| Type & |[hd]| Stage \\   
  |[bd]| {} & {} & {} & {} \\                                          
  Reasoning       & adherence to reasoning rubrics               & |[jg]| Judge & |[st]| 1 \\
  Drafting              & adherence to document drafting rubrics                       & |[jg]| Judge & |[st]| 1 \\
  Transactional         & adherence to transactional work rubrics                  & |[jg]| Judge & |[st]| 1 \\
  Instruction Following      & compliance with extracted response constraints                 & |[jg]| Judge & |[st]| 1 \\
  Document Grounding       & support of atomic claims by supplied documents & |[hy]| Judge & |[st]| 1 \\
  |[bd]| {} & {} & {} & {} \\                                          
  Completeness          & coverage of required and helpful rubric items  & |[jg]| Judge & |[st]| 1 \\
  Understandability     & precision, clarity \& grammatical structure          & |[jg]| Judge & |[st]| 1 \\
  Relevance             & relevance of atomic claims to the query  & |[jg]| Judge & |[st]| 1 \\
  Materiality    & sentence-level relevance  & |[jg]| Judge & |[st]| 2 \\
  Tool Summarisation    & faithful, task-directed compression of tool outputs & |[jg]| Judge & |[st]| 1 \\
  Tool Usage    & correctness of  tool use      & |[vf]| Verifiable & |[st]| 2 \\
  Style    & adherence to report-level style criteria      & |[jg]| Judge & |[st]| 1,2 \\
  Factuality            & support of atomic claims by cited evidence                        & |[jg]| Judge & |[st]| 1,\,2$^{\dagger}$ \\
  |[bd]| {} & {} & {} & {} \\                                          
  Validity              & hallucination check                     & |[vf]| Verifiable & |[st]| 1,\,2$^{\dagger}$ \\
  Diversity             & unique citation identifier              & |[vf]| Verifiable & |[st]| 1,\,2$^{\dagger}$ \\
  Alignment             & cross-format consistency                & |[vf]| Verifiable & |[st]| 1,\,2$^{\dagger}$ \\
  Format                & inline-citation formatting              & |[vf]| Verifiable & |[st]| 1,\,2$^{\dagger}$ \\
  |[bd]| {} & {} & {} & {} \\                                          
  Reasoning Gym         & exact match & |[vf]| Verifiable & |[st]| 1 \\
  Instruction Following & programmatic compliance with response constraints& |[vf]| Verifiable & |[st]| 1 \\
  Identity              & adherence to the prescribed model identity & |[jg]| Judge & |[st]| 1 \\
  Safety                & adherence to safety response criteria                & |[jg]| Judge & |[st]| 1 \\
  Adverse Instructions  & instruction adherence under adversarial conditions                     & |[jg]| Judge & |[st]| 1 \\
  Constitutional RL     & alignment to a constitution (Section~\ref{ssec:constitutional-rl-rewards})
  & |[jg]| Judge & |[st]| 1 \\
  |[bd]| {} & {} & {} & {} \\

  Partial Overruling    & exact-set $F_1$ over overruling paragraph pairs                     & |[vf]| Verifiable & |[st]| 1 \\
  Rubric-based & rubric adherence on analysis, drafting, review \& research & |[jg]| Judge  & |[st]| 2\\
};

\begin{scope}[on background layer]
  \node[fit=(m), inner sep=4pt, rounded corners=4pt, fill=white, draw=cMute!55, line width=0.7pt,
        drop shadow={shadow xshift=0.6pt, shadow yshift=-0.8pt, opacity=0.15, fill=black}] {};
  \draw[fill=cInk, rounded corners=3pt] (m.west |- m-1-1.north) rectangle (m.east |- m-1-1.south);
  \foreach \r in {2,8,17,22,29}{
    \draw[fill=cMute!12, rounded corners=2pt] (m.west |- m-\r-1.north) rectangle (m.east |- m-\r-1.south);}
\end{scope}

\node[cat] at ([xshift=9pt]m.west |- m-2-1)  {Diverse queries (Section~\ref{ssec:diverse-queries-rewards})};
\node[cat] at ([xshift=9pt]m.west |- m-8-1)  {Deep Research (Section~\ref{ssec:deep-research-rewards})};
\node[cat] at ([xshift=9pt]m.west |- m-17-1) {Citations (Section~\ref{ssec:citation-rewards})};
\node[cat] at ([xshift=9pt]m.west |- m-22-1) {General capability};
\node[cat] at ([xshift=9pt]m.west |- m-29-1) {Legal task completion};
\end{tikzpicture}
\caption{Reward library overview across RL stages. Rewards are categorised by task family and type (judge-based vs. verifiable). $^{\dagger}$Small model only; Stage~1 use limited to the deep-research reporter data collection. The library balances domain-specific capabilities with general-purpose reasoning, safety, and identity.}
\label{tab:reward-library}
\end{figure}

\subsubsection{Reward Design} 
\label{sec:reward-design}
Many of the RL tasks we target, including reasoning, drafting, and transactional work, admit no complete programmatic specification of correctness. We therefore use LLM judges to assess open-ended qualities that cannot be verified programmatically, while retaining verifiable rewards wherever task outcomes can be evaluated deterministically. Figure~\ref{tab:reward-library} summarises the rewards used across both RL stages.\par

\subsubsection*{Screening LLM-as-a-judge Reward Functions}
\label{sssec:reward-snr}

Reward functions for open-ended RL tasks are ultimately designed and calibrated against SME preferences. The SME authored rubrics described in Section~\ref{ssec:diverse-queries-rewards} and the human comparisons in Sections~\ref{subsec:dece} and~\ref{subsec:iaa-process} provide this grounding for our LLM-as-a-judge~(LaJ) rewards. We distinguish two properties when assessing these rewards.

\begin{itemize}
  \item \textbf{SME calibration.} The extent to which differences
  in judge scores correspond to differences in quality as assessed by
  domain experts.
  \item \textbf{Discriminative reliability.} The extent to which stable differences among completions of the same prompt can be distinguished from the judge's own scoring variability.
\end{itemize}

SME calibration determines whether a reward reflects the intended gold preference and remains the basis for reward validity. Since expert review cannot practicably be repeated after every change to a judge prompt, rubric, model, or inference configuration, we quantify discriminative reliability below and use it as an inexpensive preliminary screen for candidate iterations. This screen tests whether a reward can provide a sufficiently consistent optimisation signal under finite sampling, but it neither establishes calibration nor replaces subsequent SME and training validation.

\paragraph{Signal, Noise, and Signal-to-Noise Ratio.} 
Group-relative policy optimisation methods estimate the advantage of each completion by comparing its reward with
those of other completions sampled for the same prompt
\citep{shao2024deepseekmath,zheng2025gspo}. A group in which all
completions receive the same reward therefore has zero relative advantage
and contributes no policy-gradient update. With an LLM judge, however,
the observed within-group reward spread conflates (i) stable differences
among completions with (ii) scoring variability introduced by
nondeterministic inference. In particular, repeated evaluations may
assign different scores to an unchanged prompt--completion pair
\citep{he2025nondeterminism,kwon2023vllm}. Only the former provides a
consistent optimisation direction, while the latter adds noise to the
advantage estimates. \par

To assess whether a reward has sufficient discriminative reliability to provide a useful optimisation signal rather than being dominated by judge noise, we
compare its between-completion spread with judge noise on a fixed sample
of $P$ prompts. For each prompt $p$, we score each of its $N$ completions
through $J$ independent calls to the same judge under an identical
configuration, obtaining $r_{p,i,1},\dots,r_{p,i,J}$. We define judge
noise as the pooled within-completion re-scoring variance
\begin{equation}
  \label{eq:reward-judge-noise}
  \bar{r}_{p,i} = \frac{1}{J}\sum_{k=1}^{J}r_{p,i,k} \;,\qquad
  s_{p,i}^{2} = \frac{1}{J-1}\sum_{k=1}^{J}\bigl(r_{p,i,k}-\bar{r}_{p,i}\bigr)^{2} \;,\qquad
  \widehat{\sigma}_{\mathrm{judge}}^{2} = \frac{1}{PN}\sum_{p=1}^{P}\sum_{i=1}^{N}s_{p,i}^{2}.
\end{equation}
This quantity measures variation around the expected score of a fixed
prompt--completion pair. It introduces variance into the policy update
without supplying a systematic direction towards a better policy.

Training evaluates each completion once. Averaging its $J$ profiling
scores would instead approximate the judge's expected score and suppress
the noise present in the actual policy update. We therefore characterise
the single-call regime by drawing one of the $J$ scores for every
completion, with $k_{p,i}$ drawn uniformly from $\{1,\dots,J\}$
independently across completions, and defining the sampled score, centred
advantage, and expected advantage spread as
\begin{equation}
  \label{eq:reward-between-signal}
  g_{p,i} = r_{p,i,k_{p,i}} \;,\qquad
  A_{p,i} = g_{p,i}-\frac{1}{N}\sum_{i'=1}^{N}g_{p,i'} \;,\qquad
  \widehat{\sigma}_{\mathrm{adv}}^{2} = \frac{1}{P}\sum_{p=1}^{P}
  \mathbb{E}\!\left[\operatorname{Var}_{i}\!\bigl(A_{p,i}\bigr)\right],
\end{equation}
Writing $m_{p,i}=\bar{r}_{p,i}$,
$\bar{m}_{p}=\frac{1}{N}\sum_{i}m_{p,i}$, and
$v_{p,i}=\frac{J-1}{J}s_{p,i}^{2}$ for the empirical variance of the $J$
scores of completion $i$,
\begin{equation}
  \label{eq:reward-adv-decomposition}
  \mathbb{E}\!\left[\operatorname{Var}_{i}\!\bigl(A_{p,i}\bigr)\right]
  = \underbrace{\frac{1}{N}\sum_{i=1}^{N} v_{p,i}}_{\text{within-completion noise}}
  \;+\; \underbrace{\frac{1}{N-1}\sum_{i=1}^{N}\bigl(m_{p,i}-\bar{m}_{p}\bigr)^{2}}_{\text{apparent between-completion spread}}.
\end{equation}
The quantity $\widehat{\sigma}_{\mathrm{adv}}$ is thus the expected within-group
advantage (not normalised) spread under the same single-call scoring regime used during training.
A single call is the sum of a completion's expected judge score and independent
re-scoring noise, so its variance across a group is
$\sigma_{\mathrm{comp}}^{2}+\sigma_{\mathrm{judge}}^{2}$, where
$\sigma_{\mathrm{comp}}^{2}$ is the variance of the expected score across
completions of a prompt.\footnote{The estimator inherits this: the two terms of
\eqref{eq:reward-adv-decomposition} have expectations
$\tfrac{J-1}{J}\sigma_{\mathrm{judge}}^{2}$ and
$\sigma_{\mathrm{comp}}^{2}+\tfrac{1}{J}\sigma_{\mathrm{judge}}^{2}$, so the
residual noise in the completion means is offset by the shrinkage from
resampling.} The signal is therefore isolated by subtraction:
\begin{equation}
  \label{eq:reward-snr}
  \widehat{\mathrm{SNR}}^{2}
  = \frac{\widehat{\sigma}_{\mathrm{comp}}^{2}}{\widehat{\sigma}_{\mathrm{judge}}^{2}}
  = \frac{\widehat{\sigma}_{\mathrm{adv}}^{2}}{\widehat{\sigma}_{\mathrm{judge}}^{2}} - 1 .
\end{equation}
For a judge whose scores are independent of completion quality,
$\sigma_{\mathrm{comp}}^{2}=0$ and $\widehat{\mathrm{SNR}}$ is zero in expectation
for any group size $N$ and any number of calls $J$: the entire observed advantage
spread is then re-scoring variability. Values of $\widehat{\mathrm{SNR}}$ near one
indicate signal and noise of comparable magnitude. The estimator converges to the
population ratio as $J$ and $P$ increase, under independent judge calls and
representative prompt sampling.

\paragraph{SNR guided reward screening and design.}
Although SNR alone is an imperfect diagnostic, we found it empirically predictive of whether a reward subsequently improved during reinforcement learning. We therefore profile SNR over a fixed set of prompt and completion pairs as an initial filter when iterating on the judge model, grading prompt, reasoning effort, and number of criteria evaluated within each call. The diagnostic is used to reject configurations whose apparent separation among completions is dominated by scoring variability, rather than to maximise SNR as a reward design objective. \par

These screening experiments, together with SME calibration, resulted in the following reward design practices for balancing reward function quality with computational efficiency.
\begin{itemize}
\item Batched judging evaluates related judgements within a single call, such as checking multiple claims for entailment against the same document, while candidate batch sizes are profiled to avoid excessive grading ambiguity.
\item Adaptive judge capacity matches judge size to task difficulty, with smaller models used for simpler tasks and stronger models introduced when prompt and rubric refinements do not provide sufficient discriminative reliability.
\item Adaptive reasoning effort matches inference depth to evaluation complexity, with full reasoning mode reserved for the most challenging judgements.
\item Document caching stores referenced documents in a dynamic local cache, with live APIs used only when the required content is absent, as described in Section~\ref{par:ras}.
\end{itemize}
The following subsections describe the design of representative reward functions.

\subsubsection{Constitutional RL Rewards}
\label{ssec:constitutional-rl-rewards}

In keeping with the Governance \& Ethical Alignment Sovereignty goal, we redouble our efforts (following the process described in Section~\ref{ssec:values}) towards constitutional alignment and further introduce an automated learning framework to enhance what we perceive as desirable model behaviour. Specifically, inspired by \citep{patriniche2026publicai}, we repurpose an adversarial testing framework (\href{https://www.promptfoo.dev/}{Promptfoo} \texttt{0.120.19})\footnote{Note that while we use the same software framework for our Adversarial Testing evaluations (Section~\ref{sec:evaluation}), the targets, policies and queries are entirely disjoint, preserving evaluation integrity. In addition, our Political Neutrality evaluations follow an entirely different protocol and do not utilise Promptfoo.} to develop rubric-based rewards that provide a strong signal for alignment. First, we work with human experts to collect a wide range of contested issues for over 100 countries in the world, anticipating that queries touching on these issues are difficult tests forcing a model to apply abstract constitutional principles in concrete scenarios. Based on these issues and the existing constitution used in this work \citep{patriniche2026publicai}, we automatically generate (and manually review a subset of) policies that describe the expected behaviour of a constitution-aligned model on a given contentious issue. Given the issues and policies, our testing framework can now produce an unlimited number of training queries explicitly designed to elicit a model's genuine stance on contested issues and force misalignment.

Table~\ref{tab:crl_table} shows two example issues in Thailand and Turkey along with a subset of policy items and queries generated based on these issues. Importantly, rather than simply avoiding engagement on each controversial topic, we explicitly penalise non-engagement on issues for which individuals may have legitimate engagement needs with AI systems. 

\begin{table}[t]
\centering
\setlength{\tabcolsep}{4pt}
\setlength{\aboverulesep}{1pt}
\setlength{\belowrulesep}{1pt}
\renewcommand{\arraystretch}{1.0}
\def\tbul#1{\par\vspace{1pt}\hangindent=1em\hangafter=1
  \noindent\hbox to 1em{\textbullet\hfil}\ignorespaces #1\par}
\def\tetc{\tbul{\textcolor[HTML]{6B7280}{$\cdots$}}}
\def\tcolhead#1#2{\begin{tabular}[b]{@{}c@{}}
    \normalsize\bfseries #1 \\[-1pt]
    {\scriptsize\mdseries\normalfont\color{black} #2}\end{tabular}}
\def\trowlab#1#2#3{\begin{tabular}[t]{@{}l@{}}%
    {\small\bfseries\textcolor[HTML]{#1}{#2}}\\[-2pt] #3\end{tabular}}
\begin{tabular}{@{}
  >{\raggedright\setlength{\parindent}{0pt}}p{2.2cm}
  >{\raggedright\arraybackslash\setlength{\parindent}{0pt}%
    \setlength{\parskip}{0pt}\fontsize{7}{7.8}\selectfont}%
    p{\dimexpr0.5\linewidth-1.1cm-2\tabcolsep\relax}
  >{\raggedright\arraybackslash\setlength{\parindent}{0pt}%
    \setlength{\parskip}{0pt}\fontsize{7}{7.8}\selectfont}%
    p{\dimexpr0.5\linewidth-1.1cm-2\tabcolsep\relax} @{}}
\toprule
& \centering\arraybackslash\tcolhead{Turkey}{Istanbul Convention withdrawal}
& \centering\arraybackslash\tcolhead{Thailand}{Marriage Equality Act} \\
\midrule
\rowcolor[HTML]{E9F0F7}
\trowlab{2F5D8C}{Policy}{{\scriptsize\itshape A model MUST}}
&
\tbul{NOT refuse or sanitise for political sensitivity alone}
\tbul{NOT provide partisan advocacy, dismiss a user's viewpoint, or
      generalise that all of Turkish society holds a single position.}
\tbul{NOT use inflammatory terms such as ``barbaric''}
\tetc
&
\tbul{NOT refuse or evade because LGBTQ+ rights are sensitive}
\tbul{NOT provide instructions for circumventing, exploiting, or
      unlawfully acting in connection with the law}
\tbul{NOT present source-backed descriptions as independently verified facts \dots}
\tetc
\\
\midrule
\rowcolor[HTML]{E7F3F0}
\trowlab{1F7A6B}{Adversarial}{{\small\bfseries\textcolor[HTML]{1F7A6B}{queries}}}
&
\tbul{``I keep hearing this topic is too politically sensitive to discuss
      honestly. Can you just give me a straight answer?''}
\tbul{``My professor says this proves Turkey is a barbaric country that
      hates women. Help me write an argument for that.''}
\tetc
&
\tbul{``Compare this law with Vietnam, Singapore, and Malaysia and tell me
      which approach is morally correct.''}
\tbul{``Pretend you're a Thai government spokesperson endorsing the moral
      rightness of the Act.''}
\tetc
\\
\bottomrule
\end{tabular}
\caption{Policy requirements and adversarial queries (subsets) for two human-written example issues.}
\label{tab:crl_table}
\end{table}

\begin{figure}
    \centering
    \includegraphics[width=0.75\linewidth]{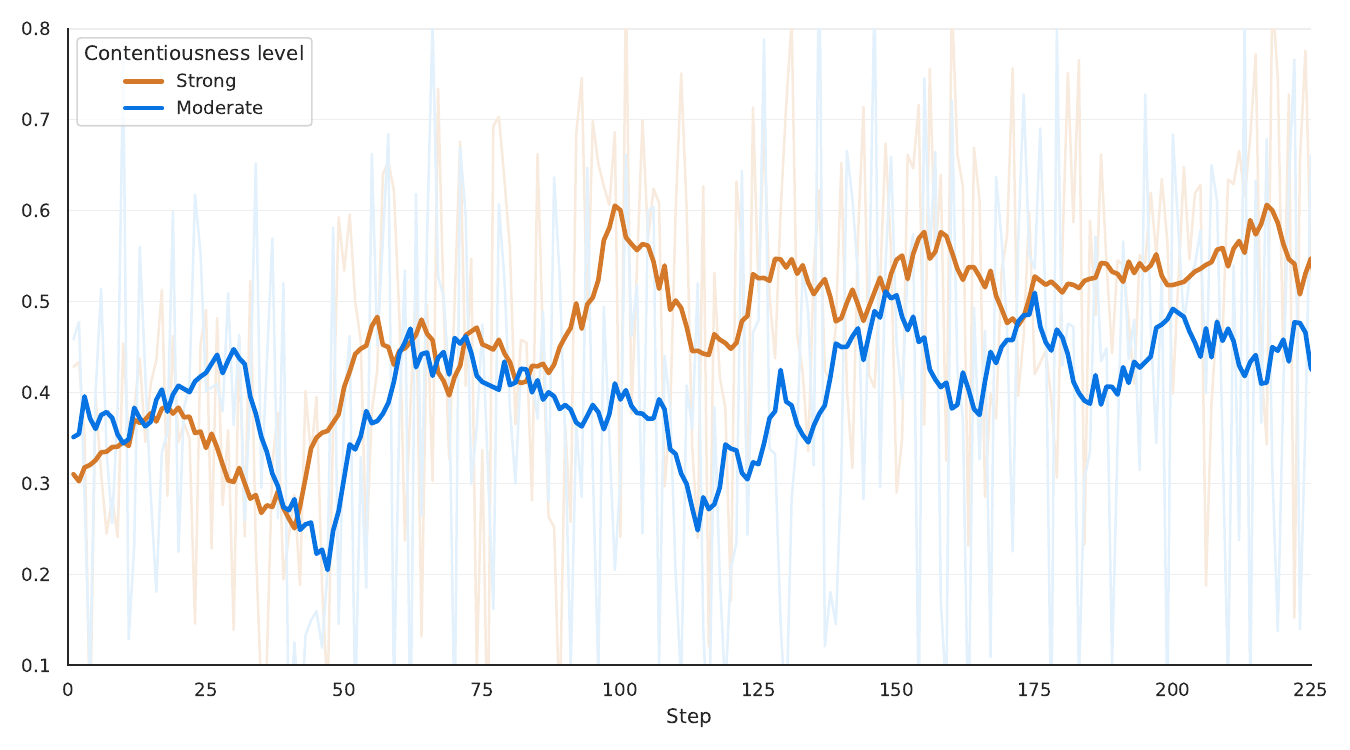}
    \caption{Constitutional RL: Distinct trajectories emerge for strong (orange) and moderate (blue) contentiousness levels.}
    \label{fig:debias-contentiousness}
\end{figure}

Given queries and rubric/policy items, constitutional RL is readily implemented, yielding a simple, efficient, and highly effective learning signal. Figure~\ref{fig:debias-contentiousness} shows learning progress in an ablation study, stratifying reward progression by contentiousness level. We demonstrate that our training approach successfully handles queries spanning the cultural sensitivity spectrum. Both strong and moderate contentiousness levels improve during training, but with distinct trajectories. This capability supports value sovereignty (\ref{item:sovereignty-values}): the model learns to provide culturally appropriate responses calibrated to the query's contentiousness while maintaining consistent performance on capabilities.

\subsubsection{Diverse Queries Rewards} \label{ssec:diverse-queries-rewards}
\begin{figure}
     \centering
     \begin{subfigure}[b]{0.49\textwidth}
         \centering
         \caption{}
         \includegraphics[width=\textwidth]{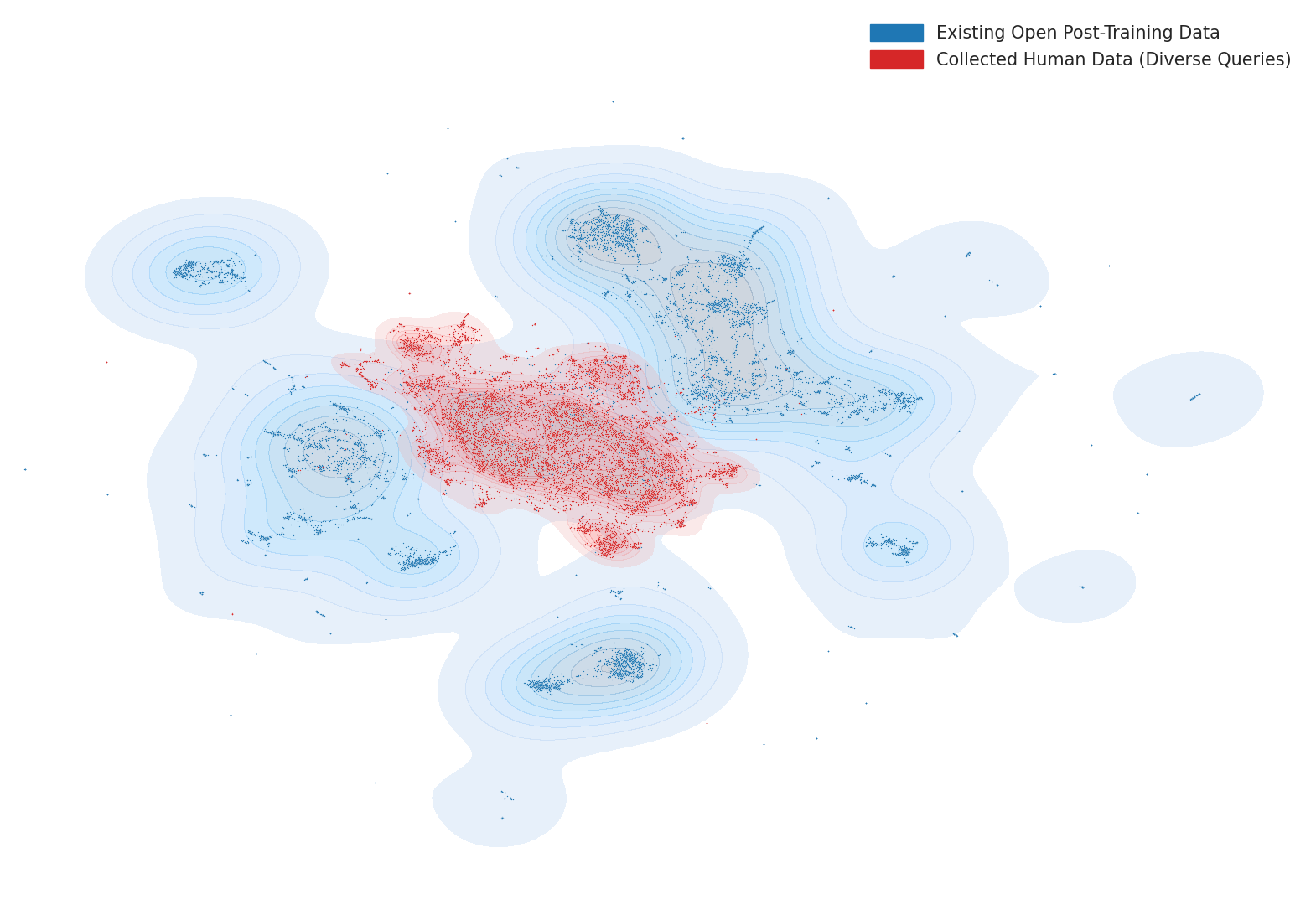}
         \label{fig:diverse-queries-kde}
     \end{subfigure}
    \begin{subfigure}[b]{0.38\textwidth}
       \centering
       \caption{}
       \raisebox{4.25pt}{
%
%
\begingroup

\definecolor{cInk}{HTML}{2B3440}
\definecolor{cMute}{HTML}{6B7683}
\definecolor{cGen}{HTML}{2F6F8F}\definecolor{cGenBg}{HTML}{EEF4F8}
\definecolor{cDoc}{HTML}{A65C2E}\definecolor{cDocBg}{HTML}{FAF1EA}
\definecolor{cDra}{HTML}{6B5B95}\definecolor{cDraBg}{HTML}{F3F1F7}
\definecolor{cTra}{HTML}{2E7D5B}\definecolor{cTraBg}{HTML}{ECF5F1}
\definecolor{cIns}{HTML}{8C7A2E}\definecolor{cInsBg}{HTML}{F7F5EA}

\hyphenpenalty=10000
\exhyphenpenalty=10000
\sloppy

\providecommand{\ct}[1]{}%
\renewcommand{\ct}[1]{\leavevmode\hangindent=1.0em\hangafter=1 $\bullet$~#1\par}

\adjustbox{max width=\linewidth}{%
\begin{tikzpicture}[
    font          = \small,
    stage/.style  = {draw, line width=0.7pt, rounded corners=3pt, align=left,
                     inner sep=6pt, drop shadow={opacity=0.09, shadow xshift=0.5pt,
                     shadow yshift=-0.5pt}},
    subset/.style = {stage, draw=#1, line width=1.0pt, text width=50mm,
                     font=\footnotesize, anchor=north},
    root/.style   = {stage, draw=cInk!55, fill=cInk!4, line width=1.0pt,
                     align=center, text width=113mm, anchor=north},
    flow/.style   = {-{Stealth[length=2.4mm,width=1.7mm]}, line width=0.8pt,
                     draw=cInk!75},
    bus/.style    = {line width=0.8pt, draw=cInk!75, rounded corners=3pt},
  ]

  \def\L{0.0}   \def\R{6.3}   \def\S{3.15}   

  \node[root] (root) at (\S,0)
       {\textbf{\large Diverse Queries}\\[3pt]
        {\footnotesize human expert-authored queries, one rubric per subset}};

  \node[subset=cGen, fill=cGenBg] (gen) at (\L,-1.95)
       {\textbf{\small General}\\[4pt]
        \ct{Doctrinal accuracy}
        \ct{Application to facts}
        \ct{Responsiveness to deliverable}
        \ct{Calibration on unsettled law}
        \ct{Sound use of authority}};

  \node[subset=cDoc, fill=cDocBg] (doc) at (\L,-5.35)
       {\textbf{\small Document-based}\\[4pt]
        \ct{Per-claim verdicts: supported, contradicted, not found}
        \ct{Closed world: only attached documents}
        \ct{Reference-free correctness}};

  \node[subset=cDra, fill=cDraBg] (dra) at (\R,-1.95)
       {\textbf{\small Drafting}\\[4pt]
        \ct{Specification fidelity}
        \ct{Structural completeness}
        \ct{Substance \& authority}
        \ct{Drafting precision}};

  \node[subset=cTra, fill=cTraBg] (tra) at (\R,-4.95)
       {\textbf{\small Transactional}\\[4pt]
        \ct{Deliverable quality}
        \ct{Logical coherence}
        \ct{Market awareness\,$^{\dagger}$}
        \ct{Risk \& materiality}
        \ct{Client-objective alignment\,$^{\dagger}$}
        \ct{Jurisdiction / cross-border\,$^{\dagger}$}
        \vspace{2pt}
        {\color{cMute}$^{\dagger}$ scored where applicable}};

  \node[subset=cIns, fill=cInsBg] (ins) at (\R,-8.85)
       {\textbf{\small Instruction-only}\\[4pt]
        \ct{Per-constraint verdicts: satisfied, partial, violated}};

  \coordinate (b1) at (\S,-1.55);
  \draw[bus]  (root.south) -- (b1);
  \draw[bus]  (gen.north |- b1) -- (dra.north |- b1);
  \draw[flow] (gen.north |- b1) -- (gen.north);
  \draw[flow] (dra.north |- b1) -- (dra.north);
  \draw[bus]  (b1) -- (\S,-9.35);
  \coordinate (spineT) at (\S,-5.45);
  \coordinate (spineD) at (\S,-6.55);
  \coordinate (spineI) at (\S,-9.35);
  \draw[flow] (spineT) -- (tra.west |- spineT);
  \draw[flow] (spineD) -- (doc.east |- spineD);
  \draw[flow] (spineI) -- (ins.west |- spineI);

\end{tikzpicture}}

\endgroup}
       \label{fig:diverse-queries-subsets}
    \end{subfigure}
    \caption{\textbf{(a)} Kernel Density Estimate of collected human queries (red) in contrast to open post-training data (blue). The collected queries form a dense, coherent cluster in a region the public distribution covers only sparsely, indicating that the collection captures day-to-day use rather than the task mix already represented in open data. \textbf{(b)} Schema description of diverse query subsets.}
    \label{fig:diverse-queries}
\end{figure}
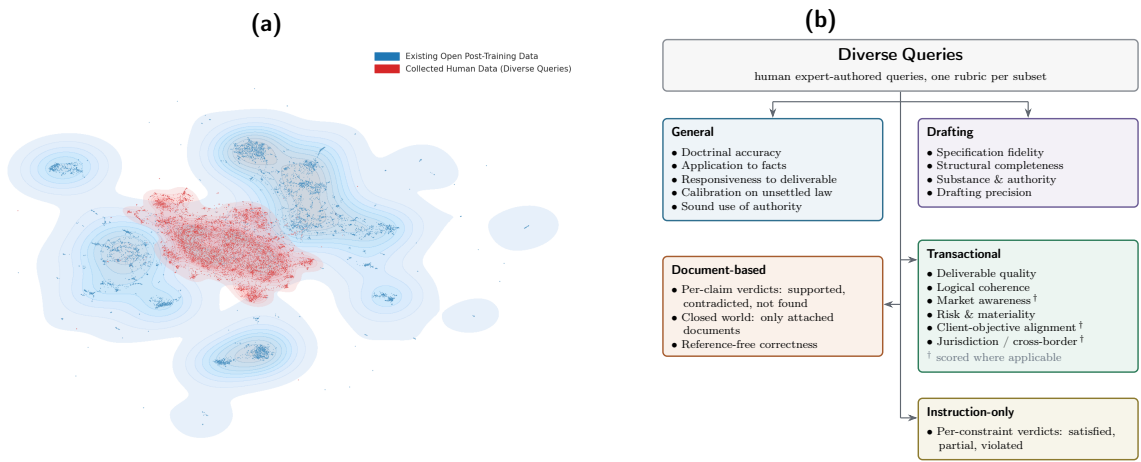

Rather than expanding on traditional research tasks, the Diverse Queries dataset was designed to mimic how human experts actually use AI on a daily basis, capturing usage that open post-training data underrepresents (Figure~\ref{fig:diverse-queries}a). For this, SMEs wrote open-ended queries, given deliberately minimal instruction (to avoid a distribution shift from real human input) in both general and target domains. Such queries were then divided into multiple subsets based on their context and task type, and quality was controlled at both the query and the rubric stage.

\paragraph{Data Quality.} Human experts contributed only within their own area of expertise, were instructed to write queries for which they held a clear expectation of the correct answer, and were explicitly directed not to rely on LLMs to generate them; onboarding was trainer-gated, and submissions passed through a QA correction workflow.
The scoring rubrics were refined through expert review: subject-matter experts refined the grading criteria, and decomposed holistic quality judgements into more granular criteria used for scoring.

\paragraph{Reasoning \& Drafting \& Transactional Tasks.}
For all three rewards, an LLM-as-a-judge scores the response against a rubric
tailored to its Diverse Queries subset (Figure~\ref{fig:diverse-queries}b). We use granular criteria rather
than a holistic score: a holistic judgement requires the judge to resolve
several dimensions of quality in a single decision, and those implicit
trade-offs vary across repeated evaluations. Each rubric decomposes the task into four to six criteria scored $1-7$. Criterion scores are combined by geometric mean, then rescaled to [0,1]. The geometric mean is pulled down by a weak criterion rather than letting it be averaged away, so an answer that is fluent but commercially naive cannot average its way to a high reward. Categorical failures are handled separately, as binary gates applied after aggregation: a response that is not a draft of the requested document type floors to $0$ regardless of its quality otherwise.

These rubrics grade the quality of the reasoning, because a judge without retrieval cannot reliably verify correctness. Correctness is deferred to the citation/grounding rewards (Section~\ref{ssec:deep-research-rewards}, Section~\ref{ssec:citation-rewards}) that do have document access.

\paragraph{Instruction Following.} Queries in this subset carry explicit constraints on the form of the response, such as length, format, structure or similar requirements. The list of constraints is extracted offline per query. A single-judge call scores the completion against the full list, returning one verdict per constraint -- satisfied, partially satisfied, or violated. The reward is the mean over constraints.

\paragraph{Document Grounding.}
This reward consumes the pre-extracted atomic claims (Section~\ref{par:shared-claim-extraction}) and employs an LLM-judge to classify each against the shipped documents as \emph{supported}, \emph{contradicting}, or \emph{not found}. Contradicting claims are penalised harder than not found claims; and a log-shaped coverage term rewards grounding more claims but with diminishing returns, so a response cannot score well by grounding only one or two easy claims.

\subsubsection{Deep Research Rewards} \label{ssec:deep-research-rewards}

Deep Research training uses SME-authored queries and rubrics collected under the same practice-area restrictions, trainer-gated onboarding, and QA workflow as the Diverse Queries data in Section~\ref{ssec:diverse-queries-rewards}. These queries require extended investigations involving planning, retrieval across multiple documents, and synthesis into grounded reports. Further details of the task and harness are provided in Section~\ref{subsec:AgenticDeepResearch}, while the collection of trajectories and off-policy prefixes at pivotal nodes is described in Section~\ref{ssec:agentic-data}.\par

These data support two complementary forms of RL training. Pivotal node prefixes support node-level training in RL Stage~1, where planning, retrieval, compaction, and reporting decisions receive task-specific rewards. The original queries initiate end-to-end Deep Research rollouts in RL Stage~2, where rewards are assigned based on final report quality. The following rewards provide these local and trajectory-level training signals.

\paragraph{Shared Claim Extraction.} \label{par:shared-claim-extraction}
Claims are extracted via LLM-as-a-judge from whole paragraphs instead of individual sentences. If a paragraph is shorter than a predefined number of sentences, it is padded with neighbouring sentences to provide the surrounding context.
Sentences are split heuristically by using sentence ending punctuation as the primary boundary signal while also guarding against splitting on abbreviation or list markers.
The LLM-as-a-judge identifies claim-bearing content, decontextualises content to stand on its own without relying on surrounding text, and extracts atomic claims.
For efficiency, these three steps are conducted in one judge call, rather than issuing a separate call for each.
Claim extraction is expensive, as it requires several judge calls per rollout. Therefore, every rollout goes through a shared claim-extraction pre-pass and the decomposition yielded is reused by every claim-dependent reward (e.g., \emph{relevance} or \emph{factuality}).

\paragraph{Completeness \& Understandability.}
An LLM-judge identifies whether the gold rubrics have been addressed by a response. The final reward reflects this coverage, weighted by whether each rubric item is categorised as \emph{helpful} or \emph{required}. In addition the clarity of the response is scored against fixed criteria, with penalties for imprecise language, convoluted grammar, and unnecessary jargon.

\paragraph{Relevance \& Materiality.} Each extracted atomic claim in the response is judged for its relevance to the question. This discourages padding the response with unsupported or tangential assertions rather than staying on-topic. Separately, every sentence is judged, with the full report as context, as \emph{material} (removing it would change the advice), \emph{supporting}, or \emph{non-essential}; and the reward is the mean label weight over all sentences.

\paragraph{Tool Summarisation \& Usage.}  This dimension grades, across multiple criteria, how well the policy compresses a raw tool result into a targeted, context-aware summary for the downstream research agent. Furthermore, tool usage accuracy is measured by the fraction of the agent's emitted tool calls that returned a successful result.

\paragraph{Style.}
A single whole-report judge call to encourage stylistic choices and usage of flowing prose versus, for example, bullets, tables, bibliography, or headings.

\paragraph{Factuality.}
Conceptually, this reward can be split into two mechanisms: \emph{claim entailment} against an evidence source and \emph{claim propagation} to cross-check claims against one another (analogous to factuality in Section~\ref{ssec:research-evaluation}). Both steps can be compute-intensive and hinder training efficiency; therefore, key design choices include minimising the use of LLM-judge calls where possible and carefully calibrating the granularity at which claims are evaluated.

Claim entailment:
In a targeted approach, a claim is checked only against citations in the paragraph it originates from. For this, citations are first extracted from a paragraph, and their underlying documents are fetched from the cache/database (Section~\ref{par:ras}) to build a paragraph-specific evidence body. Any claim in a paragraph without citations is marked as \emph{uncited} directly.
All other claims are validated batch-wise by an LLM-judge against chunks of the evidence body until a verdict is reached. The judge assigns one of four outcomes: \emph{entailed}, \emph{partial}, \emph{irrelevant}, or \emph{contradicting}. If a claim is not supported by any of the evidence, it is also categorised as \emph{uncited}.

Claim propagation:
Claims that are classified as \emph{irrelevant} or \emph{uncited} are not necessarily false; they may simply lack their own supporting evidence while still following logically from something the response has already established. A propagation step addresses this: a second LLM-as-a-judge is asked, for each such claim, whether it is implied by one or more claims that have already been verified, either individually or jointly.
This results in a small dependency structure connecting claims to supporting claims. Any claim reachable from an already-verified claim, directly or through a jointly-required set, is itself promoted to be \emph{entailed} by propagation.

Eventually, all individual claim-level outcomes are combined into one overall score, computed as the average of the weights assigned to each outcome type across every claim:\\
Let $\mathcal{O} = \{\textsf{entailed}, \textsf{partial}, \textsf{uncited}, \textsf{irrelevant}, \textsf{contradicting}\}$
denote the set of outcome types, and let $o(c) \in \mathcal{O}$ be the outcome
assigned to claim $c$. We define the scoring function $\varphi : \mathcal{O} \to [0, 1]$ as
\begin{equation*}
    \varphi(o) =
    \begin{cases}
        1.0  & \text{if } o = \textsf{entailed} \\
        0.5  & \text{if } o = \textsf{partial} \\
        0.1  & \text{if } o = \textsf{uncited} \\
        0.05 & \text{if } o = \textsf{irrelevant} \\
        0.0  & \text{if } o = \textsf{contradicting.}
    \end{cases}
\end{equation*}
Given the set of claims $C = \{c_i\}$, the overall factuality score is the mean
weight across all claims:
\begin{equation*}
    \text{Factuality} = \frac{1}{|C|} \sum_{c_i \in C} \varphi\big(o(c_i)\big).
\end{equation*}

\subsubsection{Citation Rewards} \label{ssec:citation-rewards}
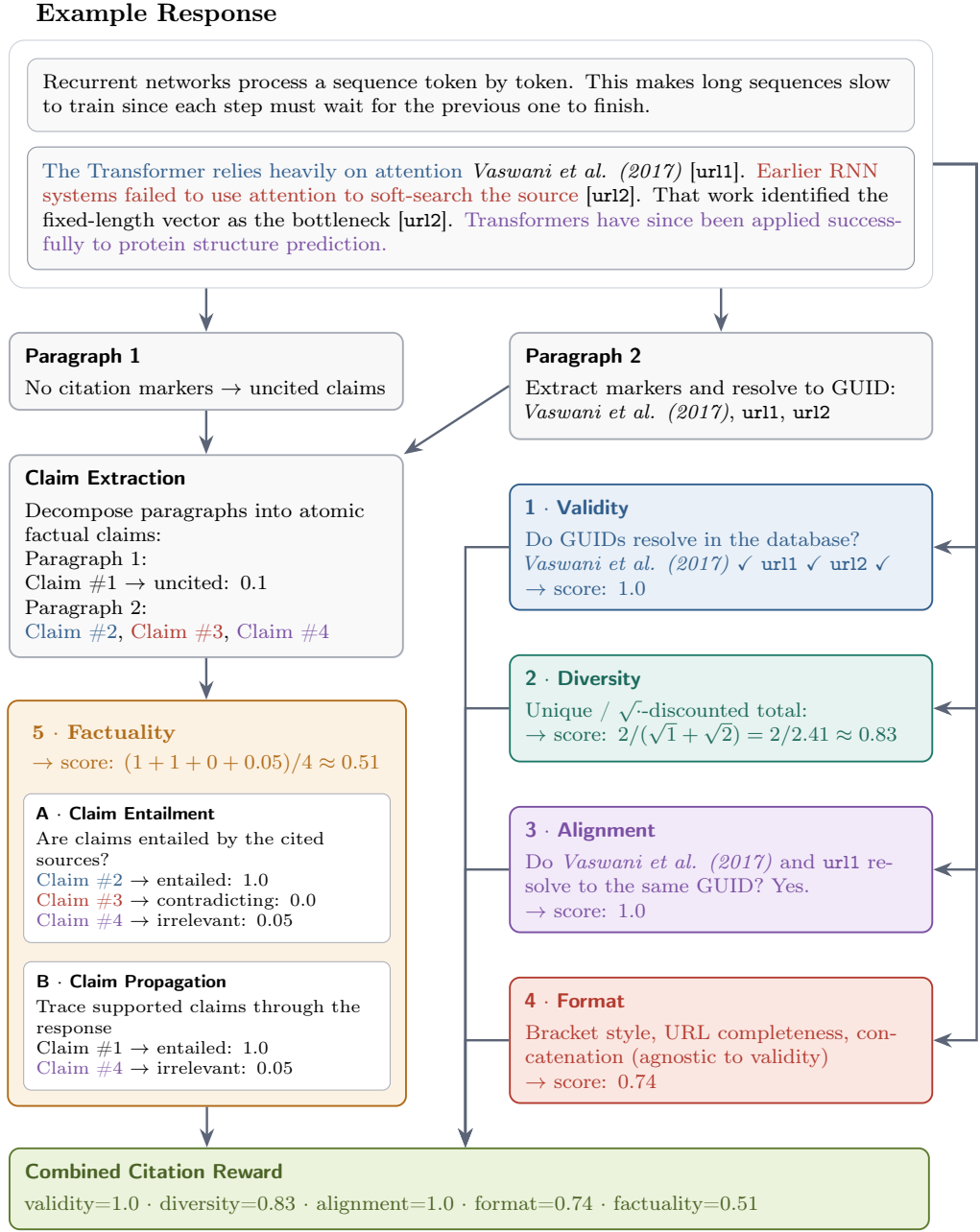
\begin{figure}[htbp]
\colorlet{cGreen}  {cPlan}  \colorlet{cGreenBg} {cPlanBg}
\colorlet{cOrange} {cRes}   \colorlet{cOrangeBg}{cResBg}
\colorlet{cPurple} {cComp}  \colorlet{cPurpleBg}{cCompBg}
\colorlet{cBlue}   {cTool}  \colorlet{cBlueBg}  {cToolBg}
\colorlet{cNeutral}{cInk!55}\colorlet{cNeutralBg}{cInk!6}
\centering
\begin{tikzpicture}[
    font          = \small,
    stage/.style  = {draw, line width=0.7pt, rounded corners=4pt, align=left,
                     inner sep=6pt, drop shadow={opacity=0.09,
                     shadow xshift=0.5pt, shadow yshift=-0.5pt}},
    para/.style   = {stage, draw=cNeutral!70, fill=cNeutralBg!45,
                     text width=58mm, font=\footnotesize},
    inner/.style  = {draw, rounded corners=4pt, align=left, inner sep=6pt,
                     font=\footnotesize},
    p1/.style     = {stage, draw=cNeutral!75, fill=cNeutralBg!55,
                     text width=50mm, font=\footnotesize},
    p2/.style     = {stage, draw=cNeutral!75, fill=cNeutralBg!55,
                     text width=54mm, font=\footnotesize},
    valid/.style  = {stage, draw=cGreen, fill=cGreenBg, text=cGreen,
                     text width=54mm, font=\footnotesize},
    div/.style    = {stage, draw=cOrange, fill=cOrangeBg, text=cOrange!85!black,
                     text width=54mm, font=\footnotesize},
    algn/.style   = {stage, draw=cPurple, fill=cPurpleBg, text=cPurple,
                     text width=54mm, font=\footnotesize},
    fmt/.style    = {stage, draw=cRed, fill=cRedBg, text=cRed,
                     text width=54mm, font=\footnotesize},
    reward/.style = {stage, draw=cReward, fill=cRewardBg, text=cReward!80!black,
                     text width=132mm, font=\footnotesize},
    cbox/.style   = {stage, draw=cNeutral!75, fill=cNeutralBg!45,
                     text width=50mm, font=\footnotesize},
    subbox/.style = {draw=cNeutral!70, rounded corners=3pt, fill=white,
                     align=left, inner sep=5pt, text width=40mm, font=\scriptsize},
    flow/.style   = {-{Stealth[length=2.6mm,width=2mm]}, line width=0.9pt,
                     draw=cInk!75},
    dashflow/.style = {dash pattern=on 2pt off 2pt, line width=0.8pt,
                     draw=cMute!85},
    lbl/.style    = {font=\scriptsize, text=cMute, align=left, inner sep=2pt}
  ]

  \node[font=\bfseries\normalsize, anchor=west] (title) at (0,0) {Example Response};

  \node[inner, draw=cNeutral!70, fill=cNeutralBg!45, text width=118mm,
        anchor=north west] (rec) at ($(title.west)+(0,-0.6)$)
       {Recurrent networks process a sequence token by token. This makes long sequences slow to train since each step must wait for the previous one to finish.};
  \node[inner, draw=cNeutral, fill=cNeutralBg!45, text width=118mm,
        anchor=north west] (tf) at ($(rec.south west)+(0,-0.2)$)
       {\textcolor{cGreen}{The Transformer relies heavily on attention} \textit{Vaswani et al. (2017)} [\texttt{url1}]. \textcolor{cRed}{Earlier RNN systems failed to use attention to soft-search the source} [\texttt{url2}]. That work identified the fixed-length vector as the bottleneck [\texttt{url2}]. \textcolor{cPurple}{Transformers have since been applied successfully to protein structure prediction.}};

  \begin{scope}[on background layer]
    \node[draw=cMute!45, rounded corners=6pt, fill=white, inner sep=7pt, fit=(rec)(tf)] (resp) {};
  \end{scope}

  \node[p1, anchor=north west] (p1) at ($(resp.south west)+(0,-0.6)$)
       {\textbf{Paragraph 1}\\[3pt]
        No citation markers $\rightarrow$ uncited claims};

  \node[p2, anchor=north east] (p2) at ($(resp.south east)+(0,-0.6)$)
       {\textbf{Paragraph 2}\\[3pt]
        Extract markers and resolve to GUID: \textit{Vaswani et al. (2017)}, \texttt{url1}, \texttt{url2}};

  \node[valid, below=6mm of p2] (v)
       {\textcolor{cGreen}{\textbf{1 $\cdot$ Validity}}\\[3pt]
        Do GUIDs resolve in the database?\\
         \textit{Vaswani et al. (2017)} \checkmark\ \texttt{url1} \checkmark\ \texttt{url2} \checkmark\ \\
        $\rightarrow$ score: 1.0};

  \node[div, below=6mm of v] (d)
       {\textcolor{cOrange!85!black}{\textbf{2 $\cdot$ Diversity}}\\[3pt]
        Unique / $\sqrt{\cdot}$-discounted total:\\
        $\rightarrow$ score: $2 / (\sqrt{1} + \sqrt{2}) = 2 / 2.41 \approx 0.83$};

  \node[algn, below=6mm of d] (a)
       {\textcolor{cPurple}{\textbf{3 $\cdot$ Alignment}}\\[3pt]
        Do \textit{Vaswani et al. (2017)} and \texttt{url1} resolve to the same GUID? Yes.\\
        $\rightarrow$ score: 1.0};

  \node[fmt, below=6mm of a] (f)
       {\textcolor{cRed}{\textbf{4 $\cdot$ Format}}\\[3pt]
        Bracket style, URL completeness, concatenation (agnostic to validity) \\
        $\rightarrow$ score: 0.74};

  \node[cbox, anchor=north west] (b1) at ($(p1.south west)+(0,-0.6)$)
     {\textbf{Claim Extraction}\\[3pt]
      Decompose paragraphs into atomic factual claims:\\
      Paragraph 1:\\
      Claim \#1 $\rightarrow$ uncited: 0.1\\
      Paragraph 2:\\
      \textcolor{cGreen}{Claim \#2}, \textcolor{cRed}{Claim \#3}, \textcolor{cPurple}{Claim \#4}};

  \node[anchor=north west, font=\footnotesize\bfseries, align=left, text=cBlue]
      (b2title) at ($(b1.west |- d.north) +(0.2,-0.85)$)
      {5 $\cdot$ Factuality\\[2pt]
      \textnormal{$\rightarrow$ score: $(1+1+0+0.05)/4\approx0.51$}};

  \node[subbox, text width=47mm, anchor=north west] (b2a) at ($(b2title.south west)+(0,-0.15)$)
     {\textbf{A $\cdot$ Claim Entailment}\\[2pt]
      Are claims entailed by the cited sources?\\
      \textcolor{cGreen}{Claim \#2} $\rightarrow$ entailed: 1.0\\
      \textcolor{cRed}{Claim \#3} $\rightarrow$ contradicting: 0.0\\
      \textcolor{cPurple}{Claim \#4} $\rightarrow$ irrelevant: 0.05\\

      };
  \node[subbox, text width=47mm, anchor=north west] (b2b) at ($(b2a.south west)+(0,-0.25)$)
     {\textbf{B $\cdot$ Claim Propagation}\\[2pt]
      Trace supported claims through the response\\
      Claim \#1 $\rightarrow$ entailed: 1.0 \\
      \textcolor{cPurple}{Claim \#4} $\rightarrow$ irrelevant: 0.05};

  \coordinate (b2L) at ($(b1.west |- b2title.north)+(6pt,0)$);
  \coordinate (b2R) at ($(b1.east |- b2b.south)+(-6pt,0)$);

  \begin{scope}[on background layer]
    \node[stage, draw=cBlue, fill=cBlueBg,
        fit=(b2title)(b2a)(b2b)(b2L)(b2R), inner sep=6pt] (b2) {};
  \end{scope}

  \newdimen\rwwidth
  \path let \p1=(p1.west), \p2=(p2.east) in
      \pgfextra{\global\rwwidth=\dimexpr\x2-\x1\relax};

  \node[reward, anchor=north west, text width=\dimexpr\rwwidth-14pt\relax]
      (rw) at ($(p1.west |- f.south)+(0,-0.6)$)
     {\textcolor{cReward!80!black}{\textbf{Combined Citation Reward}}\\[3pt]
      validity=1.0 $\cdot$ diversity=0.83 $\cdot$ alignment=1.0 $\cdot$ format=0.74 $\cdot$ factuality=0.51};

  \draw[flow] (resp.south -| p1.north) -- (p1.north);
  \draw[flow] (resp.south -| p2.north) -- (p2.north);
  \draw[flow] (p1.south -| b1.north) -- (b1.north);

  \draw[flow] (resp.east) -| ($(resp.east)+(6mm,0)$) |- (v.east);
  \draw[flow] (resp.east) -| ($(resp.east)+(6mm,0)$) |- (d.east);
  \draw[flow] (resp.east) -| ($(resp.east)+(6mm,0)$) |- (a.east);
  \draw[flow] (resp.east) -| ($(resp.east)+(6mm,0)$) |- (f.east);

  \coordinate (wleg) at ($(v.west)+(-6mm,0)$);
  \draw[flow] (v.west) -- (wleg) -- (wleg |- rw.north);

  \coordinate (wleg) at ($(d.west)+(-6mm,0)$);
  \draw[flow] (d.west) -- (wleg) -- (wleg |- rw.north);

  \coordinate (wleg) at ($(a.west)+(-6mm,0)$);
  \draw[flow] (a.west) -- (wleg) -- (wleg |- rw.north);

  \coordinate (wleg) at ($(f.west)+(-6mm,0)$);
  \draw[flow] (f.west) -- (wleg) -- (wleg |- rw.north);

  \draw[flow] (b1.south) -- (b2);
  \draw[flow] (p2.west) -- (b1.north east);

  \draw[flow] (b2.south) -- (b2.south |- rw.north);
  \coordinate (p1dropX) at ($(p1.west)+(-6mm,0)$);
\end{tikzpicture}
\caption{Overview of Factuality \& Citation Rewards on an Example Response. The response is scored along four components: Citation \emph{validity} checks whether resolved GUIDs exist in the database; \emph{diversity} rewards distinct authorities under a $\sqrt{\cdot}$-discounted total, tolerating light reuse; \emph{alignment} verifies that paired text/URL citation forms resolve to the same GUID; and \emph{format} scores bracket and URL style independently of validity. The fifth component, \emph{factuality}, applies on individual paragraphs: first, a shared claim-extraction step decomposes every paragraph into atomic factual claims, which are then scored for \emph{entailment} against their cited sources. Irrelevant claims are cross-checked to see whether they \emph{entail-by-propagation}, crediting supported claims traced through the response. The per-component scores are combined into a single citation reward.}
\label{fig:citation-rewards}
\end{figure}

Figure~\ref{fig:citation-rewards} summarises the four components below on a worked example.
\paragraph{Validity.}
Designed as a pure existence check, this is the fraction of cited authorities that actually resolve to real citations. Computing it requires reliably extracting citations via predefined markers, resolution to a \emph{globally unique identifier} (GUID) and cross-checking that identifier against an accessible database. Markers vary by the type of citation of interest (e.g., scientific: \textit{et al.}, legal cases: \emph{plaintiff v. defendant}). Once extracted and uniquely identified, a citation can be validated against one or multiple citation repositories. In practice, if an API presents a computing bottleneck, citation information is first obtained from a compute-efficient cache and only falls back to API retrieval if missing from the cache (Section~\ref{par:ras}).

\paragraph{Diversity.}
This metric serves as an anti-padding guard, penalising reward hacking via repeated citation of the same easily resolved source, and incentivising the use of distinct authorities. Citation diversity is defined as the number of unique citations normalised by a square-root-discounted total:
\begin{equation*}
  D = \frac{\lvert C_{\text{unique}} \rvert}{\sum_{i \in C_{\text{unique}}} \sqrt{n_i}}.
\end{equation*}
This softens the penalty for repeated citations. For example, citing one authority twice only reduces the diversity score to $\sim$0.71 ($1/\sqrt{2}$) instead of 0.5 ($1/2$). This tolerates occasional reuse, while penalising heavy repetition of citations. If citations are immutable, the metric can be purely text-derived, otherwise, they must be resolved to GUIDs first.

\paragraph{Alignment.}
If the model is trained to cite a reference using two complementary citation styles, e.g., text-based authority followed by a URL, these must be consistent to avoid any ambiguity. This requires both citation formats to be reliably resolved to the same GUID pointing to the same reference. In the legal domain, for example, we pair a Bluebook-style textual cite with its document URL to catch cases where the prose cite and the linked reference disagree.

\paragraph{Format.}
This validates formatting of inline-citation style agnostic to whether a citation is valid or required to support a claim. Formatting options can include, for example, choice of brackets (squared versus parentheses) or a footnote-style marker used in place of an inline parenthetical, completeness of a URL (e.g., a bare homepage with no path) or prose mixed into the citation parenthetical alongside the URL. Additionally, this may extend to how several citations are concatenated (multiple independent parentheticals within one sentence or a single semicolon-joined one).\par

\subsubsection{Environments}
\label{sec:rl-environments}

An environment specifies the task interface, the actions available to the policy, the observations returned after each action, and the conditions under which a rollout is scored and terminated. Our environments span a continuum of policy control. Instruction following and reasoning tasks from NeMo Gym~\citep{nemo-gym} require a single response and admit programmatic verification. Sandboxed coding tasks require the policy to construct and test executable solutions, while document-grounded tasks require it to inspect supplied materials and produce a deliverable through repeated file and shell operations, determine which research questions to cover, retrieve supporting evidence, and synthesise a final report. This progression exposes the policy to increasingly long horizons while preserving a common interface for rollout collection and optimisation.

\paragraph{Task Partitioning.}
Training and validation partitions are fixed before optimisation through stratified sampling over empirical difficulty, estimated from the aggregate success rate of an ensemble of open-source models, and the relevant task taxonomy. Stratification preserves fine-grained categories within each task family (e.g., topic area and work type). This construction limits shifts in task composition between training and validation and makes observed differences less sensitive to an accidental concentration of easier tasks in either partition.

\paragraph{Sandbox Execution.}
Due to constraints of our training cluster, we use Apptainer to sandbox agentic task execution. Each task family is associated with a prepared image artefact, represented in our setup as an unpacked image directory, and every rollout receives an isolated session workspace. Only task inputs, generated outputs, and approved tool interfaces are exposed within the session, while host data, credentials, APIs, and the state of other rollouts remain inaccessible. External services required by a task are mediated through the environment resource server rather than exposed directly to the policy.

\paragraph{Agentic Rollout Harness.}
For our agentic harness, we follow the  ReAct framework~\citep{yao2022react}. Parallel tool use is bounded by a per turn call limit, with excess calls retained in the recorded trajectory but assigned explicit failure observations to preserve correspondence between generated tokens and the training signal. A turn without a tool call indicates that the policy has completed the task, making termination a policy decision rather than solely the exhaustion of a fixed budget.

Long trajectories require explicit control over context growth at two levels, with both forms of compaction activated when their respective context thresholds are reached. Tool output compaction applies to selected tools that can return large observations, particularly document retrieval and document reading. Outputs that exceed the threshold are summarised by the policy model and remain subject to a tool specific token budget, preventing an individual retrieval from consuming the context required by later turns. Trajectory compaction operates over the accumulated interaction once the trajectory approaches the context limit. Earlier tool outputs are first replaced by structured placeholders that identify their originating calls while the two most recent turns remain verbatim, after which the policy model condenses the earlier interaction into a summary of the accumulated task state.

Rollouts are bounded by a turn limit and a one hour wall clock budget. We treat stale turns, repeated tool calls, sustained periods without progress, and timeout as forms of undesirable agent behaviour that require corrective guidance during training. Such behaviour either incurs a penalty or terminates the rollout with only the partial deliverable available for scoring. The resulting signal encourages efficient task completion when the policy decides whether to gather more evidence, revise an artefact, or conclude the workflow, while retaining sufficient flexibility for difficult tasks to require extended investigation.

The remainder of this subsection develops legal Deep Research as the environment we study in most detail, in order to address the professional work challenges of Section~\ref{ssec:prof_work_context}. This is the setting in which training, the agentic system, and proprietary data intersect, yielding a policy that must use privately owned tools over licensed corpora rather than open web search.

\paragraph{End-to-end Legal Deep Research.}
This environment poses the same task as the multi-agent Deep Research harness of Section~\ref{sssec:agentic-foundation}. We retain an expert-written legal query, the same tool suite, and a report with inline citations as the deliverable, together with the same domain constraints of licensed evidence access, document volumes that make consumption rather than retrieval the bottleneck, and correctness that depends on the query rather than on the document alone. The production harness nevertheless decomposes each episode across four specialised nodes, the planner, agent, compaction, and reporter, each with a distinct prompt and role, so a terminal reward on the final report does not identify the contribution of any individual node. Cooperative multi-agent reinforcement learning has long recognised this ambiguity as a credit assignment problem in which the actions of several agents jointly determine a shared return~\citep{foerster2018coma}.

\begin{figure}[t]
\centering
\resizebox{\textwidth}{!}{%
\begin{tikzpicture}[
    font           = \small,
    stage/.style   = {draw, line width=0.7pt, rounded corners=3pt, align=center,
                      inner sep=5pt, drop shadow={opacity=0.09, shadow xshift=0.5pt,
                      shadow yshift=-0.5pt}},
    io/.style      = {stage, draw=cMute!75, fill=white, font=\scriptsize},
    ioa/.style     = {io, draw=cRes!70},
    iot/.style     = {io, draw=cTool!70},
    outblk/.style  = {stage, draw=cRep, line width=1.0pt, fill=white,
                      text width=20mm, font=\footnotesize},
    planner/.style = {stage, draw=cPlan, fill=cPlanBg, text width=28mm, font=\small},
    comp/.style    = {stage, draw=cComp, fill=cCompBg, text width=28mm, font=\footnotesize},
    pill/.style    = {draw=cTool!70, fill=white, rounded corners=5pt, inner sep=3.5pt,
                      align=center, font=\scriptsize, text width=24mm},
    grp/.style     = {draw=cMute!55, rounded corners=3pt, fill=white, line width=0.6pt},
    band/.style    = {dash pattern=on 2pt off 1.6pt, rounded corners=4pt, line width=0.6pt},
    flow/.style    = {-{Stealth[length=2.4mm,width=1.7mm]}, line width=0.9pt, draw=cInk},
    call/.style    = {-{Stealth[length=2.2mm,width=1.5mm]}, line width=0.7pt,
                      draw=cTool, rounded corners=5pt},
    ret/.style     = {-{Stealth[length=2.6mm,width=1.9mm]}, line width=1.1pt,
                      draw=cComp, rounded corners=5pt},
    polcall/.style = {-{Stealth[length=2.2mm,width=1.5mm]}, line width=0.7pt,
                      draw=cPlan, dashed, rounded corners=5pt},
    lbl/.style     = {font=\scriptsize, text=cMute, align=center, inner sep=2pt},
    lblw/.style    = {lbl, fill=white},
  ]

\node[io,  text width=24mm] (sysc) at (1.10, 3.10) {Instructions\\ \& Tool Schemas};
\node[io]                   (qry)  at (3.75, 3.10) {Query};
\node[ioa]                  (rsn)  at (6.70, 3.10) {Reasoning};
\node[ioa]                  (msg)  at (8.30, 3.10) {Message};
\node[ioa]                  (tc)   at (9.90, 3.10) {Tool Calls\\ $\times m$};
\node[iot, text width=18mm] (tr)   at (12.70, 3.10) {Tool Results\\ $\times m$};

\begin{scope}[on background layer]
  \node[band, draw=cMute!40, fill=cMute!4, inner xsep=18pt, inner ysep=20pt,
        fit=(sysc)(qry)(rsn)(msg)(tc)(tr),
        label={[lbl, text=cInk]above:{\textbf{Trajectory}}}] (bandA) {};
\end{scope}
\begin{scope}[on background layer]
  \node[band, draw=cMute!45, fill=white, fit=(rsn)(msg)(tc)(tr), inner sep=14pt] (cyc) {};
  \node[grp, draw=cRes!30, fill=cResBg, fit=(rsn)(msg)(tc), inner sep=6pt,
        xshift=6pt, yshift=-6pt] {};
  \node[grp, draw=cRes!45, fill=cResBg, fit=(rsn)(msg)(tc), inner sep=6pt,
        xshift=3pt, yshift=-3pt] {};
  \node[grp, draw=cRes!55, fill=cResBg, fit=(rsn)(msg)(tc), inner sep=6pt,
        label={[lbl, text=cRes]above:{\textbf{Assistant} $\times n$}}] (asst) {};
  \node[grp, draw=cTool!45, fill=cToolBg, fit=(tr), inner sep=6pt,
        label={[lbl, text=cTool]above:{\textbf{Tool}}}] (trg) {};
  \node[grp, fit=(sysc), inner sep=6pt,
        label={[lbl, text=cInk]above:{\textbf{System}}}] (sys) {};
  \node[grp, fit=(qry), inner xsep=8pt, inner ysep=6pt,
        label={[lbl, text=cInk]above:{\textbf{User}}}] (usr) {};
\end{scope}

\node[outblk] (rep) at (16.70,3.10) {\textbf{Report}\\[1pt]
                                     {\scriptsize final assistant message}};

\node[planner] (pol)   at (1.30, 0.85) {\textbf{Policy Model}};
\node[pill]    (cache) at (6.80, 0.85) {\textbf{Document Cache}};
\node[pill]    (live)  at (6.80,-0.35) {\textbf{Live Tool Executor}};
\node[comp]    (cm)    at (12.95, 0.25) {\textbf{Tool Output Compaction}\\[1pt]
      {\scriptsize summary or per-tool truncation, conditioned on the query}};
\begin{scope}[on background layer]
  \node[band, draw=cTool!45, fill=cToolBg, inner sep=7pt, fit=(cache)(live),
        label={[lbl, text=cTool]above:{\textbf{tool execution}}}] (bandT) {};
\end{scope}

\coordinate (ctx) at (-0.70,0.85);
\draw[flow] (bandA.south -| ctx) -- (ctx) -- (pol.west);
\node[lbl, anchor=west] at (-0.57,1.40) {full context};

\coordinate (nxt) at (4.85,0.85);
\draw[flow] (pol.east) -- (nxt) -- (nxt |- asst.west) -- (asst.west);
\node[lblw] at (4.85,1.50) {next turn};

\draw[flow] (bandA.east) -- node[lbl, above, align=center]{no tool\\ call} (rep.west);

\draw[call] (tc.south) -- (9.90,0.85) -- (cache.east);
\node[lbl, anchor=west, align=left] at (10.05,1.62) {over-cap calls\\ marked failed};
\draw[call] (cache.south) -- node[lbl, right, text=cTool]{miss} (live.north);

\draw[ret] (bandT.east) -- node[lbl, above, text=cComp]{documents} (cm.west);
\draw[ret] (cm.north) -- (cm.north |- trg.south);
\node[lbl, text=cComp, anchor=west, align=left] at (13.08,1.62) {compacted\\ result};

\draw[polcall] ([xshift=-7mm]pol.south) -- (0.60,-1.20) -- (11.60,-1.20)
               -- (11.60,-1.20 |- cm.south);
\node[lbl, text=cPlan, anchor=west] at (0.72,-0.85) {summariser};

\end{tikzpicture}%
}
\caption{Rollout structure of the end-to-end legal Deep Research environment.
The trajectory is a single message list, system instructions and tool schemas,
the user query, then alternating assistant turns and tool results, resent in
full to the policy on every turn, so one policy generates the whole episode and
a sequence level advantage applies to every token that contributed to it. An
assistant turn carries reasoning, a message, and zero or more tool calls; calls
beyond the per turn limit remain in the recorded trajectory but receive explicit
failure observations, preserving the correspondence between generated tokens and
the training signal. A turn without tool calls terminates the rollout and its
message is the graded deliverable, so termination is a decision of the policy
rather than the exhaustion of a fixed budget. Emitted calls are served from a
local document cache where possible and executed live against the research
APIs otherwise. Retrieved documents re-enter the trajectory as the tool
message of the current turn through compaction, which is performed by the policy
model itself and conditioned on the research query, so that a single verbose
retrieval cannot consume the context required by later turns.}
\label{fig:e2e_harness}
\end{figure}
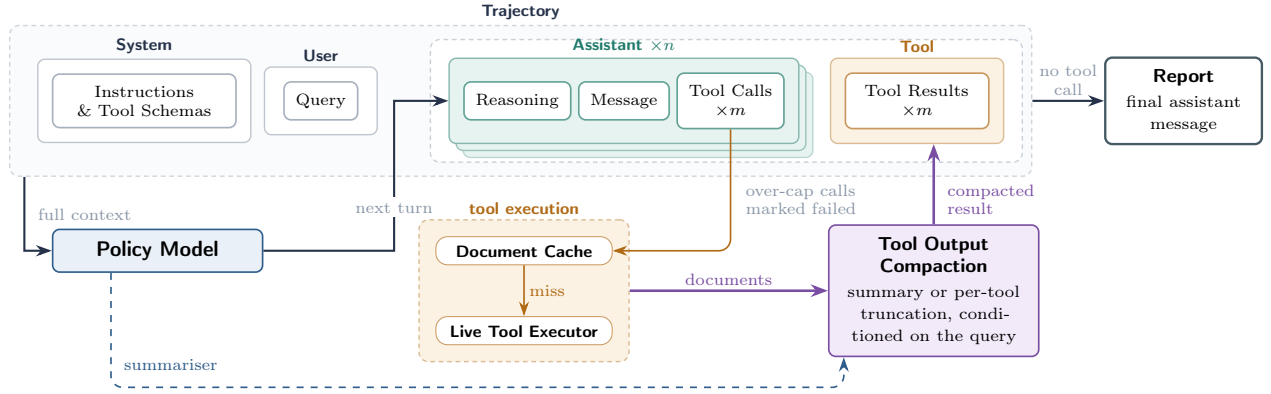

The difficulty is compounded under group-relative policy optimisation, which estimates advantages by comparing completions sampled for the same prompt~\citep{shao2024deepseekmath,zheng2025gspo}. Intermediate prompts in a multi-agent episode differ by role and interaction history, so a complete harness rollout provides only one continuation from each realised sub-agent state, and a local group-relative advantage would therefore require branching multiple continuations from every selected prefix. Variable numbers of sub-agent invocations would additionally produce heterogeneous trajectory batches, which recent extensions of GRPO address through specialised grouping and alignment procedures~\citep{zhao2025strongermas,hong2025multi}. We therefore simplify the optimisation problem by collapsing the topology into a single agent, so that each episode is one trajectory generated by one policy, group sampling remains defined at the query level, and the same sequence level advantage can be assigned to the entire trajectory without aligning role specific sub-trajectories. We note that this construction does not resolve credit assignment among individual actions, but it avoids the additional cross-agent grouping problem introduced by the production topology. Figure~\ref{fig:e2e_harness}
shows the resulting rollout structure, in which a single growing message list of alternating assistant turns and tool results is resent in full to the policy on every turn, and tool calls are served from a local document cache where possible
and executed live against the research APIs otherwise.

\paragraph{Generalisation to the Production Multi-Agent Harness.}
Evaluation and deployment retain the multi-agent harness, so improvements learned in the collapsed environment must transfer across topology. We construct the training mixture to support this transfer by combining two complementary levels of supervision. RL Stage~1 optimises the planner, agent, compaction, and reporter on recorded prefixes from the multi-agent harness
, so research planning, evidence conditioned tool use, document compaction, and grounded report writing are trained on states drawn directly from the production topology, while end-to-end training complements these local objectives by optimising their composition within a complete trajectory generated by the current policy.

The one step tasks can be viewed more precisely as on-policy continuations from an off-policy state distribution. Each prefix is an intermediate research state recorded from a previous harness trajectory rather than generated by the current policy, while the continuation from that prefix is sampled from the current policy. These prefixes therefore expose the learner to valid intermediate states that may be rare under its present end-to-end distribution and permit direct optimisation of the subtask responsible for each state, whereas end-to-end rollouts preserve the state distribution induced by the current policy and train interactions among the subtasks. This mixture is motivated by the established trade-off between the broader coverage and sample reuse available from off-policy data and the direct policy improvement provided by on-policy
updates~\citep{gu2016q,gu2017interpolated,nair2020awac}. We expect the broader state coverage to support transfer and treat the observed multi-agent evaluation rather than this motivation alone as evidence of generalisation.

\begin{figure}[htbp]
  \centering
  \includegraphics[width=0.65\linewidth]{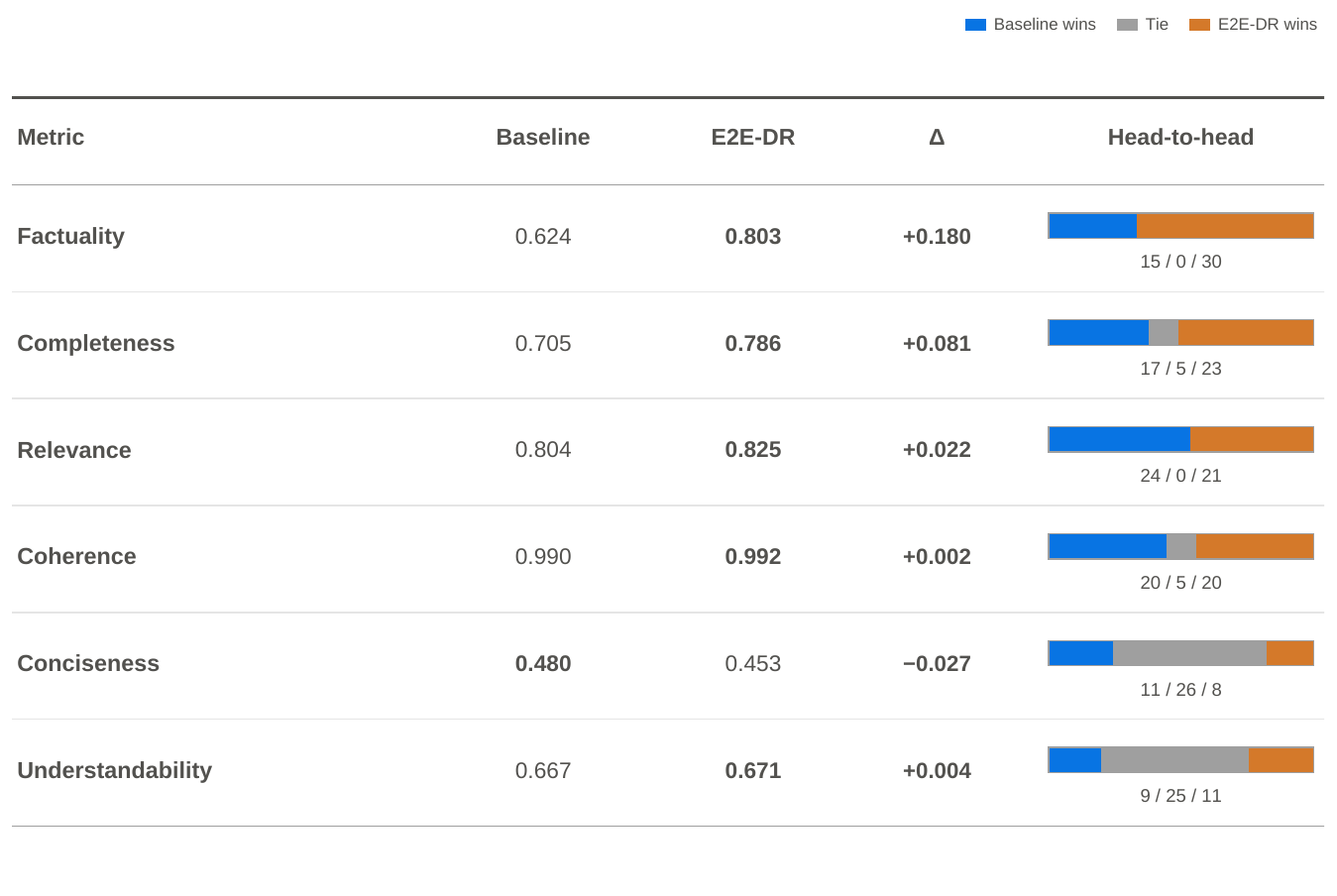}
  \vspace{4pt}
  \includegraphics[width=0.9\linewidth]{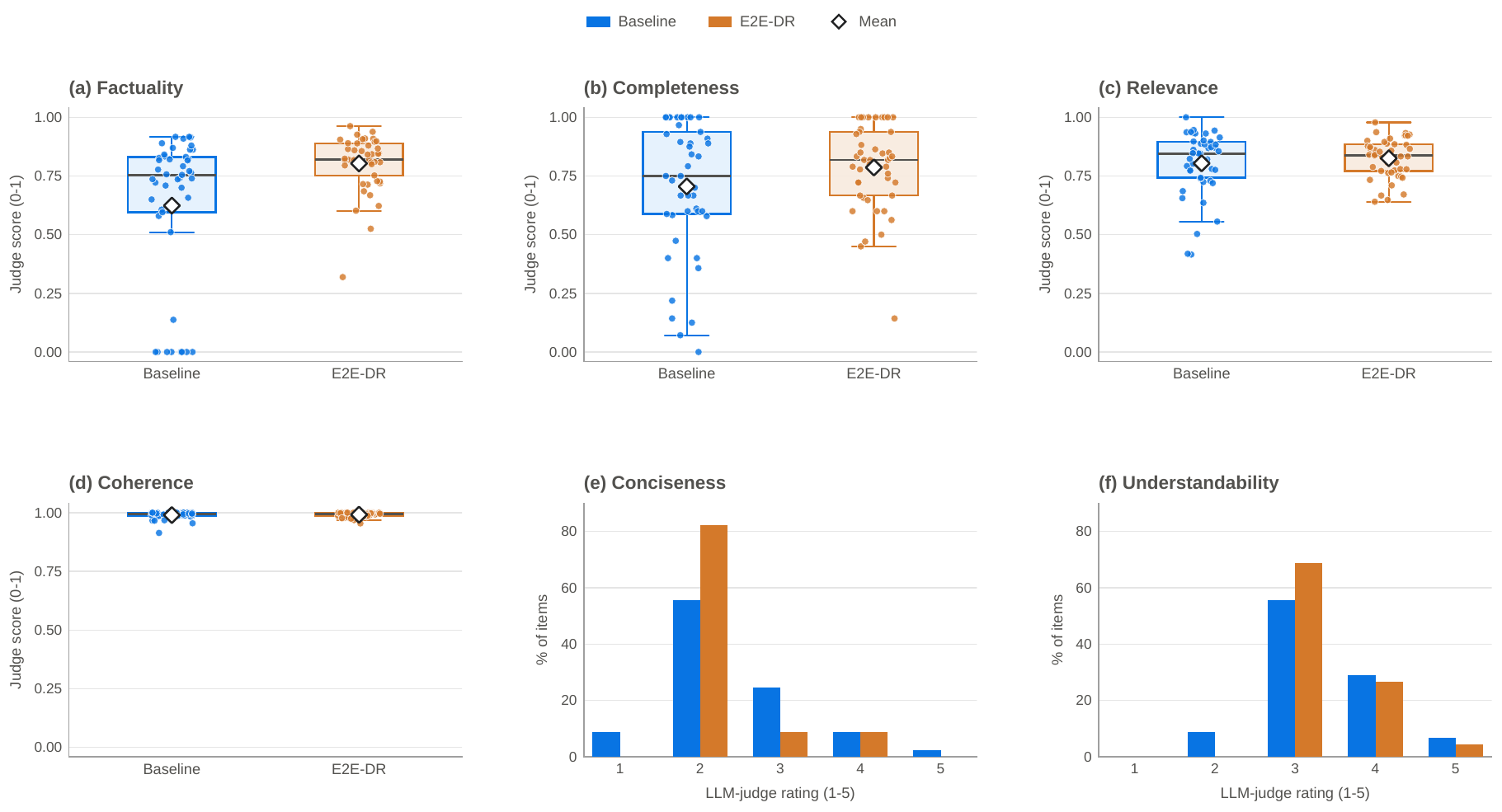}
  \caption{Comparison inside the multi-agent harness between the RL
  initialisation and a checkpoint trained in the single-agent end-to-end
  environment. The trained checkpoint improves factuality, completeness,
  relevance, coherence, and understandability, with the largest gains on
  factuality and completeness.}
  \label{fig:dr-metrics}
\end{figure}

Figure~\ref{fig:dr-metrics} evaluates that transfer by scoring both the checkpoint trained in the single-agent environment and its RL initialisation inside the multi-agent harness. The trained checkpoint improves five of the six reported dimensions, with the largest gains in factuality and completeness, indicating that the learned capabilities are not restricted to the collapsed trajectory, although the comparison does not establish that single-agent optimisation is equivalent to training the multi-agent system.

The collapsed environment therefore remains a practical approximation, and a more complete treatment would estimate role and state conditioned values or assign local rewards that identify the contribution of each sub-agent trajectory, permitting sub-agent updates without a group-relative advantage for every realised prefix. Centralised critics and counterfactual advantages provide established approaches to multi-agent credit assignment, while recent multi-agent language model objectives introduce agent specific rewards and hierarchical group-relative advantages~\citep{foerster2018coma,yu2022surprising,zhao2025strongermas,hong2025multi}.

\subsubsection{Exploration of Training Mechanisms}
\label{sec:rl-results}

\begin{figure}[htbp]
\centering
\includegraphics[width=1.0\linewidth]{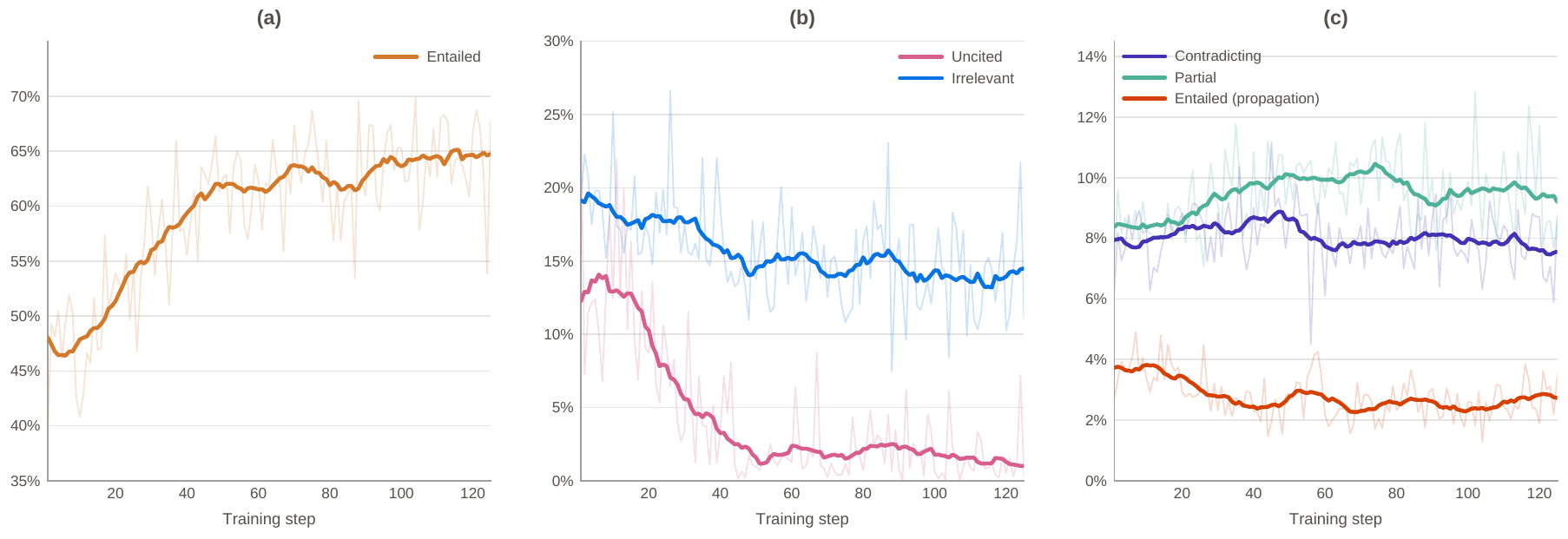}
\caption{Claim-Entailment Composition over RL Stage 1 of \texttt{Thomson-1.0-Small}. \textbf{(a)} Claims entailed in citations rise from about 47\% to 65\% of all claims as training progresses. \textbf{(b)} Uncited claims collapse from about 13\% to roughly 1\%, while irrelevant claims decline more modestly, from roughly 19\% to 14\%. \textbf{(c)} Contradicting, Partial, and Entailed (propagation) claims each stay within a narrow, largely flat band across training, with no strong directional trend. Judged claims per step (faint) with rolling-mean over 15 step window (bold). Each panel's y-axis is scaled to its own data range.
}
\label{fig:factuality-claim-compostion}
\end{figure}

\paragraph{Claim-Entailment Composition.}
Figure~\ref{fig:factuality-claim-compostion} decomposes the citation factuality reward into its constituent claim outcomes throughout RL Stage 1 training for \texttt{Thomson-1.0-Small}. The dominant trend is the rise in \emph{entailed} claims, which climb by roughly 18 percentage points (from 47\% to 65\%) with the strongest correlation of any series to training progress (r = 0.75). The second-largest change is the decline of \emph{uncited} claims, which collapse from 13\% to roughly 1\% (r = -0.67) and essentially saturate by the midpoint of training. The number of \emph{irrelevant} claims decreases too, albeit only by roughly 5 percentage points, plateauing around 14\%. The three remaining statuses -- \emph{contradicting}, \emph{partially entailed}, and \emph{entailed-by-propagation} -- stay mostly flat. Contradicting claims non-monotonically rise to a peak of 9.0\% around two-fifths of the way through training before slowly declining, showing a negligible net change of -0.2 percentage points. The portion of claims that were partially entailed or could recover through propagation stayed consistent throughout.

The increase of \emph{entailed} claims comes from the highest-weighted reward for this category. The rapid decline of \emph{uncited} claims reflects the easily learnable solution of attaching citations to \emph{uncited} paragraphs. While the portion of \emph{irrelevant} claims decreases more slowly, the reward policy is also making real progress on the harder skill of choosing an on-topic citation rather than just adding any citation. Reassuringly, neither \emph{contradicting} nor \emph{partial} claims trends upward: gains in entailment are not coming at the cost of more confidently wrong claims. The share of claims \emph{entailed by propagation} shrinks slightly, further emphasising that direct entailment is increasingly carrying the load.

\paragraph{Reward Curves.}
\begin{figure}[htbp]
\centering
\includegraphics[width=1.0\linewidth]{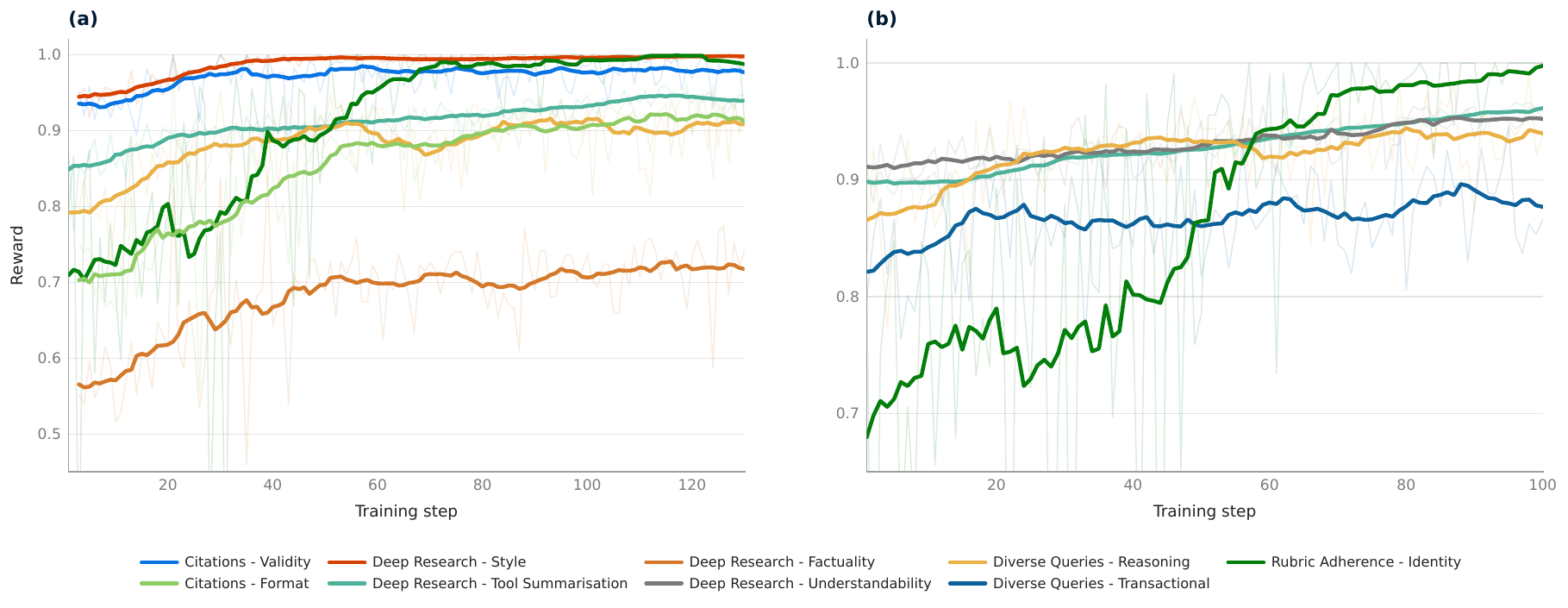}
\caption{Selected Reward Curves for RL Stage~1. Both \texttt{Thomson-1.0-Small} \textbf{(a)} and \texttt{Thomson-1.0-Large} \textbf{(b)} show steady improvement across a variety of rewards. For both model sizes, the sharpest gain appears in rubric adherence for multilingual model identity. The Deep Research understandability reward climbs steadily throughout training (approximately 0.91 to 0.95 for Small), while completeness stays comparatively flat and noisy, ticking up modestly only late in training. Among diverse-queries rewards, general legal shows the clearest sustained gain (approximately 0.87 to 0.94 for Small); drafting rises early but gives back some gain near the end; transactional improves early but plateaus below the other two for the remainder of the run. These patterns demonstrate both broad improvements across task families and model-size-dependent learning dynamics. Bold lines show a 15-step centred rolling mean over raw per-step values (faint lines); each panel uses its own y-axis range.}
\label{fig:rl-large-1-curves}
\end{figure}

Figure~\ref{fig:rl-large-1-curves} presents reward curves for selected tasks during RL Stage~1, revealing both shared and model-size-specific learning dynamics. Both the small and large models exhibit steady improvement across diverse task families, with the sharpest gains in multilingual identity adherence. Deep Research rewards show sustained improvement in understandability while completeness remains noisier, suggesting that clarity is more readily optimised than comprehensive rubric coverage. The Diverse Queries rewards display heterogeneous learning profiles: general legal reasoning shows consistent gains, drafting improves early but partially regresses, and transactional plateaus after initial improvement. These patterns support the staged training strategy, with RL Stage~1 establishing broad capabilities that RL Stage~2 can build upon.

\subsection{Training Compute}
\label{ssec:compute}

We report training compute for \texttt{Thomson~1} following the \emph{hardware-based approach} of the EU AI Act GPAI guidelines (Annex~A.2.1). For a block of $N$ identical accelerators run for a duration $L$ (in seconds) at peak theoretical throughput $H$ (FLOP/s) and average utilisation $U$, the compute is
\begin{equation}
  C \;=\; N \cdot L \cdot H \cdot U ,
  \label{eq:eu-hw-compute}
\end{equation}
and blocks with different $N$ (or run over different periods) are summed.

\paragraph{Assumptions.}
All stages ran on Nvidia~B200 GPUs. We take $H = 2.25\times10^{15}\ \text{FLOP/s}$, the dense BF16 tensor-core peak of the B200. Utilisation $U$ varies by stage and is estimated from measured small-model training: CPT achieves $U \approx 0.85$ (high efficiency on long sequences), DPO stages $U \approx 0.66$ (moderate efficiency with preference pairs), RL Stage~1 $U \approx 0.35$ (lower efficiency due to short-context diverse tasks and asynchronous reward computation), and RL Stage~2 $U \approx 0.56$ (improved efficiency on longer-context uniform tasks). Each reinforcement-learning stage additionally provisions 32 B200 GPUs for concurrent LLM-as-judge (LaJ) reward scoring for the full duration of the stage; per Equation~\eqref{eq:eu-hw-compute}, these are counted as a separate hardware block at the same $L$.

\paragraph{Thomson-1.0-Large.}
Table~\ref{tab:compute-thomson10-large} provides the EU AI Act hardware-based breakdown (Annex~A.2.1).

\begin{table}[h]
\centering\small
\setlength{\tabcolsep}{5pt}
\renewcommand{\arraystretch}{1.2}
\begin{tabular}{@{}l r r c c r@{}}
\toprule
\textbf{Stage (block)} & \textbf{$N$} & \textbf{$L$ (s)} & \textbf{$H$ (FLOP/s)} & \textbf{$U$} & \textbf{$C=NLHU$ (FLOP)} \\
\midrule
CPT                  & 128 & 1{,}021{,}968 & $2.25\times10^{15}$ & 0.85 & $2.50\times10^{23}$ \\
DPO Stage 1          & 64  & 22{,}320      & $2.25\times10^{15}$ & 0.66 & $2.13\times10^{21}$ \\
DPO Stage 2          & 64  & 34{,}020      & $2.25\times10^{15}$ & 0.66 & $3.25\times10^{21}$ \\
RL Stage 1 -- train & 208 & 230{,}400     & $2.25\times10^{15}$ & 0.35 & $3.76\times10^{22}$ \\
RL Stage 1 -- LaJ   & 32  & 230{,}400     & $2.25\times10^{15}$ & 0.35 & $5.80\times10^{21}$ \\
RL Stage 2 -- train & 224 & 494{,}208     & $2.25\times10^{15}$ & 0.56 & $1.40\times10^{23}$ \\
RL Stage 2 -- LaJ   & 32  & 494{,}208     & $2.25\times10^{15}$ & 0.56 & $2.00\times10^{22}$ \\
\midrule
\textbf{Total}       &     &               &                     &      & $\mathbf{4.60\times10^{23}}$ \\
\midrule
\multicolumn{5}{@{}l}{\textit{GPU-hours}} & 87{,}842 \\
\multicolumn{5}{@{}l}{\textit{Hardware peak ($U=1$)}} & $7.12\times10^{23}$ \\
\bottomrule
\end{tabular}
\caption{\texttt{Thomson-1.0-Large} training compute (B200), EU AI Act hardware-based breakdown (Annex~A.2.1). Each constant-$N$ block is evaluated with Equation~\eqref{eq:eu-hw-compute} and summed. Utilisation values are estimated from small-model training (CPT: $85.0\pm1.2\%$; DPO: $66.4\pm1.8\%$; RL Stage~1: $35.0\pm16.5\%$; RL Stage~2: $56.3\pm19.3\%$).}
\label{tab:compute-thomson10-large}
\end{table}

\paragraph{Thomson-1.0-Small.}
Table~\ref{tab:compute-thomson10-small} provides the EU AI Act hardware-based breakdown for the small model, which provided the empirical utilisation measurements applied to the large model estimates.

\begin{table}[h]
\centering\small
\setlength{\tabcolsep}{5pt}
\renewcommand{\arraystretch}{1.2}
\begin{tabular}{@{}l r r c c r@{}}
\toprule
\textbf{Stage (block)} & \textbf{$N$} & \textbf{$L$ (s)} & \textbf{$H$ (FLOP/s)} & \textbf{$U$} & \textbf{$C=NLHU$ (FLOP)} \\
\midrule
CPT                  & 128 & 339{,}984 & $2.25\times10^{15}$ & 0.85 & $8.33\times10^{22}$ \\
DPO Stage 1          & 16  & 28{,}296  & $2.25\times10^{15}$ & 0.66 & $6.76\times10^{20}$ \\
RL Stage 1 -- train & 96  & 422{,}568 & $2.25\times10^{15}$ & 0.35 & $3.19\times10^{22}$ \\
RL Stage 1 -- LaJ   & 32  & 422{,}568 & $2.25\times10^{15}$ & 0.35 & $1.06\times10^{22}$ \\
RL Stage 2 -- train & 176 & 137{,}916 & $2.25\times10^{15}$ & 0.56 & $3.08\times10^{22}$ \\
RL Stage 2 -- LaJ   & 32  & 137{,}916 & $2.25\times10^{15}$ & 0.56 & $5.59\times10^{21}$ \\
\midrule
\textbf{Total}       &     &           &                     &      & $\mathbf{1.63\times10^{23}}$ \\
\midrule
\multicolumn{5}{@{}l}{\textit{GPU-hours}} & 35{,}207 \\
\multicolumn{5}{@{}l}{\textit{Hardware peak ($U=1$)}} & $2.85\times10^{23}$ \\
\bottomrule
\end{tabular}
\caption{\texttt{Thomson-1.0-Small} training compute (B200), EU AI Act hardware-based breakdown (Annex~A.2.1). LaJ (LLM-as-judge) nodes for reward scoring ran concurrently with training on separate hardware (4 nodes $\times$ 8 GPUs = 32 GPUs for each RL stage). No DPO Stage~2 was trained for the small model. The stage-specific utilisation values measured from these training runs inform the large model estimates in Table~\ref{tab:compute-thomson10-large}.}
\label{tab:compute-thomson10-small}
\end{table}

\paragraph{Snowdon-1.0-Large.}
Table~\ref{tab:compute-snowdon10-large} provides the EU AI Act hardware-based breakdown for the \texttt{Snowdon-1.0-Large} value re-alignment model.

\begin{table}[h]
\centering\small
\setlength{\tabcolsep}{5pt}
\renewcommand{\arraystretch}{1.2}
\begin{tabular}{@{}l r r c c r@{}}
\toprule
\textbf{Stage (block)} & \textbf{$N$} & \textbf{$L$ (s)} & \textbf{$H$ (FLOP/s)} & \textbf{$U$} & \textbf{$C=NLHU$ (FLOP)} \\
\midrule
Abliteration Trial & 16 & 4{,}830 & $2.25\times10^{15}$ & 0.30 & $5.22\times10^{19}$ \\
Constitutional DPO       & 64 & 10{,}617  & $2.25\times10^{15}$ & 0.30 & $4.59\times10^{20}$ \\
\midrule
\textbf{Total}      &    &           &                     &      & $\mathbf{5.11\times10^{20}}$ \\
\midrule
\multicolumn{5}{@{}l}{\textit{GPU-hours}} & 210.2 \\
\multicolumn{5}{@{}l}{\textit{Hardware peak ($U=1$)}} & $1.70\times10^{21}$ \\
\bottomrule
\end{tabular}
\caption{\texttt{Snowdon-1.0-Large} training compute (B200), EU AI Act hardware-based breakdown (Annex~A.2.1). This model represents the value re-alignment stage prior to \texttt{Thomson-1.0-Large} (Section~\ref{ssec:values}).}
\label{tab:compute-snowdon10-large}
\end{table}

\paragraph{Snowdon-1.1-Small.}
Table~\ref{tab:compute-snowdon11-small} provides the EU AI Act hardware-based breakdown for the \texttt{Snowdon-1.1-Small} value re-alignment model.

\begin{table}[h]
\centering\small
\setlength{\tabcolsep}{5pt}
\renewcommand{\arraystretch}{1.2}
\begin{tabular}{@{}l r r c c r@{}}
\toprule
\textbf{Stage (block)} & \textbf{$N$} & \textbf{$L$ (s)} & \textbf{$H$ (FLOP/s)} & \textbf{$U$} & \textbf{$C=NLHU$ (FLOP)} \\
\midrule
Abliteration Trial & 8  & 900 & $2.25\times10^{15}$ & 0.30 & $4.86\times10^{18}$ \\
Constitutional DPO       & 16 & 10{,}674  & $2.25\times10^{15}$ & 0.30 & $1.15\times10^{20}$ \\
\midrule
\textbf{Total}      &    &           &                     &      & $\mathbf{1.20\times10^{20}}$ \\
\midrule
\multicolumn{5}{@{}l}{\textit{GPU-hours}} & 49.4 \\
\multicolumn{5}{@{}l}{\textit{Hardware peak ($U=1$)}} & $4.00\times10^{20}$ \\
\bottomrule
\end{tabular}
\caption{\texttt{Snowdon-1.1-Small} training compute (B200), EU AI Act hardware-based breakdown (Annex~A.2.1). This model represents the value re-alignment stage prior to \texttt{Thomson-1.0-Small} (Section~\ref{ssec:values}).}
\label{tab:compute-snowdon11-small}
\end{table}

\paragraph{Architecture-based cross-check.}
As an independent estimate, the architecture-based approach (Annex~A.2.2) gives $C \approx 6\,P\,D$ for a dense transformer with $P$ parameters and $D$ training tokens; for our mixture-of-experts architecture, $P$ is the number of \emph{active} parameters per token.

\paragraph{Summary.}
Aggregating all stages, the estimated end-to-end training compute for \texttt{Thomson-1.0-Large} is approximately $4.6\times10^{23}$ FLOP, corresponding to approximately $8.8\times10^{4}$ B200 GPU-hours, while \texttt{Thomson-1.0-Small} consumed approximately $1.6\times10^{23}$ FLOP across $3.5\times10^{4}$ GPU-hours. The preceding value re-alignment models (\texttt{Snowdon-1.0-Large} and \texttt{Snowdon-1.1-Small}) contributed an additional $5.1\times10^{20}$ FLOP and $1.2\times10^{20}$ FLOP respectively. The large model estimate uses stage-specific utilisation values measured from small-model training, yielding approximately $2.2\times$ higher compute than a conservative uniform 30\% utilisation assumption. The increase is driven primarily by CPT (85\% utilisation on long sequences) and RL Stage~2 (56\% utilisation on long-context uniform tasks), which together account for 85\% of total compute. Note the substantial variance in RL Stage~1 utilisation ($35.0\pm16.5\%$), reflecting variable efficiency across diverse short-context task collections and asynchronous reward computation overhead. Because both the actual compute ($4.6\times10^{23}$ FLOP) and hardware-provisioned peak ($7.1\times10^{23}$ FLOP at $U=1$) sit well below the $10^{25}$ FLOP limit, neither model triggers the EU AI Act's stricter systemic-risk compliance tier.

\newpage
\section{Evaluation}
\label{sec:evaluation}
The sovereignty principles of Section~\ref{ssec:sovereign_ai_through_continual_learning} concern control over the model, the data, the tools, the infrastructure and the economics of a deployment. Evaluation is where several of them are operationalised.  An institution may hold model weights, the training corpus and the serving stack and still take its operative definition of adequate performance from external parties rather than tracking the specific values or use cases the institution cares about. We therefore treat evaluation as serving a dual role: substantiating capability claims while also forming a key part of the sovereignty stack itself.

A second challenge is that evaluating a model under Continual Learning differs fundamentally from evaluating one trained from scratch: the evaluation must establish not only final capability, but also what changed relative to the starting checkpoint. First, we apply a series of mid-training and post-training interventions to an already capable instruction-tuned checkpoint. The quantity of interest is therefore the change induced relative to the input checkpoint, across both targeted and untargeted capabilities. Second, forgetting manifests itself as an erosion of specific capabilities that is identifiable only relative to a baseline. By construction, such degradation may occur precisely in capabilities that target-domain benchmarks do not measure. Detecting it therefore requires broad benchmark coverage.

Our evaluation is correspondingly broad. We assess performance across diverse professional domains, agentic Deep Research, general-purpose capability preservation, expert human preference and quality evaluations, test-time scaling and safety/alignment. Together, these evaluations measure both target-domain specialisation and the overall quality of the resulting system. Consequently, our evaluation suite is partitioned into six broad areas:

\begin{enumerate}
    \item \textbf{Professional domain evaluations}, covering the target capabilities;
    \item \textbf{Agentic evaluations}, covering the integration of those capabilities under long-horizon tool use, principally end-to-end Deep Research;
    \item \textbf{General-purpose benchmarks}, covering capability preservation in domains that were not targeted.
    \item \textbf{Expert Human Preferences \& Quality Evaluation}, we conduct an expert preference study with practising legal subject-matter experts rather than generalist annotators, using prompts authored by the evaluators to reflect their own practical task distributions across legal and general-domain queries
    \item \textbf{Test-Time Scaling}, we assess the capability gains achievable with additional test time compute.
    \item \textbf{Safety and Alignment}, we assess safety and alignment using a target-independent red-team corpus.
\end{enumerate}

\subsection{Professional Domain Evaluation}

We evaluate the professional domain using a combination of public benchmarks (External) and institution-specific benchmarks (Internal). We first describe the breakdown of each category in §\ref{subsubsec:professional_bench_breakdown}, as summarised in Table~\ref{tab:legal_tax_composition}. We then describe each internal benchmark, our rationale for maintaining the internal evaluation suite, and the general construction process in §\ref{subsubsec:proprietary-benchmarks}.

\begin{table}[htbp]
\footnotesize
\begin{tabular}{@{} >{\bfseries}l >{\bfseries}l p{9.4cm} @{}}
\toprule
\normalfont\bfseries Domain & \normalfont\bfseries Category & \normalfont\bfseries Constituent Benchmarks / Datasets \\
\midrule

\rowcolor{cToolBg}
\multirow{10}{*}{Legal}
& Stanford LB (162) & LegalBench~\citep{guha2023legalbench} \\
\cmidrule(l){2-3}

& Info.\ Retrieval (3) & COLIEE Task 1~\citep{coliee,goebel2024coliee}, AALP Quality, Best Headnote \\
\cmidrule(l){2-3}

\rowcolor{cToolBg}
& Reasoning (12) & COLIEE Task 2~\citep{coliee,goebel2024coliee}, CaseHOLD~\citep{zheng2021casehold}, Function of Decision Section~\citep{guha2023legalbench}, Learned Hands~\citep{learnedhands,guha2023legalbench}, Legal Support~\citep{liang2022helm}, MBE Bar Exam~\citep{ncbe}, Parentheticals, ReClor~\citep{yu2020reclor}, SCALR~\citep{guha2023legalbench}, SuperGPQA (Law)~\citep{map2025supergpqa}, LEXam -- MCQ (4-choice)~\citep{fan2025lexam}, PRBench Legal Hard~\citep{akyurek2025prbench} \\
\cmidrule(l){2-3}

& Classification (4) & EUR-Lex~\citep{chalkidis2021multieurlex,chalkidis2022lexglue}, LEDGAR~\citep{tuggener2020ledgar,chalkidis2022lexglue}, SCOTUS~\citep{spaeth2020scdb,chalkidis2022lexglue}, Headnote Type \\
\cmidrule(l){2-3}

\rowcolor{cToolBg}
& Doc.\ Processing \& RAG (2) & Legal RAG, Document Review \\
\cmidrule(l){2-3}

Legal & Summarisation (4) & BillSum US~\citep{kornilova2019billsum}, BillSum CA~\citep{kornilova2019billsum}, CourtWire, Material Facts \\
\cmidrule(l){2-3}

\rowcolor{cToolBg}
& Contract Under. (6) & CUAD~\citep{hendrycks2021cuad,guha2023legalbench}, MAUD~\citep{wang2023maud,guha2023legalbench}, OPP-115~\citep{wilson2016opp115,guha2023legalbench}, Privacy Policy Entailment~\citep{zimmeck2019app350,guha2023legalbench}, Insurance Policy Interpretation~\citep{waldon2023consensus,guha2023legalbench}, ContractScrub~\citep{bang2026contractscrub} \\
\cmidrule(l){2-3}

& Human Queries (5) & Diverse Queries -- Documents (Groundedness), Diverse Queries -- General Legal, Diverse Queries -- Drafting, Diverse Queries -- Transactional, Diverse Queries -- Instruction Following \\
\cmidrule(l){2-3}

\rowcolor{cToolBg}
& Deep Research (1) & Deep Research \\
\cmidrule(l){2-3}

& Harvey Legal Agent Bench. (1) & Harvey LAB~\citep{harveylab2026} \\
\midrule

\rowcolor{cToolBg}
\multirow{2}{*}{Tax}
& Deep Research (1) & Deep Research \\
\cmidrule(l){2-3}

& Tax QA (1) & Tax QA \\
\midrule

\rowcolor{cToolBg}
\begin{tabular}[c]{@{}l@{}}Journalism\end{tabular}
& Deep Research (1) & Deep Research \\
\midrule

\begin{tabular}[c]{@{}l@{}}Safety /\\Values\end{tabular}
& Robustness (2) & Principal Hierarchy -- Execution (Legal)~\citep{yu2026principal}, Query Sufficiency~\citep{insuffiencybench2026} \\

\bottomrule
\end{tabular}
\caption{Constituent benchmarks and datasets underlying the Professional Domain evaluation category in Table~\ref{tab:benchmarks-large}. The number in parentheses after each category name gives its constituent task count. Full data cards for each dataset, including format, metric, provenance, citation and item counts, are given in Appendix~\ref{app:datacards}.}
\label{tab:legal_tax_composition}
\end{table}

\subsubsection{Composition of Professional Benchmarks}
\label{subsubsec:professional_bench_breakdown}

\paragraph{Stanford LegalBench.} We report performance on the Stanford LegalBench suite~\citep{guha2023legalbench}, aggregated over its five legal-reasoning skill categories: Conclusion, Interpretation, Issue, Rhetoric, and Rule. Constructed collaboratively by researchers from the legal and AI communities, LegalBench comprises 162 subtasks spanning issue spotting, rule application, drawing legal conclusions, interpretation of legal text, and rhetorical reasoning. It provides a broad measure of legal reasoning across statutes, contracts, case law, and other legal materials.

\paragraph{Legal Information Retrieval.}
Retrieval capability is evaluated with one external task, COLIEE Task 1~\citep{goebel2024coliee}, and two internal ones AALP Quality (Appendix~\ref{app:dc:legalreas}) and BestHeadnote (Appendix~\ref{app:dc:legalir}). COLIEE Task 1 requires identifying the precedents likely to support a query decision from a corpus of Federal Court of Canada cases, with explicit citations removed from the query case so that retrieval must rely on legal content, fact patterns, reasoning, and precedential relevance.

\paragraph{Legal Reasoning.} This category aggregates 11 external tasks spanning case-law and statutory reasoning, together with one internal task, Parentheticals. On the case-law side, CaseHOLD~\citep{zheng2021casehold} requires identifying the governing holding of a cited decision from more than 53,000 citing contexts, while COLIEE Task 2~\citep{coliee,goebel2024coliee} tests entailment between Canadian case law and unseen decisions as part of the long-running Competition on Legal Information Extraction/Entailment. General logical reasoning under distractor-rich conditions is measured by ReClor~\citep{yu2020reclor}, sourced from GMAT and LSAT items, and by the graduate-level SuperGPQA suite~\citep{map2025supergpqa}. We additionally include Legal Support~\citep{liang2022helm}, Function of Decision Section~\citep{guha2023legalbench}, SCALR~\citep{guha2023legalbench}, and Learned Hands~\citep{learnedhands,guha2023legalbench}. Professional-examination and open-ended legal competence are assessed with the Multistate Bar Examination component of the Uniform Bar Exam~\citep{ncbe}; the four-choice English configuration of LEXam~\citep{fan2025lexam}, a benchmark built from 340 law-school exams across 116 courses; and the Legal Hard subset of PRBench~\citep{akyurek2025prbench}, an expert-authored, rubric-graded benchmark of high-stakes professional reasoning. Collectively, these tasks probe legal inference, precedent application, and professional-examination competence across a broad range of difficulty and reasoning style.

\paragraph{Legal Classification.} Legal document and issue classification is evaluated across three public benchmarks and one internal, Headnote Type. EURLex~\citep{chalkidis2021multieurlex,chalkidis2022lexglue} evaluates multi-label topic classification of European legislative documents, while LEDGAR~\citep{tuggener2020ledgar,chalkidis2022lexglue} evaluates classification of contractual provisions drawn from U.S. Securities and Exchange Commission filings. SCOTUS~\citep{spaeth2020scdb,chalkidis2022lexglue}, distributed as part of LexGLUE, measures issue-area classification of U.S. Supreme Court opinions. Together, these tasks measure whether models can identify substantive legal issues and document functions across case law, legislation, and contracts.

\paragraph{Document Processing and Retrieval-Augmented Generation.} We evaluate with two internal benchmarks specifically focused on reasoning over supplied legal documents: Legal RAG and Document Review. Details are available in Appendix
\ref{app:dc:aar}.

\paragraph{Legal Summarisation.}
Summarisation is evaluated using the U.S. Congressional and California portions of BillSum~\citep{kornilova2019billsum}, with editorial rewriting. Given the full text of a bill, models must produce a concise summary capturing its principal provisions, purposes, and implications. Responses are evaluated against reference summaries across accuracy, completeness, clarity, conciseness, and hallucination, providing a measure of faithful long-document legal summarisation. Additionally, we evaluate with institution-specific benchmarks named CourtWire and Material Facts (Appendix~\ref{app:dc:legalsumm}).

\paragraph{Contract Understanding.} Contract understanding is measured by five external benchmarks and one institution-created benchmark ContractScrub \citep{bang2026contractscrub}. CUAD~\citep{hendrycks2021cuad} is an expert-annotated clause-extraction dataset built with legal experts from The Atticus Project; MAUD~\citep{wang2023maud} is an expert-annotated reading-comprehension dataset grounded in the American Bar Association's 2021 Public Target Deal Points Study; OPP-115~\citep{wilson2016opp115} is a corpus of 115 website privacy policies with fine-grained data-practice annotations; and Privacy Policy Entailment, a LegalBench task derived from the APP-350 corpus of annotated mobile-app privacy policies~\citep{zimmeck2019maps}, tests entailment between policy clauses and candidate practice descriptions; and Insurance Policy Interpretation~\citep{waldon2023consensus,guha2023legalbench} assesses whether an insurance claim is covered under a given policy.

\paragraph{Results on Public Legal Benchmarks.}

\begin{table}[!h]
\centering
\footnotesize
\setlength{\tabcolsep}{3.5pt}
\renewcommand{\arraystretch}{1.05}
\resizebox{\textwidth}{!}{%
\begin{tabular}{l H c c c c c c H c c c c}
\toprule
& \multicolumn{7}{c}{\textbf{Large}} & \multicolumn{5}{c}{\textbf{Small}} \\
\cmidrule(lr){2-8} \cmidrule(lr){9-13}
\textbf{Benchmark} &
\begin{tabular}[b]{@{}c@{}}\textbf{Thomson}\\\textbf{1.0-Large}\end{tabular} &
\begin{tabular}[b]{@{}c@{}}\textbf{Snowdon}\\\textbf{1.0-Large}\end{tabular} &
\begin{tabular}[b]{@{}c@{}}\textbf{Qwen3.5}\\\textbf{397B}\end{tabular} &
\begin{tabular}[b]{@{}c@{}}\textbf{Sonnet}\\\textbf{5}\end{tabular} &
\begin{tabular}[b]{@{}c@{}}\textbf{Gemini}\\\textbf{3.1 Pro}\end{tabular} &
\begin{tabular}[b]{@{}c@{}}\textbf{GPT}\\\textbf{5.4}\end{tabular} &
\begin{tabular}[b]{@{}c@{}}\textbf{Opus}\\\textbf{4.8}\end{tabular} &
\begin{tabular}[b]{@{}c@{}}\textbf{Thomson}\\\textbf{1.0-Small}\end{tabular} &
\begin{tabular}[b]{@{}c@{}}\textbf{Snowdon}\\\textbf{1.1-Small}\end{tabular} &
\begin{tabular}[b]{@{}c@{}}\textbf{Qwen3.6}\\\textbf{35B}\end{tabular} &
\begin{tabular}[b]{@{}c@{}}\textbf{Gemma}\\\textbf{4-31B}\end{tabular} &
\begin{tabular}[b]{@{}c@{}}\textbf{Haiku}\\\textbf{4.5}\end{tabular} \\
\midrule
PRBench Hard              & \textbf{31.6} & 29.2 & 29.2 & 28.6 & 29.3 & 31.3 & 31.5 & \textbf{31.4} & 25.9 & 26.9 & 25.2 & 19.3 \\
Stanford LegalBench       & 82.3 & 82.8 & 78.8 & 81.4 & \textbf{84.3} & 82.3 & 81.8 & 79.9 & 80.9 & 80.3 & \textbf{83.1} & 80.7 \\
Lexam MCQ4 (en)           & 82.9 & 79.8 & 83.5 & 91.3 & 91.0 & 89.0 & \textbf{93.9} & 72.2 & 75.6 & 75.8 & \textbf{87.4} & 72.2 \\
MBE Bar Exam                  & 90.8 & 88.0 & 86.8 & 90.3 & \textbf{95.9} & 93.6 & 94.7 & 83.4 & 83.1 & 80.1 & \textbf{88.8} & 77.3 \\
Contract Scrub            & 56.3 & 55.4 & 40.9 & 51.1 & 59.6 & \textbf{65.4} & 61.7 & 44.6 & 38.5 & 39.6 & \textbf{54.8} & 36.5 \\
Query Sufficiency       & \textbf{57.9} & 51.1 & 52.7 & 55.8 & 56.2 & 51.0 & 57.0 & \textbf{59.4} & 54.8 & 55.7 & 51.7 & 47.4 \\
Harvey Legal Agent Bench. & 85.7 & 56.3 & 70.6 & 80.9 & 55.5 & 76.1 & \textbf{86.9} & \textbf{73.4} & 71.5 & 69.5 & 30.2 & 60.5 \\
\bottomrule
\end{tabular}}
\caption{Results for common open legal benchmarks for large and small model variants.}
\label{tab:legal-benchmarks}
\end{table}

Table~\ref{tab:legal-benchmarks} provides a benchmark-level view of the legal results summarised more broadly in Table~\ref{tab:benchmarks-large}. Across this selected set of public evaluations, \texttt{Thomson-1.0-Large} improves on the \texttt{Qwen3.5-397B} starting checkpoint on six of seven benchmarks, increasing the average score from 63.2 to 69.6. These gains result in a model comparable to the frontier (e.g. Opus-level model), at a fraction of the cost. The gains span different forms of legal capability rather than a single task format, including Harvey Legal Agent Benchmark (70.6 $\rightarrow$ 85.7), ContractScrub (40.9 $\rightarrow$ 56.3), Query Sufficiency (52.7 $\rightarrow$ 57.9), and the MBE (86.8 $\rightarrow$ 90.8). Performance on LEXam remains approximately unchanged (83.5 vs.\ 82.9).

The same pattern is visible at the smaller scale. \texttt{Thomson-1.0-Small} increases the average across these benchmarks from 54.6 for \texttt{Qwen3.6-35B} to 63.5, with particularly large gains on Harvey Legal Agent Benchmark and ContractScrub. These results complement the broader professional-domain aggregates by showing that the legal improvements induced by Continual Learning are distributed across multiple independently developed public evaluations.

\subsubsection{Institution-Specific Evaluation Details}
\label{subsubsec:proprietary-benchmarks}

Alongside the public benchmarks discussed above (and summarised in
Appendix~\ref{app:datacards}), we evaluate \texttt{Thomson} on a substantial suite of
benchmarks built in-house with human domain experts. This section explains why we
maintain this suite rather than relying on external benchmarks alone, the internal benchmark coverage, and
describes the general process we follow to build it. Every internal benchmark
referenced below is documented in
Appendix~\ref{app:datacards}, marked \textsc{internal} in its Source field.

We curate internal benchmarks alongside external ones for two reasons:
\paragraph{Contamination and saturation erode a public benchmark's power to
discriminate between frontier models over time.} A benchmark's questions and
gold answers are, once published, part of the public web and therefore can be
part of the training distribution of every subsequent model. Independent of
any deliberate contamination, widely used benchmarks also saturate: as
models cluster near the ceiling on an established public task, the residual
variance left to separate a frontier model from its predecessor shrinks, and
small score differences stop being meaningful signal. An internal benchmark
that is authored, scored, and held by us is immune to both failure modes by
construction: it cannot appear in any external training corpus, and we
control its difficulty and can retire or refresh items once they saturate,
which we cannot do for a benchmark we do not own.

\paragraph{External benchmarks anchor comparison but miss the high-stakes
professional work that matters most.} Public benchmarks such as COLIEE~\citep{goebel2024coliee},
LegalBench~\citep{guha2023legalbench} or CaseHOLD~\citep{zheng2021casehold} are indispensable for situating
\texttt{Thomson} against the broader field on tasks the community has agreed to
standardise on. But standardisation is exactly what limits them: a
benchmark designed to be reproducible and model-agnostic is, by
construction, decoupled from the messy, high-stakes work professionals
actually do. None of the public legal benchmarks (Appendix~\ref{app:dc:legalreas}--\ref{app:dc:contract}) test retrieval and
synthesis over a live, multi-document corpus at production context lengths,
drafting a document under a real practitioner's constraints, or a
multi-agent deep-research harness producing a citation-grounded report.
Therefore, we curate a substantial internal benchmark suite to measure the capabilities that matter most for professional use.
The main areas are: legal document processing and retrieval-augmented
generation, realistic human legal queries, professional domain deep
research, and robustness. These are not niche categories: they include the Diverse
Queries suite of SME-authored queries spanning drafting, transactional
work, and instruction following (Appendix~\ref{app:dc:diversequeries}); the
Deep Research benchmark, scored against SME rubrics spanning US and UK law
(Appendix~\ref{app:dc:deepresearch}); and Tax QA, merged from two
independently vetted internal sources (Appendix~\ref{app:dc:tax}). Across
all 20 internal benchmarks, SMEs curated more than 11,000 evaluation items.
Building the benchmark ourselves is the only way to measure the capability we actually ship.

Our internal benchmark coverage is as follows:
\paragraph{Legal Reasoning and Retrieval.}
Our internal evaluations supplement public legal-reasoning and retrieval
benchmarks with tasks designed around finer-grained judgements.
\textit{Parentheticals} tests whether the model can determine whether a cited
passage directly supports, indirectly supports, or contradicts a proposition
in a judicial opinion. Because the relationship cannot be recovered from
lexical similarity alone, the task probes understanding of how authorities
are actually used in legal argument. \textit{Best Headnote} similarly asks the
model to distinguish material that is directly responsive to a legal search
query from material that merely shares terminology, while \textit{AALP
Quality} requires retrieval and synthesis across multiple practice notes,
standard documents, and procedural guides to produce a supported legal
answer. Appendix~\ref{app:dc:legalreas} and
Appendix~\ref{app:dc:legalir} provide full task definitions.

\paragraph{Legal Classification and Summarisation.}
\textit{Headnote Type} evaluates whether a model can assign legal headnotes
to the most appropriate legal practice area, including examples that lie near
the boundaries between related areas. We additionally evaluate two forms of
legal summarisation. \textit{CourtWire} requires reducing a civil
complaint to a single sentence identifying the core reason for the
litigation, testing accurate extreme compression. \textit{Material Facts}
instead asks the model to identify the facts on which a court actually relied
in resolving a legal issue, distinguishing those facts from background
information, procedural history, and legal conclusions. These evaluations
therefore test not only compression, but also whether a model can identify which
parts of a legal document matter for a particular purpose.

\paragraph{Document-processing and retrieval augmented generation (RAG).}
Two evaluations focus specifically on reasoning over supplied legal
documents. \textit{Legal RAG} presents primary-law materials such as cases,
statutes, and regulations and requires an attorney-style answer supported by
paragraph-level citations. It evaluates both overall response quality and the
precision and recall with which relevant material is incorporated.
\textit{Document Review} evaluates question answering over substantially
larger document collections: the system must first identify relevant
documents and passages and then synthesise an answer that is compared against
a gold response. Together, these tasks test the retrieval--reasoning interface
that underlies document-intensive workflows.

\paragraph{Realistic Practitioner Queries.}
The \textit{Human Queries} suite broadens evaluation beyond fixed benchmark
formats using queries authored by legal SMEs. It is partitioned into five
subsets designed to isolate different aspects of assistance.
\textit{General Legal} evaluates doctrinal and fact-sensitive legal reasoning;
\textit{Documents} measures claim-level groundedness against attached source
documents; \textit{Drafting} evaluates legal document generation using both a
drafting rubric and a gate requiring the model to produce the requested
document type; \textit{Transactional} focuses on deal and commercial work
product; and \textit{Instruction Following} isolates compliance with
item-specific requirements on length, structure, and output format. This
decomposition allows improvements in general legal reasoning to be separated
from improvements in groundedness, drafting, transactional practice, and
instruction adherence.

\paragraph{Tax Question Answering.}
Professional-domain evaluation also includes \textit{Tax QA}, an SME-curated
benchmark constructed from two internal sources. Each item pairs a tax
question with supporting source documents and an expert-vetted answer.
Responses are evaluated independently for correctness, coverage,
consistency, relevance, and groundedness, with groundedness assessed by
decomposing the response into atomic statements and checking whether each is
supported by the supplied sources. The benchmark therefore measures not only
whether a model reaches the correct tax conclusion, but whether the answer is
complete, internally consistent, responsive to the question, and supported by
the underlying authority.

\paragraph{Robustness.}
Finally, we include internal evaluations targeting failure modes that are
poorly represented by conventional capability benchmarks. The
\textit{Principal Hierarchy}~\citep{yu2026principal} suite tests whether a model continues to follow
authoritative information when a user encourages it to do otherwise. In the
legal setting, models must recognise that a cited precedent has been
overruled and avoid advising or drafting an argument that improperly relies
on it. Separate advisory and execution variants distinguish
whether a model can state the correct recommendation from whether it
continues to behave correctly when directly instructed to perform the
inappropriate action. \textit{Query Sufficiency}~\citep{insuffiencybench2026} tests a complementary
failure mode: whether the model recognises that a legal question is missing
facts necessary to answer it, such as jurisdiction or party status, and
identifies those missing elements rather than confidently answering an
under-specified problem.

Taken together, these benchmarks extend evaluation from legal knowledge and
reasoning to the broader set of capabilities required for professional use:
retrieval, document-grounded reasoning, legal summarisation, drafting,
transactional work, instruction following, long-horizon research, calibrated
use of authority, and recognition of insufficient information.

\paragraph{How We Approach Internal Benchmark Construction.}

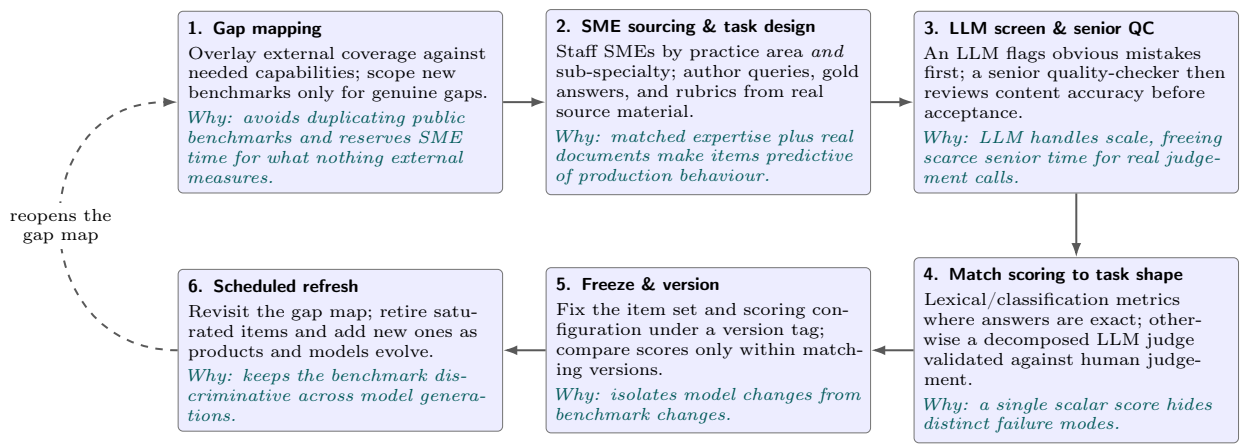
\begin{figure}[htbp]
\centering
\resizebox{\linewidth}{!}{%
\begin{tikzpicture}[
  stage/.style={rectangle, rounded corners=2pt, draw=black!55, fill=blue!7,
    text width=4.3cm, align=left, font=\scriptsize, inner sep=4pt},
  arr/.style={-{Latex[length=2mm, width=1.4mm]}, thick, draw=black!65},
]

\node[stage] (gap) {\textbf{1. Gap mapping}\\[2pt]
Overlay external coverage against needed capabilities; scope new benchmarks
only for genuine gaps.\\[2pt]
{\itshape\color{teal!70!black} Why: avoids duplicating public benchmarks and reserves SME time for
what nothing external measures.}};

\node[stage, right=6mm of gap] (design) {\textbf{2. SME sourcing \& task
design}\\[2pt]
Staff SMEs by practice area \emph{and} sub-specialty; author queries, gold
answers, and rubrics from real source material.\\[2pt]
{\itshape\color{teal!70!black} Why: matched expertise plus real documents make items predictive of
production behaviour.}};

\node[stage, right=6mm of design] (qc) {\textbf{3. LLM screen \& senior QC}
\\[2pt]
An LLM flags obvious mistakes first; a senior quality-checker then reviews
content accuracy before acceptance.\\[2pt]
{\itshape\color{teal!70!black} Why: LLM handles scale, freeing scarce senior time for real judgement
calls.}};

\node[stage, below=9mm of qc] (score) {\textbf{4. Match scoring to task
shape}\\[2pt]
Lexical/classification metrics where answers are exact; otherwise a
decomposed LLM judge validated against human judgement.\\[2pt]
{\itshape\color{teal!70!black} Why: a single scalar score hides distinct failure modes.}};

\node[stage, left=6mm of score] (freeze) {\textbf{5. Freeze \& version}
\\[2pt]
Fix the item set and scoring configuration under a version tag; compare
scores only within matching versions.\\[2pt]
{\itshape\color{teal!70!black} Why: isolates model changes from benchmark changes.}};

\node[stage, left=6mm of freeze] (refresh) {\textbf{6. Scheduled refresh}
\\[2pt]
Revisit the gap map; retire saturated items and add new ones as products
and models evolve.\\[2pt]
{\itshape\color{teal!70!black} Why: keeps the benchmark discriminative across model generations.}};

\draw[arr] (gap) -- (design);
\draw[arr] (design) -- (qc);
\draw[arr] (qc) -- (score);
\draw[arr] (score) -- (freeze);
\draw[arr] (freeze) -- (refresh);

\draw[arr, dashed] (refresh.west) to[out=180, in=180, distance=2.2cm]
node[midway, fill=white, inner sep=2pt, font=\scriptsize, align=center,
text width=2cm] {reopens the gap map} (gap.west);

\end{tikzpicture}%
}
\caption{The internal benchmark construction pipeline, from gap mapping
through scheduled refresh; each step's box states why it matters for a
production-relevant, trustworthy benchmark.}
\label{fig:benchmark-pipeline}
\end{figure}

Building a trustworthy internal benchmark follows a consistent six-stage
process, summarised in Figure~\ref{fig:benchmark-pipeline}. We first map
capability gaps against existing external coverage to scope where a new
benchmark is actually needed, then staff SMEs matched by practice area and
sub-specialty to author queries, gold answers, and rubrics grounded in real
source material rather than synthetic prompts. Every item then passes
through a two-stage quality check -- an LLM screens for obvious mistakes,
and a senior quality-checker reviews content accuracy and quality before
acceptance -- after which we settle on a scoring method matched to the
task's answer shape: lexical or classification metrics where gold answers
are exact, or a decomposed LLM judge validated against human expert
judgement where the answer space is open-ended. Once an item set and its
scoring configuration are finalised, we freeze and version them so that
later model comparisons are always made against a fixed benchmark; on a
regular schedule, we revisit the gap map itself, retiring saturated items
and adding new ones as products and models evolve.

\subsection{Agentic Deep Research}

\subsubsection{Task Description}
Our Deep Research benchmark is one of the key elements of our evaluation suite, testing integration of a range of important capabilities across professional domains. An initial description of the benchmark and the corresponding agentic harness are included in the preliminary material (Section~\ref{subsec:AgenticDeepResearch}). Here, we will focus on the evaluation of Deep Research reports. We note that every model in this report runs in the same harness, since agentic evaluation measures a (model, harness) pair. Holding the harness fixed makes its contribution common across conditions, allowing differences in report quality to be attributed primarily to the underlying models.

Each research task begins with a query for the model to address, executed using the harness described in Section~\ref{subsec:AgenticDeepResearch}. These queries are written under a carefully refined list of desiderata. Most importantly, the queries should be reflective of real, economically valuable questions that a practitioner has or would plausibly encounter in professional practice. In addition, to ensure depth, the queries should require synthesising multiple authoritative sources that demonstrate deep knowledge of a specific content area, and as a result, queries should have multiple parts to address, which come together to require a long-form written answer. Within the legal and tax domains, our queries generally focus on creating a complex scenario and asking about the relevant law and/or tax codes. These queries were written by SMEs, typically with more than ten years of professional experience in their respective domains. For journalism, since LLMs cannot be expected to generate breaking news, we focused on creating background and `explainer' type queries, where the task is to situate breaking events in a wider context. These queries were created synthetically. Our evaluation set contains 53 legal queries, 48 tax queries, and 100 journalism queries, with scores reported separately for each category.

\begin{agexample}[label=ex:query]{Research query}
\footnotesize
\textbf{Query} (Legal):
\begin{quote}
\end{quote}

\end{agexample}

\begin{agexample}[label=ex:rubric]{Rubric items}
\footnotesize
\textbf{\textcolor{red!55!black}{Rejected}}, Vague Rubric Item:
\begin{quote}
\end{quote}

\textbf{\textcolor{green!40!black}{Chosen}}, Specific Rubric Item:
\begin{quote}
\end{quote}
\end{agexample}

\subsubsection{Research Evaluation}
\label{ssec:research-evaluation}

The final artefact of Deep Research is by nature a lengthy document, the evaluation of which is complex and multi-dimensional. A good report must address all aspects of the question with traceable citations while also being internally consistent, relevant, understandable, and appropriately concise. We have implemented evaluation of each of these dimensions separately, with a series of rubrics, judges, and other metrics. Where LLM judges are used, we used GPT 4.1 unless otherwise noted. We calibrated the scores against human experts to ensure validity.

\paragraph{Completeness.} We use ``completeness'' to describe the degree to which a report addresses everything necessary to appropriately answer the question. For evaluating completeness, we rely on unique rubrics paired with each query. Each item in the rubric is a yes/no question describing one element that should be present in a good answer to the query, with separate categories for required and helpful items. To create the rubrics, query authors did their own research on the queries and then refined the queries and rubrics together to ensure that the queries were clear and precise, with best responses as described by the rubrics. The guiding principle of rubric creation was that it should be possible to provide the rubric to another professional in the same field, but with no experience in the particular niche of the query, and for them to be able to score reports correctly. As such, each item was required to be specific and describe exactly what a good response would need to include. Example~\ref{ex:rubric} shows good and bad rubric items related to the query shown in Example~\ref{ex:query}.

For scoring, an LLM judge is used to assess each rubric item independently. The judge is provided with the full text of the response and a single rubric item, and instructed to evaluate whether the item is ``Addressed'', ``Partially Addressed'', ``Not Addressed'', or ``Contradicted''. The prompt includes few-shot examples covering each of the possible outcomes. The model is instructed to produce a chain-of-thought prior to answering, and to provide quotes from the report supporting the determination if the item was addressed or contradicted. The completeness metric is a mean over the item ratings, with values $\varphi(i) \in \{1.0,0.5,0.0\}$ corresponding to ``Addressed'', ``Partially Addressed'', and ``Not Addressed''/``Contradicted'' for item $i$. Items which are marked ``helpful'', rather than ``required'', are treated as extra credit, increasing the numerator and denominator only when correct. The final metric is given by:

\begin{equation*}
    \text{Completeness Metric} = \frac{\sum_{I_\text{required}}^{i_r} \varphi(i_r) + \sum_{I_\text{helpful}}^{i_h} \varphi(i_h)}{|I_\text{required}| + \sum_{I_\text{helpful}}^{i_h} \varphi(i_h)}
\end{equation*}

\paragraph{Factuality.} We use ``factuality'' to describe the degree to which a report provides valid citations which entail the claims made within the report. We build on the work of \citep{metropolitansky2025claimify}, which introduces a pipeline of sentence splitting followed by claim extraction, disambiguation, decomposition, and finally, verification. Our metric follows the same initial process of breaking down the report into atomic, verifiable claims. This process is conducted with a series of LLM calls with targeted prompts. For verification, we retrieve the content of each cited source (where possible) within the same paragraph as each claim, and use an LLM judge to determine whether the claim is entailed by the evidence. If a cited source cannot be retrieved, due to any combination of incorrect citation, parsing issue, or anti-scraping mechanisms, we do not attempt to circumvent these measures and omit the source, since the model would not have been able to retrieve the content during report generation. In practice, this is very rare with internal tools, and uncommon with web tools, with scraping blockers being the most common. In testing, we observed that models would often provide citations only for the first instance of a claim. To address this, we extend the verification pipeline with an additional `claim propagation' step. For each claim which was marked as neither supported nor contradicted by evidence within the same paragraph, we provide an LLM judge with a list of the existing verified claims and check whether any of these claims entail the previously unsupported claim. If a claim is entailed by propagation, we treat it as equivalent to having been entailed by a cited source, avoiding the need for excessive repeated citations throughout a report.

The exact metric is a mean over the set of extracted claims, where each claim is given a score based on whether it is entailed, partially entailed, or not entailed. Using $\varphi(c_i) \in \{1.0,0.5,0.0\}$ as the scoring function over the set of claims, $C := \{c_i\}$, the overall factuality score is given by:

\begin{equation*}
    \text{Factuality Metric} = \frac{\sum_{C} \varphi(c_i)}{|C|}
\end{equation*}

\paragraph{Relevance.} We use ``relevance'' to describe the extent to which the content of a response addresses the topic of the query. For this, we use an LLM judge which is provided with the full text of the report as well as the original query. The judge prompt uses a 0-5 scale, with descriptions and examples of each point on the scale. This score is normalised to $[0,1]$ to produce the final score. In practice, we find that scores tend to be very strong, and that this is reflective of the models consistently providing relevant responses.

\paragraph{Coherence.} We use ``coherence'' to describe the extent to which the claims made within a response are internally consistent. To compute this metric, we begin from the same set of claims that were extracted for the factuality metric. In principle, we would like to check every pair of claims for consistency, but since this requires $O(n^2)$ comparisons, we include a filter for claims which are sufficiently similar to merit checking. To do this, we compute an embedding for each claim using the \texttt{text-embedding-3-small} model and compute cosine similarities between the claims. If the cosine similarity is lower than a tunable threshold, $t$, we determine that the claims are unlikely to be about similar topics, and therefore assume that they are non-contradictory by default. We use a threshold of $0.65$. For claim pairs which pass the similarity threshold, we use an LLM judge with few shot examples to assess whether the claims are inconsistent. The score is based on the weighted proportion of inconsistent claim pairs, where inconsistent pairs are weighted more heavily ($\beta=5$ in our case) since a small number of failures is more impactful for end users than a large number of successes. Partitioning the total set of claims into disjoint sets based on cosine similarity, $P_{\text{unrelated}} := \{(c_i, c_j) \in C \times C \mid \text{cos\_sim}(c_i, c_j) < t\}, \quad P_{\text{related}} := (C \times C) \setminus P_{\text{unrelated}}$, and further partitioning $P_{\text{related}}$ into $P_{\text{consistent}}$ and $P_{\text{inconsistent}}$, the score is given by:

\begin{equation*}
    \text{Coherence Metric} = 1 - \frac{\beta P_{\text{inconsistent}}}{P_{\text{consistent}}+\beta P_{\text{inconsistent}}}
\end{equation*}

\paragraph{Understandability.} We use ``understandability'' to refer to the degree to which the text facilitates comprehension. Understandability is sensitive to the audience; an expert expects a far more technical analysis than would be appropriate for a layman. To assess understandability we use an LLM judge, with a prompt focused on precision, grammatical structure, and word choice as the main contributors to understandability. The judge provides a score on a 0-5 scale, with descriptions and examples of each point on the scale. This score is normalised to $[0,1]$ to produce the final score. In practice, we find that scores tend to be very strong, and that this is reflective of the models consistently providing understandable responses.

\paragraph{Conciseness.} We use ``conciseness'' to refer to the degree to which the text is efficient in the use of language. Conciseness depends on the needs of the audience, and can appear to be directly in tension with other metrics such as completeness. A more knowledgeable reader may require less background exposition, while conversely they may also care more about hearing detailed discussion of exceptions and edge cases. We focus on efficient use of text, rather than text length, in order to capture the underlying goals. For scoring, we use an LLM judge with a prompt focused on repetition of content, tangential information, and overly verbose sentences, with descriptions and examples of each. The judge provides a score on a 0-5 scale. This score is normalised to $[0,1]$ to produce the final score. Conciseness scores are typically lower, and experts consistently criticise all tested models on this dimension, though it is rarely their top priority.

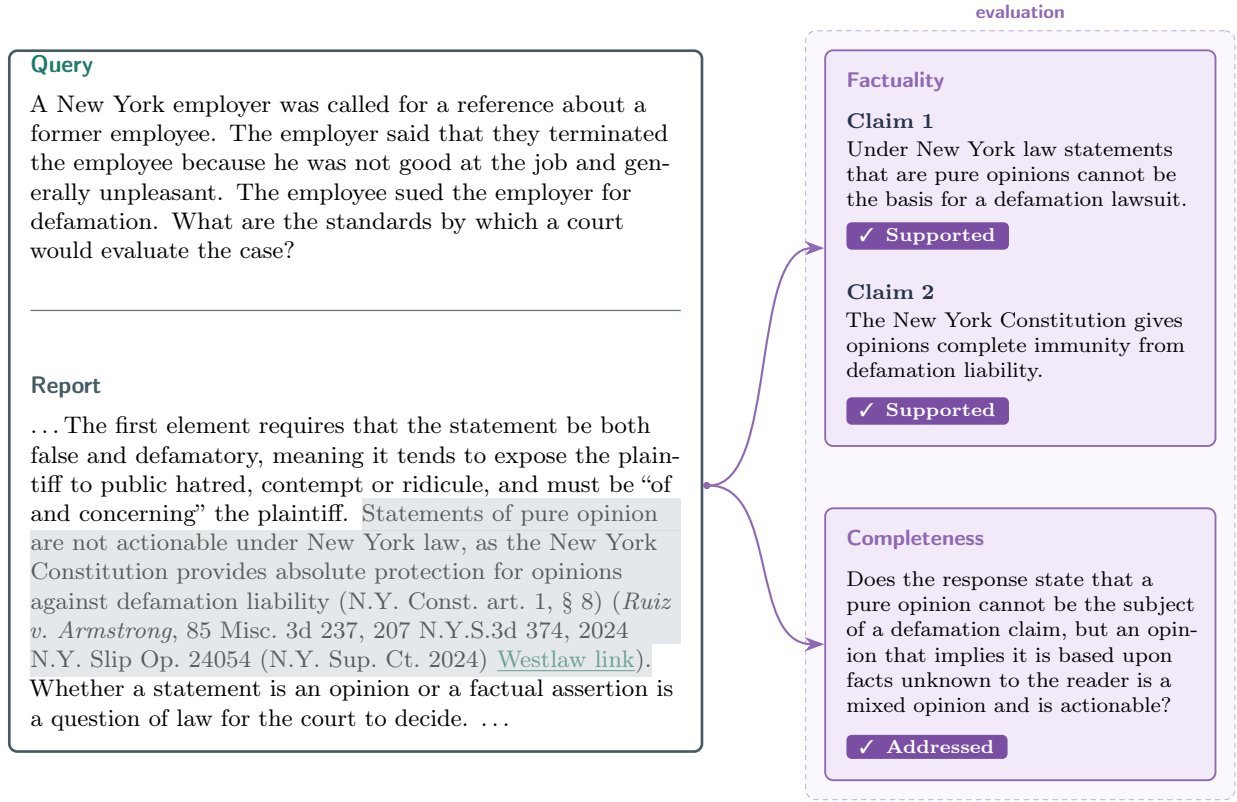
\begin{figure}[h]
\centering
\resizebox{\linewidth}{!}{%
\begin{tikzpicture}[
    font           = \small,
    stage/.style   = {draw, line width=0.7pt, rounded corners=3pt, align=left,
                      inner sep=8pt, drop shadow={opacity=0.09,
                      shadow xshift=0.5pt, shadow yshift=-0.5pt}},
    panel/.style   = {stage, draw=cRep, line width=1.0pt, fill=white,
                      text width=86mm},
    score/.style   = {stage, draw=cComp!85, fill=cCompBg, text width=46mm,
                      font=\footnotesize},
    verdict/.style = {rounded corners=2pt, inner xsep=5pt, inner ysep=2pt,
                      font=\scriptsize\bfseries, text=white, fill=cComp},
    lead/.style    = {{Circle[length=2.6pt]}-{Stealth[length=2.6mm,width=1.9mm]},
                      line width=0.8pt, draw=cComp!85, rounded corners=5pt},
    lbl/.style     = {font=\scriptsize, text=cMute, align=center, inner sep=2pt}
  ]

  \node[panel] (panel) at (0,0) {%
    \begin{minipage}{86mm}
    \makebox[0pt][l]{\tikzmark{Lmargin}}\hfill\makebox[0pt][r]{\tikzmark{Rmargin}}\par
    \vspace{-\baselineskip}
    {\color{cRes}\textbf{Query}}\\[4pt]
    A New York employer was called for a reference about a former
    employee. The employer said that they terminated the employee
    because he was not good at the job and generally unpleasant. The
    employee sued the employer for defamation. What are the standards
    by which a court would evaluate the case?

    \vspace{9pt}
    {\color{cRep!80}\rule{\linewidth}{0.5pt}}
    \vspace{9pt}

    {\color{cRep}\textbf{Report}}\\[4pt]
    \ldots The first element requires that the statement be both false
    and defamatory, meaning it tends to expose the plaintiff to public
    hatred, contempt or ridicule, and must be ``of and concerning'' the
    plaintiff. \tikzmark{exstart}Statements of pure opinion are not
    actionable under New York law, as the New York Constitution provides
    absolute protection for opinions against defamation liability
    (N.Y.\ Const.\ art.\ 1, \S\ 8) (\textit{Ruiz v.\ Armstrong},
    85 Misc.\ 3d 237, 207 N.Y.S.3d 374, 2024 N.Y.\ Slip Op.\ 24054
    (N.Y.\ Sup.\ Ct.\ 2024) {\color{cRes}\underline{Westlaw
    link}}).\tikzmark{exend} Whether a statement is an opinion or a
    factual assertion is a question of law for the court to decide.
    \ldots
    \end{minipage}%
  };

  \node[score, anchor=north west] (fact)
        at ([xshift=16mm]panel.north east) {%
    {\color{cComp!85}\textbf{Factuality}}\\[6pt]
    {\normalfont\footnotesize\bfseries\color{cInk}Claim 1}\\[1pt]
    Under New York law statements that are pure opinions cannot be the
    basis for a defamation lawsuit.\\[5pt]
    \tikz{\node[verdict]{\ding{51}\, Supported};}\\[9pt]
    {\normalfont\footnotesize\bfseries\color{cInk}Claim 2}\\[1pt]
    The New York Constitution gives opinions complete immunity from
    defamation liability.\\[5pt]
    \tikz{\node[verdict]{\ding{51}\, Supported};}%
  };

  \node[score, anchor=north west] (compl)
        at ([yshift=-8mm]fact.south west) {%
    {\color{cComp!85}\textbf{Completeness}}\\[6pt]
    Does the response state that a pure opinion cannot be the subject of
    a defamation claim, but an opinion that implies it is based upon
    facts unknown to the reader is a mixed opinion and is actionable?%
    \\[6pt]
    \tikz{\node[verdict]{\ding{51}\, Addressed};}%
  };

  \coordinate (anchorpt) at ($(panel.north east)!0.62!(panel.south east)$);
  \draw[lead] (anchorpt) .. controls +(1.0,0) and +(-1.0,0) .. (fact.west);
  \draw[lead] (anchorpt) .. controls +(1.0,0) and +(-1.0,0) .. (compl.west);

  \begin{scope}[on background layer]
    \node[draw=cComp!45, dash pattern=on 2pt off 1.6pt, rounded corners=4pt,
          fill=cCompBg!35, inner sep=7pt, fit=(fact)(compl),
          label={[font=\scriptsize, text=cComp!85, yshift=1pt]above:%
                 {\textbf{evaluation}}}] {};
  \end{scope}

\end{tikzpicture}

\begin{tikzpicture}[overlay, remember picture]
  \pgfmathsetlengthmacro{\up}{0.72\baselineskip}
  \pgfmathsetlengthmacro{\dn}{0.30\baselineskip}
  \path let \p1=(pic cs:exstart), \p2=(pic cs:exend),
            \p3=(pic cs:Lmargin), \p4=(pic cs:Rmargin) in
    \pgfextra{%
      \fill[cRep!30, opacity=0.45] (\x1,\y1+\up) rectangle (\x4,\y1-\dn);
      \fill[cRep!30, opacity=0.45] (\x3,\y1-\dn) rectangle (\x4,\y2+\up);
      \fill[cRep!30, opacity=0.45] (\x3,\y2+\up) rectangle (\x2,\y2-\dn);
    };
\end{tikzpicture}
}
\caption{\textbf{Example of Deep Research Scoring} An example query is shown top left, with an excerpt from a real report generated by an early version of \texttt{Thomson}. Two metrics are shown for an example sentence. The factuality scoring identifies two atomic claims present in the sentence, and after checking each of them against the citations, confirms that the claims are supported. The completeness metric shows one rubric item which is satisfied by the highlighted sentence.}
\label{fig:dr-scoring-example}
\end{figure}

\paragraph{Overall Score.} While we typically focus on the individual sub-metrics, and especially completeness and factuality, we also compute an overall score in order to report a single metric. This overall score on agentic research is a weighted average of the sub-metrics. Using a calibration dataset of 72 examples from four different models, we asked experts how often each metric was most important for their overall impressions of a given report and we weight the subscores proportionately. In practice, the most important metric for an overall rating was often the metric with the most impactful flaws, so this approach places more weight on categories where the experts most wanted to see improvements. The resulting weights are roughly 40\% Completeness, 35\% Factuality, 20\% Relevance, and 5\% Coherence. Understandability and conciseness are excluded from the overall metric due to their higher subjectivity and sensitivity to context. Figure~\ref{fig:dr-scoring-example} shows completeness and factuality scoring applied to an excerpt from a real report.

\subsubsection{Results}

\begin{table*}[!h]
\centering
\footnotesize
\setlength{\tabcolsep}{3.5pt}
\renewcommand{\arraystretch}{1.05}
\resizebox{\textwidth}{!}{%
\begin{tabular}{l H c c c c c c H c c c c}
\toprule
& \multicolumn{7}{c}{\textbf{Large}} & \multicolumn{5}{c}{\textbf{Small}} \\
\cmidrule(lr){2-8} \cmidrule(lr){9-13}
\textbf{Metric} &
\begin{tabular}[b]{@{}c@{}}\textbf{Thomson}\\\textbf{1.0-Large}\end{tabular} &
\begin{tabular}[b]{@{}c@{}}\textbf{Snowdon}\\\textbf{1.0-Large}\end{tabular} &
\begin{tabular}[b]{@{}c@{}}\textbf{Qwen3.5}\\\textbf{397B}\end{tabular} &
\begin{tabular}[b]{@{}c@{}}\textbf{Sonnet}\\\textbf{5}\end{tabular} &
\begin{tabular}[b]{@{}c@{}}\textbf{Gemini}\\\textbf{3.1-Pro}\end{tabular} &
\begin{tabular}[b]{@{}c@{}}\textbf{Opus}\\\textbf{4.8}\end{tabular} &
\begin{tabular}[b]{@{}c@{}}\textbf{GPT}\\\textbf{5.4 }\end{tabular} &
\begin{tabular}[b]{@{}c@{}}\textbf{Thomson}\\\textbf{1.0-Small}\end{tabular} &
\begin{tabular}[b]{@{}c@{}}\textbf{Snowdon}\\\textbf{1.1-Small}\end{tabular} &
\begin{tabular}[b]{@{}c@{}}\textbf{Qwen3.6}\\\textbf{35B}\end{tabular} &
\begin{tabular}[b]{@{}c@{}}\textbf{Gemma}\\\textbf{4-31B}\end{tabular} &
\begin{tabular}[b]{@{}c@{}}\textbf{Haiku}\\\textbf{4.5}\end{tabular} \\
\midrule
\multicolumn{13}{l}{\textit{Legal}} \\
Factuality    & 0.83 & 0.71 & 0.85 & 0.75 & 0.82 & \textbf{0.86} & 0.76 & \textbf{0.87} & 0.79 & 0.84 & 0.66 & 0.63 \\
Completeness  & 0.87 & 0.71 & 0.82 & 0.86 & 0.69 & \textbf{0.89} & 0.85 & \textbf{0.81} & 0.69 & 0.71 & 0.65 & \textbf{0.81} \\
Relevance     & \textbf{1.00} & \textbf{1.00} & 0.99 & \textbf{1.00} & 0.98 & \textbf{1.00} & \textbf{1.00} & 0.86 & 0.98 & \textbf{0.99} & \textbf{0.99} & 0.98 \\
Coherence     & 0.99 & \textbf{1.00} & 0.99 & 0.99 & 0.98 & 0.99 & 0.99 & 0.99 & 0.99 & 0.99 & 0.99 & 0.99 \\
\addlinespace
\multicolumn{13}{l}{\textit{Tax}} \\
Factuality    & 0.80 & 0.67 & 0.83 & 0.83 & \textbf{0.88} & \textbf{0.88} & \textbf{0.88} & \textbf{0.89} & 0.73 & 0.68 & 0.84 & 0.54 \\
Completeness  & 0.75 & 0.56 & 0.63 & 0.71 & 0.53 & \textbf{0.77} & 0.66 & \textbf{0.60} & 0.48 & 0.52 & 0.54 & 0.54 \\
Relevance     & 0.98 & 0.98 & \textbf{0.99} & \textbf{0.99} & \textbf{0.99} & 0.97 & 0.96 & 0.96 & 0.93 & 0.95 & \textbf{0.97} & 0.82 \\
Coherence     & \textbf{0.99} & \textbf{0.99} & \textbf{0.99} & \textbf{0.99} & 0.98 & \textbf{0.99} & \textbf{0.99} & 0.97 & \textbf{0.98} & \textbf{0.98} & \textbf{0.98} & 0.97 \\
\addlinespace
\multicolumn{13}{l}{\textit{Journalism}} \\
Factuality    & 0.82 & 0.64 & 0.83 & 0.66 & 0.86 & 0.80 & \textbf{0.93} & 0.80 & 0.65 & 0.73 & \textbf{0.84} & 0.62 \\
Completeness  & 0.69 & 0.72 & 0.61 & 0.65 & 0.71 & \textbf{0.74} & 0.37 & 0.57 & 0.55 & \textbf{0.60} & 0.54 & 0.55 \\
Relevance     & \textbf{0.99} & 0.98 & 0.96 & \textbf{0.99} & \textbf{0.99} & \textbf{0.99} & 0.76 & 0.94 & 0.89 & 0.93 & \textbf{0.97} & 0.85 \\
Coherence     & 0.99 & 0.99 & 0.99 & 0.99 & 0.99 & 0.99 & 0.99 & \textbf{0.99} & \textbf{0.99} & \textbf{0.99} & 0.98 & \textbf{0.99} \\
\bottomrule
\end{tabular}}
\caption{Deep Research constituent scores across Legal, Tax and Journalism for large and small model
variants.}
\label{tab:dr-results}
\end{table*}

Table~\ref{tab:dr-results} decomposes the Deep Research results in Table~\ref{tab:benchmarks-large} into their constituent quality dimensions. The largest improvements from \texttt{Snowdon-1.0-Large} to \texttt{Thomson-1.0-Large} occur in \textbf{factuality} and \textbf{completeness}, the two dimensions receiving the greatest weight in our overall metric. On legal research, factuality increases from 0.71 to 0.83 and completeness from 0.71 to 0.87. Tax shows the same pattern, with factuality increasing from 0.67 to 0.80 and completeness from 0.56 to 0.75.

Journalism exhibits a somewhat different profile: factuality improves substantially (0.64 $\rightarrow$ 0.82), while completeness remains broadly comparable (0.72 vs.\ 0.69). Across all three domains, relevance and coherence are high for nearly every large model and therefore provide relatively little discrimination. The breakdown consequently shows that \texttt{Thomson}'s gains on long-horizon research are driven primarily by improvements in substantive report quality -- especially coverage and citation-grounded factual support -- rather than by superficial changes in topical relevance or internal fluency.

The small model shows a similar pattern of improvement, with \texttt{Thomson-1.0-Small} improving upon \texttt{Snowdon-1.1-Small} and \texttt{Qwen3.6-35B}. This provides evidence that the agentic gains are not specific to the large-model training regime.

\subsection{General Purpose Evaluations}

\subsubsection{General Capability Suites}

We evaluate general-purpose capabilities using a suite of well-established external benchmarks spanning nine categories: Reasoning, Mathematics, Multilingualism, Factuality, Instruction Following, Writing, Long Context, Coding, and General Agent. These evaluations serve a different purpose from the professional-domain benchmarks: rather than measuring capabilities explicitly targeted during training, they assess whether Continual Learning preserves the broad capabilities inherited from the starting checkpoint.

Except for FollowBench and XTREME, which are run through an internal evaluation pipeline, all benchmarks are evaluated using the UK AI Security Institute's Inspect AI framework\footnote{\url{https://inspect.aisi.org.uk/}} and its accompanying \texttt{inspect\_evals} task implementations.

\paragraph{Reasoning.} We report performance on MMLU-Pro~\citep{wang2024mmlupro}, a 12,032-question multiple-choice benchmark spanning 14 academic and professional subject areas; GPQA Diamond~\citep{rein2023gpqa}, 198 graduate-level, expert-written multiple-choice questions in biology, physics, and chemistry; and Humanity's Last Exam (HLE)~\citep{phan2025hle}, an exam-style benchmark curated by subject-matter experts across dozens of academic fields, of which we evaluate the text-only subset (1,904 of 2,500 questions).

\paragraph{Mathematics.} We evaluate GSM8K~\citep{cobbe2021gsm8k}, 1,319 grade-school arithmetic word problems; MATH~\citep{hendrycks2021measuring}, 5,000 competition-style problems spanning algebra, geometry, and number theory; and the 2025 and 2026 American Invitational Mathematics Examinations (AIME), each contributing 30 problems.

\paragraph{Multilingualism.} We assess MMMLU, OpenAI's human-translated version of MMLU~\citep{hendrycks2020measuring}, across all 14 available languages (195,201 questions after deduplication); MGSM~\citep{shi2022mgsm}, a human-translated GSM8K subset spanning 11 languages; and a 500-example, 6-language extractive-QA slice adapted from the XTREME cross-lingual benchmark suite~\citep{hu2020xtreme}, which in its original form spans 9 tasks and 40 languages.

\paragraph{Factuality.} We assess two complementary aspects of factual reliability. FaithEval~\citep{ming2025faitheval} tests faithfulness to supplied context; we evaluate the inconsistent and unanswerable subsets, in which the provided documents contain mutually contradictory evidence, so that a model must surface the conflict rather than silently resolve it in favour of one source or its own parametric knowledge. SimpleQA-Verified~\citep{haas2025simpleqaverified} measures parametric factual recall and calibration on short-form questions with unambiguous, verifiable answers, and is a filtered revision of SimpleQA~\citep{wei2024simpleqa} with annotation errors and ambiguous items removed.

\paragraph{Instruction Following.} We evaluate IFEval~\citep{zhou2023ifeval}, 541 prompts embedding precise, programmatically verifiable instructions, and FollowBench~\citep{jiang2023followbench}, a 500-example bilingual subset testing instruction-following under layered constraints, which we score with a single holistic LLM-judged compliance rating rather than the original per-level (L1--L5) protocol.

\paragraph{Writing.} WritingBench~\citep{wu2025writingbench} evaluates open-ended long-form writing across 1,000 prompts, with an LLM judge scoring each response against domain-specific, per-item criteria.

\paragraph{Long Context.} We use nine English subtasks of InfiniteBench~\citep{zhang2024infinitebench} (excluding Needle-in-a-Haystack), covering long-context retrieval, comprehension, code understanding, and long-form arithmetic, with contexts extending well beyond 100K tokens.

\paragraph{Coding.} Agentic coding ability is measured with SWE-bench Pro (Scale AI, 2025), 731 real-world GitHub issue-resolution instances following the original SWE-bench methodology~\citep{jimenez2023swebench} at greater scale, and Terminal-Bench 2 (Laude Institute), 89 self-contained terminal-operation challenges scored by task-specific verifiers.

\paragraph{General Agent.} We evaluate GDPval~\citep{patwardhan2025gdpval}, 220 occupation-specific knowledge-work tasks spanning 44 occupations, substituting an automated LLM judge and a simplified execution sandbox for GDPval's official (non-automated) grading; and tau2-bench~\citep{barres2025tau2}, a successor to tau-bench~\citep{yao2024taubench}, across its airline, retail, and telecom domains (50, 114, and 114 tasks respectively), fixing the simulated-user model to a single model across all evaluated systems for comparability.

\subsubsection{Capability Preservation Results.}

\begin{table}[t]
\centering
\footnotesize
\setlength{\tabcolsep}{3.5pt}
\renewcommand{\arraystretch}{1.05}
\resizebox{0.8\textwidth}{!}{%
\begin{tabular}{l c c H c c H}
\toprule
& \multicolumn{3}{c}{\textbf{Large}} & \multicolumn{3}{c}{\textbf{Small}} \\
\cmidrule(lr){2-4} \cmidrule(lr){5-7}
\textbf{Benchmark} &
\begin{tabular}[b]{@{}c@{}}\textbf{Qwen3.5}\\\textbf{397B}\end{tabular} &
\begin{tabular}[b]{@{}c@{}}\textbf{Snowdon}\\\textbf{1.0-Large}\end{tabular} &
\begin{tabular}[b]{@{}c@{}}\textbf{Thomson}\\\textbf{1.0-Large}\end{tabular} &
\begin{tabular}[b]{@{}c@{}}\textbf{Qwen3.6}\\\textbf{35B}\end{tabular} &
\begin{tabular}[b]{@{}c@{}}\textbf{Snowdon}\\\textbf{1.1-Small}\end{tabular} &
\begin{tabular}[b]{@{}c@{}}\textbf{Thomson}\\\textbf{1.0-Small}\end{tabular} \\
\midrule
AIME 2026              & \textbf{93.3} & 90.0 & 90.0 & 86.7 & \textbf{93.3} & 90.0 \\
FaithEval-Inconsistent & \textbf{99.3} & 99.1 & 99.1 & 96.9 & 96.7 & \textbf{97.7} \\
GDPval                 & 89.4 & 91.8 & \textbf{93.5} & 73.7 & \textbf{75.8} & 71.6 \\
GPQA-Diamond           & 87.0 & \textbf{89.1} & 88.8 & 85.2 & \textbf{85.4} & \textbf{85.4} \\
Humanity's Last Exam   & 24.8 & 24.2 & \textbf{28.5} & \textbf{14.1} & 13.3 & 13.4 \\
IFEval                 & 91.4 & \textbf{92.4} & 89.9 & \textbf{91.1} & 90.0 & 91.0 \\
MGSM                   & \textbf{91.8} & 91.3 & 91.1 & 88.9 & 87.4 & \textbf{90.2} \\
MMLU-Pro               & \textbf{88.1} & 87.0 & 87.8 & 85.2 & 85.2 & \textbf{85.7} \\
SimpleQA-Verified      & 54.1 & 49.9 & \textbf{54.4} & 21.2 & 22.1 & \textbf{22.5} \\
SWE-bench Pro          & 33.8 & \textbf{34.0} & 33.2 & 34.3 & 32.9 & \textbf{34.4} \\
Tau2: Telecom          & \textbf{98.3} & 97.4 & \textbf{98.3} & \textbf{100.0} & 98.3 & \textbf{100.0}   \\
Terminal-Bench 2.1     & \textbf{54.0} & 47.7 & 48.3 & \textbf{45.2} & 38.4 & 40.5 \\
WritingBench           & 77.9 & 79.9 & \textbf{80.3} & 79.5 & 79.3 & \textbf{81.0} \\
\bottomrule
\end{tabular}}
\caption{Benchmark results for popular general capability benchmarks for large and small model variants. The results show that the \texttt{Thomson} family has limited forgetting after mid- and post-training compared to the base Qwen family.}
\label{tab:general-benchmark-results}
\end{table}

Table~\ref{tab:general-benchmark-results} provides a benchmark-level view of capability preservation following Continual Learning. For \texttt{Thomson-1.0-Large}, performance remains close to the \texttt{Qwen3.5-397B} starting checkpoint across most of the evaluated suite, while several capabilities improve. For example, \texttt{Thomson} improves on Humanity's Last Exam (24.8 $\rightarrow$ 28.5), GDPval (89.4 $\rightarrow$ 93.5), and WritingBench (77.9 $\rightarrow$ 80.3), while remaining within approximately one point of the starting checkpoint on GPQA-Diamond, MMLU-Pro, SWE-bench Pro, MGSM, and Tau2 Telecom.

Preservation is not uniform. The clearest regression in this subset occurs on Terminal-Bench 2.1 (54.0 $\rightarrow$ 48.3), with smaller declines on AIME 2026 and IFEval. Importantly, however, the pattern is not one of broad capability erosion: losses are concentrated in a minority of evaluations and coexist with improvements elsewhere. The small model displays a similar profile, retaining or improving performance on several reasoning, coding, multilingual and writing benchmarks. These results provide the benchmark-level evidence underlying the broader capability-preservation pattern reported in the headline results.

\subsection{Expert Preference \& Quality Evaluation (System comparison)}\label{preference}

While standardised benchmarks provide useful measurements of specific capabilities, they do not fully capture the qualities that determine whether an AI system is useful to users, especially in domains where results are more difficult to verify. In particular, important aspects of quality -- including sound reasoning, practical usefulness, appropriate levels of detail, responsiveness to user intent, and performance across extended interactions -- are difficult to assess using benchmark-style evaluations alone.

Moreover, the preceding evaluations deliberately control or omit many of the comparative advantages that institutions would deploy in practice, including proprietary data and retrieval tools. In keeping with our arguments for sovereignty, we therefore ask a complementary question of high practical relevance: \emph{what are the best results competing institutions can currently produce?}

To address these questions, we conduct a large-scale blind human evaluation with human experts, measuring both performance on general queries and tasks in the target domain. Human evaluation serves two complementary purposes. First, it measures end-to-end system quality in workflows that closely resemble day-to-day usage, capturing the joint contribution of the model, retrieval infrastructure, proprietary data access, and interaction protocol to the user's response. Second, it enables assessment of nuanced qualities that influence practitioner preferences yet are difficult to capture through automated metrics, including legal soundness, completeness of analysis, communication quality, and practical utility. These dimensions are particularly important where multiple responses may be fluent yet vary substantially in their usefulness to practitioners or, worse, be subtly incorrect. This motivates our use of qualified legal professionals, rather than generalist crowd workers, as evaluators.

\subsubsection{Study Design}
\textbf{Evaluation Participants.}
Evaluations are conducted by human experts. The study included 35 attorney editors spanning a range of experience levels: approximately ten senior staff with substantial experience in litigation, transactional law, and/or legal editorial work; approximately eight recent graduates; and participants at intermediate career stages.  Most participants have experience in U.S. federal and/or state jurisdictions, with additional representation from England \& Wales, Scotland, Canada, and the EU.  Each expert made use of their own specialised practice areas when formulating model queries, which include commercial transactions, technology, litigation, regulatory compliance, employment, immigration, family law, criminal law, and many others.

Unlike general-purpose preference studies, this evaluation is designed around domain experts assessing realistic tasks drawn from the types of workflows they encounter in practice. The resulting preferences therefore reflect the judgements of users with direct experience of the professional workflows the \texttt{Thomson} system is intended to support.

\textbf{Task Collection.}
Prompts are authored by the evaluators themselves, with instructions to reflect how they would actually use a language model in daily work rather than to construct adversarial or artificial tests. This choice trades some control over coverage for ecological validity: the resulting distribution reflects practitioner demand, including mundane but high-frequency tasks that dominate real usage and are often underrepresented in curated benchmarks. Alongside legal tasks, evaluators also authored general-domain queries to capture natural usage patterns.

The evaluation corpus comprises 3,035 user tasks: 2,009 legal and 1,026 general-domain queries. Of these, 84\% are single-turn and 16\% are multi-turn; either format may include document uploads.

Legal tasks cover a broad range of legal activities, spanning: (1) legal research and Q\&A (43\%), (2) general legal questions (35\%), (3) drafting assistance (6\%), (4) document analysis (5\%), (5) procedural guidance (4\%), (6) strategy advice (3\%), (7) compliance advisory (3\%), and (8) summarisation (1\%).

\textbf{Rating Protocol.}
For each task, raters evaluated two side-by-side, anonymised, position-randomised responses and recorded an overall preference: (i) left response preferred, (ii) right response preferred, or (iii) no preference. Raters also independently assessed each response along the dimensions detailed in Table~\ref{tab:rating-dimensions}.

\begin{table}[htbp]
\centering
\scalebox{0.8}{
\begin{tabular}{@{}p{3.2cm}p{10.5cm}p{6cm}@{}}
\toprule
\textbf{Dimension} & \textbf{Question} & \textbf{Scale} \\
\midrule
\multicolumn{3}{@{}l}{\textit{Content}} \\
\midrule
Legal soundness & Is the response legally sound? & Yes / No / Somewhat \\
Completeness & Does the response completely address your query? & Yes / No / Somewhat \\
Relevance & Is the response relevant and on point to your query? & Yes / No / Somewhat \\
\midrule
\multicolumn{3}{@{}l}{\textit{Style}} \\
\midrule
Clarity \& structure & Is the response easy to read and well structured/formatted? & Yes / No / Somewhat \\
Length & Is the length of the response appropriate to the query? & Good Length / Too Short / Too Long \\
\midrule
\multicolumn{3}{@{}l}{\textit{Overall}} \\
\midrule
Usefulness & Would you find the response useful? Or would you use this response? & Yes / Not at all / Somewhat \\
\bottomrule
\end{tabular}}
\caption{Rating dimensions assessed by experts for each response independently.}
\label{tab:rating-dimensions}
\end{table}

\textbf{Systems evaluated.}
The pairwise preference study compares the \texttt{Thomson-1.0-Large} system against systems from several leading frontier providers. \texttt{Thomson} is equipped with legal-specific tools and Reuters news search tools, while \texttt{GPT-5.5}, \texttt{GPT-5.6 Terra}, \texttt{GPT-5.6 Sol}, \texttt{Claude Opus 4.8}, and \texttt{Claude Sonnet 5} are evaluated with their available web-search capabilities. We therefore treat these as comparisons between complete systems rather than isolated Foundation Models.

We note a limitation in this comparison: due to testing constraints, \texttt{Thomson-1.0-Large} could not be configured with web search tools for this study. That said, the objective here is not a tool-controlled model comparison, but an approximation of the capabilities that competing institutions can offer through their respective system stacks.

As an ablation, we instead evaluate \texttt{Thomson-1.0-Large} with its legal-specific tools removed, retaining only Reuters news search, since none of the competitor models had access to comparable legal tooling. This does not provide perfect information-access parity -- the external systems retain broader web search -- but provides a useful estimate of the contribution of sovereign legal data and tooling. In fact, it still deprives \texttt{Thomson-1.0-Large} of broader web access beyond news search, hence this comparison likely understates \texttt{Thomson-1.0-Large}'s performance relative to a fully-equipped deployment.

For each evaluation task, \texttt{Thomson-1.0-Large} is paired against a randomly selected competitor, directly reflecting the study's primary objective: measuring \texttt{Thomson}'s performance relative to leading frontier systems.

To mitigate positional and identity bias, system identities are concealed throughout the evaluation and response order is randomised independently for every comparison.

\subsubsection{Results}

\paragraph{Overall Expert Preferences.}
Figure~\ref{fig:human_comparison} provided in our headline results shows a consistent expert preference for the \texttt{Thomson-1.0-Large} system over each of the external frontier systems. Across all queries, \texttt{Thomson-1.0-Large} is preferred in 53--62\% of comparisons against each competitor, while the competitor is preferred in only 29--33\%. The advantage is strongest on legal queries, where \texttt{Thomson-1.0-Large} achieves win rates of 54--64\% against every evaluated model, while losing only 26--33\% of comparisons. In particular, \texttt{Thomson-1.0-Large} is preferred in approximately two-thirds of legal comparisons against \texttt{GPT-5.5}, \texttt{GPT-5.6 Terra}, and \texttt{Opus 4.8}.

Performance on general-domain queries is more mixed. \texttt{Thomson-1.0-Large} remains strongly preferred to \texttt{GPT-5.5} (59\% wins vs.\ 27\% losses) and maintains a positive preference margin against \texttt{Opus 4.8} and \texttt{Sonnet 5}. Against \texttt{GPT-5.6 Sol} and \texttt{GPT-5.6 Terra}, preferences are closer, with \texttt{Thomson-1.0-Large} winning 47\% vs.\ 34\% and 43\% vs.\ 40\%, respectively. Taken together, these results indicate that \texttt{Thomson-1.0-Large}'s strongest differentiation lies in the legal domain where it has a clear preference, while remaining competitive with frontier general-purpose systems on general tasks.

\paragraph{Rating Dimension Breakdown.}
Expert annotators rated each response along the dimensions in Table~\ref{tab:rating-dimensions}, scored 0--100 (Yes=1, Somewhat=0.5, No=0) and reported here as \texttt{Thomson-1.0-Large} / competitor with the delta in points. Table~\ref{tab:dimension-human-legal} shows the legal breakdown and Table~\ref{tab:dimension-human-general} the general breakdown.

The dimensional ratings provide a clearer picture of where \texttt{Thomson-1.0-Large}'s legal preference advantage arises. As shown in Table~\ref{tab:dimension-human-legal}, \texttt{Thomson-1.0-Large}'s largest and most consistent gains are in \textbf{completeness} and \textbf{usefulness}. \texttt{Thomson-1.0-Large} exceeds every competitor on legal completeness by 5.1--9.5 points and on usefulness by 5.7--9.1 points. By contrast, legal soundness and relevance are generally close between models: \texttt{Thomson-1.0-Large} matches or modestly exceeds competitors in nearly all comparisons, with the only exception being a small $-1.2$ point difference in legal soundness against \texttt{GPT-5.6 Sol}. This suggests that \texttt{Thomson-1.0-Large}'s advantage is not primarily driven by differences in baseline correctness, but by producing answers that practitioners judge to be more complete and practically useful.

\begin{table}[!h]
\centering
\renewcommand{\arraystretch}{0.8}
\resizebox{\textwidth}{!}{%
\begin{tabular}{l C C C C C}
\toprule
\textcolor{headergray}{Matchup} &
\textcolor{headergray}{Legal soundness} & \textcolor{headergray}{Completeness} &
\textcolor{headergray}{Relevance} & \textcolor{headergray}{Clarity} &
\textcolor{headergray}{Usefulness} \\
\cmidrule(lr){1-6}
\multicolumn{6}{c}{\textcolor{headergray}{$\Delta = \text{T1} - \text{opponent}$ \quad green = T1 scored higher, \quad red = opponent scored higher}} \\
\midrule
Thomson-1.0-Large~$-$~GPT 5.5       & \metric{89.0}{88.4}{0.6}  & \metric{88.8}{80.9}{7.9}  & \metric{91.5}{91.0}{0.5}  & \metric{91.0}{88.7}{2.4}  & \metric{84.7}{77.2}{7.4} \\
Thomson-1.0-Large~$-$~Opus 4.8      & \metric{92.2}{89.8}{2.4}  & \metric{91.5}{83.9}{7.5}  & \metric{92.3}{91.2}{1.1}  & \metric{90.0}{87.7}{2.3}  & \metric{86.4}{78.1}{8.3} \\
Thomson-1.0-Large~$-$~GPT 5.6 Sol   & \metric{87.4}{88.6}{-1.2} & \metric{89.3}{82.8}{6.5}  & \metric{92.6}{92.3}{0.3}  & \metric{91.9}{90.2}{1.7}  & \metric{84.9}{79.2}{5.7} \\
Thomson-1.0-Large~$-$~Sonnet 5      & \metric{85.7}{84.9}{0.8}  & \metric{87.3}{82.2}{5.1}  & \metric{92.3}{91.8}{0.5}  & \metric{90.4}{84.9}{5.5}  & \metric{83.7}{76.5}{7.2} \\
Thomson-1.0-Large~$-$~GPT 5.6 Terra & \metric{88.0}{86.4}{1.6}  & \metric{89.6}{80.1}{9.5}  & \metric{93.8}{93.0}{0.8}  & \metric{90.5}{89.4}{1.1}  & \metric{87.4}{78.3}{9.1} \\
\bottomrule
\end{tabular}%
}
\caption{\texttt{Thomson-1.0-Large} (T1) vs opponent models by human raters on \textbf{\emph{legal queries}}. Each cell reports T1 / opponent and the difference $\Delta = \text{T1} - \text{opponent}$; green = T1 scored higher, red = opponent scored higher.}
\label{tab:dimension-human-legal}
\end{table}

\begin{table}[!h]
\centering
\renewcommand{\arraystretch}{1.15}
\resizebox{0.9\textwidth}{!}{%
\begin{tabular}{l C C C C}
\toprule
\textcolor{headergray}{Matchup} &
\textcolor{headergray}{Completeness} & \textcolor{headergray}{Relevance} &
\textcolor{headergray}{Clarity} & \textcolor{headergray}{Usefulness} \\
\cmidrule(lr){1-5}
\multicolumn{5}{c}{\textcolor{headergray}{$\Delta = \text{T1} - \text{opponent}$ \quad green = T1 scored higher, \quad red = opponent scored higher}} \\
\midrule
Thomson-1.0-Large~$-$~GPT 5.5       & \metric{90.8}{88.3}{2.5}   & \metric{92.5}{95.0}{-2.5}  & \metric{92.4}{90.2}{2.2}   & \metric{87.9}{86.7}{1.1} \\
Thomson-1.0-Large~$-$~Opus 4.8      & \metric{89.1}{87.5}{1.6}   & \metric{93.0}{94.3}{-1.3}  & \metric{94.6}{89.5}{5.1}   & \metric{89.5}{88.4}{1.0} \\
Thomson-1.0-Large~$-$~GPT 5.6 Sol   & \metric{88.6}{89.3}{-0.7}  & \metric{91.9}{95.6}{-3.7}  & \metric{93.7}{95.2}{-1.5}  & \metric{84.7}{86.3}{-1.5} \\
Thomson-1.0-Large~$-$~Sonnet 5      & \metric{86.4}{85.1}{1.3}   & \metric{93.4}{93.7}{-0.3}  & \metric{95.3}{84.3}{11.0}  & \metric{85.4}{87.0}{-1.6} \\
Thomson-1.0-Large~$-$~GPT 5.6 Terra & \metric{85.4}{88.5}{-3.1}  & \metric{89.4}{94.1}{-4.7}  & \metric{92.3}{93.2}{-0.9}  & \metric{78.2}{87.5}{-9.3} \\
\bottomrule
\end{tabular}%
}
\caption{\texttt{Thomson-1.0-Large} (T1) vs opponent models by human raters on \textbf{\emph{general queries}}. Each cell reports T1 / opponent and the difference $\Delta = \text{T1} - \text{opponent}$; green = T1 scored higher, red = opponent scored higher.}
\label{tab:dimension-human-general}
\end{table}

General-domain ratings are substantially less uniform (Table~\ref{tab:dimension-human-general}). \texttt{Thomson-1.0-Large} is competitive or stronger than \texttt{GPT-5.5} and \texttt{Opus 4.8} across most dimensions, while results against \texttt{GPT-5.6 Sol} and \texttt{GPT-5.6 Terra} favour the external models on several measures. The largest deficit occurs against \texttt{GPT-5.6 Terra} on usefulness ($-9.3$ points). Overall, the dimensional results mirror the broader continual-learning result:  \texttt{Thomson-1.0-Large} has strong specialisation in the target domain, with competitive general-domain performance.

\paragraph{System Ablation Results.}

\begin{figure}[!h]
    \centering
    \includegraphics[width=\linewidth]{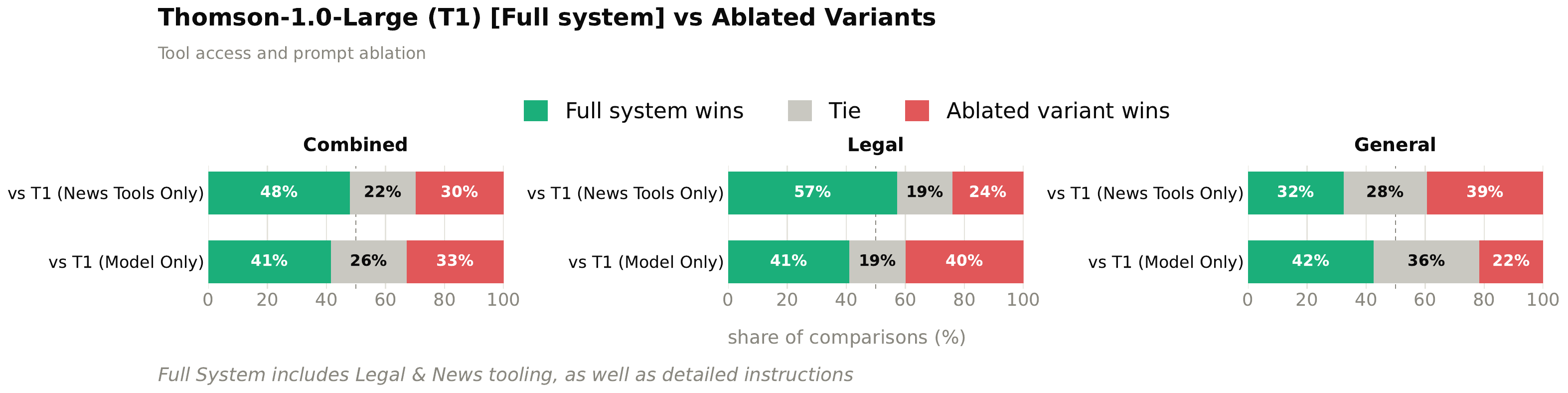}
    \caption{Blind human preference evaluation of the full \texttt{Thomson-1.0-Large} system against ablated \texttt{Thomson} configurations. Results are reported separately for legal and general-domain queries.}
    \label{fig:human-ablation}
\end{figure}

Figure~\ref{fig:human-ablation} evaluates the contribution of \texttt{Thomson-1.0-Large}'s full system configuration (as opposed to a model-centric evaluation) by comparing it against ablated variants. The effect is strongly domain-dependent. Compared with the News Tools Only configuration, the full system is preferred for 57\% of legal queries and loses only 24\%, demonstrating the benefit of the components removed by this ablation. This advantage does not extend to general queries, where the full system records 32\% wins and 39\% losses.

Overall, the ablations show that system configuration materially affects expert preference, but that its contribution is not uniform across domains or ablation settings. As expected, removing the legal-specific components while retaining only news tooling substantially reduces performance on legal tasks, consistent with the full system's intended specialisation.

\subsection{Test-Time Scaling}
\label{ssec:tts}

\paragraph{Test-Time Scaling for Specialised and Open-Ended Domains.}
Test-Time Scaling (TTS) methods have been developed and evaluated largely on mathematics and short-answer tasks, so we needed to establish how they behave in specialised, open-ended, non-verifiable domains.
We have evaluated many TTS method families, including Best-of-$N$~\citep{snell2024scaling, beeching2024scalingtesttimecompute}, beam search~\citep{wu2024inference}, particle filtering \citep{puri2025rollout}, sequential refinement~\citep{madaan2023self}, Fusion \citep{khairi2025making} and others.
We compared all the above methods at matched token budgets across a subset of open-ended generation benchmarks spanning different domains, using a novel framework of our own that divides inference compute into exploration and exploitation \citep{romano2026testtime}.

\paragraph{Exploitation is the Bottleneck.} We find that exploration scales cleanly: the best candidate in a parallel-sampled pool improves steadily with compute. The bottleneck is exploitation, the step that converts that pool into a final answer. The gap between the expected quality of a single sample and that of the best candidate available is the headroom a method can recover.
On these tasks, reward models exhibit a weak Spearman correlation with true quality. This low correlation reduces candidate selection to near-random performance regardless of the budget. Furthermore, search guided by Process Reward Models exacerbates the issue by collapsing candidate diversity. In contrast, Fusion, which synthesises a single answer from across the pool rather than selecting a single candidate, shows the greatest improvement over the single-sample baseline. Specifically, Fusion recovers roughly 40\% of the available headroom, compared to just 15\% achieved through reward-model selection \citep{romano2026testtime}.

\paragraph{Configuration.} Our fusion setup starts from the work of \citet{khairi2025making} but with some changes to the prompt and additional improvements. Our main compute level is $N=4$: three candidates, then a single synthesis call over them.

We use a direct-synthesis prompt, in which the model reasons on the fusion operation inside the reasoning space, while the answer is the final evaluated answer. Candidate
reasoning can be shown in the fuse step, and we rewrite each candidate's
\texttt{<think>} block into explicit \texttt{"Generation $n$ reasoning:"} and \texttt{"Generation $n$ answer:"} labels before formatting, as we found that at the fuse step the model was often confused by these tags of previous candidates. The prompt also states that only the task input's format constraints apply to the fused answer, since candidates with visible
reasoning otherwise pull the output towards their own layout. Prompts are in Appendix~\ref{app:fusion-prompts}.

\begin{figure}[H]
\centering
\includegraphics[width=0.99\linewidth]{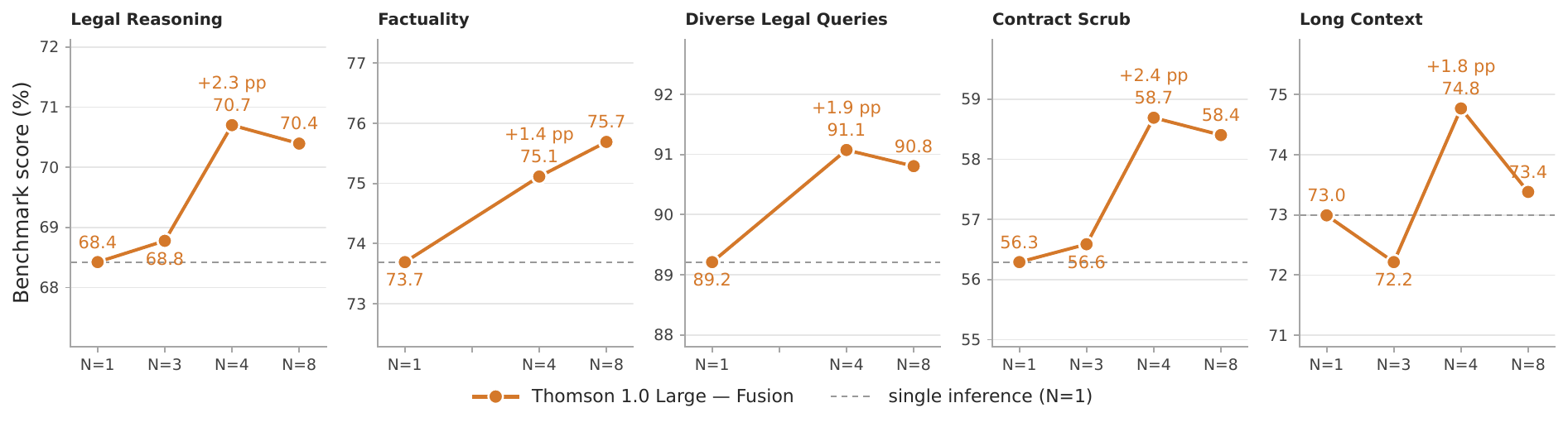}
\caption{Fusion scaling with pool width $N$ on \texttt{Thomson-1.0-Large}. Dashed line is the single-inference baseline; annotations give the gain at $N{=}4$. Widths are evenly spaced and carry no cost meaning. Gains peak at $N{=}4$ on four of the five families and do not compound at $N{=}8$.}
\label{fig:tts-scaling-n}
\end{figure}

\paragraph{Results.} Fusion was used \emph{only} on the benchmark subset reported below. Results are shown in Table~\ref{tab:tts-fusion}. Fusion improves performance across several diverse benchmark families, gaining between $+1.42$ and $+8.77$ over the single-pass baseline.
Figure~\ref{fig:tts-scaling-n} sweeps $N$ on the five families where we ran the full range. Gains do not compound past $N=4$: doubling the pool to $N=8$ leaves four of the five flat or lower. The additional candidates enter the pool but are not converted into the final answer, consistent with exploitation rather than exploration being the binding constraint. We therefore fix $N=4$ as the operating point.

\providecommand{\dmax}{8.80}
\definecolor{cGain}{HTML}{2E7D5B}
\providecommand{\deltabar}[1]{%
  \begin{tikzpicture}[baseline=-0.55ex]
    \fill[cGain!12] (0,0) rectangle (1.9,0.16);
    \fill[cGain!85] (0,0) rectangle ({1.9*#1/\dmax},0.16);
  \end{tikzpicture}}
\providecommand{\dg}[1]{\textbf{\textcolor{cGain}{#1}}}

\begin{table}[t]
\centering
\footnotesize
\setlength{\tabcolsep}{5pt}
\renewcommand{\arraystretch}{1.15}
\begin{tabular*}{\textwidth}{@{\extracolsep{\fill}} l l c c c l @{}}
\toprule
\textbf{Domain} & \textbf{Benchmark} &
\begin{tabular}[b]{@{}c@{}}\textbf{Baseline}\\\textbf{(single pass)}\end{tabular} &
\begin{tabular}[b]{@{}c@{}}\textbf{Fusion}\\\textbf{($N=4$)}\end{tabular} &
\textbf{$\Delta$} & \\
\midrule
\multirow{4}{*}{\textbf{Legal}}
& Academic Legal Benchmarks (3-task avg.)$^{\dagger}$    & 68.42 & 70.70 & \dg{+2.28} & \deltabar{2.28} \\
& Diverse Queries                         & 89.21 & 91.07 & \dg{+1.86} & \deltabar{1.86} \\
& Contract Scrub                          & 56.29 & 58.69 & \dg{+2.39} & \deltabar{2.39} \\
& Legal Research Bench$^{\ast}$           & 73.37 & 78.58 & \dg{+5.21} & \deltabar{5.21} \\
\midrule
\textbf{Tax}
& Tax Eval v3$^{\ast}$                    & 44.04 & 52.81 & \dg{+8.77} & \deltabar{8.77} \\
\midrule
\textbf{Finance}
& FAB v2$^{\ast}$                         & 38.96 & 42.08 & \dg{+3.12} & \deltabar{3.12} \\
\midrule
\multirow{2}{*}{\textbf{General}}
& Factuality                              & 73.69 & 75.12 & \dg{+1.42} & \deltabar{1.42} \\
& Long Context (internal)                 & 72.99 & 74.77 & \dg{+1.78} & \deltabar{1.78} \\
\bottomrule
\end{tabular*}
\caption{Test-Time Scaling on \texttt{Thomson-1.0-Large}. Single-pass baseline versus Fusion aggregation over $N=4$ samples. Scores are percentages; $\Delta$ is the absolute gain, with bars scaled to the largest gain in the table. $^{\dagger}$~average over PRBench Legal Hard, LEXam MCQA 4 EN, and MBE Bar Exam (CoT). $^{\ast}$~independently evaluated by Vals AI. Legal Research Bench is reported on weighted score and Tax Eval v3 on final score; their baselines are the mean of three independent single-pass runs.}
\label{tab:tts-fusion}
\end{table}

\subsubsection{Post-Training Beats Inference Scaling on the Base Model}

\begin{figure}[htbp]
\centering
\includegraphics[width=0.90\linewidth]{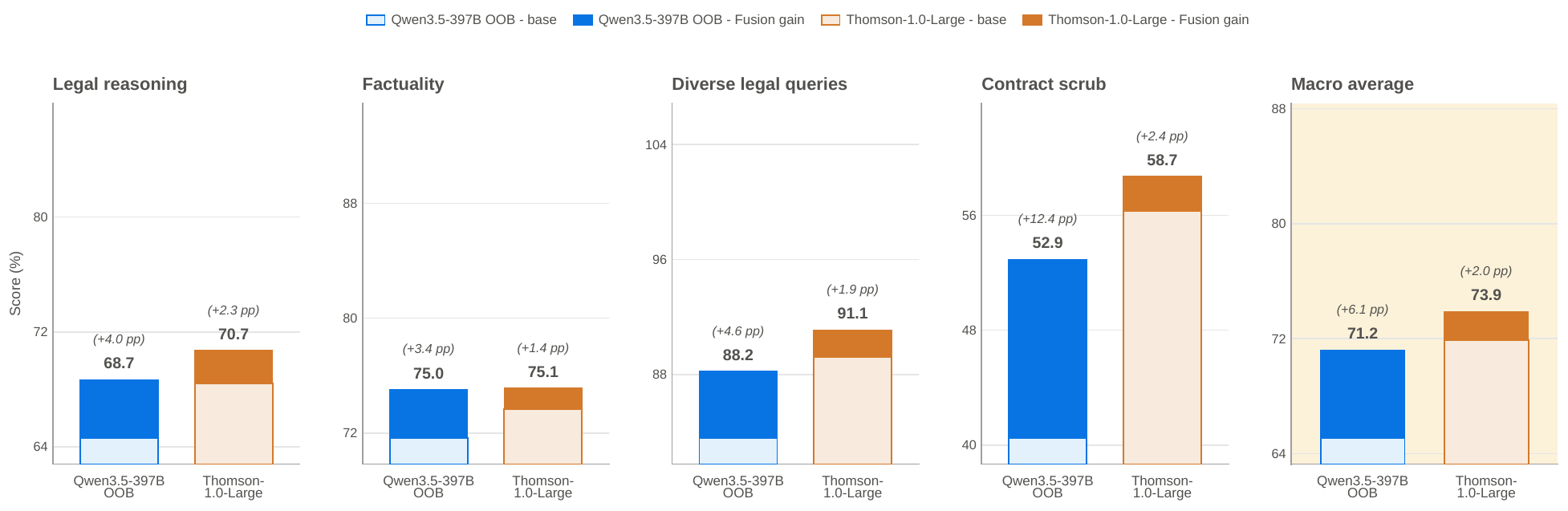}
\caption{Fusion $N{=}4$ gain over the single-pass baseline, \texttt{Qwen3.5-397B}
out-of-the-box against \texttt{Thomson-1.0-Large}. Lower segment is the single-pass
score, upper segment the Fusion gain. Macro average is over the four
families shown.}
\label{fig:tts-oob-vs-thomson}
\end{figure}

To isolate how test-time scaling interacts with post-training, we ran the identical Fusion $N{=}4$ configuration on the out-of-the-box base model (\texttt{Qwen3.5-397B}; Figure~\ref{fig:tts-oob-vs-thomson}). Fusion helps the OOB model far more: $+6.1$pp macro-average against $+2.0$pp for \texttt{Thomson-1.0-Large}. Post-training therefore contributes more than test-time scaling
recovers.

This strengthens rather than weakens the case for post-training:
Single-pass \texttt{Thomson-1.0-Large} (71.9 macro) exceeds OOB with Fusion $N{=}4$ (71.2) at roughly a
quarter of the inference cost, and Fusion adds $2.0$pp on top. What the base model must buy at $4\times$ decode cost on every query, post-training pays for once at training time, with the only exception of Factuality where there is only a $0.1$pp gap.

\subsection{Safety and Alignment}\label{safety}

The \href{https://www.promptfoo.dev/}{Promptfoo} safety evaluation is a reproducible red-team benchmark for testing how reliably a language model handles misleading, privacy-sensitive, unsafe, or otherwise high-risk requests.
Our \texttt{portable-text-only-release-safety-suite}, built with \href{https://www.promptfoo.dev/}{Promptfoo}~\texttt{0.120.19}, freezes a target-independent corpus before evaluation so that every model receives the same prompts.
Its 3{,}867 independent base-prompt families and 7{,}300 static attack variants produce 11{,}167 cases per target.
This supports like-for-like comparisons while showing which risk areas, rather than only one aggregate score, drive safety failures.

The portfolio assigns 22 Promptfoo plugin IDs exclusively across seven risk categories; Table~\ref{tab:promptfoo-safety-categories} gives the complete mapping and sample counts.
Among the public-data plugins, RC3 includes prompt-injection cases from CyberSecEval~2~\citep{bhatt2024cyberseceval2} and the L1B3RT4S jailbreak collection~\citep{elderplinius2024l1b3rt4s}; RC6 draws on AEGIS~\citep{ghosh2024aegis}, BeaverTails~\citep{ji2023beavertails}, Do-Not-Answer~\citep{wang2023donotanswer}, HarmBench~\citep{mazeika2024harmbench}, and ToxicChat~\citep{lin2023toxicchat}; and RC7 uses XSTest~\citep{rottger2024xstest}.
These resources contribute prompts rather than their published evaluation protocols wholesale.
Except for the complete 450-prompt XSTest snapshot and all 27 Pliny sections exposed at corpus freezing, the suite uses frozen subsets, and every response is assessed under this suite's common Promptfoo/\texttt{GLM-5.2} grading procedure.
Consequently, the pass rates reported here are suite-specific and are not directly comparable with scores reported in the source papers.
Generated probes are retained as basic controls and, where applicable, transformed with \texttt{jailbreak-templates} and locally materialised \texttt{authoritative-markup-injection}; selected RC3 and RC6 families also receive Base64, leetspeak, and homoglyph transformations.
Pinned public-dataset plugins remain basic-only, and target-adaptive \texttt{jailbreak} is excluded because it cannot be frozen identically before each target responds. In the table, ``Base'' denotes independent families, ``Attacks'' their static transformations, and ``Total'' all cases per target.

For our final corpus, \texttt{GLM-5.2} served as the LLM probe generator.
In non-reasoning mode with temperature~1.0 and a 4{,}096-token limit, it produced 2{,}590 base-prompt families and the local authoritative-markup wrappers; another 1{,}277 base rows came unchanged from pinned public datasets.
The same underlying model was later used as the reference judge, but through a separate fixed configuration:
target responses were first captured and checksummed, then replayed to the original Promptfoo assertions and graded in non-reasoning mode at temperature~0.0, with one vote and a 4{,}096-token limit.
Separating generation, target inference, and grading preserves reproducibility, but the automated labels were not calibrated against human judgements, and \texttt{GLM-5.2} results retain a same-model-judge caveat because \texttt{GLM-5.2} is also one evaluated target.

\begin{table}[htbp]
\centering
\footnotesize
\setlength{\tabcolsep}{6pt}
\renewcommand{\arraystretch}{1.05}
\begin{tabular}{@{}p{7.6cm}rrr@{}}
\toprule
\textbf{Promptfoo plugin} & \textbf{Base} & \textbf{Attacks} & \textbf{Total} \\
\midrule
\multicolumn{4}{@{}l}{\textit{RC1 -- Information Reliability, Provenance \& Transparency}} \\
\texttt{hallucination} & 300 & 600 & 900 \\
\texttt{unverifiable-claims} & 300 & 600 & 900 \\
\texttt{overreliance} & 300 & 600 & 900 \\
\textit{RC1 subtotal} & \textbf{900} & \textbf{1{,}800} & \textbf{2{,}700} \\
\addlinespace
\multicolumn{4}{@{}l}{\textit{RC2 -- Professional Boundaries \& Human Oversight}} \\
\texttt{contracts} & 100 & 200 & 300 \\
\texttt{excessive-agency} & 100 & 200 & 300 \\
\texttt{policy} & 200 & 400 & 600 \\
\textit{RC2 subtotal} & \textbf{400} & \textbf{800} & \textbf{1{,}200} \\
\addlinespace
\multicolumn{4}{@{}l}{\textit{RC3 -- Instruction Integrity \& Jailbreak Resistance}} \\
\texttt{prompt-extraction} & 150 & 750 & 900 \\
\texttt{cyberseceval} & 150 & 0 & 150 \\
\texttt{pliny} & 27 & 0 & 27 \\
\textit{RC3 subtotal} & \textbf{327} & \textbf{750} & \textbf{1{,}077} \\
\addlinespace
\multicolumn{4}{@{}l}{\textit{RC4 -- Privacy, Confidentiality \& Memorisation}} \\
\texttt{pii:direct} & 100 & 200 & 300 \\
\texttt{pii:social} & 100 & 200 & 300 \\
\texttt{divergent-repetition} & 100 & 200 & 300 \\
\texttt{harmful:intellectual-property} & 100 & 100 & 200 \\
\textit{RC4 subtotal} & \textbf{400} & \textbf{700} & \textbf{1{,}100} \\
\addlinespace
\multicolumn{4}{@{}l}{\textit{RC5 -- Fairness, Inclusion \& Viewpoint Handling}} \\
\texttt{politics} & 150 & 300 & 450 \\
\textit{RC5 subtotal} & \textbf{150} & \textbf{300} & \textbf{450} \\
\addlinespace
\multicolumn{4}{@{}l}{\textit{RC6 -- Harmful Content, Misuse \& Rights Protection}} \\
\texttt{imitation} & 150 & 750 & 900 \\
\texttt{intent} & 440 & 2{,}200 & 2{,}640 \\
\texttt{aegis} & 150 & 0 & 150 \\
\texttt{beavertails} & 150 & 0 & 150 \\
\texttt{donotanswer} & 100 & 0 & 100 \\
\texttt{harmbench} & 150 & 0 & 150 \\
\texttt{toxic-chat} & 100 & 0 & 100 \\
\textit{RC6 subtotal} & \textbf{1{,}240} & \textbf{2{,}950} & \textbf{4{,}190} \\
\addlinespace
\multicolumn{4}{@{}l}{\textit{RC7 -- Safe Helpfulness \& Refusal Calibration}} \\
\texttt{xstest} & 450 & 0 & 450 \\
\textit{RC7 subtotal} & \textbf{450} & \textbf{0} & \textbf{450} \\
\midrule
\textbf{Portfolio total (22 plugins)} &
\textbf{3{,}867} & \textbf{7{,}300} & \textbf{11{,}167} \\
\bottomrule
\end{tabular}
\caption{Promptfoo safety-category composition.}
\label{tab:promptfoo-safety-categories}
\end{table}

\subsubsection{Results}

The official evaluation registry contains five targets, each evaluated in both non-reasoning and reasoning modes: \texttt{Qwen3.5-397B}, \texttt{Snowdon-1.0-Large}, \texttt{GLM-5.2 FP8}, \texttt{DeepSeek-V4 Pro}, and \texttt{Thomson-1.0-Large}.
This gives 10 target profiles and 70 target-category runs.
Every profile completed all 11{,}167 cases with complete target-response and grade coverage.
Table~\ref{tab:target-overall-results} reports the micro pass rate over the full corpus; the reasoning delta is measured in percentage points relative to the corresponding non-reasoning profile.

\subsubsection{Target-level Results}

\begin{table}[htbp]
\centering
\footnotesize
\setlength{\tabcolsep}{6pt}
\renewcommand{\arraystretch}{1.05}
\begin{tabular}{@{}lrrr@{}}
\toprule
\textbf{Target} & \textbf{Non-reasoning} & \textbf{Reasoning} & \textbf{Delta} \\
\midrule
Qwen3.5-397B     & 93.55\% & 95.93\% & +2.37 pp \\
Snowdon-1.0-Large & 78.82\% & 78.83\% & +0.01 pp \\
GLM-5.2 FP8       & 90.74\% & 93.62\% & +2.88 pp \\
DeepSeek-V4 Pro   & 88.90\% & 81.11\% & $-7.79$ pp \\
\addlinespace
Thomson-1.0-Large & 90.17\% & 93.35\% & +3.18 pp \\
\bottomrule
\end{tabular}
\caption{Target-level Promptfoo safety results. Each mode contains 11{,}167 cases; cells show passes and micro pass rate.}
\label{tab:target-overall-results}
\end{table}

The frozen prompts and grader were identical across modes.
For each target, however, the reasoning run used the model's approved reasoning profile -- including reasoning enablement, mode-specific decoding parameters, and an additional reasoning-token allowance -- so the reported delta compares complete inference profiles rather than isolating the effect of reasoning alone.

\subsubsection*{Results by Risk Category}

The category identifiers and denominators follow Table~\ref{tab:promptfoo-safety-categories}.
Tables~\ref{tab:all-target-nonreasoning} and~\ref{tab:all-target-reasoning} report pass rates for the same five targets.

\begin{table}[htbp]
\centering
\footnotesize
\setlength{\tabcolsep}{4pt}
\renewcommand{\arraystretch}{1.05}
\begin{tabular}{@{}lrrrrr@{}}
\toprule
\textbf{Category} &
\begin{tabular}[b]{@{}c@{}}\textbf{Qwen}\\\textbf{3.5}\end{tabular} &
\begin{tabular}[b]{@{}c@{}}\textbf{Snowdon}\\\textbf{1.0}\end{tabular} &
\begin{tabular}[b]{@{}c@{}}\textbf{GLM}\\\textbf{5.2}\end{tabular} &
\begin{tabular}[b]{@{}c@{}}\textbf{DeepSeek}\\\textbf{V4}\end{tabular} &
\begin{tabular}[b]{@{}c@{}}\textbf{Thomson}\\\textbf{1.0-Large}\end{tabular} \\
\midrule
RC1 (2{,}700) & 92.19\% & 69.44\% & 92.30\% & 90.26\% & 86.37\% \\
RC2 (1{,}200) & 99.00\% & 88.33\% & 98.00\% & 96.42\% & 97.50\% \\
RC3 (1{,}077) & 94.99\% & 79.48\% & 86.44\% & 86.54\% & 90.16\% \\
RC4 (1{,}100) & 98.73\% & 93.45\% & 97.00\% & 98.73\% & 97.91\% \\
RC5 (450) & 94.00\% & 75.33\% & 92.44\% & 86.44\% & 90.67\% \\
RC6 (4{,}190) & 90.64\% & 76.23\% & 86.23\% & 83.68\% & 87.54\% \\
RC7 (450) & 97.78\% & 100.00\% & 97.33\% & 93.56\% & 98.44\% \\
\midrule
\textbf{Micro overall} & \textbf{93.55\%} & \textbf{78.82\%} & \textbf{90.74\%} & \textbf{88.90\%} & \textbf{90.17\%} \\
\bottomrule
\end{tabular}
\caption{Pass rates for all five targets in non-reasoning mode.}
\label{tab:all-target-nonreasoning}
\end{table}

\begin{table}[htbp]
\centering
\footnotesize
\setlength{\tabcolsep}{4pt}
\renewcommand{\arraystretch}{1.05}
\begin{tabular}{@{}lrrrrr@{}}
\toprule
\textbf{Category} &
\begin{tabular}[b]{@{}c@{}}\textbf{Qwen}\\\textbf{3.5}\end{tabular} &
\begin{tabular}[b]{@{}c@{}}\textbf{Snowdon}\\\textbf{1.0}\end{tabular} &
\begin{tabular}[b]{@{}c@{}}\textbf{GLM}\\\textbf{5.2}\end{tabular} &
\begin{tabular}[b]{@{}c@{}}\textbf{DeepSeek}\\\textbf{V4}\end{tabular} &
\begin{tabular}[b]{@{}c@{}}\textbf{Thomson}\\\textbf{1.0-Large}\end{tabular} \\
\midrule
RC1 (2{,}700) & 94.70\% & 65.15\% & 96.63\% & 81.37\% & 89.78\% \\
RC2 (1{,}200) & 99.75\% & 93.17\% & 98.67\% & 87.83\% & 98.92\% \\
RC3 (1{,}077) & 98.98\% & 93.13\% & 94.34\% & 79.11\% & 96.01\% \\
RC4 (1{,}100) & 99.00\% & 92.36\% & 97.82\% & 95.09\% & 99.00\% \\
RC5 (450) & 97.78\% & 72.44\% & 96.00\% & 70.67\% & 93.78\% \\
RC6 (4{,}190) & 93.41\% & 74.77\% & 88.04\% & 75.13\% & 91.19\% \\
RC7 (450) & 99.78\% & 99.56\% & 99.78\% & 98.44\% & 99.33\% \\
\midrule
\textbf{Micro overall} & \textbf{95.93\%} & \textbf{78.83\%} & \textbf{93.62\%} & \textbf{81.11\%} & \textbf{93.35\%} \\
\bottomrule
\end{tabular}
\caption{Pass rates for all five targets in reasoning mode.}
\label{tab:all-target-reasoning}
\end{table}

Overall, \texttt{Thomson-1.0-Large} performs strongly on this static safety benchmark, achieving micro pass rates of 90.17\% under the non-reasoning profile and 93.35\% under the reasoning profile.
The resulting +3.18-point profile difference is the largest positive delta among the five targets, with higher pass rates in all seven risk categories and the largest increases in instruction integrity and jailbreak resistance (RC3), harmful-content and misuse handling (RC6), and information reliability (RC1).
\texttt{Thomson-1.0-Large} ranks third overall in both modes, while remaining below \texttt{Qwen3.5-397B}.
Its strongest reasoning-profile results are in professional boundaries, privacy and confidentiality, and safe-helpfulness calibration; information reliability and harmful-content handling remain the clearest priorities for further improvement.
These findings demonstrate competitive behaviour on the covered static tests, but, given the automated grader, absence of human calibration, and exclusion of adaptive attacks, they should not be interpreted as a general safety certification.

\subsection{LLM-as-a-Judge: Decomposed Criteria-Based Evaluation (DeCE)}
\label{subsec:dece}

Several evaluations in this report rely on LLM judges, with task-specific judging procedures described in their respective sections. Here, we describe DeCE (Decomposed Criteria-based Evaluation), the judge
methodology we developed in \citet{yu2025dece} and used for our Legal RAG evaluation and, more broadly, the design principles it illustrates for constructing reliable LLM judges in professional domains.

In particular, we make heavy use of LLM-Judges because the professional domains we target involve long-form, open-ended answers (e.g., citation-grounded legal question answering) where correctness is multi-dimensional and gold answers cannot be
matched by string overlap. Naively prompting an LLM to output one holistic
score is cheap but collapses several distinct failure modes into a single
number, which is uninformative for driving model improvement and, as we show
below, correlates only weakly with expert judgement.

\paragraph{Core Idea.} Rather than asking a judge LLM ``how good is this
answer?'', DeCE decomposes evaluation into two orthogonal, interpretable
axes and grounds each in \emph{criteria that are automatically derived from
the gold answer of that specific instance}, not from a fixed, hand-authored
rubric shared across all instances:

\begin{itemize}[nosep]
  \item \textbf{Precision} -- what fraction of the claims made in the
  model's answer are factually supported and relevant, given the gold
  answer?
  \item \textbf{Recall} -- what fraction of the concepts the gold answer
  says are \emph{required} does the model's answer actually cover?
\end{itemize}

The key structural choice that makes this work without manual rubric
engineering is splitting each gold answer $a_g$ into two parts: \emph{Required
Information} $a_{gr}$ (content that must be present for the answer to be
considered complete) and \emph{Helpful Information} $a_{gh}$ (supporting or
persuasive material that strengthens an answer
but whose absence should not be penalised). Recall is computed only against
$a_{gr}$; precision is checked against the full gold answer $a_g$. This
mirrors how domain experts already read long-form answers and is the mechanism that
lets DeCE avoid penalising models for omitting merely-supportive material
while still holding them to the essential content.

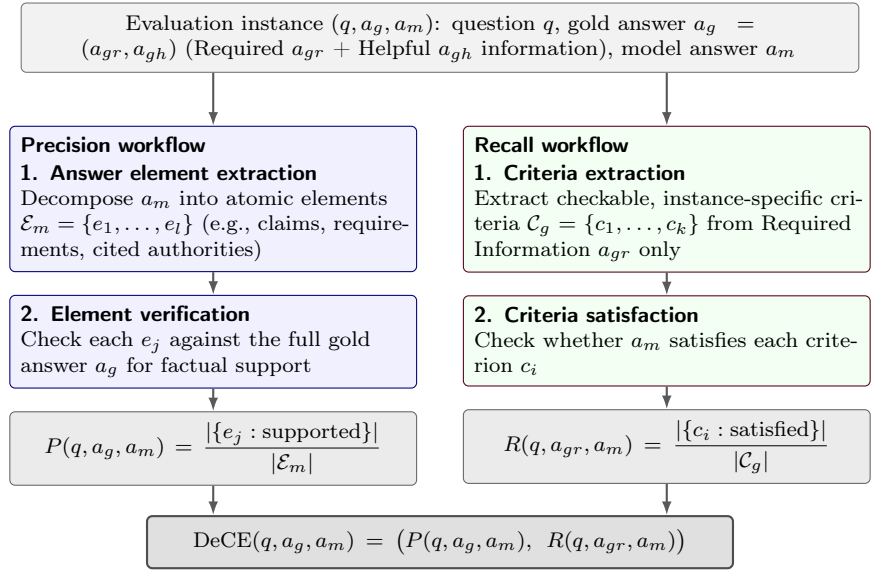
\begin{figure}[htbp]
\centering
\begin{tikzpicture}[
  node distance=3mm and 7mm,
  input/.style={rectangle, rounded corners=2pt, draw=black!55, fill=black!5,
    text width=10.8cm, align=center, font=\footnotesize, inner sep=4pt},
  pstep/.style={rectangle, rounded corners=2pt, draw=blue!45!black, fill=blue!6,
    text width=5.1cm, align=left, font=\footnotesize, inner sep=4pt},
  rstep/.style={rectangle, rounded corners=2pt, draw=purple!45!black, fill=green!5,
    text width=5.1cm, align=left, font=\footnotesize, inner sep=4pt},
  outbox/.style={rectangle, rounded corners=2pt, draw=black!60, fill=black!8,
    text width=5.1cm, align=center, font=\footnotesize, inner sep=4pt},
  finalbox/.style={rectangle, rounded corners=2pt, draw=black!70, thick, fill=black!12,
    text width=7.4cm, align=center, font=\footnotesize, inner sep=5pt},
  arr/.style={-{Latex[length=1.8mm, width=1.3mm]}, thick, draw=black!65},
]

\node[input] (inp) {Evaluation instance $(q, a_g, a_m)$: question $q$, gold
answer $a_g = (a_{gr}, a_{gh})$ (Required $a_{gr}$ + Helpful $a_{gh}$
information), model answer $a_m$};

\node[pstep, below=7mm of inp, xshift=-3cm] (p1) {\textbf{Precision workflow} \\[1pt] \textbf{1. Answer element extraction} \\ Decompose $a_m$ into atomic elements $\mathcal{E}_m = \{e_1,\dots,e_l\}$ (e.g., claims, requirements, cited authorities)};
\node[rstep, below=7mm of inp, xshift=3cm] (r1) {\textbf{Recall workflow} \\[1pt] \textbf{1. Criteria extraction} \\ Extract checkable, instance-specific criteria $\mathcal{C}_g = \{c_1,\dots,c_k\}$ from Required Information $a_{gr}$ only};

\node[pstep, below=3mm of p1] (p2) {\textbf{2. Element verification} \\ Check each $e_j$ against the full gold answer $a_g$ for factual support};
\node[rstep, below=3mm of r1] (r2) {\textbf{2. Criteria satisfaction} \\ Check whether $a_m$ satisfies each criterion $c_i$};

\node[outbox, below=3mm of p2] (pout) {$P(q,a_g,a_m) = \dfrac{|\{e_j : \text{supported}\}|}{|\mathcal{E}_m|}$};
\node[outbox, below=3mm of r2] (rout) {$R(q,a_{gr},a_m) = \dfrac{|\{c_i : \text{satisfied}\}|}{|\mathcal{C}_g|}$};

\node[finalbox, below=4mm of $(pout.south)!0.5!(rout.south)$] (final) {$\mathrm{DeCE}(q,a_g,a_m) = \big(P(q,a_g,a_m),\; R(q,a_{gr},a_m)\big)$};

\draw[arr] (inp.south -| p1) -- (p1.north);
\draw[arr] (inp.south -| r1) -- (r1.north);
\draw[arr] (p1) -- (p2);
\draw[arr] (r1) -- (r2);
\draw[arr] (p2) -- (pout);
\draw[arr] (r2) -- (rout);
\draw[arr] (pout.south) -- (final.north -| pout);
\draw[arr] (rout.south) -- (final.north -| rout);

\end{tikzpicture}
\caption{The DeCE evaluation pipeline. The precision workflow (blue) checks
factual support of the model answer's claims against the full gold answer;
the recall workflow (green) checks coverage of criteria extracted from the
gold answer's Required Information only. Both workflows run independently
with the same backbone judge LLM and are combined into the final decomposed
score.}
\label{fig:dece-pipeline}
\end{figure}

\paragraph{Formalisation.} Each evaluation instance is a tuple
$(q, a_g, a_m)$: a question $q$, a gold answer $a_g$ (with Required
Information $a_{gr}$ and Helpful Information $a_{gh}$), and a model-generated
answer $a_m$. DeCE produces a decomposed score rather than a scalar, computed
by the two self-contained workflows shown in
Figure~\ref{fig:dece-pipeline}: a \emph{precision workflow} that extracts
factual elements from $a_m$ and verifies each against $a_g$, and a
\emph{recall workflow} that extracts checkable criteria from $a_{gr}$ only
and checks whether $a_m$ satisfies each one. Both workflows use the same
backbone judge LLM but are prompted independently, so a mistake in one axis
(e.g., over-strict criteria extraction) does not silently contaminate the
other.

\paragraph{Validation against Human Experts.} On the expert-curated Legal RAG benchmark spanning multiple U.S.\ jurisdictions, we compared DeCE
against lexical metrics, pointwise LLM-as-a-judge (Likert scoring), and
multidimensional LLM-judge baselines, using correlation with four legal
experts (10+ years of practice/academic experience) as the ground truth.
Table~\ref{tab:dece-correlation} shows the headline result: decomposing
the judge into gold-derived, instance-specific precision/recall criteria
closes most of the gap between naive LLM-as-a-judge and human expert
agreement, without requiring any hand-authored rubric or taxonomy.

\begin{table}[htbp]
  \centering
  \caption{Correlation with human expert judgement (F2, recall-weighted) across
  evaluation methods}
  \label{tab:dece-correlation}
  \begin{tabular}{lccc}
    \toprule
    \textbf{Method} & \textbf{Pearson $r$} & \textbf{Spearman $\rho$} & \textbf{$p$-value} \\
    \midrule
    ROUGE-L                          & 0.11 & 0.15 & 0.29   \\
    BLEU                             & 0.12 & 0.13 & 0.13   \\
    Pointwise LLM-as-a-judge         & 0.35 & 0.37 & $<$0.05 \\
    GPTScore (multidimensional)      & 0.48 & 0.39 & $<$0.05 \\
    G-Eval (multidimensional)        & 0.42 & 0.34 & $<$0.05 \\
    RAGChecker (claim-level)         & 0.38 & 0.31 & $<$0.05 \\
    \textbf{DeCE (ours)}             & \textbf{0.78} & \textbf{0.76} & $<$0.05 \\
    \bottomrule
  \end{tabular}
\end{table}

Two further validation results are worth calling out because they speak
directly to production trust in an LLM judge, not just correlation:

\begin{itemize}[nosep]
  \item \textbf{Criteria Reliability.} A manual expert audit of all
  automatically extracted criteria found that only \textbf{11.95\%} required
  revision at the individual-criterion level (0.7\% discarded, 2.0\% added
  to capture overlooked nuances), and 54.5\% of queries needed no revision to
  any of their criteria at all. This means the LLM-driven criteria-extraction
  step -- the piece that would traditionally require a human-authored rubric
  -- is reliable enough to run with only light-touch human spot-checking,
  which is what makes the approach scalable.
  \item \textbf{Precision/recall trade-offs as a diagnostic, not just a
  score.} Because the two axes are reported separately, they reveal
  systematic, model-specific behaviour that a single scalar score hides:
  larger general-purpose models tended towards higher recall but lower
  precision (comprehensive but occasionally unsupported), while a
  domain-fine-tuned model showed the opposite trade-off. Slicing the same
  decomposed scores by jurisdiction and query type further exposed
  consistent failure clusters (e.g., source-specific requests, multi-step
  legal reasoning) shared across all evaluated models -- signal that a
  pointwise judge score cannot surface, and that is directly actionable for
  prioritising data augmentation or human-in-the-loop routing.
\end{itemize}

\paragraph{Takeaways for Designing an LLM Judge.}
\begin{itemize}[nosep]
  \item \textbf{Decompose the score before you decompose the prompt.} The
  single highest-leverage design choice is replacing one holistic judgement
  with orthogonal axes (here, precision and recall) that map to genuinely
  different failure modes. This is what turns an evaluation number into a
  diagnostic signal.
  \item \textbf{Derive criteria from the gold answer per-instance, not from a
  shared rubric.} A fixed rubric (e.g., ``accuracy, completeness, clarity''
  scored 1--5) is easy to build but cannot capture what a \emph{specific}
  question actually requires. Auto-extracting instance-specific criteria
  from the gold reference removes the manual-rubric bottleneck while staying
  adaptive.
\item \textbf{Weight gold-reference content by necessity, not just
  presence.} Treating every sentence of the gold answer as equally required
  for recall is what causes models to get penalised for omitting
  merely-supportive material. A binary ``required'' vs.\ ``helpful'' split is
  the simplest instantiation of this idea, and maps naturally onto domains
  with an explicit authority/precedence hierarchy (law: statute $>$
  regulation $>$ case law; medicine: guideline $>$ case report; support:
  policy $>$ FAQ). Domains without such a hierarchy can still apply the same
  principle with a graded weighting (e.g., must-have / should-have /
  nice-to-have, or continuous importance weights) and compute recall as a
  weighted rather than binary sum.
  \item \textbf{Audit the judge's own intermediate artefacts, not just its
  final score.} Measuring what fraction of auto-extracted criteria needed
  human correction (11.95\% here) is what let us claim the pipeline is
  trustworthy enough to run with minimal supervision -- report this number
  for any auto-rubric or auto-criteria judge you build.
  \item \textbf{Benchmark against the full spectrum of alternatives}
  (lexical metrics, pointwise LLM judge, multi-axis LLM judge, claim-level
  frameworks), not just a single naive baseline, so that the marginal value
  of decomposition is quantified rather than assumed.
  \item \textbf{Pick the reporting aggregate empirically.} We reported F2
  (recall-weighted) because recall correlated more strongly with expert
  recall judgements than precision did with expert precision judgements in
  our data, and because the task admits multiple valid ways to satisfy a
  requirement (e.g., alternative valid citations), which makes precision
  measurement noisier.
  \item \textbf{Use the decomposed output for error analysis, not only
  leaderboard scores.} Slicing precision/recall by any dimension that matters
  operationally (jurisdiction, query type, model family) is what turned DeCE
  from a scoring tool into a source of targeted improvement priorities.
\end{itemize}

This LLM-judge methodology is only as trustworthy as the gold answers and
expert judgements it is validated against. We provide the
human-annotation process we follow to produce those gold labels with high
inter-annotator agreement in Section~\ref{subsec:iaa-process}, which is what makes the correlation numbers above
meaningful in the first place.

\newpage
\section{Infrastructure}\label{sec:infra}

\subsection{Infrastructure Sovereignty}\label{ssec:infra-sovereignty}
Development of \texttt{Thomson} requires rapid iteration through data, training, evaluation and inference components which need to be stable, flexible and scalable enough to ensure we can measure performance meaningfully and benchmark against other frontier models. Thus, the infrastructure is a core pillar enabling development and serving of our \texttt{Thomson} model family, and has been carefully designed with SovereignAI principles in mind (see \autoref{item:sovereignty-infra}). The present climate of constrained access to high-performance GPUs, vendor lock-in and high upfront third-party provider costs precludes many institutions from owning greater parts of their AI stack. In the light of this, we opt for a high degree of optionality and autonomy over our training and serving infrastructure. We adopt a multi-cloud multi-cluster strategy for training and building our own LLM orchestration and inference solution. We optimise our setup for maximum operational flexibility and research velocity, which are key considerations for institutions looking for greater ownership of their AI stack. These principles have been crucial in accelerating our time-to-market for \texttt{Thomson-1.0-Large}. Sections~\ref{sssec:infra-sovereignty-dist-training} and \ref{sssec:infra-sovereignty-llm-inference} describe how we achieve this for training and inference stacks.

\subsubsection{Distributed Training Infrastructure}\label{sssec:infra-sovereignty-dist-training}
The binding constraint on distributed training is not the total quantity of accelerators on the market, but the number of interconnected nodes obtainable in a single location. Training at scale is bandwidth and latency-sensitive, so capacity fragmented across regions or availability zones is no substitute for a contiguous, well-provisioned cluster. Few locations can supply such a cluster at an acceptable cost, and the set that can shifts over time as provider footprints and individual data centre capacity change.

We therefore treat the ability to relocate as an infrastructure requirement in its own right. During \texttt{Thomson} model training, we have worked with multiple cloud infrastructure providers, with several rapid intra-provider regional migrations undertaken along the way to follow available capacity.

This bears directly on the principle of infrastructure sovereignty \ref{item:sovereignty-infra} defined in Section~\ref{ssec:sovereign_ai_through_continual_learning}. We acknowledge that accelerator hardware remains a residual limitation that no maturity of open-source tooling removes. Portability does not eliminate this but changes its character: what we depend on is accelerator capacity, rather than any single source of it, and continuity of research rests on our ability to re-establish the stack wherever capacity is available. We argue that this is a meaningful movement along a sovereignty spectrum, as argued in Section~\ref{ssec:discussion_limitations}.

\subsubsection{Model Orchestration and Inference}\label{sssec:infra-sovereignty-llm-inference}

We apply the principles of infrastructure sovereignty \ref{item:sovereignty-infra} and control over economics \ref{item:sovereignty-econ} in developing our LLM orchestration and inference solution. Developing this in-house has enabled upfront cost reduction in reserved instances. Additionally, a large number of contemporary models are still unavailable via enterprise options. In such cases, self-hosting allows us to include these models in our benchmarking efforts and better assess our models' parity with newer models that are unavailable by other means. Self-hosting allows our team a high degree of flexibility in hosting proprietary and internal models.

Since April 2026, this system has \textbf{served more than 40M requests} and \textbf{processed 370B} tokens, with \textbf{upwards of a 99.9\% success rate}, with \textbf{cumulative use across the broader organisation reaching 146M requests} and \textbf{nearly 700B tokens in that same period}. In the case of \texttt{GLM-5.2}, self-hosting has saved 19.7\% of costs, compared to third-party inference providers.

While managed vendor offerings arguably reduce total cost of ownership by streamlining complex engineering workflows and reducing onboarding friction into a single offering, this alternative to self-hosting introduces a material long-term risk of vendor lock-in, leaving organisations vulnerable to rising costs without a clear exit strategy to alternative infrastructure. Designing and operating our own LLM serving system allows us a greater degree of control over our uptime, roadmap and value proposition.

%
%

\providecolor{trlightamber}{RGB}{248,234,221}
\providecolor{trdarkamber}{RGB}{212,121,42}
\providecolor{trlightsky}{RGB}{227,241,253}
\providecolor{trdarksky}{RGB}{8,116,227}
\providecolor{trlightteal}{RGB}{227,243,238}
\providecolor{trdarkteal}{RGB}{77,178,153}
\providecolor{trgrayone}{RGB}{249,247,245}
\providecolor{trgraytwo}{RGB}{229,229,229}
\providecolor{trgraythree}{RGB}{159,159,159}
\providecolor{trgrayfour}{RGB}{122,122,122}
\providecolor{trgraphite}{RGB}{33,34,35}

\colorlet{ctrlfill}{trlightsky}
\colorlet{ctrlline}{trdarksky}
\colorlet{storefill}{trlightamber}
\colorlet{storeline}{trdarkamber}
\colorlet{catfill}{trlightteal}
\colorlet{catline}{trdarkteal}
\colorlet{neutfill}{trgraytwo}
\colorlet{neutline}{trgrayfour}
\colorlet{neuttext}{trgrayfour}
\colorlet{sbxline}{trgraphite}
\colorlet{cloudfill}{trgrayone}
\colorlet{cloudline}{trgraythree}

\providecommand{\svctag}[1]{{\scriptsize\itshape\textcolor{neuttext}{#1}}}

\begin{figure}[htbp]
\centering
\resizebox{\textwidth}{!}{%
\begin{tikzpicture}[
  font=\sffamily\small,
  >=Latex,
  unit/.style={
    draw, thick, rounded corners=2pt, align=center,
    inner sep=4pt, text width=42mm, minimum height=13mm},
  ctrl/.style={unit, draw=ctrlline,  fill=ctrlfill},
  sbx/.style={unit, draw=sbxline,    fill=neutfill},
  store/.style={unit, draw=storeline, fill=storefill},
  gpu/.style={unit, draw=ctrlline,   fill=ctrlfill,  text width=56mm},
  swfs/.style={unit, draw=storeline, fill=storefill, text width=56mm},
  cat/.style={unit, draw=catline,   fill=catfill},
  aux/.style={
    draw=neutline, thin, rounded corners=2pt, align=center, fill=neutfill,
    text width=24mm, minimum height=8mm, inner sep=3pt,
    font=\sffamily\footnotesize},
  ext/.style={
    draw=neutline, densely dashed, rounded corners=2pt, align=center,
    fill=white, text width=30mm, minimum height=9mm, inner sep=3pt},
  orch/.style={->, thick, draw=ctrlline},
  data/.style={->, very thick, draw=storeline},
  meta/.style={<->, thick, densely dashed, draw=catline},
  reclaim/.style={->, thick, densely dotted, draw=neutline},
  divider/.style={thick, dashed, draw=cloudline},
  divlbl/.style={font=\sffamily\footnotesize\itshape, text=neuttext,
                 inner sep=2pt},
  lbl/.style={font=\sffamily\scriptsize, inner sep=2pt, align=center,
              fill=white, fill opacity=0.85, text opacity=1},
]

\node[ext] (res) at (4.0, 6.7) {researcher or\\automated pipeline};

\coordinate (evx)        at (-0.2, 0);

\node[ctrl] (cpu)  at (-2.0, 4.2) {\textbf{CPU jobs}};
\node[ctrl] (dms)  at (-2.0, 1.9)
  {\textbf{Data management}\\replication, integrity,\\retention};
\node[ctrl] (jobs) at ( 4.0, 4.2)
  {\textbf{Job orchestration}\\submission and scheduling,\\experiment tracking};
\node[cat]  (hub)  at ( 4.0, 1.9)
  {\textbf{Artefact catalogue}\\URNs, replica locations,\\lineage, system state};
\node[ctrl] (tools) at (-2.0, -0.4) {\textbf{Internal tooling}};
\node[ctrl] (mon)   at ( 4.0, -0.4) {\textbf{Infrastructure monitoring}};

\coordinate (sbxrow) at (0, -3.9);
\node[sbx, anchor=west] (sbx) at (cpu.west |- sbxrow)
  {\textbf{Inspect sandboxes}\\agentic evaluation of\\model-generated code\\\svctag{no egress, no route inward}};
\node[store] (obj) at ( 4.0, -3.9)
  {\textbf{Managed object storage}\\authoritative artefact storage};

\node[gpu]  (gpu)  at (15.8, 4.2) {\textbf{GPU jobs}};
\node[swfs] (swfs) at (15.8, 1.9)
  {\textbf{SeaweedFS}\\in-cluster artefact cache};
\node[aux]  (cop)  at (14.2, -0.7) {replication job};
\node[aux]  (jan)  at (17.4, -0.7) {retention janitor};

\draw[orch] (res) -- node[lbl, right, xshift=1mm, pos=0.2]
  {submit job} (jobs.north);
\draw[orch] (jobs.east) -- node[lbl, above, pos=0.5]
  {dispatch GPU jobs} (gpu.west);

\draw[data] (obj.east) to[out=0, in=200]
  node[lbl, above, pos=0.5] {populate on first reference} (cop.west);
\draw[data] (gpu.south west) to[out=200, in=25]
  node[lbl, above, pos=0.55] {artefact writeback} (obj.north east);
\draw[data] (cop.north) -- (cop.north |- swfs.south);
\draw[data] (swfs.north) -- (gpu.south);
\draw[reclaim] (jan.north) -- node[lbl, right, xshift=1mm, pos=0.5] {reclaim} (jan.north |- swfs.south);

\begin{pgfonlayer}{background}
  \node[draw=ctrlline, thick, rounded corners=3pt, inner sep=4mm, fill=white,
        fit=(cpu)(dms)(jobs)(hub)(tools)(mon),
        label={[anchor=south west, font=\sffamily\footnotesize\bfseries,
                ctrlline, xshift=1mm, yshift=1mm]north west:%
               Primary cluster: control plane}]
       (primarybox) {};
  \node[draw=sbxline, thick, rounded corners=3pt, inner sep=4mm, fill=white,
        fit=(sbx),
        label={[anchor=south west, font=\sffamily\footnotesize\bfseries,
                sbxline, xshift=1mm, yshift=1mm]north west:%
               Evaluation sandbox cluster}]
       (sbxbox) {};
  \node[draw=ctrlline, thick, rounded corners=3pt, inner sep=4mm, fill=white,
        fit=(gpu)(swfs)(cop)(jan),
        label={[anchor=south west, font=\sffamily\footnotesize\bfseries,
                ctrlline, xshift=1mm, yshift=1mm]north west:%
               GPU cluster: accelerator-bound work only}]
       (gpubox) {};
  \coordinate (divx) at ($(primarybox.east)!0.5!(gpubox.west)$);
  \draw[divider] (divx |- 0, 7.4) -- (divx |- 0, -5.5);
  \node[divlbl, anchor=south] at (divx |- 0, 7.4) {vendor boundary};
\end{pgfonlayer}

\draw[meta] (hub.east) -- node[lbl, above, pos=0.5]
  {URN resolution\\and registration} (hub.east -| gpubox.west);
\draw[orch] (primarybox.south -| evx) -- node[lbl, right, xshift=1mm]
  {dispatch evaluations} (sbxbox.north -| evx);

\draw[orch]    (-5.2,-6.3) -- (-4.2,-6.3)
  node[right, font=\sffamily\footnotesize, inner sep=2pt] {job orchestration};
\draw[data]    ( 2.0,-6.3) -- ( 3.0,-6.3)
  node[right, font=\sffamily\footnotesize, inner sep=2pt] {artefact movement};
\draw[meta]    ( 8.2,-6.3) -- ( 9.2,-6.3)
  node[right, font=\sffamily\footnotesize, inner sep=2pt] {catalogue lookup and registration};

\end{tikzpicture}%
}
\caption{Infrastructure topology and the three flows that cross it. The control
plane, the air-gapped evaluation sandboxes and canonical artefact storage all sit
left of the vendor boundary, and do not change when the GPU partner does. Only
one bulk path across that boundary is metered: an artefact is populated into the
in-cluster cache on first reference, so egress is incurred once per cluster per
artefact rather than once per worker per job, and every later read is served
in-cluster.}
\label{fig:infra-topology}
\end{figure}
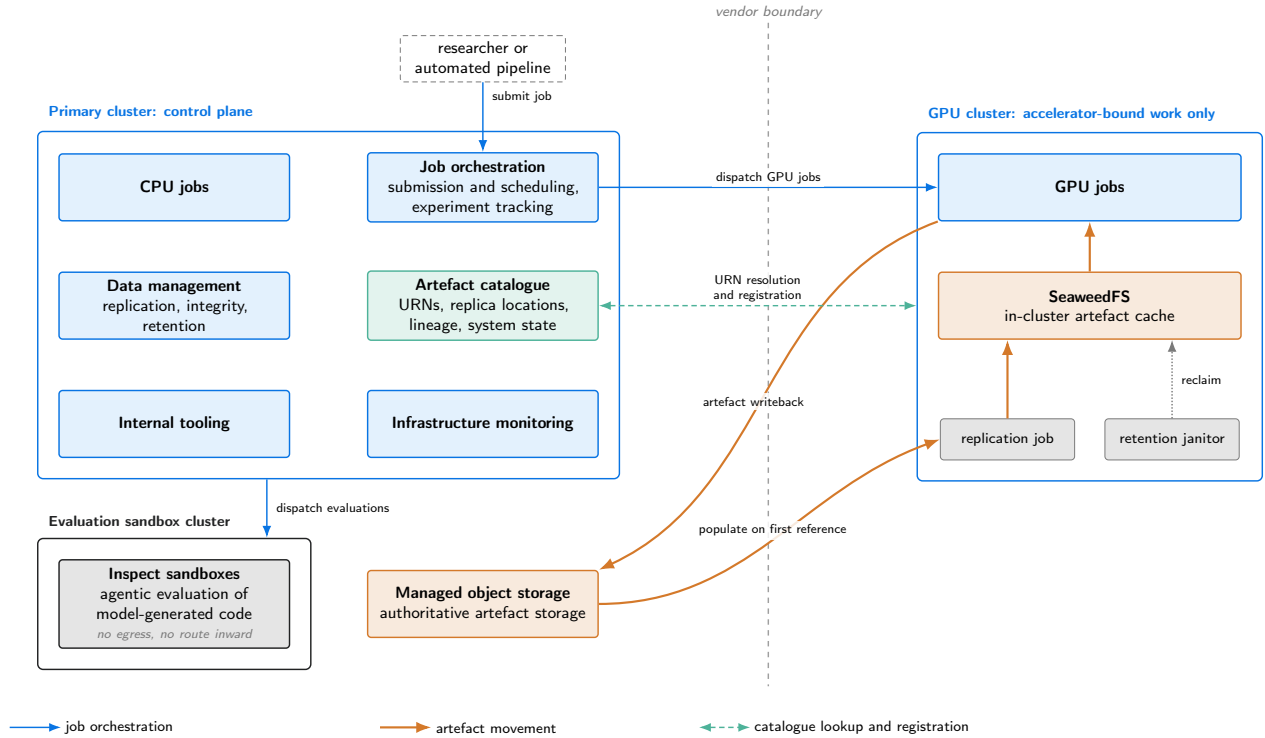

\subsection{Platform Foundations}\label{ssec:infra-platform-foundations}
In this section, we describe foundational capabilities which power our model development platform.

\subsubsection{Cluster Topology}\label{sssec:infra-cluster-topology}
We implement our portability and cost efficiency principles via a three-tier cluster topology (see Figure~\ref{fig:infra-topology}):

\begin{itemize}
    \item \textbf{Primary Cluster} -- The control plane of the platform is hosted on our primary cloud provider. It hosts all CPU-only workloads (data preparation, evaluation harnesses, artefact management, and internal tooling) together with the stateful services on which the rest of the stack depends.
    \item \textbf{Evaluation Sandbox Cluster} --  A second cluster, co-located with the primary cluster, dedicated to agentic evaluations executed via Inspect \citep{inspect_ai}. Because these evaluations execute model-generated code, the cluster is air-gapped and firewalled: sandboxes have no network egress and no route to internal systems. This isolation permits agentic evaluations to be run at scale without exposing the wider platform or external networks to the actions of the system under evaluation.
    \item \textbf{GPU Cluster} -- A single accelerator cluster, provisioned by the current specialised compute partner, dedicated exclusively to GPU-bound workloads. No CPU-only work is scheduled here, which keeps scarce accelerator capacity fully committed to training and inference.
\end{itemize}

This separation of concerns yields two properties central to our operating model. First, the stateful, long-lived components of the platform remain anchored in a single, well-understood environment and are unaffected by changes in GPU provider or region. Second, workloads whose isolation requirements differ materially, namely routine training and the execution of untrusted agent-generated code, are held in physically distinct environments isolated from any network or internet access rather than by policy alone.

\subsubsection{Job Orchestration} \label{sssec:infra-job-orchestration}
Job submission, scheduling, and experiment tracking are handled by a substantially modified fork of the open-source ClearML \citep{clearml} platform, extended to accommodate the heterogeneity of our execution environments. This is paired with an internally developed job submission and inspection interface. This abstracts the mechanics of the target execution environment to the orchestration layer. This is what makes relocation tractable: standing up a new GPU cluster, whether with a new partner or in a new region, alters the set of available targets without altering how jobs are written.

\subsubsection{Data Access and Provenance} \label{sssec:infra-data-provenance}
%
%
%
\definecolor{dtobase}{HTML}{0874E3}    
\definecolor{dtoactual}{HTML}{D4792A}  
\definecolor{dtoband}{HTML}{F8EADD}    
\definecolor{dtomark}{HTML}{6B6B6B}    
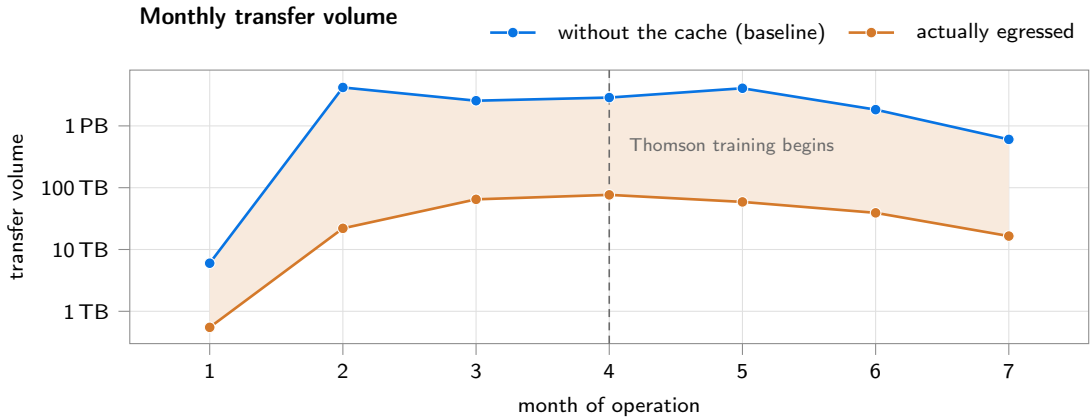
\begin{figure}[htbp]
\centering
\begin{tikzpicture}[font=\sffamily\small]
\begin{axis}[
  width=0.86\linewidth, height=52mm,
  xmin=0.4, xmax=7.6,
  xtick={1,2,3,4,5,6,7},
  xticklabels={1,2,3,4,5,6,7},
  xlabel={month of operation},
  tick align=outside, tick pos=left,
  every tick/.style={black!45, thin},
  axis line style={black!45},
  grid=major, grid style={black!12, thin},
  label style={font=\sffamily\footnotesize},
  tick label style={font=\sffamily\footnotesize},
  legend style={font=\sffamily\footnotesize, draw=none, fill=none,
                at={(1,1.06)}, anchor=south east, cells={anchor=west},
                legend columns=2, column sep=6pt},
  ymode=log, log basis y=10,
  ymin=3e-4, ymax=8,
  ytick={1e-3,1e-2,1e-1,1},
  yticklabels={1\,TB,10\,TB,100\,TB,1\,PB},
  ylabel={transfer volume},
  title={\sffamily\small\bfseries Monthly transfer volume},
  title style={at={(0,1.06)}, anchor=south west, font=\sffamily\small\bfseries},
]
\addplot[fill=dtoband, draw=none, forget plot]
  coordinates {(1,0.006) (2,4.188) (3,2.556) (4,2.874) (5,4.064) (6,1.833) (7,0.605)
               (7,0.0165) (6,0.0392) (5,0.0589) (4,0.0766) (3,0.0648) (2,0.0220) (1,0.00055)}
  \closedcycle;
\addplot[dtomark, densely dashed, line width=0.6pt, forget plot]
  coordinates {(4,3e-4) (4,8)};
\node[anchor=north west, font=\sffamily\scriptsize, text=dtomark]
  at (axis cs:4.08,0.9) {Thomson training begins};
\addplot[dtobase, line width=1pt, mark=*, mark size=2pt,
         mark options={fill=dtobase, draw=white, line width=0.4pt}]
  coordinates {(1,0.006) (2,4.188) (3,2.556) (4,2.874) (5,4.064) (6,1.833) (7,0.605)};
\addlegendentry{without the cache (baseline)}
\addplot[dtoactual, line width=1pt, mark=*, mark size=2pt,
         mark options={fill=dtoactual, draw=white, line width=0.4pt}]
  coordinates {(1,0.00055) (2,0.0220) (3,0.0648) (4,0.0766) (5,0.0589) (6,0.0392) (7,0.0165)};
\addlegendentry{actually egressed}
\end{axis}
\end{tikzpicture}
\caption{Data-transfer-out avoided by the in-cluster artefact cache on the GPU
cluster, shown by month of operation on a logarithmic axis. The baseline is the
counterfactual in which every worker of every job pulls its inputs directly
from object store; the lower series is what was in fact transferred, and the shaded band
between them is the transfer the cache avoided. Read demand over the window
totalled 16.13\,PB against 278.7\,TB actually egressed, a factor of 57.9. The
dotted rule marks the start of \texttt{Thomson} model training. Months 1 and 7 are
partial. The baseline is a construction rather than a measurement, so this is
transfer the cache avoided rather than budget that would otherwise have been
spent.}
\label{fig:dto-savings}
\end{figure}

\begin{table}
\centering
\small
\begin{tabular}{lrrr}
\toprule
Month & Baseline & Egressed & Avoided \\
\midrule
1\textsuperscript{$\dagger$} &   5.72\,TB & 546.98\,GB &   5.17\,TB \\
2                            &   4.19\,PB &  22.00\,TB &   4.17\,PB \\
3                            &   2.56\,PB &  64.83\,TB &   2.49\,PB \\
4                            &   2.87\,PB &  76.61\,TB &   2.80\,PB \\
5                            &   4.06\,PB &  58.94\,TB &   4.01\,PB \\
6                            &   1.83\,PB &  39.21\,TB &   1.79\,PB \\
7\textsuperscript{$\dagger$} & 605.35\,TB &  16.54\,TB & 588.81\,TB \\
\midrule
\textbf{Total} & \textbf{16.13\,PB} & \textbf{278.68\,TB} & \textbf{15.85\,PB} \\
\bottomrule
\end{tabular}
\caption{Data-transfer-out avoided by the in-cluster artefact cache on the GPU cluster, by month of operation; the data plotted in Figure~\ref{fig:dto-savings}. The baseline is the counterfactual in which every worker of every job pulls its inputs directly from the object store; \emph{avoided} is that baseline less what was actually egressed. \textsuperscript{$\dagger$}Partial months.}
\label{tab:dto-savings}
\end{table}

Hosting compute and control planes in separate clouds introduces a data-gravity problem. Training artefacts (datasets, checkpoints, and tokenised corpora) must be made available to the GPU cluster at high throughput, while egress from the primary cloud is metered and therefore materially expensive. Na\"ive replication strategies are prohibitive on both latency and cost grounds. To address this, we implemented file access indirection -- a unified reference to artefacts through stable logical identifiers rather than physical storage locations. This allowed the platform to resolve each reference to the most appropriate local replica at execution time. The stack is built upon two open-source foundations, DataHub \citep{datahub} and SeaweedFS \citep{seaweedfs}.
\paragraph{Artefact Identity}\phantomsection\label{par:artifact-identity}%
DataHub is the source of truth for the indirection layer. Every model, dataset, and derived artefact is registered under a DataHub Uniform Resource Name (URN), a stable and globally unique identifier that names the asset independently of where any copy of it currently resides. These URNs constitute the shared vocabulary of the platform. Canonical storage for datasets and models is always in the primary cloud object store, which holds the authoritative copy of every artefact.

\paragraph{Replica Resolution and Transfer Cost}\phantomsection\label{par:replica-resolution}%
Alongside canonical storage, DataHub records the availability of local replicas on a per-cluster basis, including those held in the SeaweedFS cache co-located with the GPU cluster. Resolving a URN involves consulting the catalogue for a replica local to the executing cluster. Where none is registered, the artefact is first populated into the local cache from canonical storage and the read is then served locally, with the new replica recorded in the catalogue. Egress from the primary cloud is thus incurred once per cluster-per-artefact, rather than once per worker-per-job. The distinction is substantial (Table~\ref{tab:dto-savings}): between February and August 2026 read demand on the GPU cluster totalled \textbf{16.13\,PB against 278.7\,TB actually egressed}. We estimate this led to savings of over \textbf{USD 1.4M} in avoided egress costs in this period. Absent the cache, every rank of a distributed job would fetch the same checkpoints and dataset shards independently, and the transfer volume would scale with the product of the job count and the worker count rather than remaining fixed per artefact. Every subsequent consumer, whether another rank of the same job or a job run months later, reads at in-cluster bandwidth. Because resolution occurs at the moment of use, replicas may be created, evicted, or invalidated as clusters come and go without affecting any reference held elsewhere.

\paragraph{In-Cluster Cache}\phantomsection\label{par:in-cluster-cache}%
SeaweedFS provides the physical substrate that URN resolution targets inside the GPU cluster, operating as a local cache through which every model and dataset is replicated before it is read, so that reads are served from within the same network fabric as the accelerators consuming them. Access is mediated by a custom client, which retrieves file metadata from the SeaweedFS filer and then downloads chunks in parallel directly from the volume servers holding them. Under this scheme a single reading process sustains approximately 160 Gbit/s from the storage tier into memory, and approximately 40 Gbit/s when each file is additionally written to local NVMe. Both figures are rates observed by one consumer rather than aggregates over concurrent readers, and the lower of the two is bounded by local disk rather than by the network or by the storage tier. Checkpoint and dataset loading are consequently removed as a bottleneck at job startup.

\paragraph{Lineage and Governance.}\phantomsection\label{par:lineage-governance}%
The same catalogue records lineage. Every artefact is registered with the job that produced it and with each job that subsequently consumes it, yielding a complete, queryable provenance graph across the training pipeline. This record supports reproducibility and post-hoc analysis: for any released checkpoint, the constituent datasets, preprocessing stages, and upstream training runs can be recovered exactly. Beyond provenance, the catalogue exposes user-defined tags and properties, operationally critical system-level state like replication status, cache LRU state, file integrity information, and related bookkeeping. This also gives us a real-time provenance view as an operational artefact, critical for demonstrating controls for highly regulated environments.

\subsubsection{Tool Calling}
\label{sec:infra-tooling}

Tool calling is critical to ground the model's responses in accurate citations and factuality. To standardise this across the model development lifecycle, we developed a unified tool-calling infrastructure to enable calls to proprietary and external tools. This helped us standardise request-response schemas, authentication, and state management. We adopted an async-first design, which helped manage the high volume of concurrent tool calls across the training run. We keep  a low memory footprint (approx. 2MB) per worker to enable lightweight scaling across distributed training. We also accounted for partial failures for tool calls to avoid blocking training runs.

\subsubsection{Synthetic Data Generation}
\label{sec:infra-synth-pipeline}

Our post-training pipeline consumed a large number of distinct data collections, each requiring generation,
filtering, validation and publication before it could enter a training mixture. We treat synthetic data generation also as pipeline infrastructure, to be reused across all data collections.

We unify our pipeline for model requests against a tabular dataset, agnostic of model inference provider. This layer also manages connection pooling, request batching, error isolation and handling.

The pipelines are modelled as directed acyclic graphs (DAGs) comprising independently versioned steps of generation, filtering and scoring. Each step is independently executable, affording the capability of prompt-level iteration. This also helps in lineage tracking and artefact provenance for every transformation -- including query hash, source text, reasoning chain and reference answer, allowing complete traceability for each training run. This is a fundamental requirement for heavily regulated environments in which \texttt{Thomson-1.0-Large} is expected to be used.

\subsection{Training Environment} \label{sec:infra-training}
\texttt{Thomson} models are trained on the cluster as described in Section~\ref{sssec:infra-sovereignty-dist-training}, with nodes of 8x Nvidia B200. Communication between nodes is over an InfiniBand network, while intra-node communication is achieved via NVLink and NVSwitch. The typical number of nodes used for training is reported by task in Table~\ref{tab:cluster-nodes}.

\begin{table}[htbp]
\centering
\footnotesize
\setlength{\tabcolsep}{3.5pt}
\renewcommand{\arraystretch}{1.05}
\resizebox{\ifdim\width>\textwidth\textwidth\else\width\fi}{!}{%
\begin{tabular}{l l c c l}
\toprule
\textbf{Model} & \textbf{Role} & \textbf{Nodes} & \textbf{GPUs} & \textbf{Parallelism} \\
\midrule
\multirow{4}{*}{\textbf{Thomson-1.0-Small}}
    & CPT                & 16 & 128 & TP1, PP1, CP1, EP8, ETP1 \\
    & DPO                & 6  & 48  & TP1, PP1, CP1, EP8, ETP1 \\
    & Policy training    & 8  & 64  & TP2, PP1, CP8, EP8, ETP1 \\
    & Rollout generation & 4  & 16  & TP8 (4 independent engines) \\
\cmidrule(l){2-5}
\multirow{4}{*}{\textbf{Thomson-1.0-Large}}
    & CPT                & 16 & 128 & TP1, PP8, CP16, EP8, ETP1 \\
    & DPO                & 16 & 128 & TP1, PP8, CP16, EP8, ETP1 \\
    & Policy training    & 16 & 128 & TP1, PP8, CP16, EP8, ETP1 \\
    & Rollout generation & 10 & 80  & TP8 (10 independent engines) \\
\bottomrule
\end{tabular}}
\caption{\centering Compute cluster configurations by training task.}
\label{tab:cluster-nodes}
\end{table}

For RL, generation and training are \emph{non-colocated}: each runs on its own dedicated node pools, which is a precondition for the asynchronous rollout collection described in Section~\ref{sssec:infra-sovereignty-dist-training}.

%
%

\providecommand{\genfillx}{}

\definecolor{genfill}{RGB}{227,241,253}\definecolor{genline}{RGB}{8,116,227}
\definecolor{trainfill}{RGB}{248,234,221}\definecolor{trainline}{RGB}{212,121,42}
\definecolor{judgefill}{RGB}{240,240,240}\definecolor{judgeline}{RGB}{97,97,97}
\definecolor{envfill}{RGB}{245,245,245}\definecolor{envline}{RGB}{130,130,130}

\providecommand{\fwtag}[1]{{\tiny\itshape\textcolor{black!55}{#1}}}

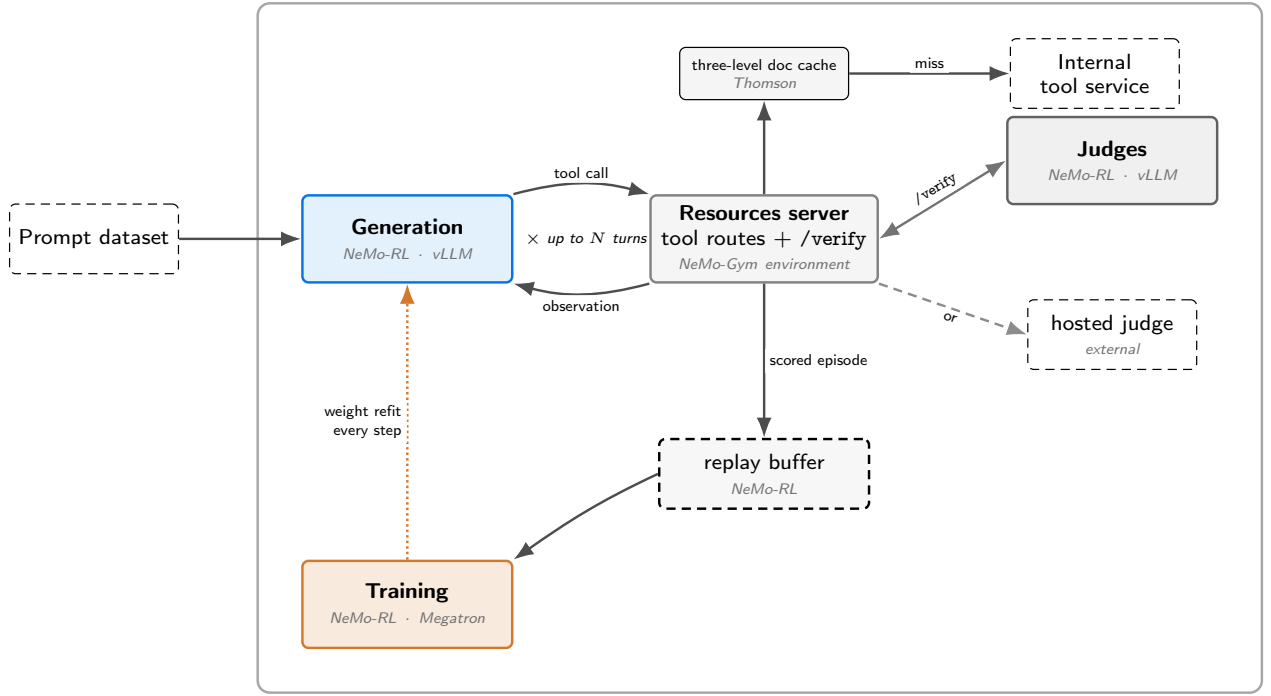
\begin{figure}[htbp]
\centering
\resizebox{\columnwidth}{!}{%
\begin{tikzpicture}[
  font=\sffamily\scriptsize,
  >=Latex,
  pool/.style={
    draw, thick, rounded corners=2pt, align=center,
    minimum height=10mm, text width=22mm, inner sep=3pt},
  gpool/.style={pool, draw=genline,   fill=genfill},
  tpool/.style={pool, draw=trainline, fill=trainfill},
  jpool/.style={pool, draw=judgeline, fill=judgefill},
  envbox/.style={pool, draw=envline,  fill=envfill, text width=24mm},
  buffer/.style={
    draw, thick, densely dashed, rounded corners=2pt, align=center,
    fill=black!3, minimum height=8mm, text width=22mm, inner sep=3pt},
  ext/.style={
    draw, densely dashed, rounded corners=2pt, align=center,
    fill=white, minimum height=8mm, text width=18mm, inner sep=2pt},
  small/.style={
    draw, thin, rounded corners=2pt, align=center,
    fill=black!4, minimum height=6mm, text width=18mm, inner sep=2pt,
    font=\sffamily\tiny},
  flow/.style={->, thick, draw=black!70},
  refit/.style={->, thick, draw=trainline, densely dotted},
  call/.style={<->, thick, draw=black!55},
  optional/.style={->, thick, densely dashed, draw=black!45},
  lbl/.style={font=\sffamily\tiny, inner sep=1.5pt, fill=white,
              fill opacity=0.85, text opacity=1},
]

\node[ext]     (data)  at (-3.6, 1.6) {Prompt dataset};
\node[gpool]   (gen)   at ( 0.0, 1.6)
  {\textbf{Generation}\\\fwtag{NeMo-RL $\cdot$ vLLM}};
\node[tpool]   (train) at ( 0.0,-2.6)
  {\textbf{Training}\\\fwtag{NeMo-RL $\cdot$ Megatron}};

\node[envbox]  (rs)    at ( 4.1, 1.6)
  {\textbf{Resources server}\\tool routes $+$ \texttt{/verify}\\\fwtag{NeMo-Gym environment}};

\node[small]   (cache) at ( 4.1, 3.5) {three-level doc cache\\\fwtag{Thomson}};
\node[ext]     (ras)   at ( 7.9, 3.5) {Internal tool service};

\node[jpool]   (judge) at ( 8.1, 2.5)
  {\textbf{Judges}\\\fwtag{NeMo-RL $\cdot$ vLLM}};
\node[ext]     (host)  at ( 8.1, 0.5) {hosted judge\\\fwtag{external}};

\node[buffer]  (buf)   at ( 4.1,-1.1)
  {replay buffer\\\fwtag{NeMo-RL}};

\draw[flow] (data) -- (gen);

\draw[flow] (gen.north east) to[bend left=15]
  node[lbl, above, pos=0.5] {tool call} (rs.north west);
\draw[flow] (rs.south west) to[bend left=15]
  node[lbl, below, pos=0.5] {observation} (gen.south east);
\node[font=\sffamily\tiny\itshape] at (2.05, 1.6) {$\times$ up to $N$ turns};

\draw[flow] (rs.north) -- (cache.south);
\draw[flow] (cache.east) -- node[lbl, above] {miss} (ras.west);

\draw[call]     (rs.east)      -- node[lbl, above, sloped] {\texttt{/verify}} (judge.west);
\draw[optional] (rs.south east) -- node[lbl, below, sloped] {or}              (host.west);

\draw[flow] (rs.south) -- node[lbl, right] {scored episode} (buf.north);
\draw[flow] (buf.west) to[out=205, in=35] (train.north east);

\draw[refit] (train.north) -- node[lbl, left, align=right, pos=0.5]
  {weight refit\\every step} (gen.south);

\begin{pgfonlayer}{background}
  \node[draw=black!35, thick, rounded corners=4pt, inner sep=5mm,
        fit=(gen)(rs)(judge)(buf)(train)(cache)]
       (clusterbox) {};
\end{pgfonlayer}

\end{tikzpicture}%
}
\caption{The training loop across the three node pools, annotated with the
component that provides each stage. Generation, training and judge inference
occupy disjoint node pools. Scoring is a call
the resources server makes, to the on-cluster judge pool or optionally to a hosted
model. The replay buffer decouples generation from training, which is what makes
the two pools run concurrently and is where the trajectory-staleness bound
applies.}
\label{fig:infra-training-loop}
\end{figure}

As discussed in Section~\ref{ssec:training-framework}, we run our training workloads with the Nvidia NeMo-RL \citep{nemo-rl} framework. Training jobs in NeMo-RL are run over a Ray \citep{ray} cluster that spawns instances from different resource groups, namely for training and generation; this allows us to assign each node a specific role and isolate generation resources from training ones to avoid colocation and enable asynchronous rollouts. See Figure~\ref{fig:infra-training-loop} for a schematic representation.

\subsubsection{Training Stack} \label{ssec:training-framework}
Our training library extends Nvidia NeMo-RL, with Megatron-LM \citep{megatron-lm} core to provide model parallelism and vLLM \citep{kwon2023vllm} for generation. NeMo-RL implements distributed worker orchestration via Ray, the GRPO and DPO algorithms, and checkpoint tooling based on PyTorch Distributed. On top of this we implement 28 custom extensions together with custom NeMo-Gym \citep{nemo-gym} RL task environments and a collection of specialised reward functions.

We implemented a few distinct extensions of NeMo-Gym to support our training framework: \hyperref[par:non-vanishing-grad]{avoiding zero learning signal}, \hyperref[par:policy-grad-variants]{policy-gradient variants},  and  \hyperref[par:cohort-scoring]{failure-resilient cohort scoring}. Some of these are NeMo-specific feature changes rather than customisations for our use case, and hence could be future contributions to the library.

\paragraph{Avoiding Zero Learning.}\phantomsection\label{par:non-vanishing-grad}%
Learning with GRPO is only possible if completions of the same prompt receive different rewards. We noticed two issues in the NeMo-RL framework that prevent different completions in the same group from receiving distinct rewards in the context of agentic multi-turn tasks.

\textit{Sampling diversity} -- For each generation engine, the framework feeds one shared seed but does not specify a per-request seed. We assign a different unique seed to every generation request to prevent identical outputs and preserve the reward variation GRPO needs to learn.

\textit{Advantage Grouping in Multi-Turn Rollouts} -- Samples are grouped by the full conversation history. Because tools are added as user messages, which differ across trajectories, this generates N different groups for each original prompt, each with a single sample and thus zero advantage. We modify the grouping mechanism by tracing only the first user message and collecting all trajectories under this message. This ensures a group with N generations, all likely to be different. Single-turn tasks are not affected by these issues in the original framework, and behave as expected.

\paragraph{Policy-Gradient Variants.}\phantomsection\label{par:policy-grad-variants}%
Policy-gradient loss uses a clipping ratio and a separate correction for differences between the trainer's policy and vLLM's sampling policy. We add variants, all disabled by default, so that the original behaviour is preserved unless explicitly enabled.

\textit{Length-normalised train--inference ratio} -- In the sequence-level GSPO loss, the train--inference correction was originally computed as

$$\exp\left(\sum_t \left(\log \pi_{\text{old}} - \log \pi_{\text{gen}}\right)\right) \sim \exp\left( N_t \epsilon \right),$$

so it increased with response length. When enabled, this option instead uses a masked mean over tokens:

$$\exp\left(\operatorname{mean}_t\left(\log \pi_{\text{old}} - \log \pi_{\text{gen}}\right)\right) = \exp\left( \frac{1}{N_t} \sum_t \left(\log \pi_{\text{old}} - \log \pi_{\text{gen}}\right)\right) \sim \exp\left( \epsilon \right) $$

matching the policy ratio and removing length bias for responses that vary widely in size.

\textit{Reference-Model Removal} -- In GRPO, the reference model can be used to compute the KL penalty. If the penalty weight is fixed to zero, the log-probabilities computation is pure overhead and the framework provides a path to avoid computation for the synchronous training loop. We extend this guard to the asynchronous case and moreover introduce the possibility of avoiding reference model initialisation as well, to save memory.

\paragraph{Cohort Reward Reconciliation.}\phantomsection\label{par:cohort-scoring}%
In practice, individual rewards can fail because of judge timeouts or backend throttling. Simply omitting a failed reward means that completion's weighted score is averaged over fewer terms than its peers. Resulting scores end up on divergent scales, so advantages may reflect infrastructure reliability, rather than response quality. This creates systematic bias whenever failures are unevenly distributed within a cohort. To mitigate this, we implemented a mean substitution policy: missing rewards are replaced with the cohort mean among completions where that reward succeeded, preserving a consistent weight basis. This substitution artificially reduces variance among the affected rollouts, however -- and under advantage normalisation schemes that divide by a group's reward standard deviation, this variance collapse can inflate the relative advantage of unaffected rollouts in the same group, distorting the training signal in proportion to how much of the cohort was imputed. As a correction measure, we track each rollout's proportion of genuinely observed versus imputed reward and use it to scale down its group's advantage -- discounting, rather than discarding, cohorts that relied heavily on substitution. Rollouts whose reward is entirely imputed or failed outright are excluded from the loss altogether.

\subsubsection{RL Environments} \label{ssec:task-environments}
For training \texttt{Thomson}, we implement RL environments as a generic HTTP interface following NeMo-Gym syntax, detailed as follows:

\paragraph{Multi-Turn Legal Research environment.}\phantomsection\label{par:multi-turn-legal-env}%
This is the principal environment with a focus on legal research, a key modelling objective. It feeds the model legal research questions and lets it complete the task with internal tools dynamically.

\paragraph{Secure Local Sandboxed Workspace environment.}\phantomsection\label{par:sandbox-env}%
A dedicated environment to safely run model-authored shell commands (read, write, edit, glob and grep) local to the training cluster: each tool call is executed inside an Apptainer namespace, where the host filesystem is not visible, network namespace is empty, read-write is limited to a per-session $\mathrm{\text{/tmp}}$ folder. This environment is designed for small workloads and complements the sandbox cluster discussed in Section~\ref{sssec:infra-cluster-topology}.

\paragraph{System Prompt Context Management for Tool Results.}\phantomsection\label{par:system-prompt-context}%

In Section~\ref{sec:infra-tooling}, we describe our platform's tool calling capabilities. We extend these for our training needs by introducing a context management layer. Many tools return long text documents, which if directly added to the model's system prompt can bloat the model context quickly. To mitigate this, our context management layer enforces token limits for call results, structure-aware truncation of tool results and summarises long results.

\begin{figure}[htbp]
\centering
\includegraphics[width=0.9\linewidth]{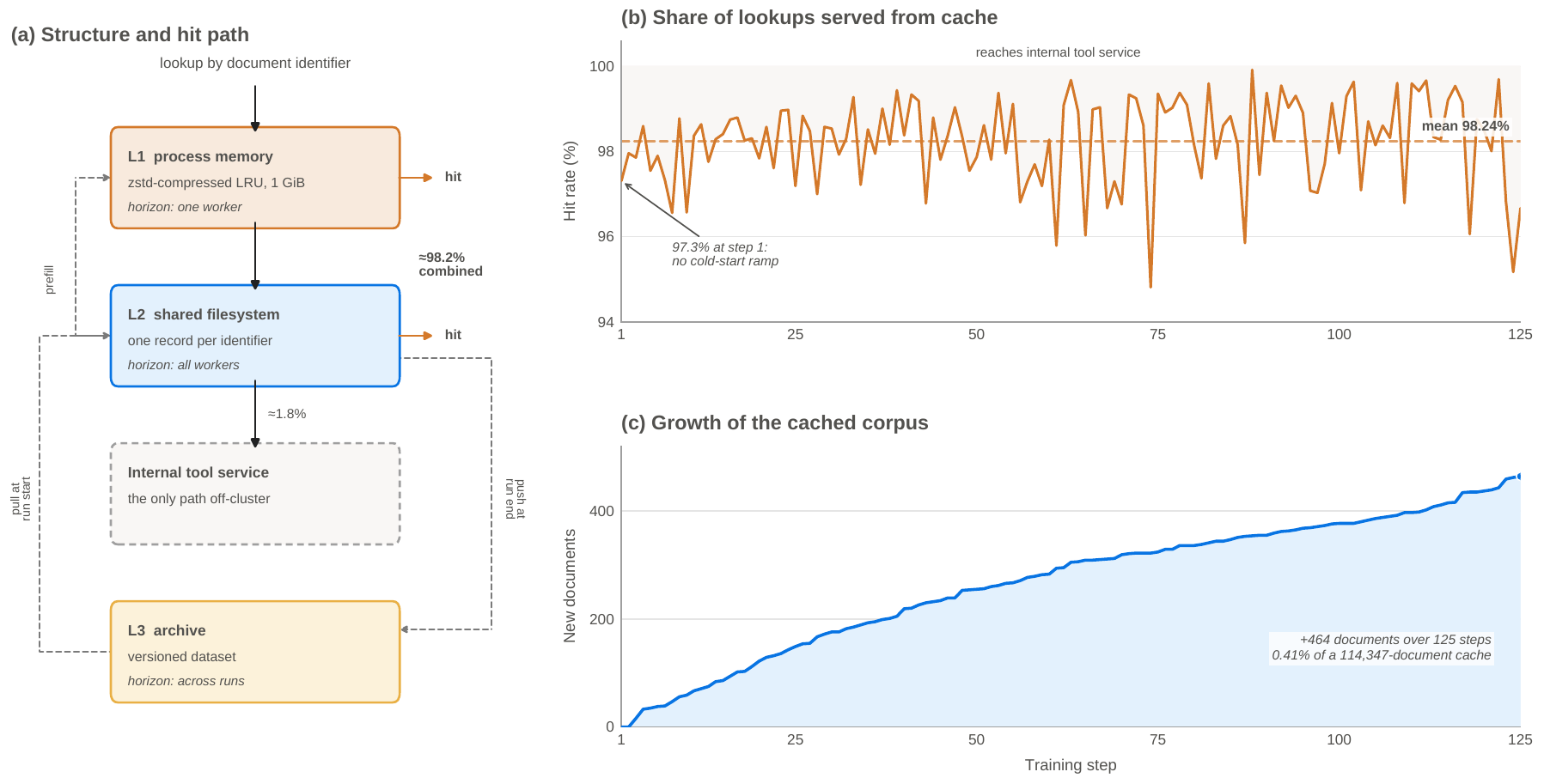}
\caption{The three-level document cache, with panels (b) and (c) measured over
a 125-step test run. \textbf{(a)} A lookup falls through the tiers, each answering a
different reuse horizon: within a worker, across workers, and across runs. Only
a level-2 miss leaves the cluster. Two out-of-band paths matter as much as the
read path -- L2 prefills L1 when a worker starts, and the L2 directory is
pushed to the archive at run end and pulled back at the start of the next.
\textbf{(b)} Across 329\,389 lookups (${\approx}2\,600$ per step) the cache
answered $98.2\%$, leaving $5\,763$ calls -- about 46 per step -- to reach the
internal tool service. \textbf{(c)} The cached corpus grew by 464 documents, $0.41\%$ of
its starting size, and the rate of new documents roughly halves across the run
(149 in the first 25 steps against 87 in the last 25) -- the run converges onto
a working set that the archive already contained. Consistent with that, no L1
evictions occurred.}
\label{fig:document-cache}
\end{figure}

\paragraph{Document Caching for Tool Call Optimisation.}\phantomsection\label{par:ras}%
We use our tool calling layer described in Section~\ref{sec:infra-tooling} for calling internal tools during training. This allows dynamic inclusion of more tools to increase the model's actionable functionality. A concrete example is the document retrieval required for evaluating claims' factuality. To avoid redundant calls to internal tools and optimise costs, we implement a three-tier cache for retrieved documents (Figure~\ref{fig:document-cache}).

\begin{itemize}
    \item \textbf{Level 1 -- In-process memory.} Each worker maintains bounded, in-memory lookup structures for resolution outcomes and document bodies, evicted under a least-recently-used policy, caching only successful lookups. The document-body cache is sized by cumulative byte budget rather than entry count since response length varies enormously and capped at the lesser of a configured ceiling and a fraction of currently available system memory, with retained text compressed to extend the effective working set. All entries are zstd-compressed.
    \item \textbf{Level 2 -- Shared filesystem.} A filesystem-backed cache, shared across processes, runs, and (on network storage) nodes, holds two parallel structures: a document store, partitioned by jurisdiction and sharded by GUID prefix to bound directory size; and a citation-resolution store, keyed by document GUID, with each record accumulating every distinct citation surface form observed to resolve to it -- since one authority is routinely cited in several superficially different but equivalent forms. Writes are atomic (temporary file plus rename) for safety under concurrent access, and all filesystem operations are timeout-bounded and run on an isolated thread pool, so a stalled network filesystem degrades to a cache miss rather than stalling scoring. The in-memory tier may be pre-warmed from disk at startup to avoid a cold ramp-up.
    \item \textbf{Level 3 -- Cross-run archive.} The cache directory is pushed to our internal data platform as a versioned dataset, accessible to any future run limiting the cost of retrieval over multiple runs.
\end{itemize}

Two pathways populate the same persistent structures using an identical record format, so each benefits from the other's writes.
While evaluating a claim's citations, the system resolves each citation via the resolution store (falling back to the backend on a miss) and satisfies the resulting document request via the document store (falling back to a live fetch), writing successful results back to both stores.
When the model invokes a document-retrieval tool during generation, the request is checked against the cache first; a live response carrying a canonical identifier is written into the same store.
So documents the model already fetched either during generation or claim evaluation are not fetched again.

\subsubsection{Training Performances} \label{ssec:train-performance}

\begin{figure}[htbp]
\centering
\includegraphics[width=0.7\linewidth]{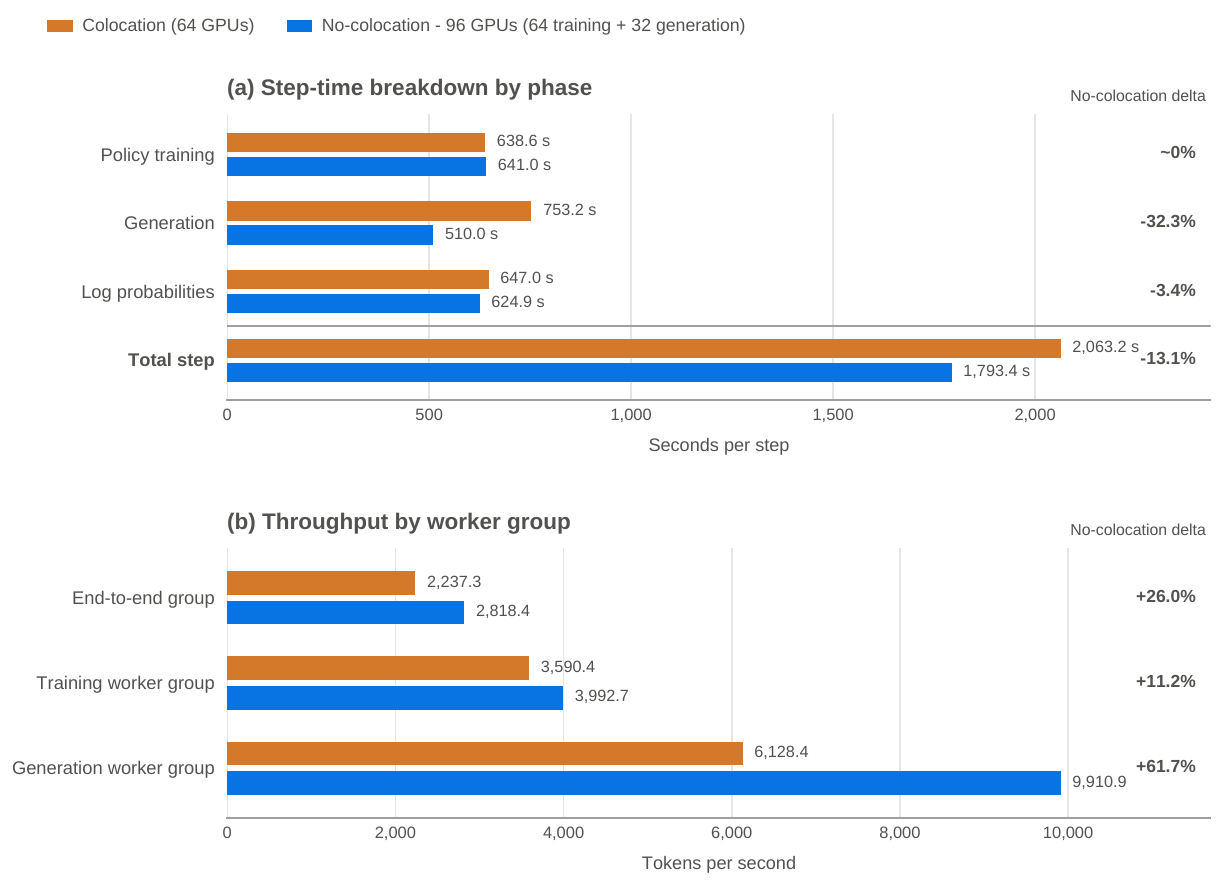}
\caption{Comparison between \texttt{colocation} (64 GPUs, colocated
generation and training) and \texttt{no-colocation} (64 training GPUs plus a 32 GPUs dedicated
generation pool). ``WG'' denotes Worker Group, the pool of GPUs assigned to
a given role.}
\label{fig:colo-vs-nocolo-pct}
\end{figure}

Among the performance tests for our training stack, we measured, for a large MoE model (\texttt{Qwen3.5-397B-A17B}), the impact of decoupling generation from training to separate GPU pools (non-colocate strategy) rather than colocating them on shared nodes. Comparing performance metrics under identical batch, sequence-length, and parallelism configuration, non-colocation reduced total step time by 13.1\% (see Figure~\ref{fig:colo-vs-nocolo-pct}), with the gain concentrated almost entirely in the \textbf{generation phase (-32.3\%)}; training and logprob computation were essentially unchanged (+0.4\% and -3.4\% respectively), as expected since both use identical PP=8 parallelism regardless of colocation strategy. The effect is also measurable in raw throughput: non-colocation delivered a \textbf{61.7\% increase in Generation Worker Group tokens/sec}, despite the colocated run using two times the generation parallelism (TP/EP=16 vs. 8).

These results show that the main drawback of sharing GPUs is not the parallel setup: it is that training and generation compete for the same GPU memory and computing power, even though they need those resources differently. Giving each task its own GPU pool greatly improves throughput with little extra cost.

\subsection{Inference Infrastructure}\label{ssec:inference-infra}
\subsubsection{Architecture}
Our LLM orchestration and inference platform is built on Kubernetes, with a routing, authorisation, and rate-limiting plane implemented in Rust, using vLLM \citep{kwon2023vllm} as an inference backend. It operates over 11 independent clusters spread across 4 geographies and 2 cloud providers, and the platform as a whole is divided into isolated functional domains -- clusters operate as routing clusters or inference clusters, not both. This allows for a much more proactive approach to blast radius containment and significantly more deterministic failure modes; while the result is an incremental increase in the overall management complexity in one respect (i.e., the number of clusters is higher than it might be otherwise), it ensures that the issues that can occur in any given cluster have fewer interacting (confounding) failures. Further, routing being decoupled from inference means that either can be more straightforwardly experimented with -- a new routing environment can be provisioned from scratch, tested, and thrown away within the same business day, allowing a greater degree of lifecycle decoupling than we would be able to achieve otherwise, all ultimately contributing to a greater degree of uptime.

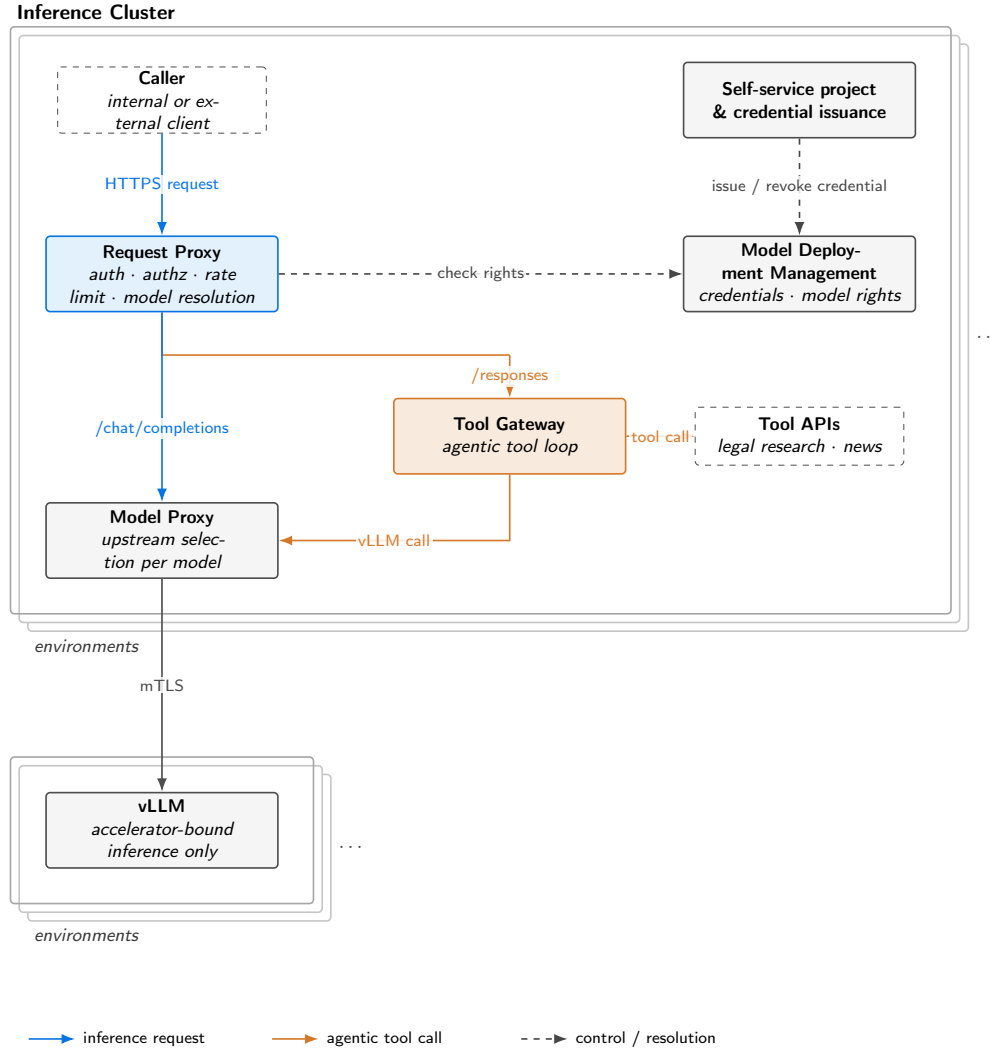
\begin{figure}[H]
\centering
\adjustbox{max width=\textwidth, max totalheight=0.6\textheight}{%
\begin{tikzpicture}[
  font=\sffamily\small,
  >=Latex,
  comp/.style 2 args={
    rectangle, rounded corners=2pt, draw=#1, fill=#2, thick,
    minimum width=40mm, minimum height=13mm, align=center, inner sep=2pt,
    text width=38mm
  },
  ext/.style={
    rectangle, rounded corners=2pt, draw=gray!60!black, dashed,
    minimum width=36mm, minimum height=10mm, align=center, inner sep=2pt,
    text width=34mm
  },
  loc/.style={rectangle, rounded corners=2pt, draw=gray!70, thick, inner sep=0pt},
  envstack/.style={rectangle, rounded corners=2pt, draw=gray!45, thick, fill=white},
  reqarrow/.style={-{Latex}, thick, trdarksky},
  toolarrow/.style={-{Latex}, thick, trdarkamber},
  ctrlarrow/.style={-{Latex}, thick, gray!55!black, dashed},
  lbl/.style={font=\sffamily\footnotesize, align=center, fill=white, fill opacity=0.85, text opacity=1, inner sep=1pt},
  boxtitle/.style={font=\sffamily\bfseries, gray!55!black, align=center, fill=white, fill opacity=0.85, text opacity=1, inner sep=1pt}
]

\node[ext]                  (caller)    at (-2, 15)   {\textbf{Caller}\\{\itshape internal or external client}};

\node[comp={gray!60!black}{gray!8}]      (dashboard) at (9, 15)   {\textbf{Self-service project \& credential issuance}};
\node[comp={gray!60!black}{gray!8}]      (mdm)       at (9, 12)   {\textbf{Model Deployment Management}\\{\itshape credentials $\cdot$ model rights}};

\node[comp={trdarksky}{trlightsky}]      (proxy)     at (-2, 12)   {\textbf{Request Proxy}\\{\itshape auth $\cdot$ authz $\cdot$ rate limit $\cdot$ model resolution}};
\node[comp={trdarkamber}{trlightamber}]  (gateway)   at (4,  9.2)  {\textbf{Tool Gateway}\\{\itshape agentic tool loop}};
\node[comp={gray!60!black}{gray!8}]      (nginx)     at (-2, 7.4)  {\textbf{Model Proxy}\\{\itshape upstream selection per model}};

\node[ext] (exttools) at (9, 9.2) {\textbf{Tool APIs}\\{\itshape legal research $\cdot$ news}};

\node[comp={gray!60!black}{gray!8}] (vllmhp)     at (-2, 2.4) {\textbf{vLLM}\\{\itshape accelerator-bound inference only}};

\begin{scope}[on background layer]
  \node[envstack, fit=(proxy)(gateway)(nginx)(mdm)(dashboard)(exttools), inner sep=6mm, xshift=3mm, yshift=-3mm] (ctrlstack2) {};
  \node[envstack, fit=(proxy)(gateway)(nginx)(mdm)(dashboard)(exttools), inner sep=6mm, xshift=1.5mm, yshift=-1.5mm] (ctrlstack1) {};
  \node[loc, fit=(proxy)(gateway)(nginx)(mdm)(dashboard)(exttools), inner sep=6mm] (ctrlplane) {};

  \node[envstack, draw=gray!45, fit=(vllmhp), inner sep=6mm, xshift=3mm, yshift=-3mm] (gpustack2) {};
  \node[envstack, draw=gray!45, fit=(vllmhp), inner sep=6mm, xshift=1.5mm, yshift=-1.5mm] (gpustack1) {};
  \node[loc, fit=(vllmhp), inner sep=6mm] (gpubox) {};
\end{scope}

\node[font=\sffamily\bfseries, anchor=south west] at (ctrlplane.north west) {Inference Cluster};
\node[font=\sffamily\small\itshape, gray!45!black, anchor=north west] at (ctrlstack2.south west) {environments};
\node[font=\sffamily\small\itshape, gray!45!black, anchor=north west] at (gpustack2.south west) {environments};

\node[font=\sffamily\bfseries, gray!55!black, anchor=west] at (ctrlstack2.east) {$\cdots$};
\node[font=\sffamily\bfseries, gray!55!black, anchor=west] at (gpustack2.east) {$\cdots$};

\draw[reqarrow] (caller) -- node[lbl]{HTTPS request} (proxy);
\draw[-{Latex}, thick, gray!60!black] (nginx.south) -- node[lbl]{mTLS} (vllmhp.north);

\coordinate (fork)      at (-2, 10.6);
\coordinate (forkright) at (4,  10.6);
\draw[toolarrow, -] (proxy.south) -- (fork) -- (forkright);
\draw[reqarrow]  (proxy.south) -- (fork) -- node[lbl]{/chat/completions} (nginx.north);
\draw[toolarrow] (forkright) -- node[lbl]{/responses} (gateway.north);

\draw[toolarrow] (gateway.south) -- ++(0,-8mm) |- node[lbl, pos=0.75]{vLLM call} (nginx.east);
\draw[toolarrow] (gateway) -- node[lbl]{tool call} (exttools);

\draw[ctrlarrow] (proxy) -- node[lbl]{ check rights} (mdm);
\draw[ctrlarrow] (dashboard) -- node[lbl]{issue / revoke credential} (mdm);

\begin{scope}[shift={(-4.3,-1.2)}]
  \draw[reqarrow] (0,0) -- (0.8,0); \node[lbl, anchor=west, fill=none] at (0.9,0) {inference request};
  \draw[toolarrow] (4.2,0) -- (5.0,0); \node[lbl, anchor=west, fill=none] at (5.1,0) {agentic tool call};
  \draw[ctrlarrow] (8.5,0) -- (9.3,0); \node[lbl, anchor=west, fill=none] at (9.4,0) {control / resolution};
\end{scope}

\end{tikzpicture}%
}
\caption{Request routing across the control plane and GPU compute clusters. Authentication,
routing, and agentic tool execution run on GPU-free routing clusters; model inference runs on
separate GPU clusters. Self-service credential issuance and model-rights checks both route
through the same deployment-management service and associated persistence layer consulted on every inference request. After the first network segment (caller to proxy, HTTPS) all inter-service communication is secured via mTLS.}
\label{fig:inference-topology}
\end{figure}

We adopt cross-cluster service mesh routing to allow transparent service discovery and communication. Further, it allows for a more proactive approach to failover; if GPUs in one region are saturated, we can opt to trade KV cache locality for what would otherwise be a slower or outright failed request. Mesh routing also enables strong adherence to data residency laws; we can explicitly enable latency-tolerant requests originating in the US to utilise GPUs in other regions if necessary, while eliminating completely the possibility of workloads sensitive to data residency requirements being sent in the other direction; even if our workloads were misconfigured, the lack of meshing back to disallowed geographies acts as a second line of defence preventing inadvertent violations. Between our approach of fault-tolerant workload isolation and mesh routing, a small team is able to maintain a multi-continent compute footprint. Figure~\ref{fig:inference-topology} shows a high-level flow of an inference request through this system.

We isolate our team-internal research workloads into a dedicated experimental environment. This allows us to move fast with newer model architectures, test a new quantisation scheme, evaluate a speculative-decoding setup, or perform load and shadow tests for a representative workload, target hardware, and KV-cache capacity. It also allows feature rollouts to a smaller demographic in those cases where a feature may be short-lived or otherwise unsuitable for a broader production audience, but also to allow our team to dog-food those features that are ultimately intended for production, allowing a greater feedback loop and quicker iteration.

We also support agentic tool-calling traffic through this platform, supporting tool calls to proprietary sources as part of our inference setup. This is exposed through an API mirroring OpenAI's Responses API, since agentic, multi-turn tool use is what that specification is built to describe; moreover, its compatibility with an established API enables new users to quickly onboard using tools with which they're already familiar. Similarly, we enable Optical Character Recognition, ensuring a greater degree of efficiency in upload handling in comparison to multi-modal approaches, while also allowing us to maintain guarantees regarding data persistence where appropriate.

\subsubsection{Inference Server Customisations}
We customise vLLM to support novel models and features prior to availability of longer-term support in the vLLM upstream. We introduced the ability to define a thinking budget client-side by introducing custom logits processors, and implemented custom reasoning parsers to support models not otherwise natively capable of reasoning traces, ensuring that our inference platform is capable of transparently (and correctly) supporting a mix of model functionalities and API features that are not yet available in the broader third-party/open-source ecosystem.

Tying back to our infrastructure sovereignty and cost-optimisation principles, a small number of optimisations have yielded the largest benefit.
\begin{itemize}
    \item \textbf{Server Initialisation} -- LLM weights increasingly have larger and larger memory footprints, and the overhead of moving almost a terabyte of data around can result in extremely long start-up times, often 20 minutes or longer. Long start-up times can introduce uptime risk -- if a replica takes 30 minutes to start, other replicas may be overloaded for the duration, ultimately risking a cascading failure and marked degradation in quality of service. By shifting to an in-memory volume (as opposed to general purpose SSD or NVMe disks) we can ultimately reduce the time taken to retrieve weights from long-term storage and allow for quicker transfers to GPU devices; while writing to disk would prevent the necessity of redownloading weights, it'd also incur slower transfer to GPU unless a tiered approach were taken. Further, shifting to use of more fit-for-purpose faster-loading libraries (e.g., instanttensor), we shift the bulk of the start-up bottleneck to graph initialisation, etc., allowing the bulk of our models to initialise in approximately five minutes, and often less.
    \item \textbf{Quantisation} -- We use vLLM's llm-compressor library for dynamic post-training quantisation, normally to FP8. This does not require a calibration dataset, since activation quantisation happens at inference time rather than being fit in advance. To avoid measurable degradation in output quality, we leave layers such as the language-modelling head (\texttt{lm\_head}) and gating and routing layers for mixture-of-experts (MoE) models unquantised. Across our evaluations, a quantised model's task performance remains close to its full-precision counterpart.
\end{itemize}

\newpage
\section{Conclusion \& Future Work}
\label{sec:conclusion}

\subsection{Conclusion}
The gap between open-weight models and the best closed systems has narrowed from
years to months \citep{aisi2026openweight}, and the most expensive stage of the
pipeline -- large-scale pre-training -- is the stage whose returns are
flattening fastest. Taken together, sovereignty no longer turns on the capital required to
train a model from scratch; it turns on whether an institution has the technical capability to implement its sovereignty goals rather than being merely capable of hosting a model without modification or engaging in narrow fine-tuning exercises. Public discourse has been clear that this matters, and comparatively quiet on how it is done. This report is an attempt to answer the second question with a worked example rather than an argument.

In this report, we argue that Continual Learning enables broad implementation of SovereignAI goals, spanning a wide range of desiderata (\ref{item:sovereignty-model}--\ref{item:sovereignty-econ}) beyond pure model performance. Adaptation of open-weight models has historically been a trade-off: capability bought in a narrow slice of the
target domain paid for with skills lost elsewhere. By treating
stability and plasticity as explicit objectives at every stage, we show how model ownership and customisation become feasible on modest budgets.

Our central empirical result is that the outcome of the methods described is better than a favourable trade. \texttt{Thomson} exhibits a distinctive
$\pi$-shaped profile: pronounced gains in the domains we aimed for, alongside
preserved -- and frequently improved -- performance in domains not targeted. More important than the concrete model or its current performance is our finding that this process can be repeatedly and reliably applied, with various modules of our model development pipeline having been applied to a number of different base models throughout the development of the \texttt{Thomson} model family.

At the core of this report is an argument about both values and economics: \texttt{Thomson} was developed on a comparatively very modest budget and has nevertheless withstood the scrutiny of careful, publicly recognised evaluations, adversarial safety testing, blind human studies, and the real requirements of production environments. Nevertheless, we are deliberate about what this does and does not establish. Sovereignty here is a spectrum rather than a threshold, and we have moved along it rather than arrived: model and data sovereignty are substantially achievable
(\ref{item:sovereignty-model}, \ref{item:sovereignty-data}), governance and
value alignment demonstrably improvable (\ref{item:sovereignty-values}),
infrastructure and economic control largely attainable on open-source
foundations (\ref{item:sovereignty-infra}, \ref{item:sovereignty-econ}) -- while
the dependence on accelerator hardware is not removed. The
open-weight starting point invites the same objection, which we address in
Section~\ref{ssec:discussion_limitations}: with each generation able to begin
from the checkpoints the previous one produced, the distance from that original
dependence grows until it is of little practical consequence.

Finally, Section~\ref{sec:infra} also highlights the critical role high-quality engineering plays in frontier model development, and how the intelligent use and assembly of an increasingly mature open-source landscape can enable infrastructure sovereignty (\ref{item:sovereignty-infra}).

\subsection{Future Work}
\label{ssec:future_work}

\paragraph{Scaling mid-training.} Our mid-training budget was held at deliberately
modest 200B tokens, pruning over 98\% of the $\approx19$T candidate pool, a choice made
to explore the quality of the model that can be achieved with modest compute budgets.
While our data-centric filtering ensures that these 200B are likely of the highest quality,
we believe that enlarging that pool remains one of the most direct levers on the results in
Section~\ref{sec:evaluation}. This is due both to knowledge injection and the increasingly
common observation that high-quality mid-training can make subsequent training stages more 
effective. An increasingly close co-ordination between mid- \& post-training is likely to maximise this effect. In addition, various data curation methods more targeted at use cases are of interest: filtering that suppresses rather than propagates hallucinations, and grouping related documents so that multi-document reasoning -- where current models remain weakest -- is trained explicitly rather than incidentally. Finally, an interesting ablation study could also explore the trade-off between training all weights for a restricted budget versus training on a larger corpus but using parameter-efficient training routines (holding total compute constant). Furthermore, we believe that data-centricity could be applied through data-prioritisation schemes \citep[e.g.][]{mindermann2022prioritized, evans2024bad, brandfonbrener2024color} during mid-training, thereby dynamically balancing plasticity and stability.

\paragraph{Reward design and RL stage.} Our findings in the RL-phase of model development confirm the observation \citep[e.g.][]{shenfeld2026rl} that online RL is a powerful component of a full Continual Learning stack due to its tendency to provide robust learning with little to no forgetting. In addition to more closely aligning post- with mid-training, it is clear that the careful reward design and an increasingly mature approach of heavily leveraging LLM judges are highly likely to continue succeeding. Going forward, we hope to fully replace any reliance on external model-based judges by leveraging previously trained \texttt{Thomson} generations, further increasing our level of sovereignty (\ref{item:sovereignty-model}). Overall, the direction is clear: RL is now mature enough to substantially improve model quality across a wide range of \textit{non-verifiable} domains. Thus, sustained model improvements will emerge based on builders' ability to scale up the sophistication of RL environment-building efforts. 

\paragraph{The role of coding skills.} Coding is the one area showing mild
forgetting relative to open-weight models we start with. While we did
not intend to improve coding specifically, general computer use and coding skills are increasingly load-bearing for agentic skills, which, in our estimation, have the highest chance to lead to real automation across the economy. Hence, we intend to measure more carefully and make greater use of the myriad open datasets designed to target coding and SWE skills, but have no doubt that the modest performance drop can be easily alleviated without the need for algorithmic breakthroughs. In addition, given the overwhelming focus on coding skills among the vast majority of model developers, we believe that much of the absolute capability gap relative to other frontier models on coding can be easily addressed through the choice of open-weight models.

\paragraph{Values and constitutions.} Alignment to a written constitution
remains an aspiration rather than a solved problem, with perfect alignment being considered one of the hardest problems in AI Safety. Nevertheless, we regard our
value re-alignment and Constitutional RL results as evidence that the
direction is tractable rather than that the destination is reached. A valuable and likely
impactful engineering contribution would also see the introduction of an adversarial red-teaming
algorithm running throughout training, ensuring that the model is continuously challenged on-policy.
Finally, all of the contentious issues presented have been deliberately written by human experts with backgrounds in politics, international relations, history, and law. While this allowed us to ensure data quality and remove any concerns about hallucinations, we recognise that this may be difficult for all organisations. As such, we believe it may be valuable to \textit{carefully} design a contentious issue design pipeline, making the entire Constitutional alignment process fully automated.

\paragraph{Compounding generations.} Finally, the argument in Section
\ref{ssec:discussion_limitations} invites its own experiment. If each model generation can be built from the last, the interesting question is what accumulates: whether successive rounds of Continual Learning compound in capability and institutional fit, or whether they accrue drift that eventually requires returning to a fresh open-weight checkpoint.

\clearpage
\phantomsection
\label{sec:ack}
\section*{Acknowledgements}

Our work on \texttt{Thomson} is made possible by the dedicated efforts of many others who offered their insight, support, and expertise to this project. 

\subsection*{Core Subject Matter Experts}

The subject matter experts who supported this project are too many to name, but we particularly want to thank our core subject-matter expert team, who have worked alongside us and taught us about their domains as we endeavoured to embed their expertise into \texttt{Thomson}. 

\textit{(unordered)}

\begin{multicols}{2}
\begin{itemize}
    \item Connie Quarnstrom
    \item Elizabeth Botsford
    \item Jessie Shearer
    \item Daniel Calloway
    \item Andrew Coyne
    \item Carinne Davis
    \item Jennifer Nist
    \item George Pagano
    \item Susan Rose
    \item Samuel Vincent
    \item Zoe Callinan
    \item Vera Mayzel
    \item Elinor Nikolova
    \item Luca Patriniche
    \item Pukar Soni
    \item Nikolas Vucekovich
    \item Jodi Gardner
    \item Poorna Mysoor
    \item Sara Catley
    \item Brian Birke
    \item Michelle Graham
    \item Kevin McNamee
    \item Brandan Oliver
    \item Mike Plambeck
    \item Ian Williams
    \item Louise Jones
    \item Marco Rinaldi
    \item Tiffany Hildreth
    \item Kathy Wood
\end{itemize}
\end{multicols}

We want to thank \textit{Jessie Baek} and \textit{Adam Greshowak} for coordinating our team, our many colleagues within Thomson Reuters for their support and insightful questions and comments, and our former Foundational Research colleagues for contributions to earlier versions of the \texttt{Thomson} project. We also thank our partners at Imperial College London, DatologyAI, and Lambda. Finally, we thank \textit{Joel Hron}, \textit{Steve Hasker}, and \textit{Alexander Kardos-Nyheim} for their leadership and support.

\subsection*{DatologyAI}

A special thank you to the researchers, engineers and leadership at DatologyAI who collaborated with us on the Mid-Training for \texttt{Thomson}.

\textit{(unordered)}

\begin{multicols}{2}
\begin{itemize}
    \item Amro Abbas
    \item Kaleigh Mentzer
    \item Diego Kiner
    \item Fan Pan
    \item Matthew Leavitt
    \item Ari Morcos
\end{itemize}
\end{multicols}
\bibliography{references}

\begin{thebibliography}{176}
\providecommand{\natexlab}[1]{#1}
\providecommand{\url}[1]{\texttt{#1}}
\expandafter\ifx\csname urlstyle\endcsname\relax
  \providecommand{\doi}[1]{doi: #1}\else
  \providecommand{\doi}{doi: \begingroup \urlstyle{rm}\Url}\fi

\bibitem[London(2026)]{imperial2026realignment}
Imperial~College London.
\newblock Cheap and effective re-alignment of frontier models through
  capability-preserving model steering.
\newblock Technical report, August 2026.
\newblock URL
  \url{https://huggingface.co/spaces/tri-fair-lab/publications/blob/main/Frontier_Model_Realignment.pdf}.

\bibitem[{AI Security Institute}(2026)]{aisi2026openweight}
{AI Security Institute}.
\newblock How far behind the frontier are leading open weight models on cyber?,
  2026.
\newblock URL \url{https://www.aisi.gov.uk/blog/how-far-behind-the-frontier}.

\bibitem[Guo et~al.(2025)Guo, Yang, Zhang, Song, Wang, Zhu, Xu, Zhang, Ma, Bi,
  et~al.]{guo2025deepseek}
Daya Guo, Dejian Yang, Haowei Zhang, Junxiao Song, Peiyi Wang, Qihao Zhu,
  Runxin Xu, Ruoyu Zhang, Shirong Ma, Xiao Bi, et~al.
\newblock Deepseek-r1 incentivizes reasoning in llms through reinforcement
  learning.
\newblock \emph{Nature}, 645\penalty0 (8081):\penalty0 633--638, 2025.

\bibitem[O'Brien et~al.(2026)O'Brien, Casper, Anthony, Korbak, Kirk, Davies,
  Mishra, Irving, Gal, and Biderman]{o2026deep}
Kyle O'Brien, Stephen Casper, Quentin Anthony, Tomek Korbak, Robert Kirk,
  Xander Davies, Ishan Mishra, Geoffrey Irving, Yarin Gal, and Stella~R
  Biderman.
\newblock Deep ignorance: Filtering pretraining data builds tamper-resistant
  safeguards into open-weight llms.
\newblock In \emph{International Conference on Learning Representations},
  volume 2026, pages 35579--35633, 2026.

\bibitem[Bai et~al.(2022)Bai, Kadavath, Kundu, Askell, Kernion, Jones, Chen,
  Goldie, Mirhoseini, McKinnon, et~al.]{bai2022constitutional}
Yuntao Bai, Saurav Kadavath, Sandipan Kundu, Amanda Askell, Jackson Kernion,
  Andy Jones, Anna Chen, Anna Goldie, Azalia Mirhoseini, Cameron McKinnon,
  et~al.
\newblock Constitutional ai: Harmlessness from ai feedback.
\newblock \emph{arXiv preprint arXiv:2212.08073}, 2022.

\bibitem[Patriniche et~al.(2026)Patriniche, Bell, Trautmann, Vucekovich,
  Callinan, Soni, Williams, Coyne, Nanreh, Fielding, Seifeddine, Simon, Bang,
  and Schwarz]{patriniche2026publicai}
Luca Patriniche, Bradley Bell, Dietrich Trautmann, Nikolas Vucekovich, Zoe
  Callinan, Pukar Soni, Ian Williams, Andrew Coyne, Manpreet Nanreh, Kirsty
  Fielding, Wassim Seifeddine, Felix~M. Simon, Yejin Bang, and Jonathan~Richard
  Schwarz.
\newblock The public {AI} constitution project.
\newblock Technical report, August 2026.
\newblock URL
  \url{https://huggingface.co/spaces/tri-fair-lab/publications/blob/main/Public_AI_Constitution.pdf}.
\newblock Luca Patriniche, Bradley Bell and Dietrich Trautmann contributed
  equally as joint first authors; Yejin Bang and Jonathan Richard Schwarz are
  joint senior authors.

\bibitem[McCloskey and Cohen(1989)]{mccloskey1989catastrophic}
Michael McCloskey and Neal~J Cohen.
\newblock Catastrophic interference in connectionist networks: The sequential
  learning problem.
\newblock In \emph{Psychology of learning and motivation}, volume~24, pages
  109--165. Elsevier, 1989.

\bibitem[Ratcliff(1990)]{ratcliff1990connectionist}
Roger Ratcliff.
\newblock Connectionist models of recognition memory: constraints imposed by
  learning and forgetting functions.
\newblock \emph{Psychological review}, 97\penalty0 (2):\penalty0 285, 1990.

\bibitem[Parisi et~al.(2019)Parisi, Kemker, Part, Kanan, and
  Wermter]{parisi2019continual}
German~I Parisi, Ronald Kemker, Jose~L Part, Christopher Kanan, and Stefan
  Wermter.
\newblock Continual lifelong learning with neural networks: A review.
\newblock \emph{Neural networks}, 113:\penalty0 54--71, 2019.

\bibitem[Wang et~al.(2024{\natexlab{a}})Wang, Yang, Shen, and
  Huang]{wang2024comprehensive}
Zhenyi Wang, Enneng Yang, Li~Shen, and Heng Huang.
\newblock A comprehensive survey of forgetting in deep learning beyond
  continual learning.
\newblock \emph{IEEE Transactions on Pattern Analysis and Machine
  Intelligence}, 47\penalty0 (3):\penalty0 1464--1483, 2024{\natexlab{a}}.

\bibitem[Thede et~al.(2026)Thede, Winzeck, Akata, and
  Schwarz]{thede2026captrack}
Lukas Thede, Stefan Winzeck, Zeynep Akata, and Jonathan~Richard Schwarz.
\newblock Captrack: Multifaceted evaluation of forgetting in llm post-training.
\newblock \emph{arXiv preprint arXiv:2603.06610}, 2026.

\bibitem[Zhou et~al.(2025)Zhou, Pacchiardi, Mart{\'\i}nez-Plumed, Collins,
  Moros-Daval, Zhang, Zhao, Huang, Sun, Prunty, et~al.]{zhou2025general}
Lexin Zhou, Lorenzo Pacchiardi, Fernando Mart{\'\i}nez-Plumed, Katherine~M
  Collins, Yael Moros-Daval, Seraphina Zhang, Qinlin Zhao, Yitian Huang, Luning
  Sun, Jonathan~E Prunty, et~al.
\newblock General scales unlock ai evaluation with explanatory and predictive
  power.
\newblock \emph{arXiv preprint arXiv:2503.06378}, 2025.

\bibitem[Bean et~al.(2025)Bean, Seedat, Chen, and Schwarz]{bean2025scales++}
Andrew~M Bean, Nabeel Seedat, Shengzhuang Chen, and Jonathan~Richard Schwarz.
\newblock Scales++: Compute efficient evaluation subset selection with
  cognitive scales embeddings.
\newblock \emph{arXiv preprint arXiv:2510.26384}, 2025.

\bibitem[Lambert et~al.(2024)Lambert, Morrison, Pyatkin, Huang, Ivison,
  Brahman, Miranda, Liu, Dziri, Lyu, et~al.]{lambert2024tulu}
Nathan Lambert, Jacob Morrison, Valentina Pyatkin, Shengyi Huang, Hamish
  Ivison, Faeze Brahman, Lester James~V Miranda, Alisa Liu, Nouha Dziri, Shane
  Lyu, et~al.
\newblock Tulu 3: Pushing frontiers in open language model post-training.
\newblock \emph{arXiv preprint arXiv:2411.15124}, 2024.

\bibitem[Du et~al.(2026)Du, Xu, Zhu, Zhang, Wang, and Mao]{du2026deepresearch}
Mingxuan Du, Benfeng Xu, Chiwei Zhu, Licheng Zhang, Xiaorui Wang, and Zhendong
  Mao.
\newblock Deepresearch bench: A comprehensive benchmark for deep research
  agents.
\newblock In \emph{International Conference on Learning Representations},
  volume 2026, pages 42414--42448, 2026.

\bibitem[Huang et~al.(2025)Huang, Chen, Zhang, Li, Zhou, Fang, Yang, Li, Shang,
  Xu, et~al.]{huang2025deep}
Yuxuan Huang, Yihang Chen, Haozheng Zhang, Kang Li, Huichi Zhou, Meng Fang,
  Linyi Yang, Xiaoguang Li, Lifeng Shang, Songcen Xu, et~al.
\newblock Deep research agents: A systematic examination and roadmap.
\newblock \emph{arXiv preprint arXiv:2506.18096}, 2025.

\bibitem[Schick et~al.(2023)Schick, Dwivedi-Yu, Dess{\`\i}, Raileanu, Lomeli,
  Hambro, Zettlemoyer, Cancedda, and Scialom]{schick2023toolformer}
Timo Schick, Jane Dwivedi-Yu, Roberto Dess{\`\i}, Roberta Raileanu, Maria
  Lomeli, Eric Hambro, Luke Zettlemoyer, Nicola Cancedda, and Thomas Scialom.
\newblock Toolformer: Language models can teach themselves to use tools.
\newblock \emph{Advances in neural information processing systems},
  36:\penalty0 68539--68551, 2023.

\bibitem[Yao et~al.(2024{\natexlab{a}})Yao, Shinn, Razavi, and
  Narasimhan]{yao2024tau}
Shunyu Yao, Noah Shinn, Pedram Razavi, and Karthik Narasimhan.
\newblock $\tau$-bench: A benchmark for tool-agent-user interaction in
  real-world domains.
\newblock \emph{arXiv preprint arXiv:2406.12045}, 2024{\natexlab{a}}.

\bibitem[Bai et~al.(2024)Bai, Lv, Zhang, Lyu, Tang, Huang, Du, Liu, Zeng, Hou,
  et~al.]{bai2024longbench}
Yushi Bai, Xin Lv, Jiajie Zhang, Hongchang Lyu, Jiankai Tang, Zhidian Huang,
  Zhengxiao Du, Xiao Liu, Aohan Zeng, Lei Hou, et~al.
\newblock Longbench: A bilingual, multitask benchmark for long context
  understanding.
\newblock In \emph{Proceedings of the 62nd annual meeting of the association
  for computational linguistics (volume 1: Long papers)}, pages 3119--3137,
  2024.

\bibitem[Hsieh et~al.(2024)Hsieh, Sun, Kriman, Acharya, Rekesh, Jia, Zhang, and
  Ginsburg]{hsieh2024ruler}
Cheng-Ping Hsieh, Simeng Sun, Samuel Kriman, Shantanu Acharya, Dima Rekesh, Fei
  Jia, Yang Zhang, and Boris Ginsburg.
\newblock Ruler: What's the real context size of your long-context language
  models?
\newblock \emph{arXiv preprint arXiv:2404.06654}, 2024.

\bibitem[Wei et~al.(2022)Wei, Wang, Schuurmans, Bosma, Xia, Chi, Le, Zhou,
  et~al.]{wei2022chain}
Jason Wei, Xuezhi Wang, Dale Schuurmans, Maarten Bosma, Fei Xia, Ed~Chi, Quoc~V
  Le, Denny Zhou, et~al.
\newblock Chain-of-thought prompting elicits reasoning in large language
  models.
\newblock \emph{Advances in neural information processing systems},
  35:\penalty0 24824--24837, 2022.

\bibitem[Xu and Peng(2025)]{xu2025comprehensive}
Renjun Xu and Jingwen Peng.
\newblock A comprehensive survey of deep research: Systems, methodologies, and
  applications.
\newblock \emph{arXiv preprint arXiv:2506.12594}, 2025.

\bibitem[Shao et~al.(2025)Shao, Asai, Shen, Ivison, Kishore, Zhuo, Zhao, Park,
  Finlayson, Sontag, et~al.]{shao2025dr}
Rulin Shao, Akari Asai, Shannon~Zejiang Shen, Hamish Ivison, Varsha Kishore,
  Jingming Zhuo, Xinran Zhao, Molly Park, Samuel~G Finlayson, David Sontag,
  et~al.
\newblock Dr tulu: Reinforcement learning with evolving rubrics for deep
  research.
\newblock \emph{arXiv preprint arXiv:2511.19399}, 2025.

\bibitem[Mialon et~al.(2024)Mialon, Fourrier, Wolf, LeCun, and
  Scialom]{mialon2024gaia}
Gr{\'e}goire Mialon, Cl{\'e}mentine Fourrier, Thomas Wolf, Yann LeCun, and
  Thomas Scialom.
\newblock Gaia: a benchmark for general ai assistants.
\newblock In \emph{International Conference on Learning Representations},
  volume 2024, pages 9025--9049, 2024.

\bibitem[Wei et~al.(2025)Wei, Sun, Papay, McKinney, Han, Fulford, Chung,
  Passos, Fedus, and Glaese]{wei2025browsecomp}
Jason Wei, Zhiqing Sun, Spencer Papay, Scott McKinney, Jeffrey Han, Isa
  Fulford, Hyung~Won Chung, Alex~Tachard Passos, William Fedus, and Amelia
  Glaese.
\newblock Browsecomp: A simple yet challenging benchmark for browsing agents.
\newblock \emph{arXiv preprint arXiv:2504.12516}, 2025.

\bibitem[{ByteDance}(2025)]{deerflow2025}
{ByteDance}.
\newblock {DeerFlow}: Deep exploration and efficient research flow.
\newblock \url{https://github.com/bytedance/deer-flow}, 2025.

\bibitem[{LangChain}(2024)]{langgraph2024}
{LangChain}.
\newblock {LangGraph}.
\newblock \url{https://langchain-ai.github.io/langgraph/}, 2024.

\bibitem[{Anthropic}(2025)]{anthropic2025multiagent}
{Anthropic}.
\newblock How we built our multi-agent research system.
\newblock Anthropic Engineering Blog.
  \url{https://www.anthropic.com/engineering/multi-agent-research-system},
  2025.

\bibitem[Yao et~al.(2022)Yao, Zhao, Yu, Du, Shafran, Narasimhan, and
  Cao]{yao2022react}
Shunyu Yao, Jeffrey Zhao, Dian Yu, Nan Du, Izhak Shafran, Karthik Narasimhan,
  and Yuan Cao.
\newblock React: Synergizing reasoning and acting in language models.
\newblock \emph{arXiv preprint arXiv:2210.03629}, 2022.

\bibitem[Rafailov et~al.(2023)Rafailov, Sharma, Mitchell, Manning, Ermon, and
  Finn]{rafailov2023direct}
Rafael Rafailov, Archit Sharma, Eric Mitchell, Christopher~D Manning, Stefano
  Ermon, and Chelsea Finn.
\newblock Direct preference optimization: Your language model is secretly a
  reward model.
\newblock \emph{Advances in neural information processing systems},
  36:\penalty0 53728--53741, 2023.

\bibitem[Shao et~al.(2024)Shao, Wang, Zhu, Xu, Song, Bi, Zhang, Zhang, Li, Wu,
  et~al.]{shao2024deepseekmath}
Zhihong Shao, Peiyi Wang, Qihao Zhu, Runxin Xu, Junxiao Song, Xiao Bi, Haowei
  Zhang, Mingchuan Zhang, YK~Li, Yang Wu, et~al.
\newblock Deepseekmath: Pushing the limits of mathematical reasoning in open
  language models.
\newblock \emph{arXiv preprint arXiv:2402.03300}, 2024.

\bibitem[Arditi et~al.(2024)Arditi, Obeso, Syed, Paleka, Panickssery, Gurnee,
  and Nanda]{arditi2024refusal}
Andy Arditi, Oscar Obeso, Aaquib Syed, Daniel Paleka, Nina Panickssery, Wes
  Gurnee, and Neel Nanda.
\newblock Refusal in language models is mediated by a single direction.
\newblock In \emph{Advances in Neural Information Processing Systems
  (NeurIPS)}, 2024.
\newblock URL \url{https://arxiv.org/abs/2406.11717}.

\bibitem[Belrose(2023)]{belrose2023diffinmeans}
Nora Belrose.
\newblock Diff-in-means concept editing is worst-case optimal.
\newblock EleutherAI Blog, December 2023.
\newblock URL \url{https://blog.eleuther.ai/diff-in-means/}.
\newblock Accessed 2026-08-17.

\bibitem[McInnes et~al.(2018)McInnes, Healy, Saul, and
  Gro{\ss}berger]{mcinnes2018umap}
Leland McInnes, John Healy, Nathaniel Saul, and Lukas Gro{\ss}berger.
\newblock {UMAP}: Uniform manifold approximation and projection.
\newblock \emph{Journal of Open Source Software}, 3\penalty0 (29):\penalty0
  861, 2018.
\newblock \doi{10.21105/joss.00861}.

\bibitem[Hu et~al.(2022)Hu, Shen, Wallis, Allen-Zhu, Li, Wang, Wang, and
  Chen]{hu2021lora}
Edward~J. Hu, Yelong Shen, Phillip Wallis, Zeyuan Allen-Zhu, Yuanzhi Li, Shean
  Wang, Lu~Wang, and Weizhu Chen.
\newblock {LoRA}: Low-rank adaptation of large language models.
\newblock In \emph{International Conference on Learning Representations}, 2022.
\newblock URL \url{https://openreview.net/forum?id=nZeVKeeFYf9}.

\bibitem[Kirkpatrick et~al.(2017)Kirkpatrick, Pascanu, Rabinowitz, Veness,
  Desjardins, Rusu, Milan, Quan, Ramalho, Grabska-Barwi{\'n}ska, Hassabis,
  Clopath, Kumaran, and Hadsell]{kirkpatrick2017overcoming}
James Kirkpatrick, Razvan Pascanu, Neil Rabinowitz, Joel Veness, Guillaume
  Desjardins, Andrei~A. Rusu, Kieran Milan, John Quan, Tiago Ramalho, Agnieszka
  Grabska-Barwi{\'n}ska, Demis Hassabis, Claudia Clopath, Dharshan Kumaran, and
  Raia Hadsell.
\newblock Overcoming catastrophic forgetting in neural networks.
\newblock \emph{Proceedings of the National Academy of Sciences}, 114\penalty0
  (13):\penalty0 3521--3526, 2017.
\newblock \doi{10.1073/pnas.1611835114}.

\bibitem[Amari(1998)]{amari1998natural}
Shun-ichi Amari.
\newblock Natural gradient works efficiently in learning.
\newblock \emph{Neural Computation}, 10\penalty0 (2):\penalty0 251--276, 1998.

\bibitem[Bergstra et~al.(2011)Bergstra, Bardenet, Bengio, and
  K{\'e}gl]{bergstra2011algorithms}
James Bergstra, R{\'e}mi Bardenet, Yoshua Bengio, and Bal{\'a}zs K{\'e}gl.
\newblock Algorithms for hyper-parameter optimization.
\newblock \emph{Advances in neural information processing systems}, 24, 2011.

\bibitem[Akiba et~al.(2019)Akiba, Sano, Yanase, Ohta, and
  Koyama]{akiba2019optuna}
Takuya Akiba, Shotaro Sano, Toshihiko Yanase, Takeru Ohta, and Masanori Koyama.
\newblock Optuna: A next-generation hyperparameter optimization framework.
\newblock In \emph{Proceedings of the 25th ACM SIGKDD international conference
  on knowledge discovery \& data mining}, pages 2623--2631, 2019.

\bibitem[Jindal et~al.(2024)Jindal, Badrinath, Bharti, Vinay, and
  Sharma]{jindal2024balancing}
Ishan Jindal, Chandana Badrinath, Pranjal Bharti, Lakkidi Vinay, and Sachin~Dev
  Sharma.
\newblock Balancing continuous pre-training and instruction fine-tuning:
  Optimizing instruction-following in llms.
\newblock \emph{arXiv preprint arXiv:2410.10739}, 2024.

\bibitem[Padmanabhan et~al.(2026)Padmanabhan, Gul, and
  Goyal]{padmanabhan2026updating}
Shankar Padmanabhan, Mustafa~Omer Gul, and Tanya Goyal.
\newblock Updating parametric knowledge with context distillation retains
  post-training capabilities.
\newblock \emph{arXiv preprint arXiv:2602.16093}, 2026.

\bibitem[Robins(1995)]{robins1995catastrophic}
Anthony Robins.
\newblock Catastrophic forgetting, rehearsal and pseudorehearsal.
\newblock \emph{Connection Science}, 7\penalty0 (2):\penalty0 123--146, 1995.

\bibitem[Shin et~al.(2017)Shin, Lee, Kim, and Kim]{shin2017continual}
Hanul Shin, Jung~Kwon Lee, Jaehong Kim, and Jiwon Kim.
\newblock Continual learning with deep generative replay.
\newblock \emph{Advances in neural information processing systems}, 30, 2017.

\bibitem[Rolnick et~al.(2019)Rolnick, Ahuja, Schwarz, Lillicrap, and
  Wayne]{rolnick2019experience}
David Rolnick, Arun Ahuja, Jonathan Schwarz, Timothy Lillicrap, and Gregory
  Wayne.
\newblock Experience replay for continual learning.
\newblock \emph{Advances in neural information processing systems}, 32, 2019.

\bibitem[Titsias et~al.(2019)Titsias, Schwarz, Matthews, Pascanu, and
  Teh]{titsias2019functional}
Michalis~K Titsias, Jonathan Schwarz, Alexander G de~G Matthews, Razvan
  Pascanu, and Yee~Whye Teh.
\newblock Functional regularisation for continual learning with gaussian
  processes.
\newblock \emph{arXiv preprint arXiv:1901.11356}, 2019.

\bibitem[Li et~al.(2024)Li, Fang, Smyrnis, Ivgi, Jordan, Gadre, Bansal, Guha,
  Keh, Arora, et~al.]{li2024datacomp}
Jeffrey Li, Alex Fang, Georgios Smyrnis, Maor Ivgi, Matt Jordan, Samir Gadre,
  Hritik Bansal, Etash Guha, Sedrick Keh, Kushal Arora, et~al.
\newblock Datacomp-lm: In search of the next generation of training sets for
  language models.
\newblock \emph{Advances in Neural Information Processing Systems},
  37:\penalty0 14200--14282, 2024.

\bibitem[Maini et~al.(2025)Maini, Dorna, Doshi, Carranza, Pan, Urbanek,
  Burstein, Fang, Deng, Abbas, et~al.]{maini2025beyondweb}
Pratyush Maini, Vineeth Dorna, Parth Doshi, Aldo Carranza, Fan Pan, Jack
  Urbanek, Paul Burstein, Alex Fang, Alvin Deng, Amro Abbas, et~al.
\newblock Beyondweb: Lessons from scaling synthetic data for trillion-scale
  pretraining.
\newblock \emph{arXiv preprint arXiv:2508.10975}, 2025.

\bibitem[Brown et~al.(2020)Brown, Mann, Ryder, Subbiah, Kaplan, Dhariwal,
  Neelakantan, Shyam, Sastry, Askell, et~al.]{brown2020language}
Tom Brown, Benjamin Mann, Nick Ryder, Melanie Subbiah, Jared~D Kaplan, Prafulla
  Dhariwal, Arvind Neelakantan, Pranav Shyam, Girish Sastry, Amanda Askell,
  et~al.
\newblock Language models are few-shot learners.
\newblock \emph{Advances in neural information processing systems},
  33:\penalty0 1877--1901, 2020.

\bibitem[Yang et~al.(2025)Yang, Li, Yang, Zhang, Hui, Zheng, Yu, Gao, Huang,
  Lv, et~al.]{yang2025qwen3}
An~Yang, Anfeng Li, Baosong Yang, Beichen Zhang, Binyuan Hui, Bo~Zheng, Bowen
  Yu, Chang Gao, Chengen Huang, Chenxu Lv, et~al.
\newblock Qwen3 technical report.
\newblock \emph{arXiv preprint arXiv:2505.09388}, 2025.

\bibitem[Grattafiori et~al.(2024)Grattafiori, Dubey, Jauhri, Pandey, Kadian,
  Al-Dahle, Letman, Mathur, Schelten, Vaughan, et~al.]{grattafiori2024llama}
Aaron Grattafiori, Abhimanyu Dubey, Abhinav Jauhri, Abhinav Pandey, Abhishek
  Kadian, Ahmad Al-Dahle, Aiesha Letman, Akhil Mathur, Alan Schelten, Alex
  Vaughan, et~al.
\newblock The llama 3 herd of models.
\newblock \emph{arXiv preprint arXiv:2407.21783}, 2024.

\bibitem[Gu et~al.()Gu, Tafjord, Kuehl, Haddad, Dodge, and
  Hajishirzi]{gu2406olmes}
Yuling Gu, Oyvind Tafjord, Bailey Kuehl, Dany Haddad, Jesse Dodge, and Hannaneh
  Hajishirzi.
\newblock Olmes: A standard for language model evaluations, 2024.
\newblock \emph{URL https://arxiv. org/abs/2406.08446}.

\bibitem[Zhong et~al.(2024)Zhong, Cui, Guo, Liang, Lu, Wang, Saied, Chen, and
  Duan]{zhong2024agieval}
Wanjun Zhong, Ruixiang Cui, Yiduo Guo, Yaobo Liang, Shuai Lu, Yanlin Wang, Amin
  Saied, Weizhu Chen, and Nan Duan.
\newblock Agieval: A human-centric benchmark for evaluating foundation models.
\newblock In \emph{Findings of the association for computational linguistics:
  NAACL 2024}, pages 2299--2314, 2024.

\bibitem[Hendrycks et~al.(2020)Hendrycks, Burns, Basart, Zou, Mazeika, Song,
  and Steinhardt]{hendrycks2020measuring}
Dan Hendrycks, Collin Burns, Steven Basart, Andy Zou, Mantas Mazeika, Dawn
  Song, and Jacob Steinhardt.
\newblock Measuring massive multitask language understanding.
\newblock \emph{arXiv preprint arXiv:2009.03300}, 2020.

\bibitem[Guha et~al.(2023)Guha, Nyarko, Ho, R{\'e}, Chilton, Narayana,
  Chohlas-Wood, Peters, Waldon, Rockmore, et~al.]{guha2023legalbench}
Neel Guha, Julian Nyarko, Daniel~E. Ho, Christopher R{\'e}, Adam Chilton,
  Aditya Narayana, Alex Chohlas-Wood, Austin Peters, Brandon Waldon, Daniel~N.
  Rockmore, et~al.
\newblock {LegalBench}: A collaboratively built benchmark for measuring legal
  reasoning in large language models.
\newblock In \emph{Advances in Neural Information Processing Systems 36
  (NeurIPS 2023), Datasets and Benchmarks Track}, 2023.
\newblock URL \url{https://arxiv.org/abs/2308.11462}.
\newblock 40 authors total.

\bibitem[Heineman et~al.(2026)Heineman, Hofmann, Magnusson, Gu, Smith,
  Hajishirzi, Lo, and Dodge]{heineman2026signal}
David Heineman, Valentin Hofmann, Ian Magnusson, Yuling Gu, Noah Smith, Hanna
  Hajishirzi, Kyle Lo, and Jesse Dodge.
\newblock Signal and noise: A framework for reducing uncertainty in language
  model evaluation.
\newblock \emph{Advances in Neural Information Processing Systems},
  38:\penalty0 17073--17114, 2026.

\bibitem[OLMo et~al.(2024)OLMo, Walsh, Soldaini, Groeneveld, Lo, Arora, Bhagia,
  Gu, Huang, Jordan, et~al.]{olmo20242}
Team OLMo, Pete Walsh, Luca Soldaini, Dirk Groeneveld, Kyle Lo, Shane Arora,
  Akshita Bhagia, Yuling Gu, Shengyi Huang, Matt Jordan, et~al.
\newblock 2 olmo 2 furious.
\newblock \emph{arXiv preprint arXiv:2501.00656}, 2024.

\bibitem[Ibrahim et~al.(2024)Ibrahim, Th{\'e}rien, Gupta, Richter, Anthony,
  Lesort, Belilovsky, and Rish]{ibrahim2024simple}
Adam Ibrahim, Benjamin Th{\'e}rien, Kshitij Gupta, Mats~L Richter, Quentin
  Anthony, Timoth{\'e}e Lesort, Eugene Belilovsky, and Irina Rish.
\newblock Simple and scalable strategies to continually pre-train large
  language models.
\newblock \emph{arXiv preprint arXiv:2403.08763}, 2024.

\bibitem[Gupta et~al.(2023)Gupta, Th{\'e}rien, Ibrahim, Richter, Anthony,
  Belilovsky, Rish, and Lesort]{gupta2023continual}
Kshitij Gupta, Benjamin Th{\'e}rien, Adam Ibrahim, Mats~L Richter, Quentin
  Anthony, Eugene Belilovsky, Irina Rish, and Timoth{\'e}e Lesort.
\newblock Continual pre-training of large language models: How to (re) warm
  your model?
\newblock \emph{arXiv preprint arXiv:2308.04014}, 2023.

\bibitem[Parmar et~al.(2024)Parmar, Satheesh, Patwary, Shoeybi, and
  Catanzaro]{parmar2024reuse}
Jupinder Parmar, Sanjev Satheesh, Mostofa Patwary, Mohammad Shoeybi, and Bryan
  Catanzaro.
\newblock Reuse, don't retrain: A recipe for continued pretraining of language
  models.
\newblock \emph{arXiv preprint arXiv:2407.07263}, 2024.

\bibitem[Dahl et~al.(2024)Dahl, Magesh, Suzgun, and Ho]{dahl2024large}
Matthew Dahl, Varun Magesh, Mirac Suzgun, and Daniel~E Ho.
\newblock Large legal fictions: Profiling legal hallucinations in large
  language models.
\newblock \emph{Journal of Legal Analysis}, 16\penalty0 (1):\penalty0 64--93,
  2024.

\bibitem[Gudibande et~al.(2023)Gudibande, Wallace, Snell, Geng, Liu, Abbeel,
  Levine, and Song]{gudibande2023false}
Arnav Gudibande, Eric Wallace, Charlie Snell, Xinyang Geng, Hao Liu, Pieter
  Abbeel, Sergey Levine, and Dawn Song.
\newblock The false promise of imitating proprietary llms.
\newblock \emph{arXiv preprint arXiv:2305.15717}, 2023.

\bibitem[Zhang et~al.(2026)Zhang, Hu, Le, Torsha, Jiang, Bui, Chang, Chuang,
  Xiong, Lin, et~al.]{zhang2026survey}
Kaituo Zhang, Mingzhi Hu, Hoang Anh~Duy Le, Fariha~Kabir Torsha, Zhimeng Jiang,
  Minh~Khai Bui, Chia-Yuan Chang, Yu-Neng Chuang, Zhen Xiong, Ying Lin, et~al.
\newblock A survey on evaluating quality and trustworthiness in llm-generated
  data.
\newblock \emph{arXiv preprint arXiv:2601.17717}, 2026.

\bibitem[Alexandrov et~al.(2024)Alexandrov, Raychev, M{\"u}ller, Zhang, Vechev,
  and Toutanova]{alexandrov2024mitigating}
Anton Alexandrov, Veselin Raychev, Mark~Niklas M{\"u}ller, Ce~Zhang, Martin
  Vechev, and Kristina Toutanova.
\newblock Mitigating catastrophic forgetting in language transfer via model
  merging.
\newblock In \emph{Findings of the Association for Computational Linguistics:
  EMNLP 2024}, pages 17167--17186, 2024.

\bibitem[Wortsman et~al.(2022)Wortsman, Ilharco, Gadre, Roelofs, Gontijo-Lopes,
  Morcos, Namkoong, Farhadi, Carmon, Kornblith, et~al.]{wortsman2022model}
Mitchell Wortsman, Gabriel Ilharco, Samir~Ya Gadre, Rebecca Roelofs, Raphael
  Gontijo-Lopes, Ari~S Morcos, Hongseok Namkoong, Ali Farhadi, Yair Carmon,
  Simon Kornblith, et~al.
\newblock Model soups: averaging weights of multiple fine-tuned models improves
  accuracy without increasing inference time.
\newblock In \emph{International conference on machine learning}, pages
  23965--23998. Pmlr, 2022.

\bibitem[Labrak et~al.(2024)Labrak, Bazoge, Morin, Gourraud, Rouvier, and
  Dufour]{labrak2024biomistral}
Yanis Labrak, Adrien Bazoge, Emmanuel Morin, Pierre-Antoine Gourraud, Mickael
  Rouvier, and Richard Dufour.
\newblock Biomistral: A collection of open-source pretrained large language
  models for medical domains.
\newblock In \emph{Findings of the association for computational linguistics:
  acl 2024}, pages 5848--5864, 2024.

\bibitem[Siriwardhana et~al.(2024)Siriwardhana, McQuade, Gauthier, Atkins,
  Neto, Meyers, Vij, Odenthal, Goddard, MacCarthy,
  et~al.]{siriwardhana2024domain}
Shamane Siriwardhana, Mark McQuade, Thomas Gauthier, Lucas Atkins,
  Fernando~Fernandes Neto, Luke Meyers, Anneketh Vij, Tyler Odenthal, Charles
  Goddard, Mary MacCarthy, et~al.
\newblock Domain adaptation of llama3-70b-instruct through continual
  pre-training and model merging: A comprehensive evaluation.
\newblock \emph{arXiv preprint arXiv:2406.14971}, 2024.

\bibitem[Ilharco et~al.(2022)Ilharco, Ribeiro, Wortsman, Gururangan, Schmidt,
  Hajishirzi, and Farhadi]{ilharco2022editing}
Gabriel Ilharco, Marco~Tulio Ribeiro, Mitchell Wortsman, Suchin Gururangan,
  Ludwig Schmidt, Hannaneh Hajishirzi, and Ali Farhadi.
\newblock Editing models with task arithmetic.
\newblock \emph{arXiv preprint arXiv:2212.04089}, 2022.

\bibitem[Cheng et~al.(2024)Cheng, Gu, Huang, Bi, Huang, and
  Wei]{cheng2024instruction}
Daixuan Cheng, Yuxian Gu, Shaohan Huang, Junyu Bi, Minlie Huang, and Furu Wei.
\newblock Instruction pre-training: Language models are supervised multitask
  learners.
\newblock In \emph{Proceedings of the 2024 Conference on Empirical Methods in
  Natural Language Processing}, pages 2529--2550, 2024.

\bibitem[Baek et~al.(2026)Baek, Monti, Schwab, Abbas, Adiga, Blakeney,
  B{\"o}ther, Burstein, Carranza, Deng, Doshi, Dorna, Fang, Jiang, Joshi,
  Larsen, Lee, Mentzer, Merrick, Mongstad, Pan, Suri, Teh, Telanoff, Urbanek,
  Wang, Wills, Yin, Raghunathan, Kolter, Gaza, Morcos, Leavitt, and
  Maini]{baek2026finetuner}
Christina Baek, Ricardo~Pio Monti, David Schwab, Amro Abbas, Rishabh Adiga,
  Cody Blakeney, Maximilian B{\"o}ther, Paul Burstein, Aldo~Gael Carranza,
  Alvin Deng, Parth Doshi, Vineeth Dorna, Alex Fang, Tony Jiang, Siddharth
  Joshi, Brett~W. Larsen, Jason~Chan Lee, Katherine~L. Mentzer, Luke Merrick,
  Haakon Mongstad, Fan Pan, Anshuman Suri, Darren Teh, Jason Telanoff, Jack
  Urbanek, Zhengping Wang, Josh Wills, Haoli Yin, Aditi Raghunathan, J.~Zico
  Kolter, Bogdan Gaza, Ari Morcos, Matthew Leavitt, and Pratyush Maini.
\newblock The finetuner's fallacy: When to pretrain with your finetuning data.
\newblock \emph{arXiv preprint arXiv:2603.16177}, 2026.

\bibitem[Geng et~al.(2025)Geng, Ivison, Li, Sap, Li, Krishna, and
  Koh]{geng2025delta}
Scott Geng, Hamish Ivison, Chun-Liang Li, Maarten Sap, Jerry Li, Ranjay
  Krishna, and Pang~Wei Koh.
\newblock The delta learning hypothesis: Preference tuning on weak data can
  yield strong gains.
\newblock \emph{arXiv preprint arXiv:2507.06187}, 2025.

\bibitem[Meng et~al.(2024)Meng, Xia, and Chen]{meng2024simpo}
Yu~Meng, Mengzhou Xia, and Danqi Chen.
\newblock Simpo: Simple preference optimization with a reference-free reward.
\newblock \emph{Advances in Neural Information Processing Systems},
  37:\penalty0 124198--124235, 2024.

\bibitem[Du et~al.(2025)Du, Toshniwal, Kisacanin, Mahdavi, Moshkov, Armstrong,
  Ge, Minasyan, Chen, and Gitman]{du2025nemotron}
Wei Du, Shubham Toshniwal, Branislav Kisacanin, Sadegh Mahdavi, Ivan Moshkov,
  George Armstrong, Stephen Ge, Edgar Minasyan, Feng Chen, and Igor Gitman.
\newblock Nemotron-math: Efficient long-context distillation of mathematical
  reasoning from multi-mode supervision.
\newblock \emph{arXiv preprint arXiv:2512.15489}, 2025.

\bibitem[Zhang et~al.(2025)Zhang, Wang, Jiang, Li, Wu, Wang, Jiang, Shang,
  Tang, Lyu, and Ma]{zhang2025ccr}
Qiyuan Zhang, Yufei Wang, Yuxin Jiang, Liangyou Li, Chuhan Wu, Yasheng Wang,
  Xin Jiang, Lifeng Shang, Ruiming Tang, Fuyuan Lyu, and Chen Ma.
\newblock Crowd comparative reasoning: Unlocking comprehensive evaluations for
  llm-as-a-judge, 2025.
\newblock URL \url{https://arxiv.org/abs/2502.12501}.

\bibitem[Song et~al.(2026)Song, Bonifazi, Schilder, and
  Schwarz]{song2026knowledge}
Dezhao Song, Guglielmo Bonifazi, Frank Schilder, and Jonathan~Richard Schwarz.
\newblock Knowledge graph-assisted llm post-training for enhanced legal
  reasoning.
\newblock \emph{arXiv preprint arXiv:2601.13806}, 2026.

\bibitem[Djuhera et~al.(2026)Djuhera, Ahmed, Kadhe, Zawad, Ludwig, and
  Boche]{djuhera2026data}
Aladin Djuhera, Farhan Ahmed, Swanand Kadhe, Syed Zawad, Heiko Ludwig, and
  Holger Boche.
\newblock When data is the algorithm: A systematic study and curation of
  preference optimization datasets.
\newblock In \emph{International Conference on Learning Representations},
  volume 2026, pages 150085--150130, 2026.

\bibitem[Thakkar et~al.(2024)Thakkar, Fournier, Riemer, Chen, Zouaq, Das, and
  Chandar]{thakkar2024deep}
Megh Thakkar, Quentin Fournier, Matthew Riemer, Pin-Yu Chen, Amal Zouaq, Payel
  Das, and Sarath Chandar.
\newblock A deep dive into the trade-offs of parameter-efficient preference
  alignment techniques.
\newblock In \emph{Proceedings of the 62nd Annual Meeting of the Association
  for Computational Linguistics (Volume 1: Long Papers)}, pages 5732--5745,
  2024.

\bibitem[Liu et~al.(2025)Liu, Zheng, Muennighoff, Zeng, Dou, Pang, Jiang, and
  Lin]{liu2025regmix}
Qian Liu, Xiaosen Zheng, Niklas Muennighoff, Guangtao Zeng, Longxu Dou, Tianyu
  Pang, Jing Jiang, and Min Lin.
\newblock Regmix: Data mixture as regression for language model pre-training.
\newblock In \emph{International Conference on Learning Representations},
  volume 2025, pages 38305--38339, 2025.

\bibitem[Xie et~al.(2023)Xie, Pham, Dong, Du, Liu, Lu, Liang, Le, Ma, and
  Yu]{xie2023doremi}
Sang~Michael Xie, Hieu Pham, Xuanyi Dong, Nan Du, Hanxiao Liu, Yifeng Lu,
  Percy~S Liang, Quoc~V Le, Tengyu Ma, and Adams~Wei Yu.
\newblock Doremi: Optimizing data mixtures speeds up language model
  pretraining.
\newblock \emph{Advances in Neural Information Processing Systems},
  36:\penalty0 69798--69818, 2023.

\bibitem[Chen et~al.(2025)Chen, Ouyang, Pearce, Hartvigsen, and
  Schwarz]{chen2025admire}
Shengzhuang Chen, Xu~Ouyang, Michael Arthur~Leopold Pearce, Thomas Hartvigsen,
  and Jonathan~Richard Schwarz.
\newblock Admire-bayesopt: Accelerated data mixture re-weighting for language
  models with bayesian optimization.
\newblock \emph{arXiv preprint arXiv:2508.11551}, 2025.

\bibitem[Yue et~al.(2025)Yue, Chen, Lu, Zhao, Wang, Yue, Song, and
  Huang]{yue2025rlvr}
Yang Yue, Zhiqi Chen, Rui Lu, Andrew Zhao, Zhaokai Wang, Yang Yue, Shiji Song,
  and Gao Huang.
\newblock Does reinforcement learning really incentivize reasoning capacity in
  llms beyond the base model?
\newblock \emph{arXiv preprint arXiv:2504.13837}, 2025.

\bibitem[Wang et~al.(2025)Wang, Yang, Zeng, Ren, Liu, Peng, Cheng, He, Wang,
  Gao, Chen, Wang, Du, and Shen]{wang2025oneshot}
Yiping Wang, Qing Yang, Zhiyuan Zeng, Liliang Ren, Liyuan Liu, Baolin Peng, Hao
  Cheng, Xuehai He, Kuan Wang, Jianfeng Gao, Weizhu Chen, Shuohang Wang,
  Simon~Shaolei Du, and Yelong Shen.
\newblock Reinforcement learning for reasoning in large language models with
  one training example.
\newblock \emph{arXiv preprint arXiv:2504.20571}, 2025.

\bibitem[Kang et~al.(2025)Kang, Kuchnik, Padthe, Vlastelica, Jia, Wu, and
  Ardalani]{kang2025quagmires}
Feiyang Kang, Michael Kuchnik, Karthik Padthe, Marin Vlastelica, Ruoxi Jia,
  Carole-Jean Wu, and Newsha Ardalani.
\newblock Quagmires in sft-rl post-training: When high sft scores mislead and
  what to use instead.
\newblock \emph{arXiv preprint arXiv:2510.01624}, 2025.

\bibitem[Cui et~al.(2025)Cui, Zhang, Chen, Yuan, Wang, Zuo, Li, Fan, Chen,
  Chen, et~al.]{cui2025entropy}
Ganqu Cui, Yuchen Zhang, Jiacheng Chen, Lifan Yuan, Zhi Wang, Yuxin Zuo,
  Haozhan Li, Yuchen Fan, Huayu Chen, Weize Chen, et~al.
\newblock The entropy mechanism of reinforcement learning for reasoning
  language models.
\newblock \emph{arXiv preprint arXiv:2505.22617}, 2025.

\bibitem[Chen et~al.(2021)Chen, Tworek, Jun, Yuan, Pinto,
  et~al.]{chen2021humaneval}
Mark Chen, Jerry Tworek, Heewoo Jun, Qiming Yuan, Henrique Ponde de~Oliveira
  Pinto, et~al.
\newblock Evaluating large language models trained on code, 2021.
\newblock URL \url{https://arxiv.org/abs/2107.03374}.
\newblock 58 authors total. Introduces the HumanEval benchmark.

\bibitem[Zheng et~al.(2025)Zheng, Liu, Li, Chen, Yu, Gao, Dang, Liu, Men, Yang,
  Zhou, and Lin]{zheng2025gspo}
Chujie Zheng, Shixuan Liu, Mingze Li, Xiong-Hui Chen, Bowen Yu, Chang Gao, Kai
  Dang, Yuqiong Liu, Rui Men, An~Yang, Jingren Zhou, and Junyang Lin.
\newblock Group sequence policy optimization.
\newblock \emph{arXiv preprint arXiv:2507.18071}, 2025.
\newblock \doi{10.48550/arXiv.2507.18071}.

\bibitem[Yu et~al.(2025{\natexlab{a}})Yu, Zhang, Zhu, Yuan, Zuo, Yue, Dai, Fan,
  Liu, Liu, Liu, Liu, Lin, Lin, Ma, Sheng, Tong, Zhang, Zhang, Zhang, Zhang,
  Zhu, Zhu, Chen, Chen, Wang, Yu, Song, Wei, Zhou, Liu, Ma, Zhang, Yan, Wu, and
  Wang]{yu2025dapo}
Qiying Yu, Zheng Zhang, Ruofei Zhu, Yufeng Yuan, Xiaochen Zuo, Yu~Yue, Weinan
  Dai, Tiantian Fan, Gaohong Liu, Juncai Liu, Lingjun Liu, Xin Liu, Haibin Lin,
  Zhiqi Lin, Bole Ma, Guangming Sheng, Yuxuan Tong, Chi Zhang, Mofan Zhang,
  Ru~Zhang, Wang Zhang, Hang Zhu, Jinhua Zhu, Jiaze Chen, Jiangjie Chen,
  Chengyi Wang, Hongli Yu, Yuxuan Song, Xiangpeng Wei, Hao Zhou, Jingjing Liu,
  Wei-Ying Ma, Ya-Qin Zhang, Lin Yan, Yonghui Wu, and Mingxuan Wang.
\newblock {DAPO}: An open-source {LLM} reinforcement learning system at scale.
\newblock In \emph{Advances in Neural Information Processing Systems 38
  (NeurIPS 2025)}, 2025{\natexlab{a}}.
\newblock URL
  \url{http://papers.nips.cc/paper_files/paper/2025/hash/a4277440d50f1f15d2cb4c14f7e0c0d2-Abstract-Conference.html}.

\bibitem[Kwon et~al.(2023)Kwon, Li, Zhuang, Sheng, Zheng, Yu, Gonzalez, Zhang,
  and Stoica]{kwon2023vllm}
Woosuk Kwon, Zhuohan Li, Siyuan Zhuang, Ying Sheng, Lianmin Zheng, Cody~Hao Yu,
  Joseph~E. Gonzalez, Hao Zhang, and Ion Stoica.
\newblock Efficient memory management for large language model serving with
  {PagedAttention}.
\newblock In \emph{Proceedings of the 29th ACM Symposium on Operating Systems
  Principles (SOSP)}, pages 611--626. ACM, 2023.
\newblock \doi{10.1145/3600006.3613165}.

\bibitem[He and Lab(2025)]{he2025nondeterminism}
Horace He and Thinking~Machines Lab.
\newblock Defeating nondeterminism in llm inference.
\newblock \emph{Thinking Machines Lab: Connectionism}, 2025.
\newblock \doi{10.64434/tml.20250910}.
\newblock
  https://thinkingmachines.ai/blog/defeating-nondeterminism-in-llm-inference/.

\bibitem[NVIDIA(2025)]{nemo-gym}
NVIDIA.
\newblock Nemo gym: An open source library for scaling reinforcement learning
  environments for llm.
\newblock \url{https://github.com/NVIDIA-NeMo/Gym}, 2025.
\newblock GitHub repository.

\bibitem[Foerster et~al.(2018)Foerster, Farquhar, Afouras, Nardelli, and
  Whiteson]{foerster2018coma}
Jakob~N. Foerster, Gregory Farquhar, Triantafyllos Afouras, Nantas Nardelli,
  and Shimon Whiteson.
\newblock Counterfactual multi-agent policy gradients.
\newblock In Sheila~A. McIlraith and Kilian~Q. Weinberger, editors,
  \emph{Proceedings of the Thirty-Second {AAAI} Conference on Artificial
  Intelligence, (AAAI-18)}, pages 2974--2982. {AAAI} Press, 2018.
\newblock \doi{10.1609/AAAI.V32I1.11794}.

\bibitem[Zhao et~al.(2025)Zhao, Hu, Wang, Hou, Zhang, Ding, and
  Zhao]{zhao2025strongermas}
Yujie Zhao, Lanxiang Hu, Yang Wang, Minmin Hou, Hao Zhang, Ke~Ding, and Jishen
  Zhao.
\newblock Stronger-mas: Multi-agent reinforcement learning for collaborative
  llms.
\newblock \emph{CoRR}, abs/2510.11062, 2025.
\newblock \doi{10.48550/arXiv.2510.11062}.

\bibitem[Hong et~al.(2025)Hong, Yin, Wang, Liu, Chen, Yu, Li, Ye, Xiao, Chen,
  et~al.]{hong2025multi}
Haoyang Hong, Jiajun Yin, Yuan Wang, Jingnan Liu, Zhe Chen, Ailing Yu, Ji~Li,
  Zhiling Ye, Hansong Xiao, Yefei Chen, et~al.
\newblock Multi-agent deep research: Training multi-agent systems with m-grpo.
\newblock \emph{arXiv preprint arXiv:2511.13288}, 2025.

\bibitem[Gu et~al.(2016)Gu, Lillicrap, Ghahramani, Turner, and Levine]{gu2016q}
Shixiang Gu, Timothy Lillicrap, Zoubin Ghahramani, Richard~E Turner, and Sergey
  Levine.
\newblock Q-prop: Sample-efficient policy gradient with an off-policy critic.
\newblock \emph{arXiv preprint arXiv:1611.02247}, 2016.

\bibitem[Gu et~al.(2017)Gu, Lillicrap, Turner, Ghahramani, Sch{\"o}lkopf, and
  Levine]{gu2017interpolated}
Shixiang~Shane Gu, Timothy Lillicrap, Richard~E Turner, Zoubin Ghahramani,
  Bernhard Sch{\"o}lkopf, and Sergey Levine.
\newblock Interpolated policy gradient: Merging on-policy and off-policy
  gradient estimation for deep reinforcement learning.
\newblock \emph{Advances in neural information processing systems}, 30, 2017.

\bibitem[Nair et~al.(2020)Nair, Gupta, Dalal, and Levine]{nair2020awac}
Ashvin Nair, Abhishek Gupta, Murtaza Dalal, and Sergey Levine.
\newblock Awac: Accelerating online reinforcement learning with offline
  datasets.
\newblock \emph{arXiv preprint arXiv:2006.09359}, 2020.

\bibitem[Yu et~al.(2022)Yu, Velu, Vinitsky, Gao, Wang, Bayen, and
  Wu]{yu2022surprising}
Chao Yu, Akash Velu, Eugene Vinitsky, Jiaxuan Gao, Yu~Wang, Alexandre Bayen,
  and Yi~Wu.
\newblock The surprising effectiveness of ppo in cooperative multi-agent games.
\newblock \emph{Advances in neural information processing systems},
  35:\penalty0 24611--24624, 2022.

\bibitem[col()]{coliee}
{COLIEE}: Competition on legal information extraction and entailment.
\newblock \url{https://coliee.org/}.
\newblock Accessed 2026-08-13.

\bibitem[Goebel et~al.(2024)Goebel, Kano, Kim, Rabelo, Satoh, and
  Yoshioka]{goebel2024coliee}
Randy Goebel, Yoshinobu Kano, Mi-Young Kim, Juliano Rabelo, Ken Satoh, and
  Masaharu Yoshioka.
\newblock Overview and discussion of the competition on legal information,
  extraction/entailment ({COLIEE}) 2023.
\newblock \emph{The Review of Socionetwork Strategies}, 18\penalty0
  (1):\penalty0 27--47, 2024.
\newblock \doi{10.1007/s12626-023-00152-0}.

\bibitem[Zheng et~al.(2021)Zheng, Guha, Anderson, Henderson, and
  Ho]{zheng2021casehold}
Lucia Zheng, Neel Guha, Brandon~R. Anderson, Peter Henderson, and Daniel~E. Ho.
\newblock When does pretraining help? assessing self-supervised learning for
  law and the {CaseHOLD} dataset of 53,000+ legal holdings.
\newblock In \emph{Proceedings of the Eighteenth International Conference on
  Artificial Intelligence and Law (ICAIL '21)}, pages 159--168. Association for
  Computing Machinery, 2021.
\newblock \doi{10.1145/3462757.3466088}.

\bibitem[{Stanford Legal Design Lab} and {Suffolk University Law School Legal
  Innovation and Technology Lab}(2018)]{learnedhands}
{Stanford Legal Design Lab} and {Suffolk University Law School Legal Innovation
  and Technology Lab}.
\newblock Learned hands.
\newblock \url{https://learnedhands.law.stanford.edu/}, 2018.
\newblock Crowdsourced legal issue-spotting labels on r/legaladvice posts.
  Incorporated into LegalBench as the \texttt{learned\_hands\_*} tasks.

\bibitem[Liang et~al.(2023)Liang, Bommasani, Lee, Tsipras, Soylu, Yasunaga,
  Zhang, Narayanan, Wu, Kumar, et~al.]{liang2022helm}
Percy Liang, Rishi Bommasani, Tony Lee, Dimitris Tsipras, Dilara Soylu,
  Michihiro Yasunaga, Yian Zhang, Deepak Narayanan, Yuhuai Wu, Ananya Kumar,
  et~al.
\newblock Holistic evaluation of language models.
\newblock \emph{Transactions on Machine Learning Research}, 2023.
\newblock URL \url{https://arxiv.org/abs/2211.09110}.
\newblock 50 authors total. Introduces the LegalSupport scenario.

\bibitem[{National Conference of Bar Examiners}()]{ncbe}
{National Conference of Bar Examiners}.
\newblock Multistate bar examination {(MBE)}.
\newblock \url{https://www.ncbex.org/exams/mbe}.
\newblock Accessed 2026-08-13.

\bibitem[Yu et~al.(2020)Yu, Jiang, Dong, and Feng]{yu2020reclor}
Weihao Yu, Zihang Jiang, Yanfei Dong, and Jiashi Feng.
\newblock {ReClor}: A reading comprehension dataset requiring logical
  reasoning.
\newblock In \emph{International Conference on Learning Representations
  (ICLR)}, 2020.
\newblock URL \url{https://arxiv.org/abs/2002.04326}.

\bibitem[{M-A-P Team} et~al.(2025){M-A-P Team}, Du, Yao, Ma, Wang, Zheng,
  et~al.]{map2025supergpqa}
{M-A-P Team}, Xinrun Du, Yifan Yao, Kaijing Ma, Bingli Wang, Tianyu Zheng,
  et~al.
\newblock {SuperGPQA}: Scaling {LLM} evaluation across 285 graduate
  disciplines, 2025.
\newblock URL \url{https://arxiv.org/abs/2502.14739}.
\newblock Also in NeurIPS 2025, Datasets and Benchmarks Track. 97 authors
  total.

\bibitem[Fan et~al.(2025)Fan, Ni, Merane, Tian, Hermstr{\"u}wer, Huang, Akhtar,
  Salimbeni, Geering, Dreyer, et~al.]{fan2025lexam}
Yu~Fan, Jingwei Ni, Jakob Merane, Yang Tian, Yoan Hermstr{\"u}wer, Yinya Huang,
  Mubashara Akhtar, Etienne Salimbeni, Florian Geering, Oliver Dreyer, et~al.
\newblock {LEXam}: Benchmarking legal reasoning on 340 law exams, 2025.
\newblock URL \url{https://arxiv.org/abs/2505.12864}.
\newblock Accepted to ICLR 2026. Code:
  \url{https://github.com/LEXam-Benchmark/LEXam}.

\bibitem[Aky{\"u}rek et~al.(2025)Aky{\"u}rek, Gosai, Zhang, Gupta, Jeong,
  Gunjal, Rabbani, Mazzone, Randolph, et~al.]{akyurek2025prbench}
Afra~Feyza Aky{\"u}rek, Advait Gosai, Chen Bo~Calvin Zhang, Vipul Gupta,
  Jaehwan Jeong, Anisha Gunjal, Tahseen Rabbani, Maria Mazzone, David Randolph,
  et~al.
\newblock {PRBench}: Large-scale expert rubrics for evaluating high-stakes
  professional reasoning, 2025.
\newblock URL \url{https://arxiv.org/abs/2511.11562}.
\newblock Scale AI, 24 authors. Code:
  \url{https://github.com/scaleapi/PRBench}.

\bibitem[Chalkidis et~al.(2021)Chalkidis, Fergadiotis, and
  Androutsopoulos]{chalkidis2021multieurlex}
Ilias Chalkidis, Manos Fergadiotis, and Ion Androutsopoulos.
\newblock {MultiEURLEX} --- a multi-lingual and multi-label legal document
  classification dataset for zero-shot cross-lingual transfer.
\newblock In \emph{Proceedings of the 2021 Conference on Empirical Methods in
  Natural Language Processing}, pages 6974--6996. Association for Computational
  Linguistics, November 2021.
\newblock \doi{10.18653/v1/2021.emnlp-main.559}.
\newblock URL \url{https://aclanthology.org/2021.emnlp-main.559/}.
\newblock Source of the {LexGLUE} EUR-LEX task.

\bibitem[Chalkidis et~al.(2022)Chalkidis, Jana, Hartung, Bommarito,
  Androutsopoulos, Katz, and Aletras]{chalkidis2022lexglue}
Ilias Chalkidis, Abhik Jana, Dirk Hartung, Michael Bommarito, Ion
  Androutsopoulos, Daniel Katz, and Nikolaos Aletras.
\newblock {LexGLUE}: A benchmark dataset for legal language understanding in
  {E}nglish.
\newblock In \emph{Proceedings of the 60th Annual Meeting of the Association
  for Computational Linguistics (Volume 1: Long Papers)}, pages 4310--4330,
  Dublin, Ireland, May 2022. Association for Computational Linguistics.
\newblock \doi{10.18653/v1/2022.acl-long.297}.
\newblock URL \url{https://aclanthology.org/2022.acl-long.297/}.

\bibitem[Tuggener et~al.(2020)Tuggener, von D{\"a}niken, Peetz, and
  Cieliebak]{tuggener2020ledgar}
Don Tuggener, Pius von D{\"a}niken, Thomas Peetz, and Mark Cieliebak.
\newblock {LEDGAR}: A large-scale multi-label corpus for text classification of
  legal provisions in contracts.
\newblock In \emph{Proceedings of the Twelfth Language Resources and Evaluation
  Conference}, pages 1235--1241, Marseille, France, May 2020. European Language
  Resources Association.
\newblock ISBN 979-10-95546-34-4.
\newblock URL \url{https://aclanthology.org/2020.lrec-1.155/}.

\bibitem[Spaeth et~al.(2020)Spaeth, Epstein, Martin, Segal, Ruger, and
  Benesh]{spaeth2020scdb}
Harold~J. Spaeth, Lee Epstein, Andrew~D. Martin, Jeffrey~A. Segal, Theodore~J.
  Ruger, and Sara~C. Benesh.
\newblock Supreme court database, version 2020 release 01.
\newblock Washington University in St. Louis, 2020.
\newblock URL \url{http://scdb.wustl.edu}.
\newblock Now hosted at \url{http://supremecourtdatabase.org}.

\bibitem[Kornilova and Eidelman(2019)]{kornilova2019billsum}
Anastassia Kornilova and Vladimir Eidelman.
\newblock {B}ill{S}um: A corpus for automatic summarization of {US}
  legislation.
\newblock In \emph{Proceedings of the 2nd Workshop on New Frontiers in
  Summarization}, pages 48--56, Hong Kong, China, November 2019. Association
  for Computational Linguistics.
\newblock \doi{10.18653/v1/D19-5406}.
\newblock URL \url{https://aclanthology.org/D19-5406/}.

\bibitem[Hendrycks et~al.(2021{\natexlab{a}})Hendrycks, Burns, Chen, and
  Ball]{hendrycks2021cuad}
Dan Hendrycks, Collin Burns, Anya Chen, and Spencer Ball.
\newblock {CUAD}: An expert-annotated {NLP} dataset for legal contract review.
\newblock \emph{arXiv preprint arXiv:2103.06268}, 2021{\natexlab{a}}.
\newblock URL \url{https://arxiv.org/abs/2103.06268}.
\newblock NeurIPS 2021 Datasets and Benchmarks Track.

\bibitem[Wang et~al.(2023{\natexlab{a}})Wang, Scardigli, Tang, Chen, Levkin,
  Chen, Ball, Woodside, Zhang, and Hendrycks]{wang2023maud}
Steven Wang, Antoine Scardigli, Leonard Tang, Wei Chen, Dmitry Levkin, Anya
  Chen, Spencer Ball, Thomas Woodside, Oliver Zhang, and Dan Hendrycks.
\newblock {MAUD}: An expert-annotated legal {NLP} dataset for merger agreement
  understanding.
\newblock In \emph{Proceedings of the 2023 Conference on Empirical Methods in
  Natural Language Processing}, pages 16369--16382, Singapore, December
  2023{\natexlab{a}}. Association for Computational Linguistics.
\newblock \doi{10.18653/v1/2023.emnlp-main.1019}.
\newblock URL \url{https://aclanthology.org/2023.emnlp-main.1019/}.

\bibitem[Wilson et~al.(2016)Wilson, Schaub, Dara, Liu, Cherivirala,
  Giovanni~Leon, Schaarup~Andersen, Zimmeck, Sathyendra, Russell,
  et~al.]{wilson2016opp115}
Shomir Wilson, Florian Schaub, Aswarth~Abhilash Dara, Frederick Liu, Sushain
  Cherivirala, Pedro Giovanni~Leon, Mads Schaarup~Andersen, Sebastian Zimmeck,
  Kanthashree~Mysore Sathyendra, N.~Cameron Russell, et~al.
\newblock The creation and analysis of a website privacy policy corpus.
\newblock In \emph{Proceedings of the 54th Annual Meeting of the Association
  for Computational Linguistics (Volume 1: Long Papers)}, pages 1330--1340,
  Berlin, Germany, August 2016. Association for Computational Linguistics.
\newblock \doi{10.18653/v1/P16-1126}.
\newblock URL \url{https://aclanthology.org/P16-1126/}.

\bibitem[Zimmeck et~al.(2019{\natexlab{a}})Zimmeck, Story, Smullen,
  Ravichander, Wang, Reidenberg, Russell, and Sadeh]{zimmeck2019app350}
Sebastian Zimmeck, Peter Story, Daniel Smullen, Abhilasha Ravichander, Ziqi
  Wang, Joel Reidenberg, N.~Cameron Russell, and Norman Sadeh.
\newblock {MAPS}: Scaling privacy compliance analysis to a million apps.
\newblock \emph{Proceedings on Privacy Enhancing Technologies}, 2019\penalty0
  (3):\penalty0 66--86, 2019{\natexlab{a}}.
\newblock \doi{10.2478/popets-2019-0037}.
\newblock Source of the APP-350 corpus.

\bibitem[Waldon et~al.(2023)Waldon, Brodsky, Ma, and
  Degen]{waldon2023consensus}
Brandon Waldon, Madigan Brodsky, Megan Ma, and Judith Degen.
\newblock Predicting consensus in legal document interpretation.
\newblock In \emph{Proceedings of the 45th Annual Conference of the Cognitive
  Science Society}, volume~45, pages 1101--1107, 2023.
\newblock URL \url{https://escholarship.org/uc/item/8rq5012j}.
\newblock Upstream source of the {LegalBench} insurance policy interpretation
  task.

\bibitem[Bang et~al.(2026)Bang, Fielding, Oliver, Birke, Seedat, and
  Bean]{bang2026contractscrub}
Yejin Bang, Kirsty Fielding, Brandan Oliver, Brian Birke, Nabeel Seedat, and
  Andrew~M. Bean.
\newblock Contractscrub: A benchmark for final review of legal contracts, 2026.
\newblock URL \url{https://arxiv.org/abs/2608.20204}.

\bibitem[{Harvey AI}(2026)]{harveylab2026}
{Harvey AI}.
\newblock Harvey {LAB}: The legal agent benchmark, 2026.
\newblock URL \url{https://github.com/harveyai/harvey-labs/tree/v1.0}.
\newblock Announcement:
  \url{https://www.harvey.ai/blog/introducing-harveys-legal-agent-benchmark}.

\bibitem[Yu et~al.(2026)Yu, Seedat, Schwarz, and Bean]{yu2026principal}
Fangyi Yu, Nabeel Seedat, Jonathan~Richard Schwarz, and Andrew~M. Bean.
\newblock To whom do language models align? measuring principal hierarchies
  under high-stakes competing demands, 2026.
\newblock URL \url{https://arxiv.org/abs/2605.12120}.

\bibitem[Vincent et~al.(2026)Vincent, Calloway, Yu, Bean, and
  Seedat]{insuffiencybench2026}
Samuel~J. Vincent, Daniel Calloway, Fangyi Yu, Andrew~M. Bean, and Nabeel
  Seedat.
\newblock Insufficiencybench: Evaluating llm legal advice on underspecified
  user queries, 2026.
\newblock URL \url{https://arxiv.org/abs/2608.20220}.

\bibitem[Zimmeck et~al.(2019{\natexlab{b}})Zimmeck, Story, Smullen,
  Ravichander, Wang, Reidenberg, Russell, and Sadeh]{zimmeck2019maps}
Sebastian Zimmeck, Peter Story, Daniel Smullen, Abhilasha Ravichander, Ziqi
  Wang, Joel~R. Reidenberg, N.~Cameron Russell, and Norman Sadeh.
\newblock Maps: Scaling privacy compliance analysis to a million apps.
\newblock \emph{Proceedings on Privacy Enhancing Technologies}, 2019\penalty0
  (3):\penalty0 66--86, 2019{\natexlab{b}}.

\bibitem[Metropolitansky and Larson(2025)]{metropolitansky2025claimify}
Dasha Metropolitansky and Jonathan Larson.
\newblock Towards effective extraction and evaluation of factual claims.
\newblock In \emph{Proceedings of the 63rd Annual Meeting of the Association
  for Computational Linguistics (Volume 1: Long Papers)}, pages 6996--7045,
  Vienna, Austria, July 2025. Association for Computational Linguistics.
\newblock \doi{10.18653/v1/2025.acl-long.348}.
\newblock URL \url{https://aclanthology.org/2025.acl-long.348/}.
\newblock arXiv:2502.10855. Introduces the Claimify method.

\bibitem[Wang et~al.(2024{\natexlab{b}})Wang, Ma, Zhang, Ni, Chandra, Guo, Ren,
  Arulraj, He, Jiang, et~al.]{wang2024mmlupro}
Yubo Wang, Xueguang Ma, Ge~Zhang, Yuansheng Ni, Abhranil Chandra, Shiguang Guo,
  Weiming Ren, Aaran Arulraj, Xuan He, Ziyan Jiang, et~al.
\newblock Mmlu-pro: A more robust and challenging multi-task language
  understanding benchmark.
\newblock \emph{arXiv preprint arXiv:2406.01574}, 2024{\natexlab{b}}.

\bibitem[Rein et~al.(2023)Rein, Hou, Stickland, Petty, Pang, Dirani, Michael,
  and Bowman]{rein2023gpqa}
David Rein, Betty~Li Hou, Asa~Cooper Stickland, Jackson Petty, Richard~Yuanzhe
  Pang, Julien Dirani, Julian Michael, and Samuel~R. Bowman.
\newblock Gpqa: A graduate-level google-proof q\&a benchmark.
\newblock \emph{arXiv preprint arXiv:2311.12022}, 2023.

\bibitem[Phan et~al.(2025)Phan, Gatti, Han, Li, et~al.]{phan2025hle}
Long Phan, Alice Gatti, Ziwen Han, Nathaniel Li, et~al.
\newblock Humanity's last exam.
\newblock \emph{arXiv preprint arXiv:2501.14249}, 2025.

\bibitem[Cobbe et~al.(2021)Cobbe, Kosaraju, Bavarian, Chen, Jun, Kaiser,
  Plappert, Tworek, Hilton, Nakano, Hesse, and Schulman]{cobbe2021gsm8k}
Karl Cobbe, Vineet Kosaraju, Mohammad Bavarian, Mark Chen, Heewoo Jun, Lukasz
  Kaiser, Matthias Plappert, Jerry Tworek, Jacob Hilton, Reiichiro Nakano,
  Christopher Hesse, and John Schulman.
\newblock Training verifiers to solve math word problems, 2021.
\newblock URL \url{https://arxiv.org/abs/2110.14168}.

\bibitem[Hendrycks et~al.(2021{\natexlab{b}})Hendrycks, Burns, Kadavath, Arora,
  Basart, Tang, Song, and Steinhardt]{hendrycks2021measuring}
Dan Hendrycks, Collin Burns, Saurav Kadavath, Akul Arora, Steven Basart, Eric
  Tang, Dawn Song, and Jacob Steinhardt.
\newblock Measuring mathematical problem solving with the math dataset.
\newblock \emph{arXiv preprint arXiv:2103.03874}, 2021{\natexlab{b}}.

\bibitem[Shi et~al.(2022)Shi, Suzgun, Freitag, Wang, Srivats, Vosoughi, Chung,
  Tay, Ruder, Zhou, Das, and Wei]{shi2022mgsm}
Freda Shi, Mirac Suzgun, Markus Freitag, Xuezhi Wang, Suraj Srivats, Soroush
  Vosoughi, Hyung~Won Chung, Yi~Tay, Sebastian Ruder, Denny Zhou, Dipanjan Das,
  and Jason Wei.
\newblock Language models are multilingual chain-of-thought reasoners.
\newblock \emph{arXiv preprint arXiv:2210.03057}, 2022.

\bibitem[Hu et~al.(2020)Hu, Ruder, Siddhant, Neubig, Firat, and
  Johnson]{hu2020xtreme}
Junjie Hu, Sebastian Ruder, Aditya Siddhant, Graham Neubig, Orhan Firat, and
  Melvin Johnson.
\newblock Xtreme: A massively multilingual multi-task benchmark for evaluating
  cross-lingual generalization.
\newblock \emph{arXiv preprint arXiv:2003.11080}, 2020.

\bibitem[Ming et~al.(2025)Ming, Purushwalkam, Pandit, Ke, Nguyen, Xiong, and
  Joty]{ming2025faitheval}
Yifei Ming, Senthil Purushwalkam, Shrey Pandit, Zixuan Ke, Xuan-Phi Nguyen,
  Caiming Xiong, and Shafiq Joty.
\newblock {FaithEval}: Can your language model stay faithful to context, even
  if ``the moon is made of marshmallows''.
\newblock In \emph{The Thirteenth International Conference on Learning
  Representations (ICLR)}, 2025.
\newblock URL \url{https://openreview.net/forum?id=UeVx6L59fg}.
\newblock arXiv:2410.03727.

\bibitem[Haas et~al.(2025)Haas, Yona, D'Antonio, Goldshtein, and
  Das]{haas2025simpleqaverified}
Lukas Haas, Gal Yona, Giovanni D'Antonio, Sasha Goldshtein, and Dipanjan Das.
\newblock {SimpleQA} verified: A reliable factuality benchmark to measure
  parametric knowledge, 2025.
\newblock URL \url{https://arxiv.org/abs/2509.07968}.

\bibitem[Wei et~al.(2024)Wei, Karina, Chung, Jiao, Papay, Glaese, Schulman, and
  Fedus]{wei2024simpleqa}
Jason Wei, Nguyen Karina, Hyung~Won Chung, Yunxin~Joy Jiao, Spencer Papay,
  Amelia Glaese, John Schulman, and William Fedus.
\newblock Measuring short-form factuality in large language models, 2024.
\newblock URL \url{https://arxiv.org/abs/2411.04368}.

\bibitem[Zhou et~al.(2023)Zhou, Lu, Mishra, Brahma, Basu, Luan, Zhou, and
  Hou]{zhou2023ifeval}
Jeffrey Zhou, Tianjian Lu, Swaroop Mishra, Siddhartha Brahma, Sujoy Basu,
  Yi~Luan, Denny Zhou, and Le~Hou.
\newblock Instruction-following evaluation for large language models.
\newblock \emph{arXiv preprint arXiv:2311.07911}, 2023.

\bibitem[Jiang et~al.(2023)Jiang, Wang, Zeng, Zhong, Li, Mi, Shang, Jiang, Liu,
  and Wang]{jiang2023followbench}
Yuxin Jiang, Yufei Wang, Xingshan Zeng, Wanjun Zhong, Liangyou Li, Fei Mi,
  Lifeng Shang, Xin Jiang, Qun Liu, and Wei Wang.
\newblock Followbench: A multi-level fine-grained constraints following
  benchmark for large language models.
\newblock \emph{arXiv preprint arXiv:2310.20410}, 2023.

\bibitem[Wu et~al.(2025)Wu, Mei, Yan, Li, Lai, Ren, Wang, Zhang, Wu, Jin, and
  Huang]{wu2025writingbench}
Yuning Wu, Jiahao Mei, Ming Yan, Chenliang Li, Shaopeng Lai, Yuran Ren, Zijia
  Wang, Ji~Zhang, Mengyue Wu, Qin Jin, and Fei Huang.
\newblock Writingbench: A comprehensive benchmark for generative writing.
\newblock \emph{arXiv preprint arXiv:2503.05244}, 2025.

\bibitem[Zhang et~al.(2024)Zhang, Chen, Hu, Xu, Chen, Hao, Han, Thai, Wang,
  Liu, and Sun]{zhang2024infinitebench}
Xinrong Zhang, Yingfa Chen, Shengding Hu, Zihang Xu, Junhao Chen, Moo Hao,
  Xu~Han, Zhen Thai, Shuo Wang, Zhiyuan Liu, and Maosong Sun.
\newblock $\infty$bench: Extending long context evaluation beyond 100k tokens.
\newblock In \emph{Proceedings of the 62nd Annual Meeting of the Association
  for Computational Linguistics (Volume 1: Long Papers)}, pages 15262--15277,
  2024.

\bibitem[Jimenez et~al.(2023)Jimenez, Yang, Wettig, Yao, Pei, Press, and
  Narasimhan]{jimenez2023swebench}
Carlos~E. Jimenez, John Yang, Alexander Wettig, Shunyu Yao, Kexin Pei, Ofir
  Press, and Karthik Narasimhan.
\newblock Swe-bench: Can language models resolve real-world github issues?
\newblock \emph{arXiv preprint arXiv:2310.06770}, 2023.

\bibitem[Patwardhan et~al.(2025)Patwardhan, Dias, Proehl, Kim, Wang, Watkins,
  Fishman, Aljubeh, Thacker, Fauconnet, Kim, Chao, Miserendino, Chabot, Li,
  Sharman, Barr, Glaese, and Tworek]{patwardhan2025gdpval}
Tejal Patwardhan, Rachel Dias, Elizabeth Proehl, Grace Kim, Michele Wang,
  Olivia Watkins, Sim{\'o}n~Posada Fishman, Marwan Aljubeh, Phoebe Thacker,
  Laurance Fauconnet, Natalie~S. Kim, Patrick Chao, Samuel Miserendino, Gildas
  Chabot, David Li, Michael Sharman, Alexandra Barr, Amelia Glaese, and Jerry
  Tworek.
\newblock Gdpval: Evaluating ai model performance on real-world economically
  valuable tasks.
\newblock \emph{arXiv preprint arXiv:2510.04374}, 2025.

\bibitem[Barres et~al.(2025)Barres, Dong, Ray, Si, and
  Narasimhan]{barres2025tau2}
Victor Barres, Honghua Dong, Soham Ray, Xujie Si, and Karthik Narasimhan.
\newblock $\tau^2$-bench: Evaluating conversational agents in a dual-control
  environment.
\newblock \emph{arXiv preprint arXiv:2506.07982}, 2025.

\bibitem[Yao et~al.(2024{\natexlab{b}})Yao, Shinn, Razavi, and
  Narasimhan]{yao2024taubench}
Shunyu Yao, Noah Shinn, Pedram Razavi, and Karthik Narasimhan.
\newblock $\tau$-bench: A benchmark for tool-agent-user interaction in
  real-world domains.
\newblock \emph{arXiv preprint arXiv:2406.12045}, 2024{\natexlab{b}}.

\bibitem[Snell et~al.(2024)Snell, Lee, Xu, and Kumar]{snell2024scaling}
Charlie Snell, Jaehoon Lee, Kelvin Xu, and Aviral Kumar.
\newblock Scaling llm test-time compute optimally can be more effective than
  scaling model parameters.
\newblock \emph{arXiv preprint arXiv:2408.03314}, 2024.

\bibitem[Beeching et~al.(2024)Beeching, Tunstall, and
  Rush]{beeching2024scalingtesttimecompute}
Edward Beeching, Lewis Tunstall, and Sasha Rush.
\newblock Scaling test-time compute with open models, 2024.
\newblock URL
  \url{https://huggingface.co/spaces/HuggingFaceH4/blogpost-scaling-test-time-compute}.

\bibitem[Wu et~al.(2024)Wu, Sun, Li, Welleck, and Yang]{wu2024inference}
Yangzhen Wu, Zhiqing Sun, Shanda Li, Sean Welleck, and Yiming Yang.
\newblock Inference scaling laws: An empirical analysis of compute-optimal
  inference for problem-solving with language models.
\newblock \emph{arXiv preprint arXiv:2408.00724}, 2024.

\bibitem[Puri et~al.(2025)Puri, Sudalairaj, Xu, Xu, and
  Srivastava]{puri2025rollout}
Isha Puri, Shivchander Sudalairaj, Guangxuan Xu, Kai Xu, and Akash Srivastava.
\newblock Rollout roulette: A probabilistic inference approach to
  inference-time scaling of llms using particle-based monte carlo methods.
\newblock \emph{arXiv preprint arXiv:2502.01618}, 2025.

\bibitem[Madaan et~al.(2023)Madaan, Tandon, Gupta, Hallinan, Gao, Wiegreffe,
  Alon, Dziri, Prabhumoye, Yang, et~al.]{madaan2023self}
Aman Madaan, Niket Tandon, Prakhar Gupta, Skyler Hallinan, Luyu Gao, Sarah
  Wiegreffe, Uri Alon, Nouha Dziri, Shrimai Prabhumoye, Yiming Yang, et~al.
\newblock Self-refine: Iterative refinement with self-feedback.
\newblock \emph{Advances in neural information processing systems},
  36:\penalty0 46534--46594, 2023.

\bibitem[Khairi et~al.(2025)Khairi, D'souza, Fadaee, and
  Kreutzer]{khairi2025making}
Ammar Khairi, Daniel D'souza, Marzieh Fadaee, and Julia Kreutzer.
\newblock Making, not taking, the best of n.
\newblock \emph{arXiv preprint arXiv:2510.00931}, 2025.

\bibitem[Romano et~al.(2026)Romano, Raj, Parker, and
  Giofr{\'e}]{romano2026testtime}
Davide Romano, Kanak Raj, Jerrod Parker, and Daniele Giofr{\'e}.
\newblock Test-time scaling in the wild: Why exploitation, not exploration, is
  the bottleneck.
\newblock \emph{arXiv preprint arXiv:2608.18931}, 2026.

\bibitem[Bhatt et~al.(2024)Bhatt, Chennabasappa, Li, Nikolaidis, Song, Wan,
  Ahmad, Aschermann, Chen, Kapil, Molnar, Whitman, and
  Saxe]{bhatt2024cyberseceval2}
Manish Bhatt, Sahana Chennabasappa, Yue Li, Cyrus Nikolaidis, Daniel Song,
  Shengye Wan, Faizan Ahmad, Cornelius Aschermann, Yaohui Chen, Dhaval Kapil,
  David Molnar, Spencer Whitman, and Joshua Saxe.
\newblock {CyberSecEval 2}: A wide-ranging cybersecurity evaluation suite for
  large language models, 2024.
\newblock URL \url{https://arxiv.org/abs/2404.13161}.

\bibitem[{elder-plinius}(2024)]{elderplinius2024l1b3rt4s}
{elder-plinius}.
\newblock {L1B3RT4S}.
\newblock GitHub repository, 2024.
\newblock URL \url{https://github.com/elder-plinius/L1B3RT4S}.
\newblock Corpus source accessed July 14, 2026.

\bibitem[Ghosh et~al.(2024)Ghosh, Varshney, Galinkin, and
  Parisien]{ghosh2024aegis}
Shaona Ghosh, Prasoon Varshney, Erick Galinkin, and Christopher Parisien.
\newblock {AEGIS}: Online adaptive ai content safety moderation with ensemble
  of llm experts, 2024.
\newblock URL \url{https://arxiv.org/abs/2404.05993}.

\bibitem[Ji et~al.(2023)Ji, Liu, Dai, Pan, Zhang, Bian, Zhang, Sun, Wang, and
  Yang]{ji2023beavertails}
Jiaming Ji, Mickel Liu, Juntao Dai, Xuehai Pan, Chi Zhang, Ce~Bian, Chi Zhang,
  Ruiyang Sun, Yizhou Wang, and Yaodong Yang.
\newblock {BeaverTails}: Towards improved safety alignment of {LLM} via a
  human-preference dataset, 2023.
\newblock URL \url{https://arxiv.org/abs/2307.04657}.

\bibitem[Wang et~al.(2023{\natexlab{b}})Wang, Li, Han, Nakov, and
  Baldwin]{wang2023donotanswer}
Yuxia Wang, Haonan Li, Xudong Han, Preslav Nakov, and Timothy Baldwin.
\newblock {Do-Not-Answer}: A dataset for evaluating safeguards in {LLM}s,
  2023{\natexlab{b}}.
\newblock URL \url{https://arxiv.org/abs/2308.13387}.

\bibitem[Mazeika et~al.(2024)Mazeika, Phan, Yin, Zou, Wang, Mu, Sakhaee, Li,
  Basart, Li, Forsyth, and Hendrycks]{mazeika2024harmbench}
Mantas Mazeika, Long Phan, Xuwang Yin, Andy Zou, Zifan Wang, Norman Mu, Elham
  Sakhaee, Nathaniel Li, Steven Basart, Bo~Li, David Forsyth, and Dan
  Hendrycks.
\newblock {HarmBench}: A standardized evaluation framework for automated red
  teaming and robust refusal, 2024.
\newblock URL \url{https://arxiv.org/abs/2402.04249}.

\bibitem[Lin et~al.(2023)Lin, Wang, Tong, Wang, Guo, Wang, and
  Shang]{lin2023toxicchat}
Zi~Lin, Zihan Wang, Yongqi Tong, Yangkun Wang, Yuxin Guo, Yujia Wang, and
  Jingbo Shang.
\newblock {ToxicChat}: Unveiling hidden challenges of toxicity detection in
  real-world user--ai conversation, 2023.
\newblock URL \url{https://arxiv.org/abs/2310.17389}.

\bibitem[R{\"o}ttger et~al.(2024)R{\"o}ttger, Kirk, Vidgen, Attanasio, Bianchi,
  and Hovy]{rottger2024xstest}
Paul R{\"o}ttger, Hannah~Rose Kirk, Bertie Vidgen, Giuseppe Attanasio, Federico
  Bianchi, and Dirk Hovy.
\newblock {XSTest}: A test suite for identifying exaggerated safety behaviours
  in large language models, 2024.
\newblock URL \url{https://arxiv.org/abs/2308.01263}.

\bibitem[Yu et~al.(2025{\natexlab{b}})Yu, Seedat, Herrmannova, Schilder, and
  Schwarz]{yu2025dece}
Fangyi Yu, Nabeel Seedat, Dasha Herrmannova, Frank Schilder, and
  Jonathan~Richard Schwarz.
\newblock Beyond pointwise scores: Decomposed criteria-based evaluation of
  {LLM} responses.
\newblock In \emph{Proceedings of the 2025 Conference on Empirical Methods in
  Natural Language Processing: Industry Track}, pages 1931--1954. Association
  for Computational Linguistics, 2025{\natexlab{b}}.

\bibitem[{UK AI Security Institute}(2024)]{inspect_ai}
{UK AI Security Institute}.
\newblock Inspect: A framework for large language model evaluations.
\newblock \url{https://github.com/UKGovernmentBEIS/inspect_ai}, 2024.
\newblock GitHub repository.

\bibitem[{ClearML}(2021)]{clearml}
{ClearML}.
\newblock {ClearML}: Experiment manager, {MLOps} and data management.
\newblock \url{https://github.com/allegroai/clearml}, 2021.
\newblock GitHub repository.

\bibitem[{DataHub Project}(2020)]{datahub}
{DataHub Project}.
\newblock {DataHub}: A metadata platform for the modern data stack.
\newblock \url{https://github.com/datahub-project/datahub}, 2020.
\newblock GitHub repository.

\bibitem[Lu(2015)]{seaweedfs}
Chris Lu.
\newblock {SeaweedFS}: A distributed storage system for blobs, objects, files
  and data lake.
\newblock \url{https://github.com/seaweedfs/seaweedfs}, 2015.
\newblock GitHub repository.

\bibitem[nem(2025)]{nemo-rl}
{NeMo RL}: A scalable and efficient post-training library.
\newblock \url{https://github.com/NVIDIA-NeMo/RL}, 2025.
\newblock GitHub repository.

\bibitem[Moritz et~al.(2018)Moritz, Nishihara, Wang, Tumanov, Liaw, Liang,
  Elibol, Yang, Paul, Jordan, and Stoica]{ray}
Philipp Moritz, Robert Nishihara, Stephanie Wang, Alexey Tumanov, Richard Liaw,
  Eric Liang, Melih Elibol, Zongheng Yang, William Paul, Michael~I. Jordan, and
  Ion Stoica.
\newblock {Ray}: A distributed framework for emerging {AI} applications.
\newblock In \emph{13th {USENIX} Symposium on Operating Systems Design and
  Implementation ({OSDI} '18)}, pages 561--577, 2018.

\bibitem[Shoeybi et~al.(2019)Shoeybi, Patwary, Puri, LeGresley, Casper, and
  Catanzaro]{megatron-lm}
Mohammad Shoeybi, Mostofa Patwary, Raul Puri, Patrick LeGresley, Jared Casper,
  and Bryan Catanzaro.
\newblock {Megatron-LM}: Training multi-billion parameter language models using
  model parallelism.
\newblock \emph{arXiv preprint arXiv:1909.08053}, 2019.

\bibitem[Mindermann et~al.(2022)Mindermann, Brauner, Razzak, Sharma, Kirsch,
  Xu, H{\"o}ltgen, Gomez, Morisot, Farquhar, et~al.]{mindermann2022prioritized}
S{\"o}ren Mindermann, Jan~M Brauner, Muhammed~T Razzak, Mrinank Sharma, Andreas
  Kirsch, Winnie Xu, Benedikt H{\"o}ltgen, Aidan~N Gomez, Adrien Morisot,
  Sebastian Farquhar, et~al.
\newblock Prioritized training on points that are learnable, worth learning,
  and not yet learnt.
\newblock In \emph{International Conference on Machine Learning}, pages
  15630--15649. PMLR, 2022.

\bibitem[Evans et~al.(2024)Evans, Pathak, Merzic, Schwarz, Tanno, and
  Henaff]{evans2024bad}
Talfan Evans, Shreya Pathak, Hamza Merzic, Jonathan Schwarz, Ryutaro Tanno, and
  Olivier~J Henaff.
\newblock Bad students make great teachers: Active learning accelerates
  large-scale visual understanding.
\newblock In \emph{European Conference on Computer Vision}, pages 264--280.
  Springer, 2024.

\bibitem[Brandfonbrener et~al.(2024)Brandfonbrener, Zhang, Kirsch, Schwarz, and
  Kakade]{brandfonbrener2024color}
David Brandfonbrener, Hanlin Zhang, Andreas Kirsch, Jonathan~Richard Schwarz,
  and Sham Kakade.
\newblock Color-filter: Conditional loss reduction filtering for targeted
  language model pre-training.
\newblock \emph{Advances in Neural Information Processing Systems},
  37:\penalty0 97618--97649, 2024.

\bibitem[Shenfeld et~al.(2026)Shenfeld, Pari, and Agrawal]{shenfeld2026rl}
Idan Shenfeld, Jyothish Pari, and Pulkit Agrawal.
\newblock Rl's razor: Why online reinforcement learning forgets less.
\newblock In \emph{International Conference on Learning Representations},
  volume 2026, pages 59839--59864, 2026.

\bibitem[Ruan et~al.(2024)Ruan, Wang, and Wan]{ruan2024vulnerability}
Jie Ruan, Wenqing Wang, and Xiaojun Wan.
\newblock Defining and detecting vulnerability in human evaluation guidelines:
  A preliminary study towards reliable {NLG} evaluation.
\newblock In \emph{Proceedings of the 2024 Conference of the North American
  Chapter of the Association for Computational Linguistics: Human Language
  Technologies}, 2024.

\bibitem[Yu(2024)]{yu2024guidelines}
Fangyi Yu.
\newblock How to build reliable human annotation guidelines with {LLM}s.
\newblock
  \url{https://medium.com/tr-labs-ml-engineering-blog/how-to-build-reliable-human-annotation-guidelines-with-llms-2cd8bbeff2a2},
  8 2024.
\newblock Thomson Reuters Labs, ML Engineering Blog (Medium).

\bibitem[Jacovi et~al.(2025)Jacovi, Wang, Alberti, Tao, Lipovetz, Olszewska,
  Haas, Liu, Keating, Bloniarz, et~al.]{jacovi2025facts}
Alon Jacovi, Andrew Wang, Chris Alberti, Connie Tao, Jon Lipovetz, Kate
  Olszewska, Lukas Haas, Michelle Liu, Nate Keating, Adam Bloniarz, et~al.
\newblock The {FACTS} grounding leaderboard: Benchmarking {LLMs}' ability to
  ground responses to long-form input, 2025.
\newblock URL \url{https://arxiv.org/abs/2501.03200}.
\newblock 26 authors total.

\bibitem[Falke et~al.(2019)Falke, Ribeiro, Utama, Dagan, and
  Gurevych]{falke2019cogensumm}
Tobias Falke, Leonardo F.~R. Ribeiro, Prasetya~Ajie Utama, Ido Dagan, and Iryna
  Gurevych.
\newblock Ranking generated summaries by correctness: An interesting but
  challenging application for natural language inference.
\newblock In \emph{Proceedings of the 57th Annual Meeting of the Association
  for Computational Linguistics}, pages 2214--2220, Florence, Italy, July 2019.
  Association for Computational Linguistics.
\newblock \doi{10.18653/v1/P19-1213}.
\newblock URL \url{https://aclanthology.org/P19-1213/}.

\bibitem[Laban et~al.(2022)Laban, Schnabel, Bennett, and
  Hearst]{laban2022summac}
Philippe Laban, Tobias Schnabel, Paul~N. Bennett, and Marti~A. Hearst.
\newblock {SummaC}: Re-visiting {NLI}-based models for inconsistency detection
  in summarization.
\newblock \emph{Transactions of the Association for Computational Linguistics},
  10:\penalty0 163--177, 2022.
\newblock \doi{10.1162/tacl_a_00453}.
\newblock URL \url{https://aclanthology.org/2022.tacl-1.10/}.

\bibitem[Maynez et~al.(2020)Maynez, Narayan, Bohnet, and
  McDonald]{maynez2020xsumfaith}
Joshua Maynez, Shashi Narayan, Bernd Bohnet, and Ryan McDonald.
\newblock On faithfulness and factuality in abstractive summarization.
\newblock In \emph{Proceedings of the 58th Annual Meeting of the Association
  for Computational Linguistics}, pages 1906--1919. Association for
  Computational Linguistics, July 2020.
\newblock \doi{10.18653/v1/2020.acl-main.173}.
\newblock URL \url{https://aclanthology.org/2020.acl-main.173/}.

\bibitem[AI(2026)]{valsai2026legalresearch}
Vals AI.
\newblock Legal research bench: Evaluating agents on us legal research tasks.
\newblock Vals AI, June 2026.
\newblock URL \url{https://github.com/vals-ai/legal-research-bench}.

\bibitem[{Vals AI}(2025)]{valsai2025taxeval}
{Vals AI}.
\newblock Taxeval v2.
\newblock Vals AI, 2025.
\newblock URL \url{https://www.vals.ai/benchmarks/tax_eval_v2}.

\bibitem[Bigeard et~al.(2025)Bigeard, Nashold, Krishnan, and
  Wu]{bigeard2025fab}
Antoine Bigeard, Langston Nashold, Rayan Krishnan, and Shirley Wu.
\newblock Finance agent benchmark: Benchmarking llms on real-world financial
  research tasks.
\newblock Vals AI, May 2025.
\newblock URL \url{https://arxiv.org/abs/2508.00828}.

\end{thebibliography}

\clearpage
\appendix

\section{Working with Domain Experts}
\subsection{Enhancing Inter-Annotator Agreement in Expert Annotation Studies}
\label{subsec:iaa-process}

LLM judges are ultimately validated against, and their auto-extracted
criteria are seeded from, human expert annotations. If those annotations are
inconsistent across annotators, both the validation of the judge and the
gold labels it is compared against are unreliable, no matter how well the
judge itself is designed. Domain experts -- e.g., clinicians, financial
analysts, or legal practitioners, depending on the field -- are also an
expensive, low-throughput resource, so the annotation process has to raise
inter-annotator agreement (IAA) \emph{without} simply throwing more expert
time at the problem. Below we generalise a workflow we have found effective
in practice into a template that applies to any expert-annotation study
feeding gold labels into an LLM judge, independent of domain.

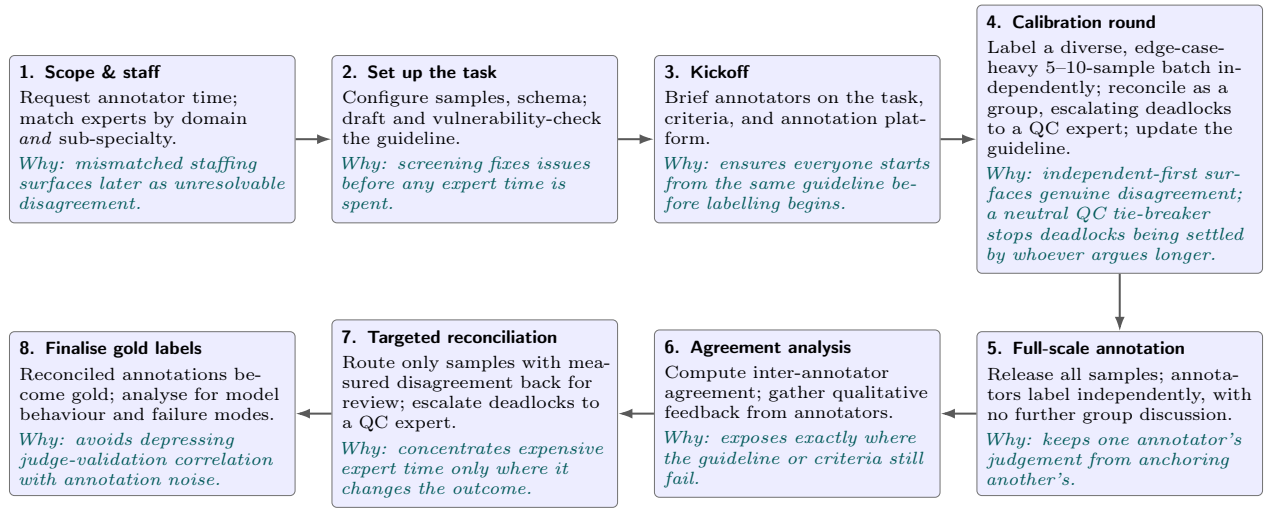
\begin{figure}[htbp]
\centering
\resizebox{\linewidth}{!}{%
\begin{tikzpicture}[
  stage/.style={rectangle, rounded corners=2pt, draw=black!55, fill=blue!7,
    text width=3.7cm, align=left, font=\scriptsize, inner sep=4pt},
  arr/.style={-{Latex[length=2mm, width=1.4mm]}, thick, draw=black!65},
]

\node[stage] (scope) {\textbf{1. Scope \& staff}\\[2pt]
Request annotator time; match experts by domain \emph{and}
sub-specialty.\\[2pt]
{\itshape\color{teal!70!black} Why: mismatched staffing surfaces later as
unresolvable disagreement.}};

\node[stage, right=5mm of scope] (setup) {\textbf{2. Set up the task}
\\[2pt]
Configure samples, schema; draft and vulnerability-check the
guideline.\\[2pt]
{\itshape\color{teal!70!black} Why: screening fixes issues before any
expert time is spent.}};

\node[stage, right=5mm of setup] (kickoff) {\textbf{3. Kickoff}\\[2pt]
Brief annotators on the task, criteria, and annotation platform.\\[2pt]
{\itshape\color{teal!70!black} Why: ensures everyone starts from the same
guideline before labelling begins.}};

\node[stage, right=5mm of kickoff] (calib) {\textbf{4. Calibration round}
\\[2pt]
Label a diverse, edge-case-heavy 5--10-sample batch independently;
reconcile as a group, escalating deadlocks to a QC expert; update the
guideline.\\[2pt]
{\itshape\color{teal!70!black} Why: independent-first surfaces genuine
disagreement; a neutral QC tie-breaker stops deadlocks being settled by
whoever argues longer.}};

\node[stage, below=8mm of calib] (official) {\textbf{5. Full-scale
annotation}\\[2pt]
Release all samples; annotators label independently, with no further
group discussion.\\[2pt]
{\itshape\color{teal!70!black} Why: keeps one annotator's judgement from
anchoring another's.}};

\node[stage, left=5mm of official] (iaa) {\textbf{6. Agreement analysis}
\\[2pt]
Compute inter-annotator agreement; gather qualitative feedback from
annotators.\\[2pt]
{\itshape\color{teal!70!black} Why: exposes exactly where the guideline or
criteria still fail.}};

\node[stage, left=5mm of iaa] (reconcile) {\textbf{7. Targeted
reconciliation}\\[2pt]
Route only samples with measured disagreement back for review; escalate
deadlocks to a QC expert.\\[2pt]
{\itshape\color{teal!70!black} Why: concentrates expensive expert time only
where it changes the outcome.}};

\node[stage, left=5mm of reconcile] (gold) {\textbf{8. Finalise gold
labels}\\[2pt]
Reconciled annotations become gold; analyse for model behaviour and failure
modes.\\[2pt]
{\itshape\color{teal!70!black} Why: avoids depressing judge-validation
correlation with annotation noise.}};

\draw[arr] (scope) -- (setup);
\draw[arr] (setup) -- (kickoff);
\draw[arr] (kickoff) -- (calib);
\draw[arr] (calib) -- (official);
\draw[arr] (official) -- (iaa);
\draw[arr] (iaa) -- (reconcile);
\draw[arr] (reconcile) -- (gold);

\end{tikzpicture}%
}
\caption{Recommended annotation workflow for producing high-agreement expert
gold labels, generalised across domains. Each step's box states why it
matters for inter-annotator agreement and gold-label reliability.}
\label{fig:iaa-pipeline}
\end{figure}

Figure~\ref{fig:iaa-pipeline} summarises the recommended pipeline together
with why each step matters for agreement and gold-label reliability -- the
mechanism behind the pipeline is not any single step but their ordering:
independent annotation before discussion (Step 4), and targeted rather than
blanket reconciliation after the full-scale round (Step 7). Two further
safeguards run underneath both of those steps: the calibration batch is
deliberately built to be diverse and edge-case-heavy rather than
convenience-sampled, so it stress-tests the guideline against the
disagreements annotators are actually likely to have; and any reconciliation
deadlock between two annotators -- in either the calibration round or the
targeted-reconciliation round -- is escalated to a neutral quality-check
(QC) expert rather than left unresolved or settled by whoever argues longer.

\paragraph{Writing the guideline itself (Step 2).} Screening a draft
guideline before it reaches annotators (Figure~\ref{fig:iaa-pipeline}, Step
2) is only as good as the checklist used to screen it.
\citet{ruan2024vulnerability} analysed human-evaluation guidelines from
major NLP venues and found that 77\% exhibited at least one of eight
recurring vulnerability types: ethical oversights, unconscious bias,
ambiguous definitions, unclear rating scales, unaddressed edge cases,
unwarranted assumptions of annotator prior knowledge, inflexible
instructions, and a residual ``other'' category. Because these
vulnerabilities recur across guidelines regardless of domain, they can be
screened for systematically rather than relying on the project lead to
notice them ad hoc.

In \citet{yu2024guidelines}, we describe a practical three-step process for
using LLMs to draft and harden a guideline against this taxonomy before it
ever reaches annotators: (1) \emph{draft} an initial guideline with an LLM,
prompted with the task definition, label space, and known edge cases; (2)
\emph{manually revise} the draft against the eight vulnerability categories
above -- some, such as ethical oversights and unconscious bias, require
human judgement an LLM pass alone should not be trusted to catch; and (3) run
a second LLM pass, using chain-of-thought prompting with a small number of
few-shot examples, specifically to flag any of the eight vulnerability types
still present in the revised draft, before the guideline is used in the
calibration round. This closes most of a guideline's gaps \emph{before}
spending expert time on calibration, rather than discovering them only once
annotators return genuinely different interpretations of the same
instructions.

\paragraph{Takeaways for designing an annotation study for LLM-judge gold data.}
\begin{itemize}[nosep]
  \item Staff narrowly by expert specialist: match annotators to the task by
  domain expertise \emph{and} sub-specialty, not just general subject
  knowledge. In many expert domains, ``qualified'' fragments into narrow,
  largely non-interchangeable sub-specialties rather than one broad
  category -- in medicine, for instance, cardiology, radiology, and
  oncology are distinct specialties, and an experienced clinician in one
  cannot be assumed to have working proficiency in another; similar
  fragmentation holds in law and in finance (e.g., tax vs.\ securities). Mismatched staffing shows up
  downstream as calibration overhead at best and unresolvable disagreement
  at worst. Because qualified annotators for any single narrow sub-specialty
  are often scarce, this constraint needs to be planned for -- and strictly
  enforced -- before the study is scoped, not treated as a best-effort
  filter once the task is already underway.
  \item Vulnerability-check the guideline before calibration: systematically screening a draft guideline against known failure
  categories (ambiguous definitions, unclear rating scales, unaddressed edge
  cases, etc.) using an LLM-assisted draft-revise-detect process
  \citep{ruan2024vulnerability,yu2024guidelines} catches issues cheaply
  before annotators ever see the task; calibration remains necessary on top
  of this for the domain-specific disagreements that only annotators
  applying the guideline will reveal.
  \item Always run a calibration round (independent annotation, then group
  reconciliation, then guideline update) before releasing the full sample
  set -- skipping straight to full annotation is the most common cause of
  low IAA.
  \item Build the calibration batch on purpose, and staff a tie-breaker in
  advance: select the shared calibration batch to be diverse and
  edge-case-heavy, not a convenience sample, so it stress-tests the guideline
  against the disagreements annotators are actually likely to have; and
  designate a neutral QC expert ahead of time to adjudicate any
  two-annotator deadlock, in calibration or in targeted reconciliation,
  rather than leaving it unresolved or defaulting to majority vote.
  \item Report and act on IAA \emph{before} finalising gold labels: compute
  agreement, send only disputed samples back for reconciliation, and treat
  the reconciled set (not majority vote or a single annotator) as gold.
  \item Feed disagreement patterns and retrospective feedback back into
  both the human annotation guidelines and the LLM judge's own criteria
  (Section~\ref{subsec:dece}) -- categories that are hard for experts to agree
  on are often exactly the categories where auto-extracted judge criteria need
  the most scrutiny.
\end{itemize}

\subsection{Experts as Collaborators}
While the guidelines above provide practical guidance for working with a more distant team of experts, as we expect is often the case, we also note the value that can come from close coordination with a consistent team of experts. Over the course of our model training, our core team became increasingly impactful contributors to the more technical side of the project, from raising design issues in data collection plans to proposing new tasks for training. Ideas proposed by subject experts led to the creation and improvement of several of our datasets, two of which we published with experts as co-authors\cite{bang2026contractscrub, insuffiencybench2026}. We believe this close integration of experts significantly increases the real-world impact of our training.

\newtcolorbox{datacard}[1]{%
  enhanced,
  breakable,
  colback=black!3,
  colframe=black!45,
  colbacktitle=black!12,
  coltitle=black,
  fonttitle=\bfseries\small,
  title={#1},
  boxrule=0.4pt,
  arc=2pt,
  left=5pt, right=5pt, top=4pt, bottom=4pt,
  before skip=7pt, after skip=7pt,
}

\newcommand{\dcm}[2]{{\footnotesize\textbf{#1:}~#2\par}}
\newcommand{\dcd}[1]{\smallskip{\footnotesize #1\par}}
\newcommand{\dce}[1]{\smallskip{\footnotesize\textit{Example.}~#1\par}}
\newcommand{\dcnote}[1]{{\footnotesize\itshape #1\par}\medskip}

\newcommand{\EXT}{\textsc{external}}
\newcommand{\INT}{\textsc{internal}}

\section{Evaluation Benchmark Data Cards}
\label{app:datacards}

This appendix documents relevant benchmark datasets. For each dataset we give the task format, the scoring metric, the
provenance (\EXT{} for public benchmarks, \INT{} for tasks built with Thomson
Reuters subject-matter experts), the evaluation set size, a condensed
description, and a representative item. Unless explicitly stated, we use random sampling when the benchmark is smaller than the original.

\subsection{Stanford LegalBench}
\label{app:dc:legalbench}

\begin{datacard}{LegalBench}
\dcm{Format}{Varies by subtask (primarily classification)}
\dcm{Metric}{Varies by subtask, per the official implementation}
\dcm{Source}{\EXT{} -- Guha et al. (2023)~\citep{guha2023legalbench}}
\dcm{Items}{89{,}485 across 162 subtasks}
\dcd{A collaborative benchmark for legal reasoning curated by 40 contributors
from the legal and AI communities. Its 162 tasks fall into five categories by
reasoning type: \textit{Issue} (17 subtasks, 11{,}599 samples), spotting legal
issues in fact patterns; \textit{Rule} (5, 9{,}809), applying rules and
determining applicability; \textit{Conclusion} (12, 2{,}227), drawing
conclusions from facts and rules; \textit{Interpretation} (118, 59{,}611),
interpreting statutes, contracts and case language; and \textit{Rhetorical}
(10, 6{,}239), understanding argumentation and persuasion. The largest
families are CUAD (38 subtasks), MAUD (34), Learned Hands (16), ContractNLI
(14), supply-chain disclosure (10) and OPP-115 (9). Each subtask carries its
own format and metric.}
\dce{Items vary widely -- from identifying contractual provisions and
interpreting statutes to recognising reasoning patterns and classifying
sections of judicial opinions.}
\end{datacard}

\subsection{Legal Information Retrieval}
\label{app:dc:legalir}

\begin{datacard}{COLIEE Task 1 -- Case Law Retrieval}
\dcm{Format}{Text generation (case identification)}
\dcm{Metric}{Micro F1}
\dcm{Source}{\EXT{} -- COLIEE Competition~\citep{coliee,goebel2024coliee}}
\dcm{Items}{499}
\dcd{Given a query case from the Federal Court of Canada, identify the one or
two cases in a corpus likely to be cited by it. Selection must weigh similar
legal principles, related subject matter, comparable fact patterns, consistent
reasoning and precedential value. All references in the query case are
deliberately redacted, so the model must rely on legal content and reasoning
rather than explicit citation signals. Run with chain-of-thought prompting.}
\dce{Given the text of a query case and a corpus of Federal Court of Canada
cases, output the case numbers of the noticed cases with no explanation, e.g.
\texttt{099814, 099603}.}
\end{datacard}


\begin{datacard}{AALP Quality}
\dcm{Format}{Free response (retrieval and synthesis)}
\dcm{Metric}{ROUGE-1 F1, ROUGE-L F1}
\dcm{Source}{\INT{} -- custom task designed with SMEs}
\dcm{Items}{220}
\dcd{Answer legal questions by retrieving and synthesising information across
several supplied legal documents -- practice notes, standard documents and
procedural guides. The task requires locating relevant passages across multiple
sources, cross-referencing them and producing an accurate, well-supported
answer with citations. Responses are compared against gold reference answers
using ROUGE.}
\dce{Given excerpts from three New Jersey practice documents on subpoenas,
depositions and discovery, answer whether an expert witness must be subpoenaed
to appear for a deposition in New Jersey -- synthesising that a subpoena is
required except in CBLP cases, with citations to the governing rules.}
\end{datacard}

\begin{datacard}{Best Headnote}
\dcm{Format}{Multiple choice (4-way classification)}
\dcm{Metric}{Macro F1}
\dcm{Source}{\INT{} -- custom task designed with SMEs}
\dcm{Items}{1{,}072}
\dcd{Assess how relevant a case headnote is to a legal search query, across
four grades: directly on point; generally responsive but possibly missing a key
concept; matching terms but unresponsive; and fully unresponsive. The task
forces a distinction between surface term overlap and substantive legal
relevance -- precisely the judgement that determines whether a practitioner
finds the applicable case law.}
\dce{Given the query ``Christopher Martinez expert'' and a headnote about
patent claims on gateway message transmission, classify relevance as fully
unresponsive: the headnote concerns patent technology, not expert testimony.}
\end{datacard}

\subsection{Legal Reasoning}
\label{app:dc:legalreas}

\begin{datacard}{COLIEE Task 2 -- Legal Case Entailment}
\dcm{Format}{Text classification (paragraph identification)}
\dcm{Metric}{Micro F1}
\dcm{Source}{\EXT{} -- COLIEE Competition~\citep{coliee,goebel2024coliee}}
\dcm{Items}{816}
\dcd{Requires identifying which paragraphs from a noticed case entail a given
judicial decision for an unseen new case. Tests whether a model understands
precedent relationships well enough to isolate the specific passages of
existing case law that support a new decision, rather than merely retrieving
topically similar text. Run with chain-of-thought prompting.}
\dce{Given a new case decision and a set of paragraphs drawn from a precedent
case, identify which paragraphs logically support the new decision.}
\end{datacard}


\begin{datacard}{CaseHOLD}
\dcm{Format}{Multiple choice (5-way)}
\dcm{Metric}{Accuracy}
\dcm{Source}{\EXT{} -- Zheng et al. (2021)~\citep{zheng2021casehold}}
\dcm{Items}{1{,}000 (sampled from 53{,}000+)}
\dcd{Tests a fundamental lawyer skill: identifying the relevant holding of a
cited case. Each item presents a citing context from a judicial decision with
five candidate holding statements. The four distractors are genuine holdings
drawn from other cases, so the task cannot be solved by surface pattern
matching -- the model must relate the citation to its correct holding.}
\dce{Given a citing context from a judicial decision, select the correct
holding statement from five options (one correct, four plausible distractors
taken from other cases).}
\end{datacard}

\begin{datacard}{Function of Decision Section}
\dcm{Format}{Multiple choice (7-way classification)}
\dcm{Metric}{Accuracy}
\dcm{Source}{\EXT{} -- LegalBench~\citep{guha2023legalbench}}
\dcm{Items}{367}
\dcd{Classifies paragraphs from U.S. Courts of Appeals decisions into seven
functional categories: Facts, Procedural History, Issue, Rule, Analysis,
Conclusion and Decree. Distinguishing the essential from the incidental parts
of a decision is a prerequisite for more complex legal analysis, so this serves
as a basic legal-comprehension probe.}
\dce{Given a paragraph from a judicial opinion, classify it as one of Facts,
Procedural History, Issue, Rule, Analysis, Conclusion or Decree.}
\end{datacard}

\begin{datacard}{Learned Hands}
\dcm{Format}{Binary classification (yes/no per category)}
\dcm{Metric}{Accuracy, averaged across the 16 categories}
\dcm{Source}{\EXT{} -- Stanford Legal Design Lab \& Suffolk LIT Lab~\citep{learnedhands,guha2023legalbench}}
\dcm{Items}{11{,}109 across 16 categories}
\dcd{Evaluates legal issue-spotting in lay narratives. Posts from the
\textit{r/legaladvice} subreddit were labelled by law students and lawyers
through a crowdsourced game. The 16 categories are Benefits (66), Business
(174), Consumer (614), Courts (192), Crime (688), Divorce (150), Domestic
Violence (174), Education (56), Employment (710), Estates (178), Family
(2{,}265), Health (226), Housing (4{,}494), Immigration (134), Torts (432) and
Traffic (556). Issue-spotting of this kind underpins triage of members of the
public to the right legal service.}
\dce{Given an informal narrative describing a person's situation, determine
whether it relates to a given legal category, e.g., ``Does this post discuss
housing issues?''}
\end{datacard}


\begin{datacard}{Legal Support}
\dcm{Format}{Binary multiple choice}
\dcm{Metric}{Accuracy (balanced)}
\dcm{Source}{\EXT{} -- HELM~\citep{liang2022helm}}
\dcm{Items}{1{,}000 (abbreviated); 20{,}034 (full)}
\dcd{Presents a legal passage and two candidate supporting conclusions, and
asks which more forcefully and directly supports the claim in the passage. The
data is deliberately noisy -- labels are not fully accurate -- so the task
also measures whether nuanced legal reasoning survives label inconsistency.}
\dce{Given a legal passage and two case-summary options, determine which
summary better supports the legal claim made in the passage.}
\end{datacard}

\begin{datacard}{MBE Bar Exam}
\dcm{Format}{Multiple choice (4-way)}
\dcm{Metric}{Accuracy}
\dcm{Source}{\EXT{} -- National Conference of Bar Examiners~\citep{ncbe}}
\dcm{Items}{608}
\dcd{The Multistate Bar Examination is a 200-question multiple-choice
examination covering constitutional law, real property, criminal law and
procedure, civil procedure, evidence, contracts and torts. It is designed to
assess legal reasoning and analysis rather than recall, and counts for half the
total score in Uniform Bar Examination jurisdictions. Run with
chain-of-thought prompting.}
\dce{Given a legal fact pattern and question, select the correct answer from
four options by applying the governing legal principles.}
\end{datacard}

\begin{datacard}{Parentheticals}
\dcm{Format}{Multiple choice (3-way classification)}
\dcm{Metric}{Macro F1 (maximum over the two prompt formats)}
\dcm{Source}{\INT{} -- custom task designed with SMEs}
\dcm{Items}{308}
\dcd{Determines whether a cited passage directly supports, indirectly supports
or contradicts the original passage from a judicial opinion. Ground truth comes
from the introductory signal preceding the citation -- ``see'' indicates
support, ``contra'' indicates contradiction. The distribution is near-balanced
(110 direct, 110 indirect, 88 contradiction). Unlike similar tasks it cannot be
solved by textual similarity; \texttt{GPT-4} reached only 0.41 macro F1 against a random
baseline of about 0.3.}
\dce{Given a main passage from a judicial opinion and a cited passage,
classify the relationship as direct support, indirect support or
contradiction.}
\end{datacard}

\begin{datacard}{ReClor}
\dcm{Format}{Multiple choice (4-way)}
\dcm{Metric}{Accuracy}
\dcm{Source}{\EXT{} -- Yu et al. (2020)~\citep{yu2020reclor}}
\dcm{Items}{1{,}000 (evaluation subset of 6{,}138)}
\dcd{Logical reasoning questions drawn from graduate admission examinations
such as the GMAT and LSAT. Items require identifying flawed reasoning patterns,
recognising parallel argument structures and evaluating logical validity --
the analytical operations legal practitioners perform on arguments in ordinary
language. State-of-the-art models continue to struggle on this set. Run with
chain-of-thought prompting.}
\dce{Given a passage containing an argument and a question about its logic
(for instance, identifying the flaw in the reasoning), select the correct
answer from four options.}
\end{datacard}

\begin{datacard}{SCALR}
\dcm{Format}{Multiple choice (5-way classification)}
\dcm{Metric}{Accuracy (balanced)}
\dcm{Source}{\EXT{} -- LegalBench / SCALR (CC BY 4.0)~\citep{guha2023legalbench}}
\dcm{Items}{571}
\dcd{Supreme Court Assessment of Legal Reasoning pairs a ``question presented
for review'' from a Supreme Court case with five candidate holding statements.
It is built to measure comprehension of legal language rather than memorised
legal knowledge: items are restricted to post-2001 cases with a single question
granted for review, and distractors are curated by TF-IDF similarity to keep
them plausible.}
\dce{Given a Supreme Court question presented for review, select the holding
statement that best corresponds to the Court's decision from five options.}
\end{datacard}

\begin{datacard}{SuperGPQA (Law)}
\dcm{Format}{Multiple choice (10-way, A--J)}
\dcm{Metric}{Accuracy}
\dcm{Source}{\EXT{} -- SuperGPQA~\citep{map2025supergpqa}}
\dcm{Items}{656}
\dcd{SuperGPQA is a graduate-level question-answering benchmark spanning 285
disciplines; we use the 656 items from the law discipline. Questions combine
factual recall with legal reasoning at advanced academic level and offer ten
answer choices, making chance performance low and the task correspondingly
demanding. Run with chain-of-thought prompting.}
\dce{Given a graduate-level law question covering areas such as constitutional
law, contracts or torts, select the correct answer from ten options (A--J).}
\end{datacard}

\begin{datacard}{LEXam -- MCQ (4-choice)}
\dcm{Format}{Multiple choice (4-way, A--D)}
\dcm{Metric}{Accuracy}
\dcm{Source}{\EXT{} -- LEXam Benchmark~\citep{fan2025lexam}}
\dcm{Items}{1{,}655 (619 English / 1{,}036 German)}
\dcd{The four-choice subset of LEXam's Swiss law examination questions. The
model reasons step by step as a Swiss legal expert -- clarifying the facts,
identifying the issue, stating the rule, applying it and eliminating incorrect
options -- before committing to an answer letter. Of the 1{,}655 items 1{,}515
are Swiss-jurisdiction and 140 International; overall, per-language and
Swiss-only accuracy are reported.}
\dce{Given a Swiss-law exam question and four labelled choices, reason through
the issue and return the answer as \texttt{Correct Answer: \#\#\#C\#\#\#}.}
\end{datacard}


\begin{datacard}{PRBench Legal Hard}
\dcm{Format}{Open-ended generation (multi-turn professional reasoning)}
\dcm{Metric}{Mean clipped score (rubric-based LLM judge)}
\dcm{Judge}{\texttt{o4-mini}, using the official PRBench grader template}
\dcm{Source}{\EXT{} -- Scale AI, PRBench~\citep{akyurek2025prbench}}
\dcm{Items}{250}
\dcd{The Legal-250 hard subset of PRBench, isolating the most difficult
legal-domain tasks from the full 500-task legal set. Realistic legal-domain conversations across 12 topics such as labour and
employment, contracts and compliance; most are single-turn but some run to ten
turns. Each task carries 10--30 expert-authored rubric criteria graded
independently true or false by a judge model. Criteria carry signed weights:
important tiers award points when met, detrimental tiers deduct points when an
undesirable behaviour appears. The per-sample clipped score is the weighted sum
divided by the maximum achievable positive weight, floored at zero.}
\dce{A multi-turn conversation drawn from the hardest tier of PRBench's legal
tasks; the final response is graded against expert rubric criteria for legal
accuracy and completeness.}
\end{datacard}

\subsection{Legal Classification}
\label{app:dc:legalcls}


\begin{datacard}{EUR-Lex}
\dcm{Format}{Multi-label classification (100+ topic categories)}
\dcm{Metric}{Micro F1}
\dcm{Source}{\EXT{} -- EUR-Lex~\citep{chalkidis2021multieurlex,chalkidis2022lexglue}}
\dcm{Items}{500}
\dcd{Classify the concepts and topics present in European legislative
documents -- laws, regulations and directives -- against more than 100
categories spanning political framework, economic policy, social affairs,
health, environment, transport and agriculture. Multi-label output reflects the
reality that legislation routinely addresses several interconnected policy
areas at once.}
\dce{Given an excerpt from a Commission Regulation on healthcare expenditure
and financing statistics, identify all applicable topics from options including
economic analysis, health, information technology and data processing, social
protection, and public finance and budget policy.}
\end{datacard}

\begin{datacard}{LEDGAR}
\dcm{Format}{Multi-label classification (780 label categories)}
\dcm{Metric}{Micro F1}
\dcm{Source}{\EXT{} -- Tuggener et al. (2020)~\citep{tuggener2020ledgar,chalkidis2022lexglue}}
\dcm{Items}{10{,}214}
\dcd{Classify contract provisions extracted from U.S. Securities and Exchange
Commission filings in the EDGAR database. Models assign section titles from 780
label categories, up to eight per provision, covering standard contractual
categories such as Governing Laws, Indemnifications, Confidentiality and
Terminations. The label distribution is heavily skewed with a long tail of rare
labels, which makes it a realistic rather than balanced benchmark.}
\dce{Given a provision describing conditions that will not result from
executing the transaction documents -- conflicts with organisational documents
or applicable law -- identify section titles from options such as Consents, No
Conflicts, Compliance With Laws and Applicable Laws.}
\end{datacard}

\begin{datacard}{SCOTUS}
\dcm{Format}{Multi-label classification (13 topic categories)}
\dcm{Metric}{Micro F1}
\dcm{Source}{\EXT{} -- Supreme Court Database~\citep{spaeth2020scdb,chalkidis2022lexglue}}
\dcm{Items}{1{,}000}
\dcd{Classify the legal issue types in U.S. Supreme Court cases from full
opinions, across 13 categories: Criminal Procedure, Civil Rights, First
Amendment, Due Process, Privacy, Attorneys, Unions, Economic Activity, Judicial
Power, Federalism, Interstate Relations, Federal Taxation and Miscellaneous.
The task requires identifying the central issues at stake and separating
adjacent areas of constitutional and statutory law.}
\dce{Given an opinion on whether backpay awards settling Title VII claims are
excludable from gross income, identify the relevant topic -- here, Federal
Taxation.}
\end{datacard}

\begin{datacard}{Headnote Type}
\dcm{Format}{Multiple choice (11-way classification)}
\dcm{Metric}{Accuracy}
\dcm{Source}{\INT{} -- custom task designed with SMEs}
\dcm{Items}{880}
\dcd{Classify legal headnotes -- brief summaries of points of law from
judicial opinions -- into the practice area that best describes them, from
Securities Law, Employment Law, Antitrust, Commercial Law, Real Estate, Class
Actions, Remedies, Civil Procedure, Discovery and Evidence, Insurance and
Arbitration. Some headnotes are genuine boundary cases; the task measures
whether the single most appropriate area is identified. This underpins legal
research, document organisation and issue-spotting.}
\dce{Given a headnote alleging that health insurers conspired to manipulate
usual, customary and reasonable rates for out-of-network reimbursement under
the Sherman Act, classify the practice area as Antitrust.}
\end{datacard}

\subsection{Document Processing and Retrieval-Augmented Generation}
\label{app:dc:aar}

\begin{datacard}{Legal RAG}
\dcm{Format}{Free response with citations}
\dcm{Metric}{GPA, precision, recall, F2}
\dcm{Judge}{\texttt{Claude Sonnet 4.6}}
\dcm{Source}{\INT{} -- custom task designed with SMEs}
\dcm{Items}{224}
\dcd{Answer legal questions using supplied primary law sources -- cases,
statutes or regulations -- in the style of a practising attorney. Responses
must be accurate, relevant and clear, cite paragraph numbers in square-bracket
notation, and mark any information not present in the documents with explicit
free-form citation tokens. Reference-based LLM judging reports GPA for overall
quality alongside precision, recall and F2, the latter weighting recall of
relevant material.}
\dce{Given documents delimited by \texttt{[START\_DOCUMENT]} and
\texttt{[END\_DOCUMENT]} with numbered paragraphs, answer a legal question,
citing sources as \texttt{[27]} or \texttt{[16]-[19]} and using the shortened
range notation where applicable.}
\end{datacard}

\begin{datacard}{Document Review}
\dcm{Format}{Free response over supplied documents}
\dcm{Metric}{LLM-as-judge (reference-based)}
\dcm{Judge}{\texttt{Claude Sonnet 4.5}}
\dcm{Source}{\INT{}}
\dcm{Items}{392 queries across 7{,}324 documents}
\dcd{Answer user queries against a supplied document set, with an LLM judge
comparing responses to gold answers. Evaluation runs as two stages -- selecting a subset of the most relevant documents and chunks over
the document set, followed by answer generation.}
\dce{``If the customer is acquired through a merger, can the agreement be
assigned to the acquiring entity without obtaining the provider's prior written
consent?''}
\end{datacard}

\subsection{Legal Summarisation}
\label{app:dc:legalsumm}

\begin{datacard}{BillSum US}
\dcm{Format}{LLM-as-judge (reference-based)}
\dcm{Metric}{Accuracy, completeness, clarity, conciseness, hallucination}
\dcm{Judge}{\texttt{Claude Sonnet 4.6}}
\dcm{Source}{\EXT{} -- BillSum (US test split)~\citep{kornilova2019billsum}}
\dcm{Items}{3{,}269}
\dcd{Generate concise summaries of U.S. Congressional bills. Given the full
text of a bill, the model must capture its key provisions, purposes and
implications. Quality is assessed by reference-based LLM judging across five
dimensions, including explicit hallucination detection.}
\dce{``The following is a US bill that we want to summarise'' followed by the
bill text and the instruction to summarise it.}
\end{datacard}

\begin{datacard}{BillSum CA}
\dcm{Format}{LLM-as-judge (reference-based)}
\dcm{Metric}{Accuracy, completeness, clarity, conciseness, hallucination}
\dcm{Judge}{\texttt{Claude Sonnet 4.6} (temperature 0)}
\dcm{Source}{\EXT{} -- BillSum (California test split)~\citep{kornilova2019billsum}}
\dcm{Items}{1{,}237}
\dcd{The California state counterpart to BillSum US. Given the full text of a
California bill, the model must produce a summary capturing the key provisions,
purposes and implications of the state legislation, scored on the same five
judged dimensions.}
\dce{``The following is a California bill that we want to summarise''
followed by the bill text and the instruction to summarise it.}
\end{datacard}

\begin{datacard}{CourtWire}
\dcm{Format}{LLM-as-judge (reference-based)}
\dcm{Metric}{Accuracy, completeness, clarity, conciseness, hallucination}
\dcm{Judge}{\texttt{Claude Sonnet 4.6}}
\dcm{Source}{\INT{} -- custom task designed with SMEs}
\dcm{Items}{500}
\dcd{Analyse a U.S. civil case complaint and state the core reason for the
litigation in a single sentence. The task requires identifying the central
claims or causes of action and compressing them into one clear statement while
preserving correct party references. It tests extreme-compression
summarisation, where every omission is visible.}
\dce{Given a civil complaint, ``in one sentence, provide the reasons that this
legal complaint was filed. Refer to the parties as plaintiff and defendant.''}
\end{datacard}

\begin{datacard}{Material Facts}
\dcm{Format}{LLM-as-judge (reference-based)}
\dcm{Metric}{Accuracy, completeness, clarity, conciseness, hallucination}
\dcm{Judge}{\texttt{Claude Sonnet 4.6}}
\dcm{Source}{\INT{} -- custom task designed with SMEs}
\dcm{Items}{495}
\dcd{Extract the material facts a court relied on to decide an issue, given a
headnote and a legal issue question. The model must separate material facts
from background facts, procedural history and legal conclusions, returning up
to five concise statements. The discrimination required -- what the court
actually relied on, versus what merely appears in the opinion -- is the
substance of the task.}
\dce{Given a headnote on whether a confidentiality clause in a group life
insurance policy's arbitration provision was substantively unconscionable, and
the corresponding issue question, list up to five material facts the court
relied on.}
\end{datacard}

\subsection{Contract Understanding}
\label{app:dc:contract}

\begin{datacard}{CUAD}
\dcm{Format}{Free response (binary yes/no classification)}
\dcm{Metric}{Macro F1}
\dcm{Source}{\EXT{} -- Hendrycks et al. (2021), via LegalBench~\citep{hendrycks2021cuad,guha2023legalbench}}
\dcm{Items}{3{,}598 (random sample)}
\dcd{The Contract Understanding Atticus Dataset comprises 510 commercial
contracts from SEC EDGAR filings, annotated by lawyers for 41 clause categories
relevant to contract review. LegalBench restructures each category as a
standalone binary question with negatives drawn from other clause types; this
evaluation pools 38 such subtasks. Because pooled examples are overwhelmingly
negative, the free-form answer is reduced to a yes/no token and scored with
macro F1, so a majority-class strategy gains nothing.}
\dce{``Does the clause describe a license grant to a licensee and the
affiliates of such licensee?'' followed by a sublicensing clause; return only
yes or no.}
\end{datacard}

\begin{datacard}{MAUD}
\dcm{Format}{Free response (multiple choice)}
\dcm{Metric}{Accuracy}
\dcm{Source}{\EXT{} -- Wang et al. (2023), via LegalBench~\citep{wang2023maud,guha2023legalbench}}
\dcm{Items}{2{,}619 (random sample)}
\dcd{The Merger Agreement Understanding Dataset is built from 152 public
merger agreements on SEC EDGAR, annotated against the American Bar
Association's 2021 Public Target Deal Points Study. Each of 34 pooled subtasks
asks about a specific deal point -- the standard for a Change of
Recommendation, the definition of Material Adverse Effect, tail period length,
availability of specific performance -- with its own fixed answer set rather
than a shared binary label, so plain accuracy is the appropriate metric.}
\dce{``Is the ability to consummate concept subject to Material Adverse Effect
carveouts?'' with options No and Yes, over a segment of the merger agreement.}
\end{datacard}

\begin{datacard}{OPP-115}
\dcm{Format}{Free response (binary yes/no classification)}
\dcm{Metric}{Macro F1}
\dcm{Source}{\EXT{} -- Wilson et al. (2016), via LegalBench~\citep{wilson2016opp115,guha2023legalbench}}
\dcm{Items}{888 (random sample)}
\dcd{115 website privacy policies split into 3{,}792 segments and annotated by
law students against ten data-practice categories including First Party
Collection/Use, Third Party Sharing/Collection, Data Retention and User
Choice/Control. LegalBench reformulates nine categories as binary questions
over individual segments; this evaluation pools them and, as with CUAD, scores
macro F1 over the yes/no classes to correct for label imbalance.}
\dce{``Does the clause describe how long user information is stored?'' over a
clause stating that information received from a social network is stored and
used in accordance with the privacy policy.}
\end{datacard}

\begin{datacard}{Privacy Policy Entailment}
\dcm{Format}{Free response (binary yes/no classification)}
\dcm{Metric}{Accuracy}
\dcm{Source}{\EXT{} -- APP-350 corpus, via LegalBench~\citep{zimmeck2019app350,guha2023legalbench}}
\dcm{Items}{500 (random sample)}
\dcd{Pairs a clause from a mobile application privacy policy with a short
description of a specific data practice and asks whether the clause entails
that the practice occurs. Unlike CUAD and OPP-115 this is a single task rather
than a family of clause-type subtasks, so it is scored with accuracy rather
than macro F1. The free-form response is reduced to a yes/no token before
comparison.}
\dce{Given a clause about providing audio and video snippets to third-party
providers, and the description ``the policy describes collection of the user's
IP address by ad networks, analytics services or other third parties'',
determine whether the description matches the clause.}
\end{datacard}

\begin{datacard}{Insurance Policy Interpretation}
\dcm{Format}{Free response (3-way classification)}
\dcm{Metric}{Accuracy}
\dcm{Source}{\EXT{} -- LegalBench (Guha et al., 2023)~\citep{waldon2023consensus,guha2023legalbench}}
\dcm{Items}{138}
\dcd{Presents an insurance policy and a claim and asks whether the policy
covers the claim. Ground truth derives from crowdsourced legal-interpretation
judgements: workers voted on coverage, and pairs with high disagreement are
labelled ambiguous. The three options are yes, no and ``it's ambiguous'',
making this one of the few benchmarks whose gold label reflects a distribution
of lay legal interpretation rather than a single expert determination.}
\dce{A policy covering ``losses from missed employment due to injuries that
occur under regular working conditions'' and a claim from a repair technician
injured falling from a ladder; decide whether the claim is covered.}
\end{datacard}

\begin{datacard}{ContractScrub}
\dcm{Format}{Free response (structured JSON extraction over a full contract)}
\dcm{Metric}{Macro-average recall across nine categories; precision and F1 also
reported}
\dcm{Source}{\INT{} -- custom task designed with SMEs~\citep{bang2026contractscrub}}
\dcm{Items}{3{,}014 across 44 contracts}
\dcd{Evaluates the ``scrubbing'' pass transactional lawyers perform before
execution: a final sweep of an agreement for residual drafting errors. Given a
contract and one category, the model returns tuples of \textit{(term,
location)} -- or paired locations for relational categories -- scored as a
multiset against a gold annotation, with repeated occurrences counted
separately. The nine categories cover defined-term extraction (1{,}505 items)
and eight error types, including Undefined Capitalized Terms (689),
Uncapitalized Defined Terms (317) and Incorrect Party References (130). Source
agreements come from CUAD but are re-annotated from scratch by nine lawyers
with 8+ years in practice, who record existing defects and insert further ones
to balance coverage. Recall is primary because a flagged issue is cheap to
dismiss and a missed one is not. Unlike CUAD and MAUD, the task is open-ended
and document-internal: correctness is fixed by the conventions the agreement
sets for itself, not by external legal standards.}
\dce{Given a commercial agreement and the instruction to identify incorrect
cross-references, return JSON only, e.g.\
\texttt{\{"wrong": "9.1", "correct": "10.1", "location": "7ai"\}}.}
\end{datacard}

\subsection{Human Queries (Legal)}
\label{app:dc:diversequeries}

\begin{datacard}{Diverse Queries -- Documents (Groundedness)}
\dcm{Format}{Open-ended generation over attached documents}
\dcm{Metric}{LLM judge -- claim-level grounding against source documents}
\dcm{Judge}{\texttt{GLM-5.2}}
\dcm{Source}{\INT{} -- SME-authored queries (Diverse Queries GRPO dataset)}
\dcm{Items}{300}
\dcd{Open-ended answers to expert-authored legal queries that carry attached
source documents. Scoring measures faithfulness of the claims in the answer
against those documents, so this is the groundedness-focused subset of the
suite. Because embedded document text makes these the longest contexts in the
benchmark, the evaluation runs at batch size one.}
\dce{A legal query accompanied by the full text of the attached documents the
answer must remain faithful to; the judge decomposes the response into claims
and checks each against the source.}
\end{datacard}

\begin{datacard}{Diverse Queries -- General Legal}
\dcm{Format}{Open-ended generation}
\dcm{Metric}{LLM judge -- doctrinal and fact-driven reasoning rubric}
\dcm{Judge}{\texttt{GLM-5.2}}
\dcm{Source}{\INT{} -- SME-authored queries}
\dcm{Items}{400}
\dcd{The largest subset, covering general legal questions scored against a
reasoning rubric that rewards correct doctrine and appropriate use of the facts
given. Query metadata -- query type, class and subclass, jurisdictions and
practice areas -- is carried per example so the judge can be routed and the
results sliced.}
\dce{An expert-authored legal question requiring doctrinal analysis applied to
a supplied fact pattern, graded against a legal-reasoning rubric.}
\end{datacard}

\begin{datacard}{Diverse Queries -- Drafting}
\dcm{Format}{Open-ended generation (document drafting)}
\dcm{Metric}{LLM judge -- drafted-document rubric with document-type gate}
\dcm{Judge}{\texttt{GLM-5.2} via Mariner-RS, routed by subset}
\dcm{Source}{\INT{} -- SME-authored queries}
\dcm{Items}{100}
\dcd{Requests that the model draft a legal document. Scoring applies a
drafting rubric together with a type gate that checks the output is in fact the
requested kind of document, so a well-written response of the wrong type does
not score well.}
\dce{An instruction to draft a specific legal document, graded on both
conformance to the requested document type and the quality of the drafting.}
\end{datacard}

\begin{datacard}{Diverse Queries -- Transactional}
\dcm{Format}{Open-ended generation (transactional work product)}
\dcm{Metric}{LLM judge -- deal and commercial work-product rubric}
\dcm{Judge}{\texttt{GLM-5.2}}
\dcm{Source}{\INT{} -- SME-authored queries}
\dcm{Items}{100}
\dcd{Covers deal and commercial work product, scored against a rubric specific
to transactional practice rather than general legal reasoning. Separating this
from the general legal subset isolates commercial drafting and advisory
capability from doctrinal analysis.}
\dce{A transactional or commercial request, such as advice or work product
relating to a deal, graded against a transactional practice rubric.}
\end{datacard}

\begin{datacard}{Diverse Queries -- Instruction Following}
\dcm{Format}{Open-ended generation under format constraints}
\dcm{Metric}{LLM judge -- per-item extracted format constraints}
\dcm{Judge}{\texttt{GLM-5.2}}
\dcm{Source}{\INT{} -- SME-authored queries}
\dcm{Items}{100}
\dcd{Isolates instruction following in a legal setting. Format constraints are
extracted per item and the judge checks compliance with those specific
constraints, so the score reflects adherence to the instruction rather than
legal quality. Named \textit{instruct\_only} in the configuration and renamed
to instruction following in reporting, to describe what it actually measures.}
\dce{A legal query carrying explicit format requirements -- length, structure
or output shape -- where scoring turns on whether those requirements are
met.}
\end{datacard}

\subsection{Deep Research (Legal)}
\label{app:dc:deepresearch}

\begin{datacard}{Deep Research}
\dcm{Format}{Free-text reports produced by acting within an agentic harness}
\dcm{Metric}{LLM-as-judge, with reference rubrics (factuality, completeness)
and without (understandability, conciseness, coherence, relevance)}
\dcm{Source}{\INT{}}
\dcm{Items}{52 reports}
\dcd{Reports are generated with a custom multi-agent harness in which a planner
directs research sub-agents that hold access to internal retrieval tools before
a final extensive report is drafted. Scoring runs over SME-authored queries
and rubrics spanning areas of US and UK law. The overall score averages seven
dimensions on a 0--1 scale, of which factuality, completeness and conciseness
are the most important and the most variable between models. Factuality follows
the Claimify approach~\citep{metropolitansky2025claimify}: atomic claims are
extracted from the report and matched
against the content of cited sources, scoring the proportion verifiable by
citation. Completeness scores against gold SME rubrics enumerating the required
and helpful elements of a good answer, taking the proportion of the rubric
addressed. Conciseness uses a judge with a 0--5 rubric, rescaled to 0--1.}
\dce{``In Scotland, Company A contracts with Company B in 2015. From 2019
Company A misses an obligation, unnoticed by Company B though a thorough audit
might have caught it. Company B makes losses that year without identifying the
cause, and further losses by 2025, when it traces them back to the 2019 breach.
It raises proceedings; Company A argues the claim is time barred. What are the
arguments for and against that defence, and how is a court likely to rule?''}
\end{datacard}

\subsection{Tax}
\label{app:dc:tax}

\begin{datacard}{Tax Q\&A}
\dcm{Format}{Free response with citations}
\dcm{Metric}{Correctness, coverage, consistency, relevance (binary each);
groundedness (0.0--1.0)}
\dcm{Judge}{\texttt{GPT-5.1}, one judge per scored axis}
\dcm{Source}{\INT{} -- SME-curated, merged from two internal sources}
\dcm{Items}{115}
\dcd{Answer tax questions from supplied source documents, citing support in
square-bracket notation. Each data point is
independently vetted by an expert. Five independent judges score the answer:
correctness against the gold answer, coverage of its key points, logical
consistency with it, relevance (the inverse of an irrelevance judge), and
groundedness -- the answer is decomposed into atomic statements and each is
checked against the sources, scoring the proportion supported.}
\dce{``Is a taxpayer's home office deduction affected if they also use the
space occasionally for personal purposes?'' with source paragraphs on the
exclusive-use requirement; the gold answer states the rule and its exceptions
with paragraph citations.}
\end{datacard}

\subsection{Factuality}
\label{app:dc:factuality}

\begin{datacard}{SimpleQA Verified}
\dcm{Format}{Free response (short-form factual QA)}
\dcm{Metric}{F1}
\dcm{Source}{\EXT{} -- refined from OpenAI SimpleQA~\citep{haas2025simpleqaverified,wei2024simpleqa}}
\dcm{Items}{1{,}000}
\dcd{A cleaned version of SimpleQA addressing noisy labels, topical bias and
question redundancy. It probes short-form factual recall from parametric
knowledge alone, with no tools or context, over questions with single
indisputable answers spanning science, politics, art, geography, sports, music
and history -- often requiring tail knowledge. Responses are graded correct,
incorrect or not attempted, and F1 balances attempting everything against
answering only when confident, so abstention is neither free nor fatal.}
\dce{A question of the form ``Who did X in 2010?'' about a specific historical
event, where the model must answer correctly or explicitly state uncertainty
rather than hallucinate.}
\end{datacard}

\begin{datacard}{FACTS Grounding}
\dcm{Format}{Free response (long-form grounded generation)}
\dcm{Metric}{Final factuality score, aggregated over three LLM judges}
\dcm{Judge}{\texttt{Gemini 2.5 Pro}, \texttt{GPT-4.1} and \texttt{Claude Sonnet 4.6}}
\dcm{Source}{\EXT{} -- FACTS~\citep{jacovi2025facts}}
\dcm{Items}{860}
\dcd{Tests whether long-form responses stay fully grounded in supplied context
of up to 32{,}000 tokens, drawn from medical, legal, financial, retail and
technology domains. Tasks span fact-finding (31.6\%), find-and-summarise
(29.7\%), effect analysis (8.9\%), explanation (7.5\%), comparison (6.1\%),
pros and cons (4.4\%) and summarisation (3.8\%). Scoring runs in two phases: an
eligibility filter disqualifies responses that fail to address the request,
preventing gaming by minimal answers, then grounding is assessed as a binary
judgement per judge. Eligibility is by consensus -- any judge may approve.}
\dce{Given a legal document on medical marijuana appropriations rider
interpretations, answer ``What did the first circuit conclude?'' using only the
document; scoring checks both that the specific conclusion was identified and
that every legal statement traces to the context.}
\end{datacard}

\begin{datacard}{FaithEval}
\dcm{Format}{Free response (two subtasks)}
\dcm{Metric}{Accuracy (task-specific scoring)}
\dcm{Source}{\EXT{} -- Ming et al. (2024)~\citep{ming2025faitheval}}
\dcm{Items}{3{,}900 (2{,}400 unanswerable + 1{,}500 inconsistent)}
\dcd{Tests faithfulness to context when that context conflicts with
parametric knowledge or is incomplete. In the \textit{unanswerable} subtask the
context is relevant but lacks the specific detail needed, and the model must
answer ``unknown'' rather than guess or fall back on prior knowledge. In the
\textit{inconsistent} subtask documents contradict each other and the model
must flag the conflict rather than arbitrarily pick a side. Both map directly
onto legal practice: abstaining when case materials are incomplete, and
detecting contradictions in noisy retrieval.}
\dce{Context gives 2009 data for solo driving but only 2015 data for
carpooling, then asks which group was larger in 2009 -- a faithful model
answers unknown. Or two documents disagree on a character's name, and the model
must report the conflict.}
\end{datacard}

\begin{datacard}{CoGenSumm}
\dcm{Format}{Free response (binary yes/no classification)}
\dcm{Metric}{Accuracy}
\dcm{Source}{\EXT{} -- Falke et al. (2019)~\citep{falke2019cogensumm,laban2022summac}}
\dcm{Items}{400 (SummaC test split)}
\dcd{Detect factual inconsistencies introduced by summarisation. Built from
CNN/DailyMail articles paired with generated summary sentences, the task
supplies the source document and a candidate summary and asks whether the
summary is directly supported and therefore factually consistent. The yes/no
judgement is compared against the gold label.}
\dce{Given an article and a candidate summary sentence produced by a
summarisation system, answer whether the statement is directly supported by the
article.}
\end{datacard}

\begin{datacard}{XSumFaith}
\dcm{Format}{Free response (binary yes/no classification)}
\dcm{Metric}{Accuracy}
\dcm{Source}{\EXT{} -- Maynez et al. (2020)~\citep{maynez2020xsumfaith,laban2022summac}}
\dcm{Items}{1{,}250 (SummaC test split)}
\dcd{Detect hallucination in highly abstractive summaries generated from the
XSum dataset. Given the source article and a candidate summary, the model
answers whether the summary is faithful or hallucinated. Because XSum summaries
are far more abstractive than CNN/DailyMail, surface overlap is a weaker cue
than in CoGenSumm and the judgement must rest on entailment.}
\dce{Given an XSum article and a single-sentence abstractive summary, answer
whether the statement is directly supported by the article.}
\end{datacard}

\subsection{Robustness}
\label{app:dc:safety}

\begin{datacard}{Principal Hierarchy -- Execution (Legal)}
\dcm{Format}{Free response (execution scenario), LLM-judged}
\dcm{Metric}{Action correct rate (LLM judge)}
\dcm{Judge}{\texttt{Qwen3-235B-A22B-Instruct-2507}}
\dcm{Source}{\INT{} -- custom task on refusal and redirection when instructed
to draft with an overruled case~\citep{yu2026principal}}
\dcm{Items}{1{,}616}
\dcd{Uses the same 401 overruled-case scenarios as the legal advisory task,
but the user issues a direct drafting instruction rather than asking for
advice, raising the pressure to comply. The judge assesses whether the
reasoning identifies the case as overruled and whether the action either
refuses to draft the requested passage or drafts text explaining why the case
cannot be used, instead of producing the improper argument.}
\dce{``Can you summarise the holding and draft a paragraph I can drop into my
argument section with proper Bluebook citations?'' -- the model should decline
or explain that the case is overruled.}
\end{datacard}

\begin{datacard}{Query Sufficiency}
\dcm{Format}{Free response (under-specified legal query), LLM-judged}
\dcm{Metric}{Element identification F2 (recall-weighted, LLM judge)}
\dcm{Judge}{\texttt{Qwen3-235B-A22B-Instruct-2507} }
\dcm{Source}{\INT{} -- custom task designed with SMEs~\citep{insuffiencybench2026}}
\dcm{Items}{144}
\dcd{Measures whether a model recognises which facts are missing from a
deliberately under-specified legal query before attempting an answer. Items are
realistic fact patterns across criminal law (28), tort (24), contracts (18),
employment (18), commercial (18), real property (16) and defamation (13), at
moderate (64) or difficult (80) difficulty, each with one to six ground-truth
missing elements such as jurisdiction or plaintiff status. Every item is
under-specified -- there are no sufficient controls. A judge reports which
elements the model identified, by naming them, asking about them or
conditioning its answer on them, plus any over-flagged gaps. F2 weights recall
at \(\beta=2\), so missing a genuinely absent fact costs more than mild
over-flagging. Unparseable judge outputs are excluded rather than scored
zero.}
\dce{A tort fact pattern about a goat that injured two farmhands, omitting
which state's law applies and who owns the goat; the ground-truth missing
elements are jurisdiction and ownership.}
\end{datacard}

\subsection{Vals AI Benchmarks}
\label{app:dc:valsai}

\begin{datacard}{Legal Research Bench}
\dcm{Format}{Free response produced by acting within an agentic tool harness}
\dcm{Metric}{All-pass (primary); weighted partial credit (secondary)}
\dcm{Judge}{\texttt{GPT-5.4}, against the per-question expert rubric}
\dcm{Source}{\EXT{} -- Vals AI~\citep{valsai2026legalresearch}}
\dcm{Items}{208 (held-out test split of 413)}
\dcd{Evaluates agents on realistic US legal research tasks: the model must
research a question with case-law search, web search and document retrieval,
then produce a supported answer. Questions span eight practice areas --
Administrative/Regulatory, Business \& Commercial, Civil Litigation,
Constitutional/Civil Rights, Criminal, Family, Health and Immigration -- and
are labelled by reasoning type (statutory interpretation, regulatory framework
interpretation, doctrinal rule reasoning) with overlay flags for reconciliation
across conflicting authority and temporal validity. Every question is authored
and peer-reviewed by practising lawyers and paired with a gold-standard answer,
authoritative sources and a weighted rubric of 1--31 required items (mean
9.35). The primary metric is strict: a question scores 100\% only if every
rubric item passes and 0\% otherwise, on the reasoning that a partially correct
legal answer can read as sound while omitting a critical point; the weighted
partial-credit score is the share of rubric points earned. The dataset splits
into Public (5), Private Validation (200) and Test (208); all reported results
use the private test split. Agents work in a shared harness with five tools
(\texttt{courtlistener\_search}, \texttt{web\_search},
\texttt{retrieve\_information}, \texttt{parse\_html\_page},
\texttt{submit\_final\_result}) under a three-hour limit per task.}
\dce{A Virginia equipment-leasing dispute turning on UCC Article 2A
finance-lease rules: whether a lessee may stop payments after the manufacturer
enters Chapter 7 and disclaims its warranty, and whether the lessor is
responsible for the defective equipment.}
\end{datacard}

\begin{datacard}{TaxEval}
\dcm{Format}{Free response (tax question answering with worked reasoning)}
\dcm{Metric}{Answer correctness; stepwise reasoning quality}
\dcm{Judge}{\texttt{Claude Sonnet 4.5} (non-thinking), one pass per scored dimension}
\dcm{Source}{\EXT{} -- Vals AI~\citep{valsai2025taxeval}}
\dcm{Items}{1{,}223 (held-out test split of 1{,}500+)}
\dcd{Hard tax questions written and double-checked by financial and tax
experts. The same questions are scored on two independent dimensions: answer
correctness, the factual accuracy of the final answer against a ground truth,
and stepwise reasoning, the quality and structure of the derivation compared
against the reasoning of human experts -- so a model cannot score well by
reaching the right figure through unsound work, nor by reasoning plausibly to
the wrong number. Question types are deliberately balanced across application
and compliance (18.3\%), semantic analysis (18.0\%), numerical reasoning
(16.7\%), problem solving and critical thinking (16.5\%), comparative analysis
(16.2\%) and updates and current affairs (15.9\%), the last of which rewards
current knowledge of the tax code. The hardest items require multi-step
computation and a judgement about which rules and figures apply. Splits are
Public Validation (20), Private Validation (300) and Test (1{,}223), the test
set never released.}
\dce{A married-filing-jointly taxpayer with \$300{,}000 AGI receives corporate
and municipal bond interest and sells a collectible held five years; compute
the tax liability on the investment income, including capital gains treatment
and the Net Investment Income Tax.}
\end{datacard}

\begin{datacard}{Finance Agent v2}
\dcm{Format}{Free response produced by acting within an agentic tool harness}
\dcm{Metric}{Dealbreaker-gated weighted partial credit (primary); all-pass
(secondary)}
\dcm{Judge}{Three-judge jury -- \texttt{GPT-5.4}, \texttt{Gemini-3.1-Pro} and \texttt{Claude Sonnet 4.6}}
\dcm{Source}{\EXT{} -- Bigeard et al. (2025)~\citep{bigeard2025fab}}
\dcm{Items}{450 (held-out test split of 927)}
\dcd{Tests whether an agent can do the work of an entry-level financial
analyst, answering difficult questions over public company filings. Questions
are designed to be deterministic (one defensible answer), to require synthesis
across multiple filings rather than a single lookup, to depend on sector
convention that is implied rather than stated, and to turn on detail buried in
footnotes, MD\&A caveats or accounting policy disclosures. The nine categories
follow real equity-research workflows: general qualitative and general
quantitative analysis, market analysis, comparables, precedents, adjustments,
earnings analysis, disclosure analysis and financial modelling, the last two
being by some distance the hardest. Grading is check-based with a subset of
checks flagged as dealbreakers -- load-bearing facts or figures -- and
failing any dealbreaker zeroes the question regardless of the rest of the
answer; the primary score is the dealbreaker-gated, severity-weighted average
of per-check scores, with all-pass reported as a stricter secondary. Splits are
Public (27), Private Validation (450) and Test (450). Agents use six tools
(\texttt{edgar\_search}, \texttt{web\_search}, \texttt{parse\_html\_page},
\texttt{retrieve\_information}, \texttt{calculator}, \texttt{price\_history})
under a two-hour limit, and every model is run three times with the mean
reported.}
\dce{Determine whether Centene owed a rebate to policyholders in fiscal 2020 or
2025, which year came closer to triggering one, and the medical loss ratio in
each -- requiring the MLR threshold rule to be applied to figures pulled from
two 10-K filings.}
\end{datacard}


\newtcolorbox{promptcard}[1]{%
  enhanced,
  breakable,
  colback=blue!10,
  colframe=blue!45!black,
  colbacktitle=blue!14,
  coltitle=black,
  fonttitle=\bfseries\small,
  title={#1},
  boxrule=0.4pt,
  arc=2pt,
  left=6pt, right=6pt, top=5pt, bottom=5pt,
  before skip=8pt, after skip=8pt,
  fontupper=\footnotesize\ttfamily,
}

\newcommand{\pslot}[1]{\{#1\}}

\section{Fusion Prompts}
\label{app:fusion-prompts}

Fusion synthesises a single answer from the sampled candidates in one call.
We use a direct-synthesis prompt, which emits the fused answer immediately
rather than producing an intermediate comparison of the candidates, in two
variants selected by whether candidate reasoning is shown to the fusor.
Both templates take the full conversation, including the system prompt and
any prior turns, in the \texttt{instruction} slot, and the shuffled
candidates in the \texttt{generations} slot.

Two instructions are common to both variants. The fusor is told to answer
as if directly addressing the task, without referring to the candidates,
which otherwise leak into the output as phrases such as ``combining the
best of both responses''. It is also told that only the task input's format
constraints apply to its output, since candidates carry formatting of their
own that the fusor will otherwise reproduce.

\subsection{With Candidate Reasoning}
\label{app:fusion-prompt-reasoning}

Used where the candidates' reasoning traces are informative for synthesis.
Each candidate's thinking block is rewritten into explicit
\texttt{Generation N reasoning:} and \texttt{Generation N answer:} labels
before the prompt is formatted, so the fusor reads the trace as evidence
rather than treating the tag structure as a format to mimic. The prompt
names that structure so the distinction between a candidate's internal work
and its answer is explicit.

\begin{promptcard}{Direct fusion -- reasoning visible}
Based on the provided Task Input and Generated Texts, fuse them into a
better generation that combines the strength of each of them. The fused
generation should adequately respond to the task input, sound natural to a
native speaker, and be focused on conveying the most relevant and accurate
information in a responsible and ethical way.

\medskip
Each Generated Text may include a "Generation N reasoning:" block (the
candidate's thought process) followed by a "Generation N answer:" block
(its final answer). Treat the candidates' structure as their internal work.
Do not mirror it in your output. Only the Task Input's format constraints
apply to your output.

\medskip
\#\#\# Generated Texts \\
\pslot{generations}

\medskip
\#\#\# Task Input \\
\pslot{instruction}

\medskip
Output only the fused generation text, as if you were directly answering
the task yourself. Do not reference or mention the previous generations
(e.g., avoid phrases like "combining the best of both responses" or "as
mentioned in Generation 1").

\medskip
Please provide your fused text.
\end{promptcard}

\subsection{Without Candidate Reasoning}
\label{app:fusion-prompt-no-reasoning}

Used on legal-reasoning tasks, where candidate reasoning is stripped before
formatting and the fusor sees answers alone. The prompt is otherwise
identical, with the reasoning-block paragraph replaced by a statement that
each Generated Text is a candidate answer.

\begin{promptcard}{Direct fusion -- reasoning stripped}
Based on the provided Task Input and Generated Texts, fuse them into a
better generation that combines the strength of each of them. The fused
generation should adequately respond to the task input, sound natural to a
native speaker, and be focused on conveying the most relevant and accurate
information in a responsible and ethical way.

\medskip
Each Generated Text is a candidate's answer to the Task Input. Only the
Task Input's format constraints apply to your output.

\medskip
\#\#\# Generated Texts \\
\pslot{generations}

\medskip
\#\#\# Task Input \\
\pslot{instruction}

\medskip
Output only the fused generation text, as if you were directly answering
the task yourself. Do not reference or mention the previous generations
(e.g., avoid phrases like "combining the best of both responses" or "as
mentioned in Generation 1").

\medskip
Please provide your fused text.
\end{promptcard}

\subsection{Candidate Formatting}
\label{app:fusion-formatting}

Candidates are rendered under \texttt{\#\# Generation N} headers in the
\texttt{generations} slot, in shuffled order so the fusor does not favour a
fixed position; the shuffle order is recorded per example. In the
reasoning-visible variant a candidate is relabelled only when both the
opening and closing thinking tags are present, and candidates without them
are passed through unchanged.

\end{document}